\documentclass[11pt, a4paper, logo, copyright]{deepmind}

\usepackage[authoryear, sort&compress, round]{natbib}
\usepackage{cleveref}
\usepackage{caption}
\usepackage{subcaption}
\usepackage{listings}
\usepackage{dm-colors}

\hypersetup{colorlinks=false}

\title{Does Circuit Analysis Interpretability Scale? Evidence from Multiple Choice Capabilities in Chinchilla}

\correspondingauthor{tlieberum@deepmind.com}

\renewcommand{\today}{2023-07-18}

\author{Tom Lieberum}
\author{Matthew Rahtz}
\author{J\'{a}nos Kram\'{a}r}
\author{Neel Nanda}
\author{Geoffrey Irving}
\author{Rohin Shah}
\author{Vladimir Mikulik}

\affil{Google DeepMind}

\begin{document}

\begin{abstract}
\emph{Circuit analysis} is a promising technique for understanding the internal mechanisms of language models. However, existing analyses are done in small models far from the state of the art. To address this, we present a case study of circuit analysis in the 70B Chinchilla model, aiming to test the scalability of circuit analysis. In particular, we study multiple-choice question answering, and investigate Chinchilla's capability to identify the correct answer \emph{label} given knowledge of the correct answer \emph{text}. We find that the existing techniques of logit attribution, attention pattern visualization, and activation patching naturally scale to Chinchilla, allowing us to identify and categorize a small set of `output nodes' (attention heads and MLPs).

We further study the `correct letter' category of attention heads aiming to understand the semantics of their features, with mixed results. For normal multiple-choice question answers, we significantly compress the query, key and value subspaces of the head without loss of performance when operating on the answer labels for multiple-choice questions, and we show that the query and key subspaces represent an `Nth item in an enumeration' feature to at least some extent. However, when we attempt to use this explanation to understand the heads' behaviour on a more general distribution including randomized answer labels, we find that it is only a partial explanation, suggesting there is more to learn about the operation of `correct letter' heads on multiple choice question answering.

\end{abstract}

\maketitle

\section{Introduction}

\begin{figure}[ht]
     \centering
     \includegraphics{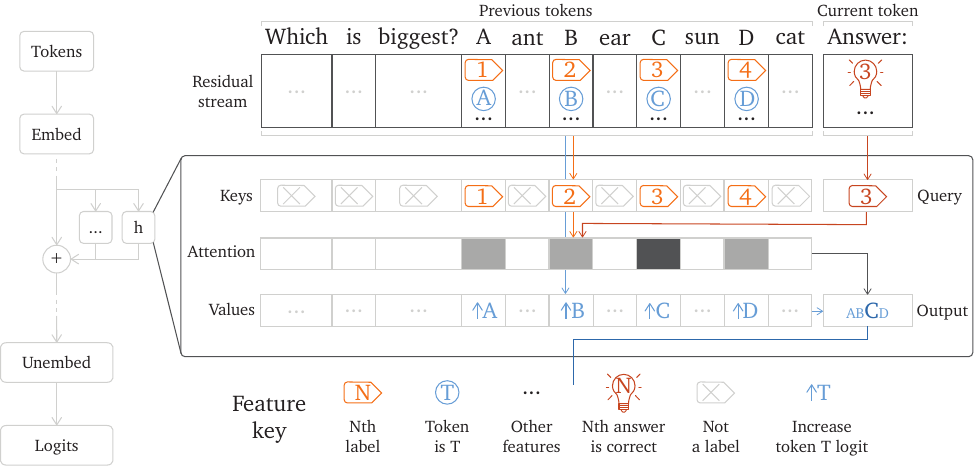}
     \caption{Overview of the most interesting attention heads we identified -- the `correct letter' heads. At the final token position, the head strongly attends to the letter A, B, C or D corresponding to the correct answer, and copies this letter to the output logits. To do this, each head computes a query consisting of two features: first, a feature encoding whether or not the token is a label to rule out tokens other than A, B, C or D; and second, based on information written to the residual stream by previous parts of the circuit, an `Nth label' feature which selects for the correct answer letter specifically. Using the resulting attention, the head focuses on the value for the correct answer letter, which increases the logit for that letter. Note that the head's operation is more messy than this diagram indicates: see~\cref{sec:understanding_nodes} for more details.}
     \label{fig:overview}
\end{figure}

Current methods for training and evaluation in large language models currently focus on the behaviour of the model~\citep{ziegler2019fine,bai2022constitutional,ouyang2022training,saunders2022self,glaese2022improving,perez2022red}. \emph{Mechanistic interpretability} aims to generate detailed knowledge of a model's internal reasoning, and thus could significantly improve upon these methods. For example, such knowledge would strengthen methods that aim to oversee models' reasoning, as in debate~\citep{irving2018ai} and process-based feedback~\citep{uesato2022solving, lightman2023lets}. Furthermore, the ability to examine models' full reasoning processes could help us detect \emph{deceptive alignment}~\citep{kenton2021alignment,hubinger2019risks}, a key source of extreme risk~\citep{openai2023gpt4,shevlane2023model} in which a model behaves well to deliberately conceal its undesirable intentions.

We focus on \emph{circuit analysis}: the identification and study of particular internal mechanisms that drive a specific subset of models' behaviour. Existing circuit analysis on language models has a variety of weaknesses, but in this work we focus on two in particular.
First, the models studied are relatively small: for example, the seminal work on transformer circuits focused on two-layer attention-only transformers~\citep{elhage2021mathematical} and research on the circuits used in grammatical identification of indirect objects was done on the 117M variant of GPT-2~\citep{wang2022interpretability}.
Second, prior work identifies which components of a model are relevant and how information flows between them, but usually does not focus as much on \emph{what} information is flowing, such that we could predict the circuit's behaviour on an expanded data distribution.

We address the first weakness by investigating a model of a significantly larger size: the 70B-parameter \emph{Chinchilla} model~\citep{hoffmann2022training}. Concretely, we investigate the circuit underlying multiple-choice question-answering in the Massive Multitask Language Understanding (MMLU) benchmark~\citep{hendrycks2020measuring}.
Typically, MMLU is considered challenging because of the vast breadth of knowledge required. However, as we show in \cref{sec:mmlu}, the difficulty for language models also derives from the algorithmic aspect: in particular, not only must the model determine which answer is correct, it must identify the letter corresponding to that correct answer and output that letter. This makes it an ideal test for testing the scalability of existing tools for circuit analysis: like other cases where circuit analysis has found success~\citep{wang2022interpretability, nanda2023progress, chan2022causal}, the task is algorithmic, and unlike previous cases, it only emerges at scale (and in particular is not present in a 7B-parameter model)\footnote{Note however that the emergence with scale is likely because multiple-choice questions are rare in the training data, rather than the task being inherently challenging for neural networks to learn.}. For this reason, we limit the scope of our investigation to the algorithmic aspect of the circuit, and leave the knowledge retrieval aspect to future work. In \cref{sec:finding_nodes}, we find that existing techniques scale successfully: through a combination of logit attribution and attention pattern visualization, we identify `correct letter' heads that perform the algorithmic task, and validate the circuit through activation patching~\citep{chan2022causal}.

To address the second weakness, we investigate a variety of techniques for generalizing our understanding of the `correct letter' heads to a broader distribution in \cref{sec:understanding_nodes}. In particular, we use singular value decomposition (SVD) to identify 3-dimensional subspaces that capture the queries, keys, and values for the head when limited to the distribution of MMLU questions, and investigate the behavior of these subspaces on mutated prompts to determine what features they represent. These analyses suggest that the query and key subspaces encode a general `n-th item in an enumeration' feature while the value subspace encodes the token identity, suggesting an overall algorithm illustrated in~\cref{fig:overview}. However, we emphasize that our results are mixed: the identified direction does not always explain the head's behaviour on broader distributions, and in particular only partially explains behaviour when the labels are randomised letters~(\cref{fig:loss_lr_attn_randomise_answer_letters}).

Overall, we see this case study as providing a data point suggesting that while algorithmic tasks can be quite interpretable, the specific features used to implement them can be quite messy in their semantics, even when limited to a distribution where we expect little superposition of features.

In summary, our contributions are as follows:
\begin{enumerate}
    \item We demonstrate that the existing circuit analysis techniques of logit attribution, attention pattern visualization, and activation patching can be readily applied to a large (70B) model to identify and understand the final nodes of the multiple-choice question-answering circuit.
    \item We investigate the high-level features used by `correct letter' heads, with mixed results: we identify a low-dimensional subspace that approximately encodes `n-th item in an enumeration', but the subspace only partially explains behaviour on a more general distribution.
\end{enumerate}

\section{Background}

\subsection{Chinchilla}

The object of this study is Chinchilla 70B~\citep{hoffmann2022training}, a compute-optimally trained large language model using a decoder-only transformer architecture. The model has 80 layers, with 64 attention heads per layer, with RMSNorm before each component (MLP and self-attention block) and before the final unembedding matrix, and linear relative positional embeddings~\citep{dai2019transformer}.

\begin{figure}[t]
    \centering
    \includegraphics[width=0.7\textwidth]{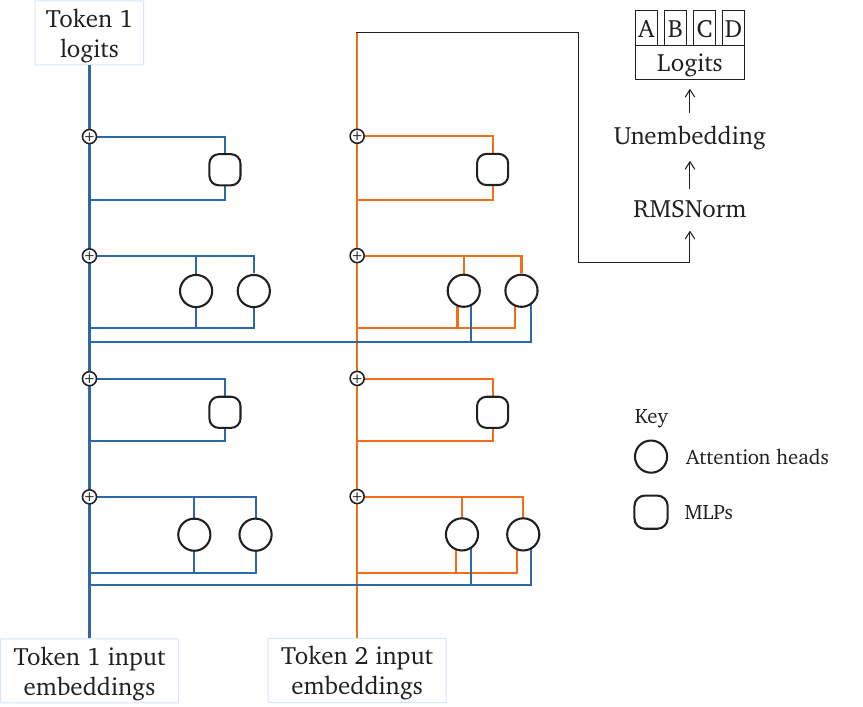}
    \caption{Circuit diagram of decoder-only transformer}
    \label{fig:transformer_diagram}
\end{figure}

RMSNorm scales its input to have unit root mean square (RMS) and then multiplies with a learned gain vector. To simplify analysis, we combine this learned gain vector with the weight matrix following the RMSNorm, such that the RMSNorm itself becomes purely a normalisation by the RMS. (See also~\citet{elhage2021mathematical} for a related discussion on LayerNorm.)

Given a fixed RMS, the residual architecture of the transformer means that we can write the output logits of the model $\mathcal{L}$ in terms of the unembedding matrix $W_U$, and for each layer $\ell$, the outputs of the MLP $m_\ell$ and the outputs of the $i$th head $h_\ell^i$:

\begin{align}
    \mathcal{L} = \operatorname{softmax}\left(W_U\frac{1}{RMS} \sum^{80}_{\ell=1}\Big[ m_\ell + \sum^{64}_{i=1}h^i_\ell \Big]\right).
\end{align}

This formulation makes it clear that in principle every component has a direct, linear connection to the logits, given the fixed RMS. Empirically, the change in final RMS contributed by any given component when patching it is small relative to the final RMS as that is dominated by the final layers. This may break down however when e.g. zero ablating components in the last few layers which contribute a majority to the final RMS.

For more details on how to conceptualize decoder-only transformers in the context of interpretability, we encourage the reader to consult~\citet{elhage2021mathematical}.

\subsection{Massive Multitask Language Understanding (MMLU)}
\label{sec:mmlu}

\begin{figure}[t]
    \centering
    \includegraphics[width=\textwidth]{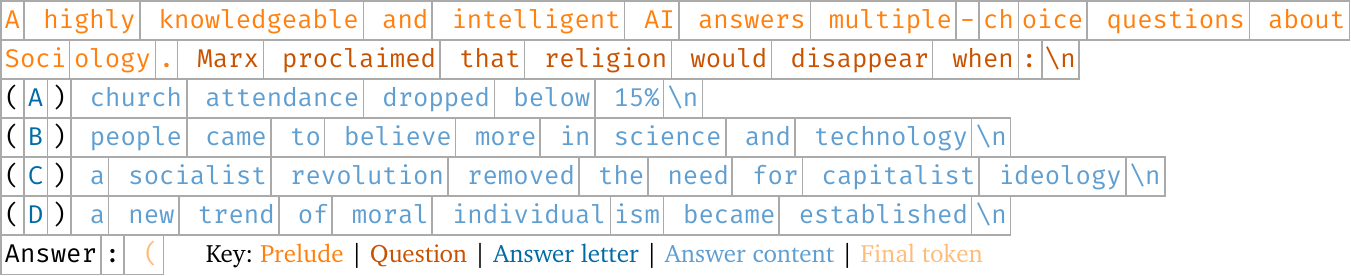}
    \caption{Example prompt from MMLU, with token boundaries indicated by grey lines. Note that the letter tokens \texttt{A}, \texttt{B}, \texttt{C} and \texttt{D} are tokenised separately.}
    \label{fig:mmlu_tokens}
\end{figure}

To study multiple-choice question-answering, we use the Massive Multitask Language Understanding  benchmark (MMLU)~\citep{hendrycks2020measuring}. The full benchmark consists of roughly 16,000 examples on topics ranging from high school biology to professional accounting. We limit ourselves to a subset of 6 topics of the benchmark which Chinchilla performs particularly well on. We process examples from the dataset into prompts as shown in~\cref{fig:mmlu_tokens}. We used the particular prompt at the end to force the model to focus on the token `\texttt{X}', rather than spreading its prediction between several almost identical tokens such as `\texttt{X}', ` \texttt{X}', ` \texttt{X}.', ` \texttt{X},', etc. Considerations such as these are unfortunately common when engaging in mechanistic interpretability, highlighting the need for exceeding care when tokenization is involved. During the analysis in~\cref{sec:finding_nodes} and~\cref{sec:understanding_nodes} we use 0-shot prompting. 

MMLU is an interesting benchmark to study because smaller models perform quite badly at it. We investigate three models of the Chinchilla family of sizes 1B, 7B and 70B with results on the standard 5-shot version of MMLU shown in~\cref{tab:chinch_mmlu_acc}. Only the 70B model is able to perform well in the standard setting. Chinchilla 7B is able to perform better than random but only if scored against the \emph{text} of the correct answer, rather than the label A, B, C or D. This suggests that Chinchilla 7B lacks the ability to perform the required symbol manipulation, while still possessing some of the relevant knowledge. To further support this claim, we investigate the performance of these models on a synthetic multiple choice dataset which does not require factual knowledge and only requires the ability to choose the option corresponding to a random token that was asked about. Of the three models, only Chinchilla 70B is able achieve better than random performance on this task; see~\cref{sec:syntactic_abcd_results} for details.

\begin{table}[ht]
    \centering
    \begin{tabular}{l||c|c}
         Model Size & Label & Text \\
         \hline\hline
         1B& 25\%& 27\% \\
         \hline
         7B& 26\%& 32\%\\
         \hline
         70B& 68\%& 65\%\\
    \end{tabular}
    \caption{Accuracy on 5-shot MMLU by various sizes of the Chinchilla family when scoring either by the label (A, B, C or D) or the content text of the correct answer.}
    \label{tab:chinch_mmlu_acc}
\end{table}

\subsection{Activation Patching}
\label{sec:patching_background}
\begin{figure}[ht]
     \centering
     \begin{subfigure}[b]{0.3\textwidth}
         \centering
         \includegraphics[width=\textwidth]{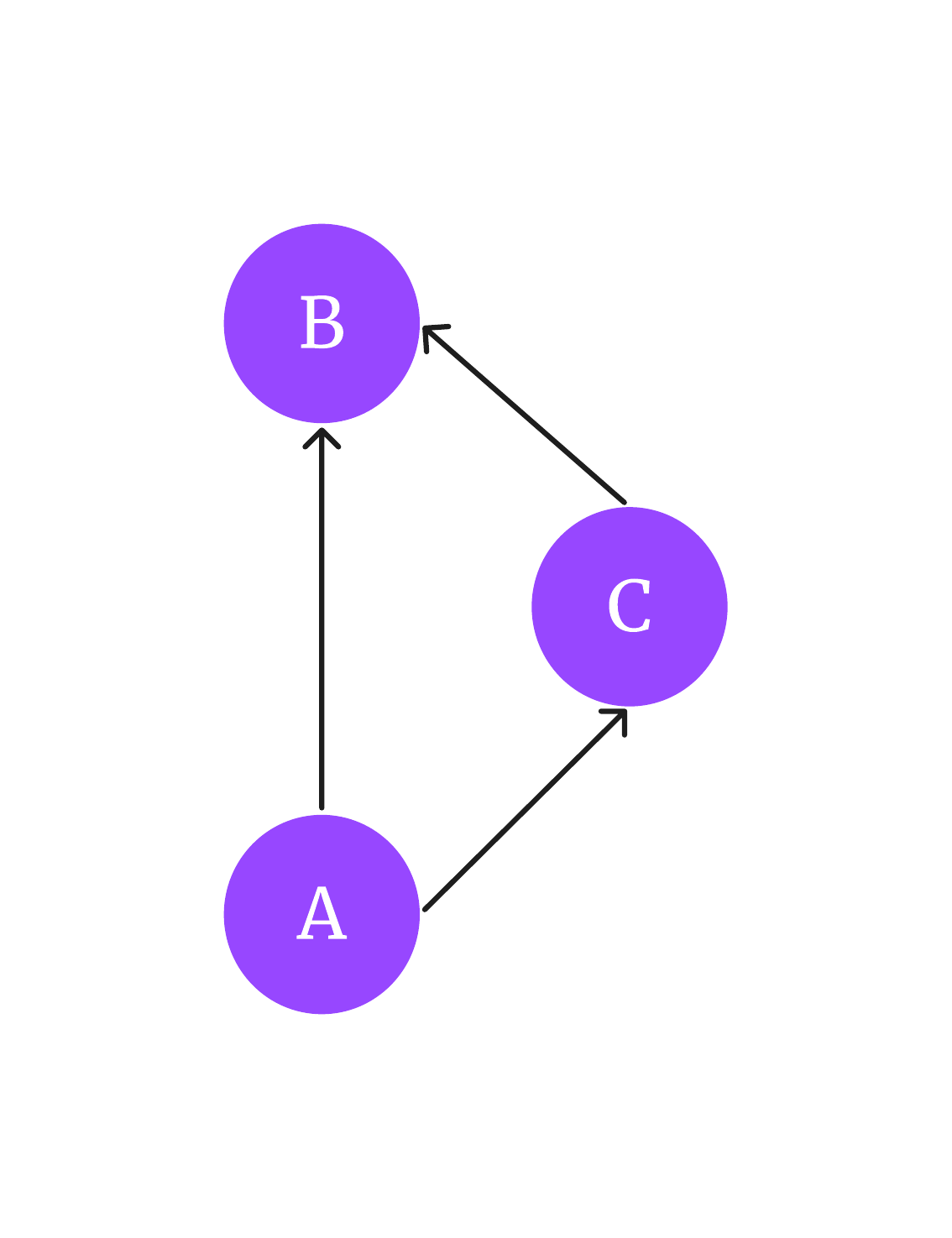}
         \caption{Clean forward pass, no intervention}
         \label{fig:toy_causal_diagram_normal}
     \end{subfigure}
     \hfill
     \begin{subfigure}[b]{0.3\textwidth}
         \centering
         \includegraphics[width=\textwidth]{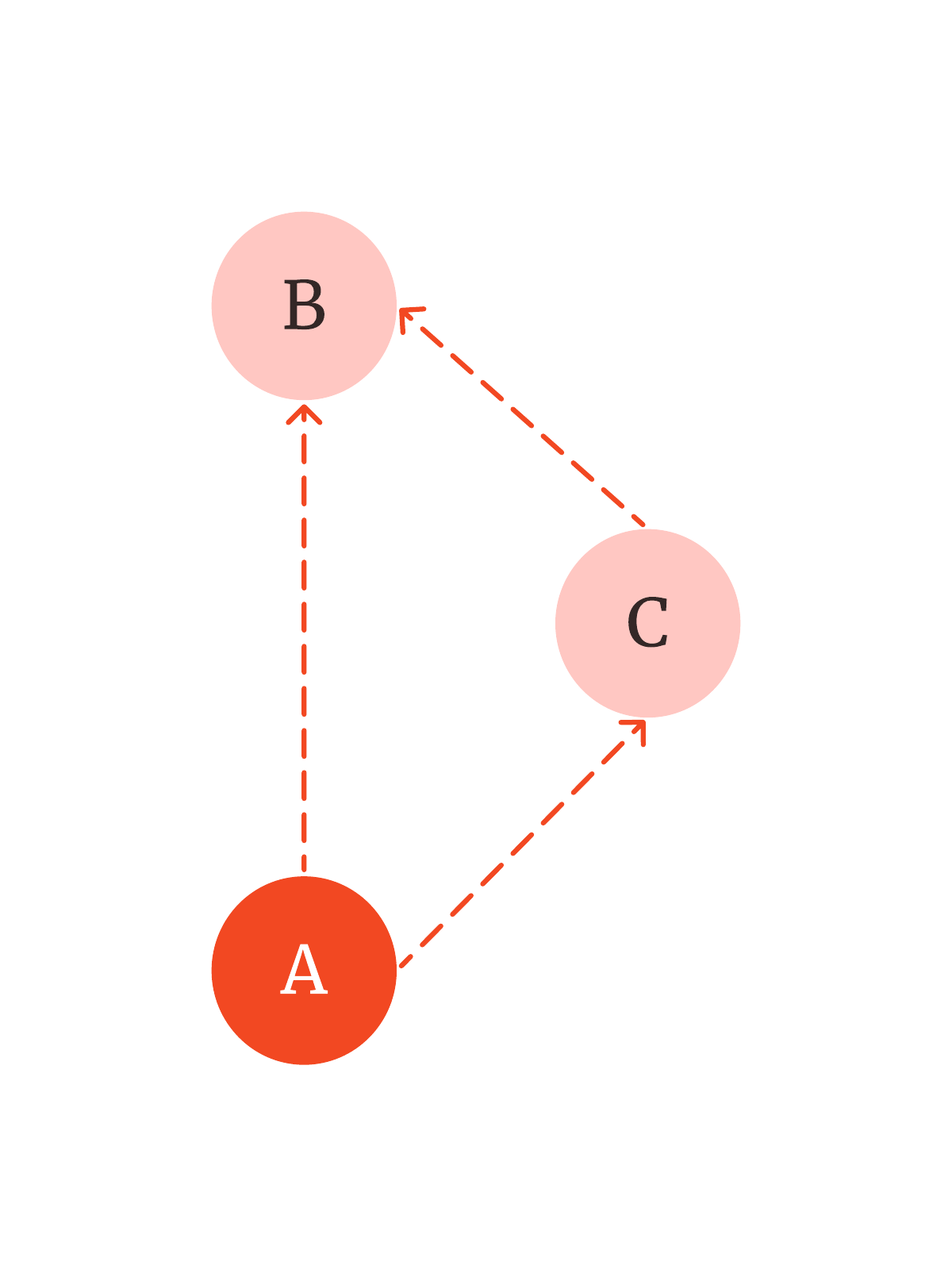}
         \caption{Intervene on A to observe \emph{total} effect on B.}
         \label{fig:toy_causal_diagram_total}
     \end{subfigure}
     \hfill
     \begin{subfigure}[b]{0.3\textwidth}
         \centering
         \includegraphics[width=\textwidth]{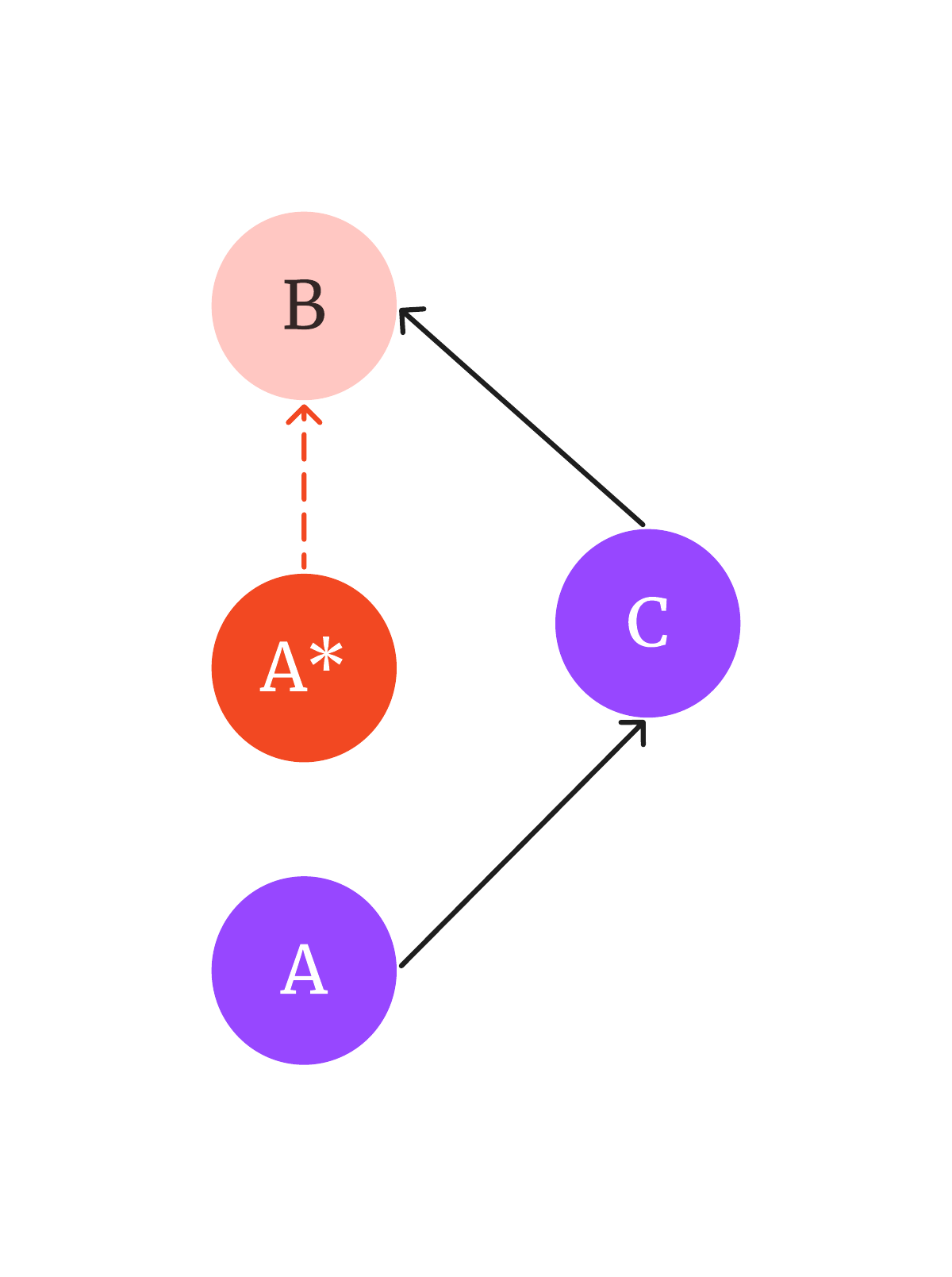}
         \caption{Intervene on the edge A$\to$B to observe \emph{direct} effect on B.}
         \label{fig:toy_causal_diagram_direct}
     \end{subfigure}
        \caption{Simple toy causal diagram. A has a direct effect on B, but also an indirect effect mediated via C.}
        \label{fig:toy_causal_diagram}
\end{figure}

To identify the relevant circuit nodes, we are focusing on a technique termed `activation patching', e.g. used by~\citet{wang2022interpretability} and~\citet{meng2022locating}, or `patching' for short. However note that the general idea has also been developed in the causal inference literature under the name of do-calculus~\citep{pearl1995causal,pearl2012calculus}. We will explain the idea on a simple causal network displayed in~\cref{fig:toy_causal_diagram}. In that network, node A has a direct effect on nodes B and C, and C has a direct effect on B, meaning that A also has an \emph{indirect} effect on B via C. 

To determine the effect that a node A has on node B, we can intervene on the node A by forcing it to have a different value than it would otherwise have and observe the resulting change in B, displayed in~\cref{fig:toy_causal_diagram_total}. This captures the \emph{total} effect of A on B, i.e. the sum of direct and indirect effects. We can also isolate the direct effect that A has on B. To do so, we replace the value that is passed on from A to B with a different value A$^*$, shown in~\cref{fig:toy_causal_diagram_direct}. This can also be seen as intervening on the edge A$\to$B~\citep{wang2022interpretability}.

\subsubsection{Intervening in Chinchilla 70B}

We view the language model in question as a causal graph, where we interpret attention heads and MLPs as the nodes or variables. Edges are implicitly given by a direct path between these nodes, e.g. each attention head will have an edge to all future nodes in the graph, since they are implicitly directly connected via the residual stream~\citep{elhage2021mathematical}. We will usually treat each attention head in each layer at a given token position as a separate node. While we will likewise treat each MLP at each layer and position as a separate node, this is simply due to us not investigating MLPs more deeply and we suggest treating each hidden neuron or possibly groups of neurons as the atomic unit instead in future work, as this seems more reflective of the semantics of the computation happening inside deep neural networks~\citep{olah2017feature,cammarata2020thread, gurnee2023finding}.

Our typical interventions take the form of \emph{resampling ablations}. (See~\cref{sec:related_work_circuit_tracing} and~\citet{chan2022causal} for discussion on other forms of ablation, such as zero ablation and mean ablation.) We start out with a forward pass of the model on a sampled prompt $p_{original}$. To intervene on a node, we replace the node's activation in that forward pass from a prompt $p_{intervention}$ resampled from the same distribution such that it differs in key details -- in our case, which answer is correct. For example, if we are interested in the total effect of node A, we can replace its activation during a forward pass on $p_{original}$ with its activation on $p_{intervention}$ and then measure the difference in loss (the difference in negative log probability of the correct answer from $p_{intervention}$). Intervening on an edge A $\to$ B can also be done straightforwardly in a transformer model, since the pre-RMSNorm input to each node is a sum of the outputs of all previous nodes. Thus, given outputs of the node A on each prompt $A(p_{original})$ and $A(p_{intervention})$, we can replace the pre-RMSNorm input $x$ to B with $x - A(p_{original}) + A(p_{intervention})$.

\section{Identifying the circuit using existing techniques}
\label{sec:finding_nodes}

In this section, we apply logit attribution, activation patching, and attention pattern visualization to identify the final nodes in the circuit. These are the nodes that have a large direct positive effect on the final logits, i.e. as a direct result of their output, the correct token is assigned a higher probability than before. We find that there is a set of 45 nodes (attention heads and MLPs) which are causally responsible for recovering almost all of the model's performance through direct effects when patched, suggesting they are located towards the end of the circuit. We provide preliminary evidence about further nodes in the circuit in~\cref{sec:rest_of_the_circuit}.

\subsection{Identifying final nodes in the circuit}
\label{sec:ident_final_nodes}
Final nodes in the circuit have as a necessary condition that their output is directly affecting final logits\footnote{This condition is not sufficient however, since in theory their indirect effect could ``cancel out'' their direct effect, resulting in a total effect of zero or even of the opposite sign, as we will discuss later.}.
In order to compute the direct effect of each model component we can exploit the fact that for a fixed scaling factor, the final logits are the sum of the individual components' contributions. Thus we can avoid having to run the model separately for every component and can get the effect for all components in parallel.

We restrict ourselves to a subset of 6 MMLU topics which Chinchilla performs particularly well on, with the assumption that this makes it easier to identify the circuit. For each prompt $p$, we run a forward pass of the model and collect the outputs $f(p)$ of each component in the circuit -- each attention head and MLP. We also save the RMS of the final residual stream value (pre-RMSNorm) as $RMS_p$. The direct contribution of each component to the logits is then given by

\begin{align}
    \Delta = \frac{1}{RMS_p} W_U f(p)
\end{align}

where $W_U$ is the unembedding matrix.

However, note that the softmax converting logits to probabilities is invariant under constant shifts. To isolate the net effect, we can either subtract the mean logit over the full vocabulary, or the mean logit over the set of possible tokens (i.e. A, B, C, or D). Both of these approaches have benefits and drawbacks. The former will also identify nodes which will decrease the probability of the correct answer token via decreasing logits of non-ABCD tokens, while leaving the ABCD logits unchanged. The latter on the other hand has the issue that it does not capture nodes which serve the task of identifying the set of relevant tokens among all possible tokens, without paying attention to which answer specifically is the correct one. A full investigation will need to consider both options. In this work we focus on the latter approach, as we are interested in the question of how the model knows the correct content text as opposed to how it knows that it should answer a multiple choice question in general.

Let $W_U^{ABCD}$ be the unembedding matrix restricted to the tokens A, B, C, D. Following the discussion above, we define the net effect on ABCD by component $f$ on prompt $p$ as

\begin{align}
    \Delta_{ABCD}(p) &= \frac{1}{RMS_p} W_U^{ABCD} f(p) \\
    \Delta_{ABCD \, net}(p) &= \Delta_{ABCD}(p) - \underset{\text{ABCD}}{\operatorname{mean}}\Big[\Delta_{ABCD}(p)\Big]
\end{align}
where the mean is taken over the token axis. Finally, to get the effect on the \emph{correct} token, we index $\Delta_{ABCD}$ with the corresponding token index. (Note therefore that the unit of this quantity is delta logits -- specifically, change in the logit of the correct answer token.)

\begin{figure}[t]
    \centering
    \includegraphics[width=0.7\textwidth]{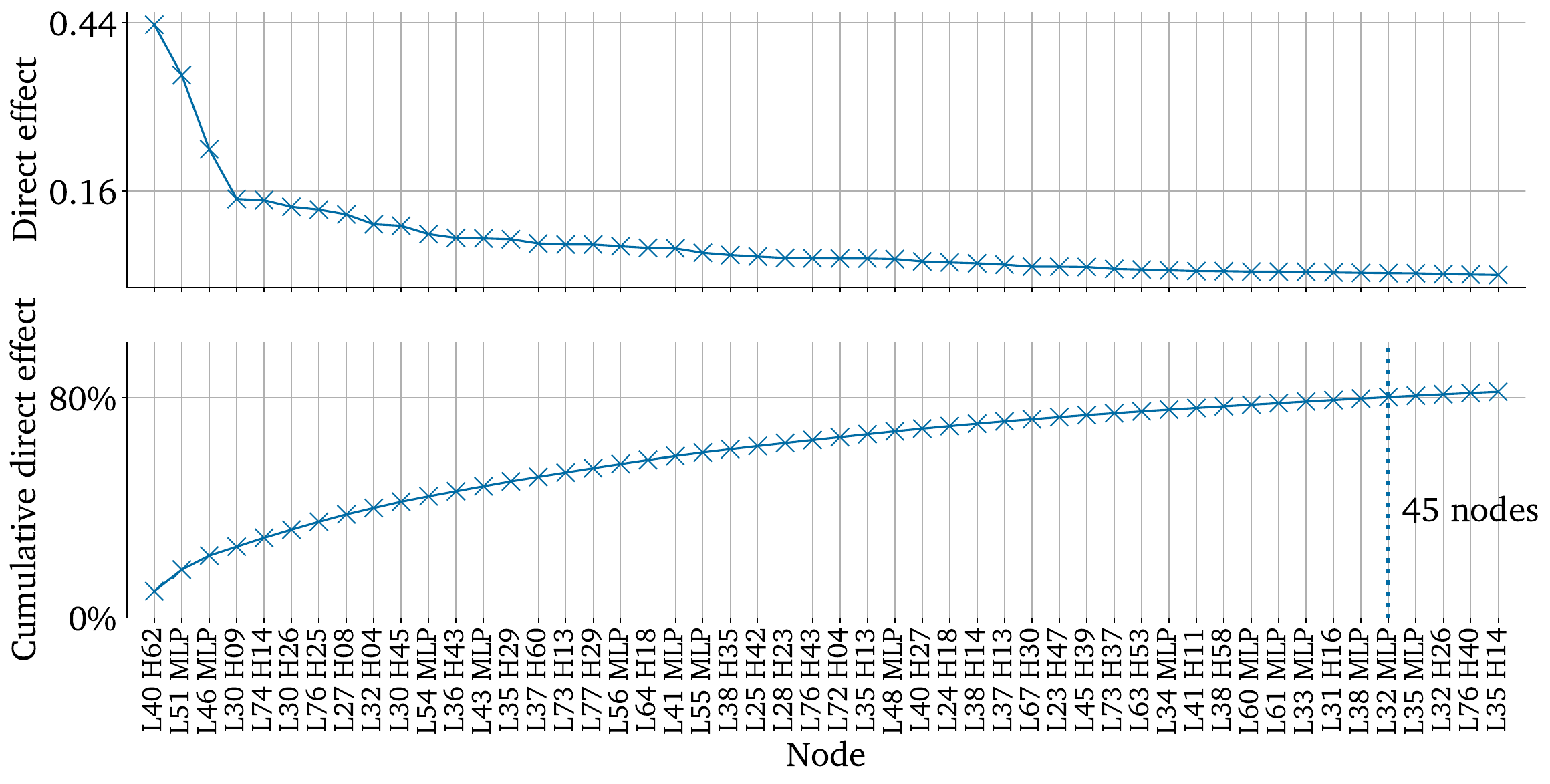}
    \caption{Net direct effect of each component and cumulative net direct effect, sorted in descending order.}
    \label{fig:net_direct_effect}
\end{figure}

We average the effect over 128 prompts, randomly sampled from the dataset. The results of this analysis are shown in~\cref{fig:net_direct_effect}.
We observe that there are a few nodes with moderately high direct effects and a long tail of small direct effects. Concretely, 45 nodes explain 80\% of the summed positive \footnote{Some nodes have a consistently \emph{negative} direct effect. We omit these when calculating the denominator of the fraction.} direct effect over all nodes, which we will analyze more closely. These 45 nodes are comprised of 32 attention heads and 13 MLPs.

\begin{figure}[t]
    \centering
    \includegraphics[width=0.7\textwidth]{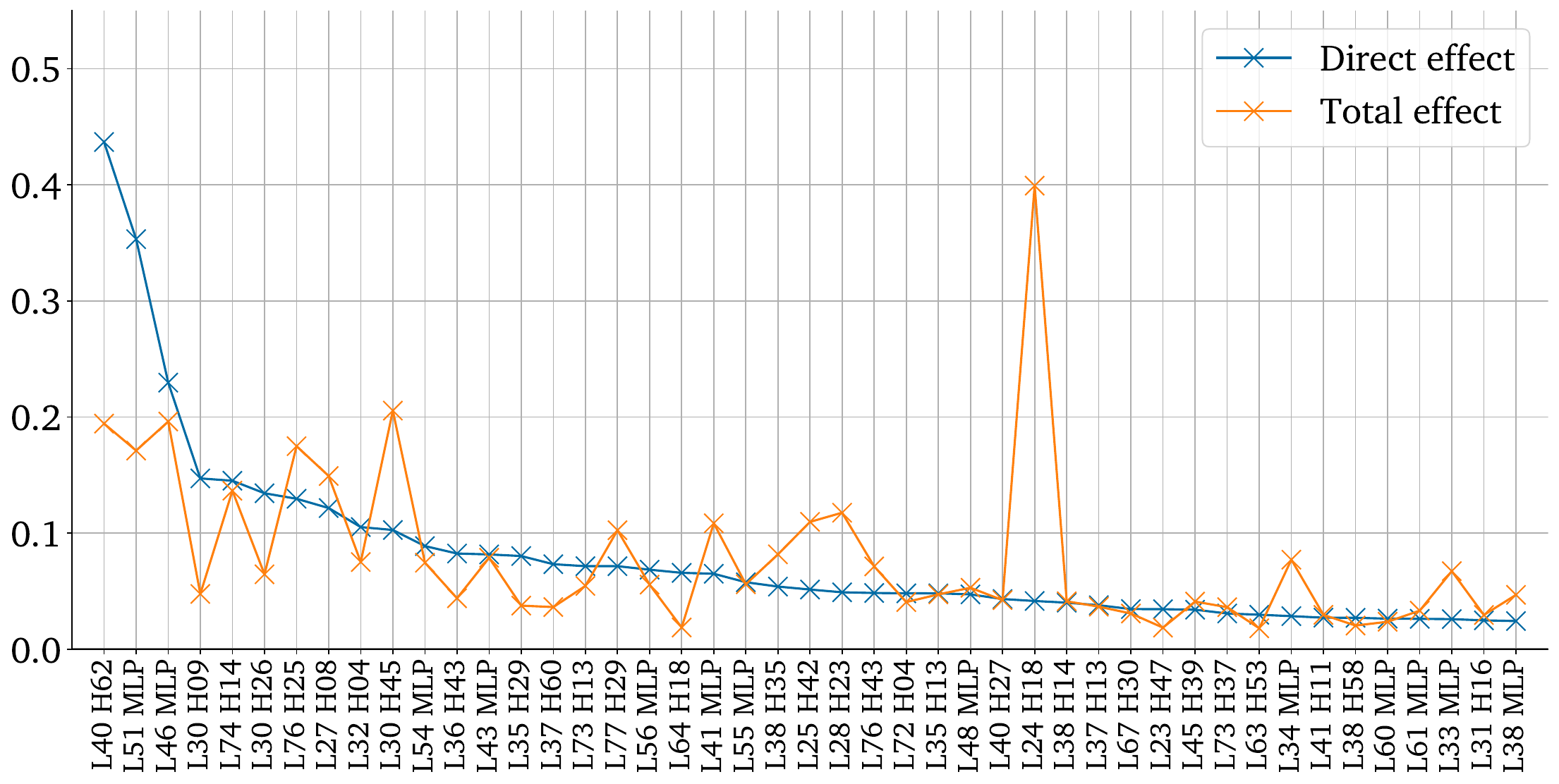}
    \caption{Direct and total effect of the nodes with highest direct effect.}
    \label{fig:total_and_direct_effect}
\end{figure}

Since direct effect does not necessarily imply total effect, we run an activation patching experiment for every one of the 45 nodes and record the average total net effect on the correct answer logit. For this, we sample two prompts $p_{original}$ and $p_{intervention}$. We patch in the activation of a node on $p_{intervention}$ into a forward pass on $p_{original}$ and record the net change in logits on the token corresponding to the correct answer letter according to $p_{intervention}$. The results of this are shown in~\cref{fig:total_and_direct_effect}. Overall the total effect seems to track the direct only somewhat. Two things stand out.

First, the top two direct effect nodes have significantly lower total effect. As of now we do not have a satisfying explanation for this. The difference may simply be attributable to the fact that the total effect considers additional pathways through the model. (In particular, we hypothesise that latter parts of the model may perform confidence calibration, weakening logits that are too strong, though we did not investigate this.) However, there may also be more subtle causes, such as unintended effects from mixing activations from two different prompts, or from the fixing the RMS in the direct effects calculation.

Second, there is a large spike in total effect at \texttt{L24 H18}. In~\cref{sec:rest_of_the_circuit} we show that this head is a crucial input to the queries of what we call correct letter heads, by moving information from the correct content tokens to the final token. Thus its total effect is dominated by the indirect effect via the correct letter heads. 

We show the total and direct effects broken down by the correct letter in~\cref{sec:nuances_identifying_output} and~\cref{sec:net_direct_effect_facetted} respectively. As we note in~\cref{sec:nuances_identifying_output}, the per letter total effect results are somewhat confusing, and in particular in contrast to our other results. We suspect that one reason for this could be that the model implements some kind of backup behavior~\citep{wang2022interpretability} distorting the effect of patching a single node.

\begin{figure}[t]
    \centering
     \begin{subfigure}[b]{0.49\textwidth}
         \centering
         \includegraphics[width=\textwidth]{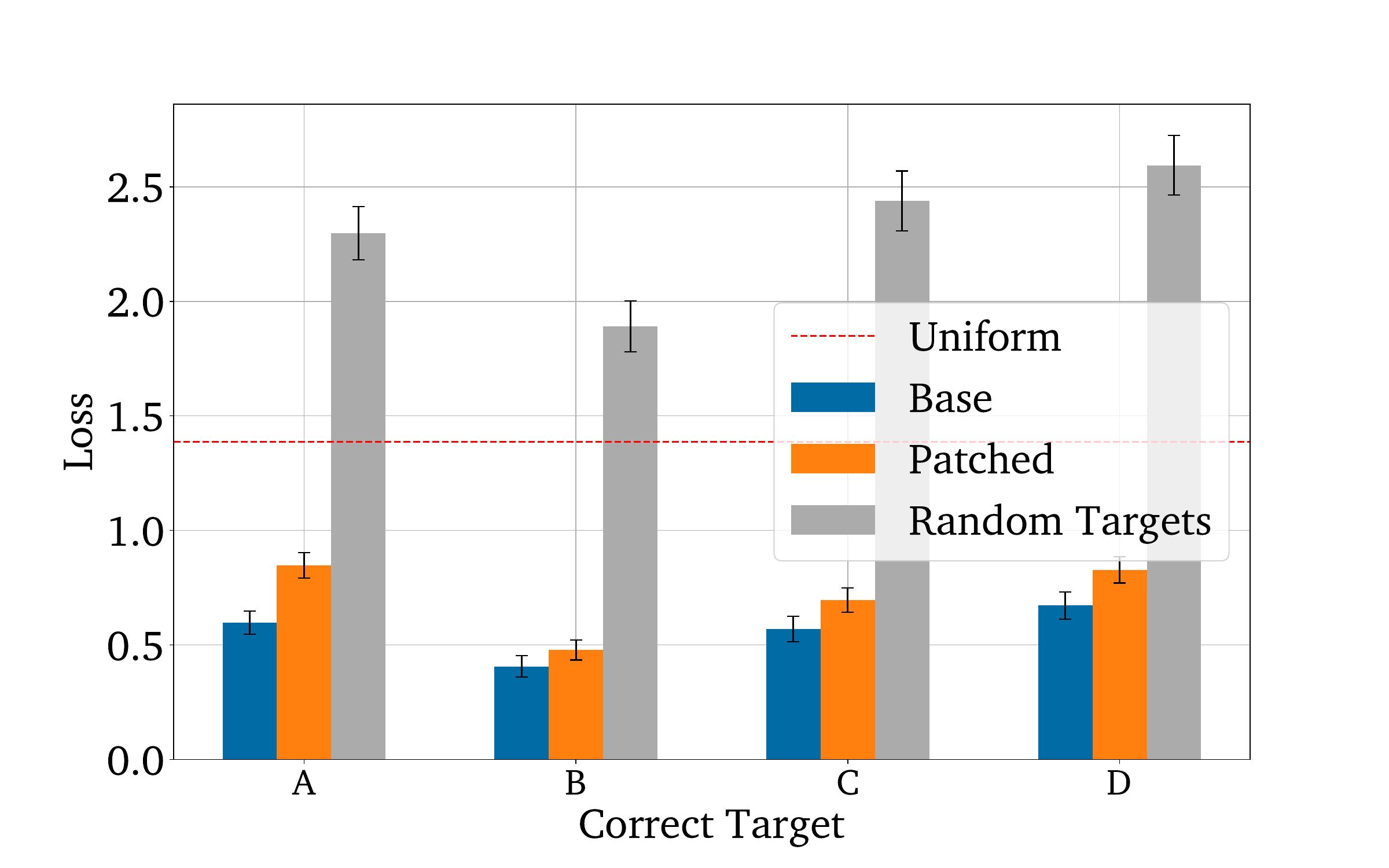}
         \caption{Loss (negative log probability of correct answer letter, taking softmax over the full vocabulary)}
     \end{subfigure}
     \hfill
     \begin{subfigure}[b]{0.49\textwidth}
         \centering
         \includegraphics[width=\textwidth]{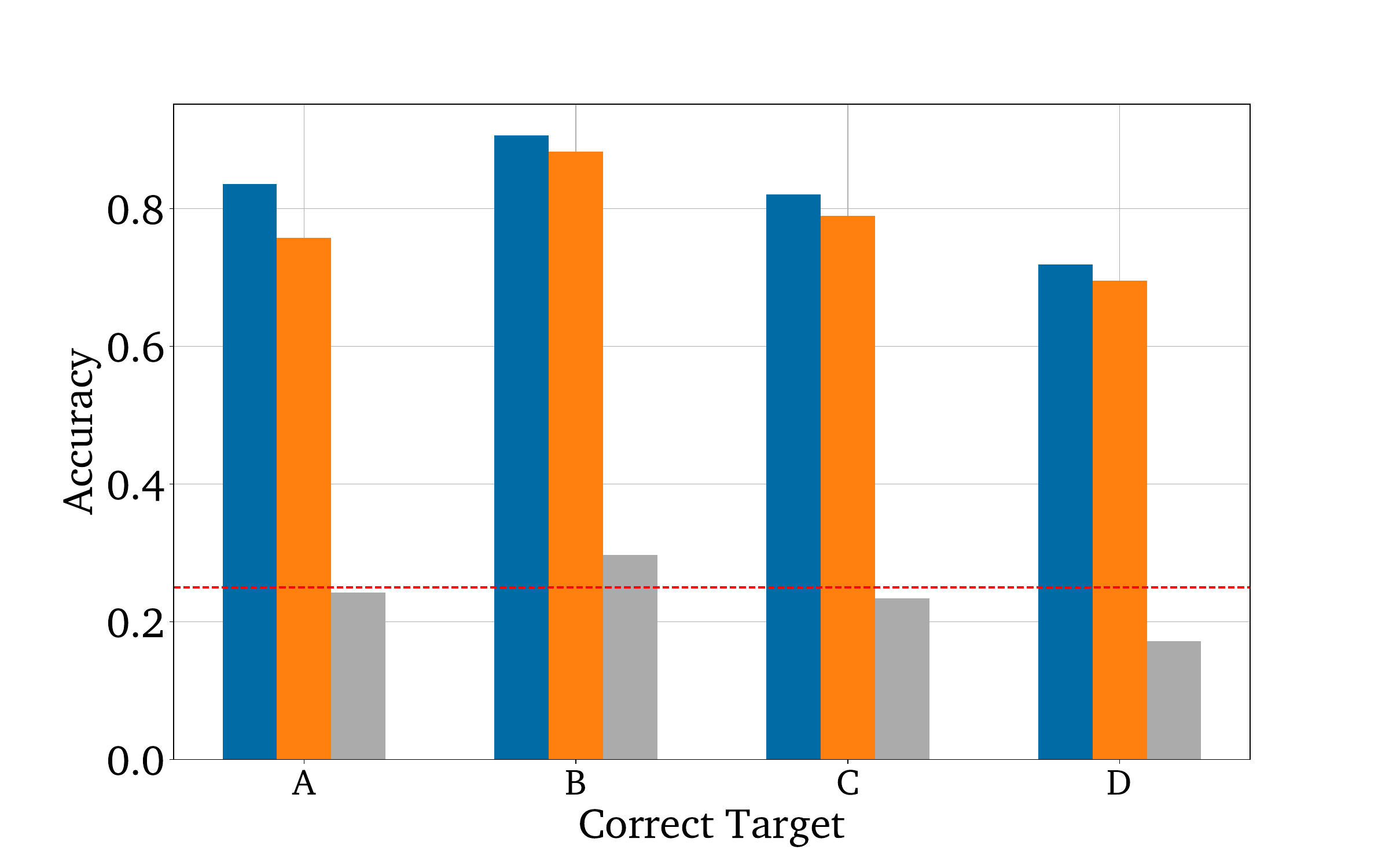}
         \caption{Accuracy (top-1 over the set ABCD)}
     \end{subfigure}
     \caption{Effect of patching all 45 MLPs and Heads that we identify as contributing directly to the output. For accuracy we report the mean and for loss the mean and standard error over 128 prompts per target.}
    \label{fig:total_effect_all_output_nodes}
\end{figure}

Finally, we can also validate the found set of nodes by patching in all of them together and record the resulting change in loss (average negative log probability of the correct answer letter). This is shown for different `targets' (correct answer letters) in~\cref{fig:total_effect_all_output_nodes}. We show the loss when the model is run on $p_{intervention}$ and evaluated according to $p_{intervention}$ (`Base'), when it is run on $p_{intervention}$ and evaluated according to a $p_{original}$ (`Random Targets') and when it is run on $p_{original}$, we patch in the 45 nodes from $p_{intervention}$ and evaluate according to $p_{intervention}$ (`Patched'). This is measuring the total effect of the set of these nodes. We observe that using these 45 nodes recovers most of the loss and accuracy on the chosen subset of MMLU. 

\subsection{MLP behaviour}

We show the net contribution to the logit of the correct label in all four cases for these MLPs in~\cref{fig:net_direct_effect_all_letters}. We observe that most MLPs are highly specialized, contributing very strongly to one or two letters if they are correct, while not contributing much or even harming net performance on other letters. This becomes especially apparent when considering the net logit relative to the other possible letters, instead of the full vocabulary.

We want to emphasize that taking a full MLP layer as the atomic node hides the number and identity of the individual neurons which are active and what their individual effects are. There is also some evidence that the correct unit of analysis is instead a group of neurons as they can encode multiple features in `superposition'~\citep{elhage2022superposition, gurnee2023finding}. A full analysis should consider these individual neurons or groups of neurons which contribute to this behavior, which we will leave for future work.

\subsection{Analysing attention patterns}
\label{sec:analysing_attention_patterns}

We analyse the attention patterns of the heads identified in~\cref{fig:net_direct_effect} to understand how they work. We find that they can roughly be clustered into 4 groups, based on their value-weighted attention patterns. Specifically, we measure the product of the attention probability and the L2 norm of the value vector at each position. We report the value-weighted attention on the prelude tokens, the label tokens and the final tokens (c.f.~\cref{fig:mmlu_tokens}). For the remaining positions we report the maximum in the column "OTHER" in each plot. 

The boundary between these groups of heads is not sharp and we encourage readers to take a look at the full array of plots in~\cref{sec:classification_output_heads} to get an overview of the different behaviors on display. The categories we found most sensible are

\begin{itemize}
\item `Correct Letter' heads, which attend from the final position to the correct label.
\item `Uniform' heads, which roughly attend uniformly to all letters.
\item `Single letter' heads, which mostly attend to a single fixed letter
\item `Amplification' heads, which we hypothesize to `amplify' information already in the residual and aggregate information from the last few tokens into the last token. We hypothesize this due to them being late in the network and due to their attention pattern.
\end{itemize}

\begin{figure}[t]
     \centering
     \begin{subfigure}[b]{0.49\textwidth}
         \centering
         \includegraphics[width=\textwidth]{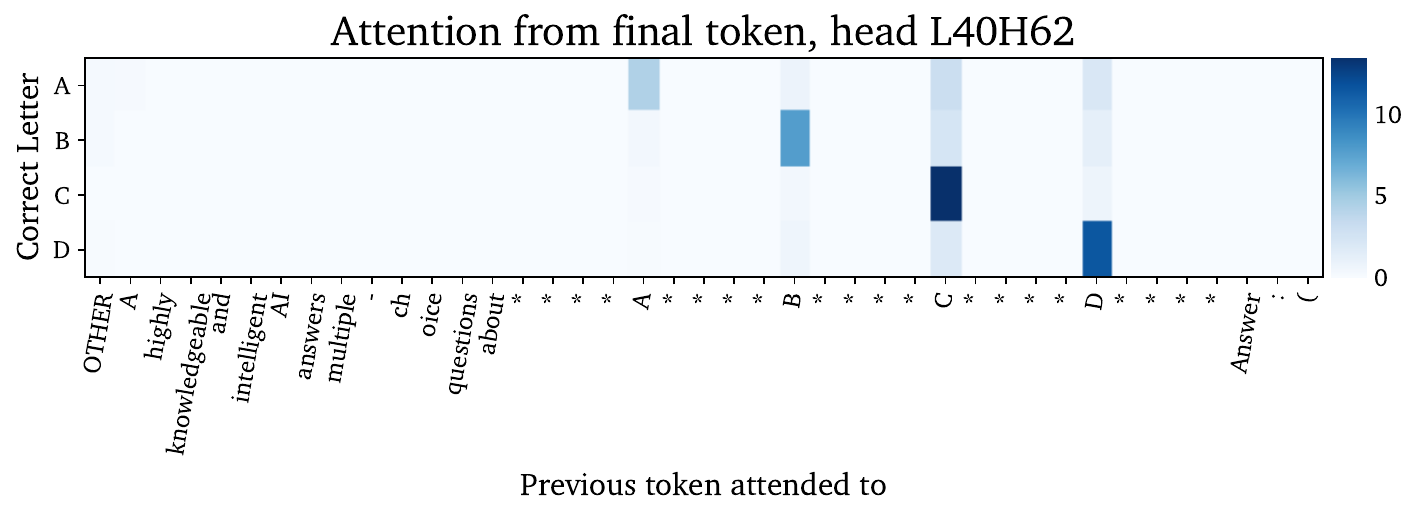}
         \caption{Correct Letter head}
     \end{subfigure}
     \hfill
     \begin{subfigure}[b]{0.49\textwidth}
         \centering
         \includegraphics[width=\textwidth]{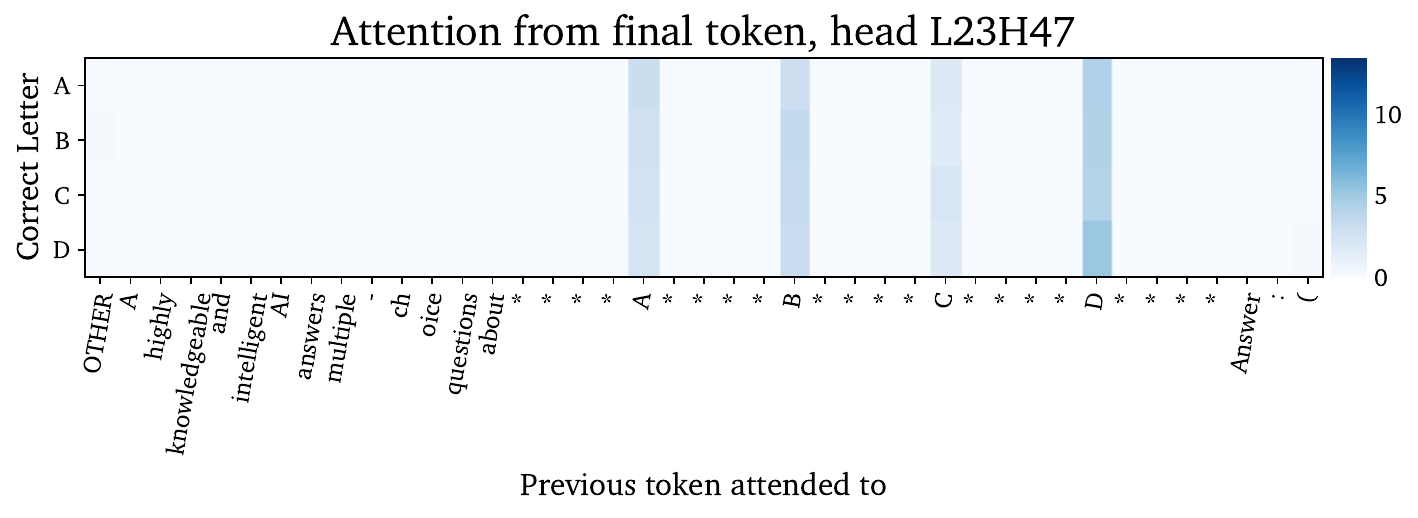}
         \caption{Constant head}
     \end{subfigure}
     \hfill
     \begin{subfigure}[b]{0.49\textwidth}
         \centering
         \includegraphics[width=\textwidth]{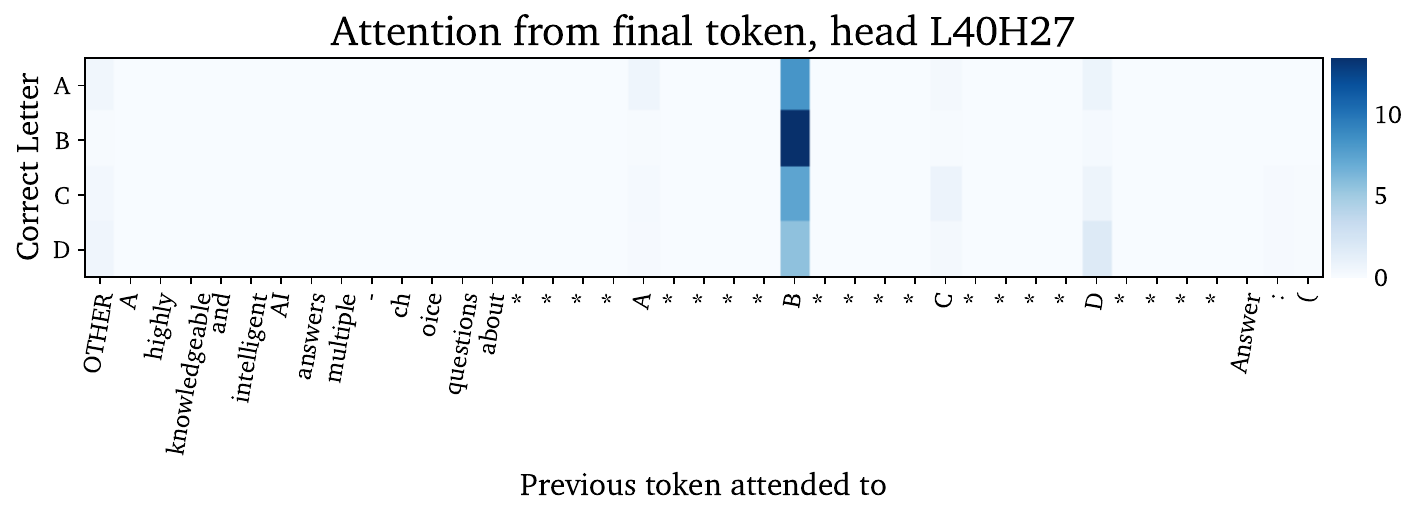}
         \caption{Single letter head}
     \end{subfigure}
     \hfill
     \begin{subfigure}[b]{0.49\textwidth}
         \centering
         \includegraphics[width=\textwidth]{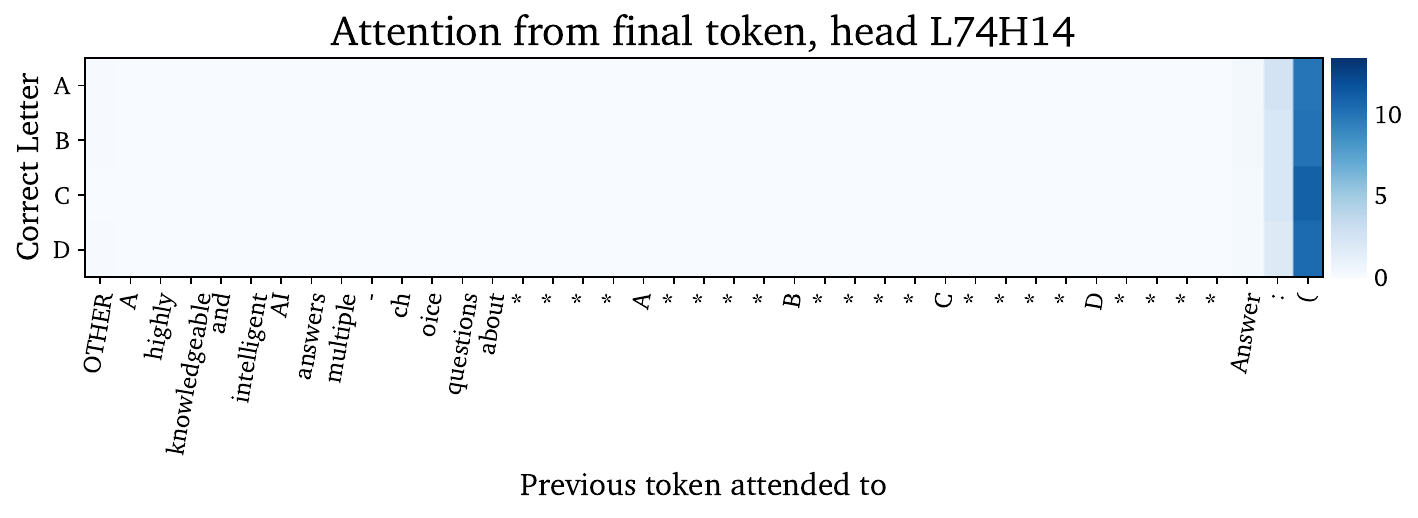}
         \caption{`Amplification' head}
     \end{subfigure}
    \caption{Value-weighted attention patterns of selected heads for each identified head class. For the exact methodology see~\cref{sec:classification_output_heads}.}
    \label{fig:head_classes_examples}
\end{figure}

We show a particularly crisp example for each type of head in~\cref{fig:head_classes_examples}. We note that most single letter heads cannot implement a generalizing algorithm by themselves since they do not seem to differ in their behavior depending on the correct answer and a single label can not contain the information about which option is correct in general (with the exception of \texttt{D})\footnote{However, they could form a generalizing algorithm in aggregate, e.g. via attention head superposition~\citep{jermyn2023attention}.}.

This analysis already reveals significant information about the circuit. A priori, we might have thought that models would move label information to the content tokens, and then attend to the content tokens to extract the correct label (a similar mechanism as in induction heads~\citep{elhage2021mathematical, olsson2022context}). However, this cannot be happening, since all heads attend to the \emph{labels} rather than the answer contents, even though the causal masking employed in transformer decoders prohibits information flow from the contents to the corresponding labels. It seems likely that the heads attend to the labels merely to identify the label corresponding to the already-determined correct answer. Note however that since labels can contain information about which of the previous contents was correct, it is also possible that the heads identify both the correct answer and its corresponding label from the keys\footnote{In particular, an attention head could implement the following algorithm. For each label A, B, C or D, the attention on it should be equal to 0 if the correct answer appeared before that label; otherwise, the attention logit should be proportional to the number of incorrect answers that have appeared before that label. The OV-circuit would then simply copy the letter identity. Indeed, it seems plausible that some variant of this mechanism is used by head \texttt{L30 H45}, cf.~\cref{fig:misc_heads_val_attn}}.

\begin{figure}[t]
    \centering
    \includegraphics{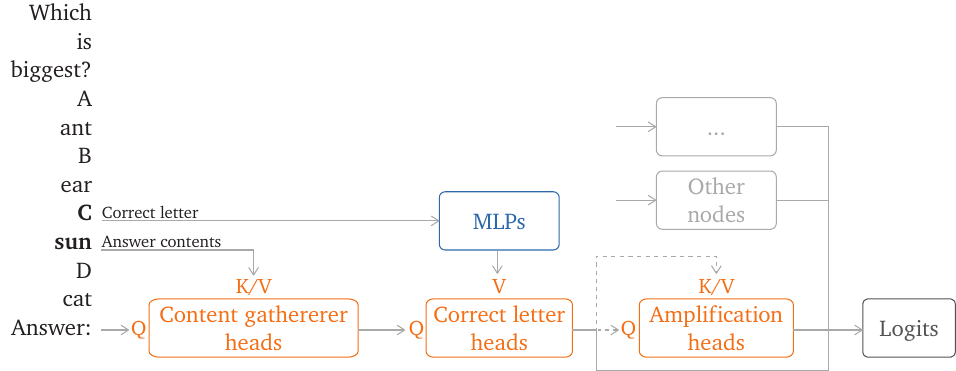}
    \caption{Information flow between circuit nodes investigated in this work. Q, K and V denote query, key and value inputs to each head, and dashed lines represent hypothesised but unproven connections. Content Gatherer heads move information from token positions corresponding to the \emph{contents} of the correct answer to the final token position. This information is used by Correct Letter heads to select the \emph{letter} of the correct answer, aided by MLPs. The Correct Letter heads then directly increase the logit of the correct letter -- we believe, further mediated by the Amplification heads.}
    \label{fig:endnodes_circuit}
\end{figure}

\subsection{Discovering more nodes}

Now that we have found the final nodes, we can recurse on our results and ask which nodes influence the final nodes. We are doing so mainly for the subset of output nodes which we term `correct letter heads' in~\cref{sec:understanding_nodes}, although we do suspect that some parts are shared by other output heads as well. As this is not the main focus of this paper, we include the results in~\cref{sec:rest_of_the_circuit}. In summary, the correct letter heads obtain the information about the correct label via a class of heads which we call `content gatherers' whose most prominent representative is head \texttt{L24 H18}. These heads attend from the final token to the content of the correct answer and thereby Q-compose~\citep{elhage2021mathematical} with the correct letter heads. The correct letter heads' OV circuit on the other hand acts upon information written at the label positions by a large set of MLPs in the early-mid layers of the network. We hypothesise an overall circuit diagram in~\cref{fig:endnodes_circuit}, though we emphasize that we have not validated all aspects of this diagram.

\section{Understanding the Semantics of the Correct Letter Heads}
\label{sec:understanding_nodes}

In this section, we focus on understanding the most interesting group of attention heads identified in~\cref{sec:analysing_attention_patterns}, the Correct Letter heads. To help guide the reader, we first present an outline of our results in this section before delving into details. 
\begin{enumerate}
    \item We show that Q and K spaces of the Correct Letter heads can be compressed into a 3D subspace without harming their performance.
    \item By varying the prompt structure and labels, we narrow down the semantics of the low-dimensional Q and K spaces used by the Correct Letter heads.
    \item Via the above we provide preliminary evidence that the Correct Letter heads seem to use both a somewhat general feature of `Nth item in a list' and a more adhoc feature based on label identity.
    \item Finally, we summarize our findings in pseudocode form, albeit with various caveats.
\end{enumerate}

The Correct Letter heads are the most interesting heads of the groups identified because they seem to be core to a generalizing algorithm of choosing the correct answer from the presented options. Further note that the head with the largest direct effect is \texttt{L40 H62} which we identify as a Correct Letter head.

To better understand these heads, we attempt to write pseudocode descriptions of how they operate, as suggested by the north star of mechanistic interpretability research to ``reverse engineer neural networks into understandable computer programs''~\citep{elhage2022solu}. Such pseudocode will still in many cases need to reference the underlying linear algebra -- though hopefully with simplifications that allow the pseudocode to be easier to reason about than the model itself. In particular, we attempt to narrow down exactly which subspaces the heads read from and write to in the residual stream (assuming these subspaces to be much smaller than the full dimensionality of the key, query and value subspaces), and determine what semantic features these subspaces correspond to.

In this section we show that the Correct Letter heads do indeed operate on subspaces of significantly lower rank than the original subspaces. In particular we show that we can losslessly reduce the Correct Letter heads' QK circuits to a 3-dimensional subspace after taking into account the query and key means. Furthermore, we provide evidence that these low-rank approximations capture features that generalize somewhat, but are also somewhat specialized to the particular case of ABCD.

\subsection{Distilling Heads}
To form low-rank approximations of the Correct Letter heads, we first sample a dataset of 1024 prompts from the previously described MMLU subset. For each head, we collect the keys and values at the label positions and the query from the final token. This gives us the sets $Q^h$, $K^h$ and $V^h$ for each Correct Letter head $h$. 

We are interested in the feature which allows the heads to distinguish between the different label positions. We can trivially decompose queries and keys into their respective means over the dataset and a prompt-specific delta term.

\begin{align}
    q(x) &= q_\mu + q_\delta(x)\\
    k(x) &= k_\mu + k_\delta(x)
\end{align}

The dot product in the argument of the self-attention mechanism can then be written as a sum of four terms

\begin{align}
    q(x) \cdot k(x) = q_\mu \cdot k_\mu \enspace + \enspace q_\delta \cdot k_\mu \enspace + \enspace q_\mu \cdot k_\delta \enspace + \enspace q_\delta \cdot k_\delta.
\end{align}

Of these terms, only the latter two can provide signal to distinguish between correct answers, since $k_\mu$ is the same across answers. In~\cref{sec:key_deltas} we show the size of the individual dot product components at the label positions under different correct labels. Empirically, we observe that only $q_\delta \cdot k_\delta$ contains significant information about the correct label, meaning that $q_\mu$ does not contain a generic ``Are you the correct label?'' query and/or that $k_\delta$ does not contain the corresponding feature. \footnote{Based on the attention being nonzero only at the label positions, we hypothesise that $q_\mu$ encodes something to the effect of ``Are you \emph{any} label (as opposed to a non-label token)?''} To identify the feature distinguishing between labels, we now apply singular value decomposition (SVD) on the union of the centered datasets $Q^h_\delta \cup K_\delta^h$. For the values we do not perform this decomposition and perform SVD on the uncentered data.

\begin{figure}[t]
    \centering
    \includegraphics{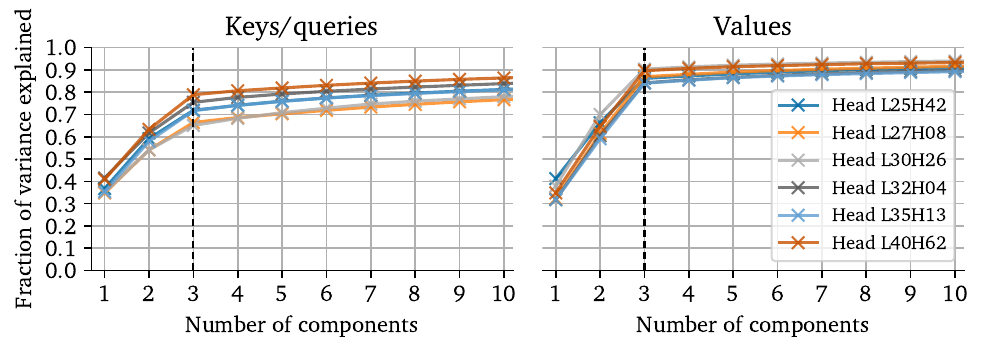}
    \caption{Cumulative explained variance for different numbers of principal components when performing PCA on the key/query and value subspaces of the Correct Letter heads. Note that for all heads, there is a knee at three components. See text for more details.}
    \label{fig:qkv_delta_scree_plots}
\end{figure}

The corresponding scree plot in~\cref{fig:qkv_delta_scree_plots} suggests that using 3 components captures roughly 65-80\% of the variance for all heads for keys and queries and 80-90\% for the values, so we choose this number of components going forward.

\begin{figure}[t]
    \centering
     \begin{subfigure}[b]{0.32\textwidth}
         \centering
         \includegraphics[width=\textwidth]{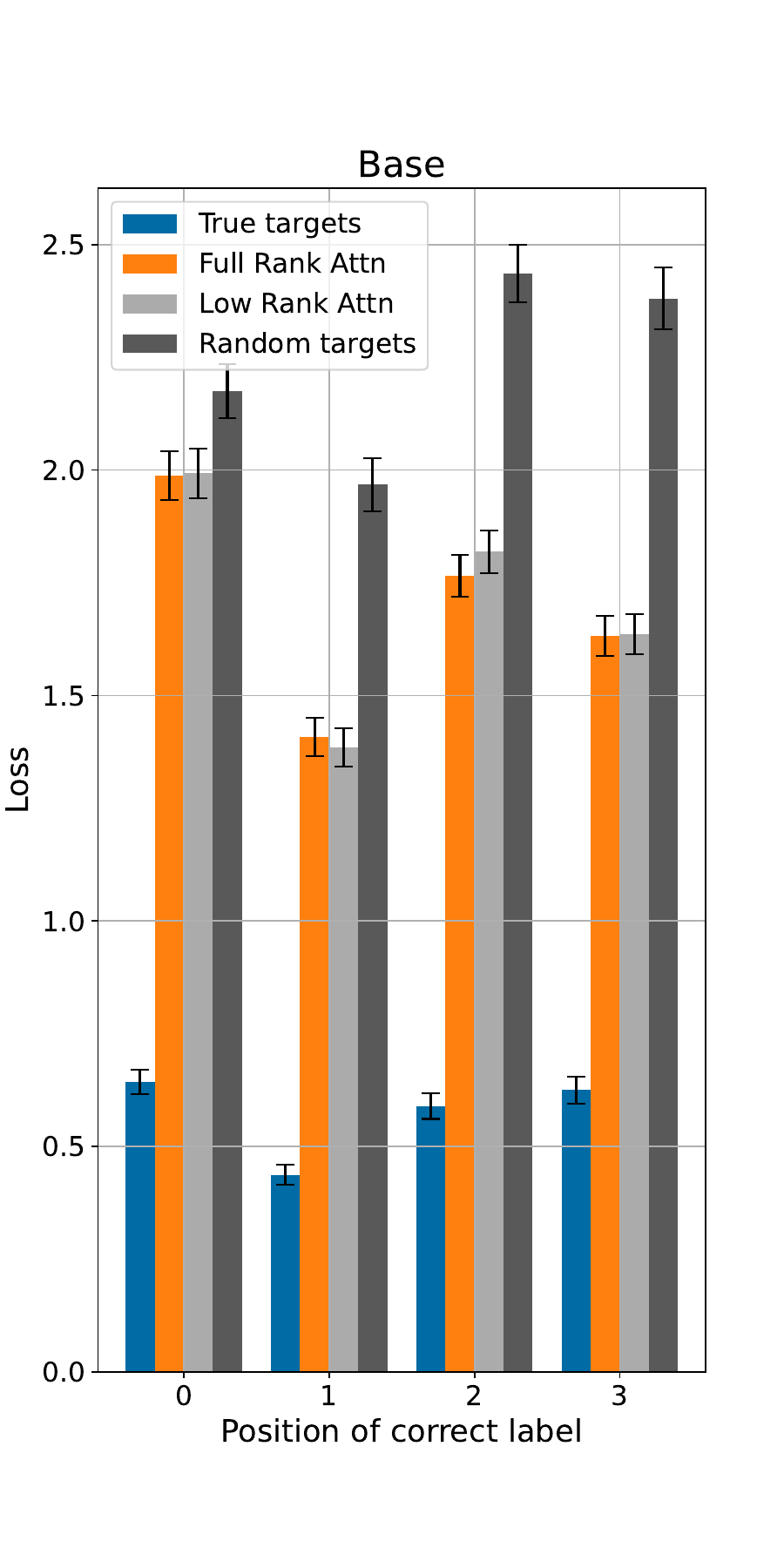}
         \caption{Labels: A, B, C, D}
         \label{fig:loss_lr_attn_base}
     \end{subfigure}
     \hfill
     \begin{subfigure}[b]{0.32\textwidth}
         \centering
         \includegraphics[width=\textwidth]{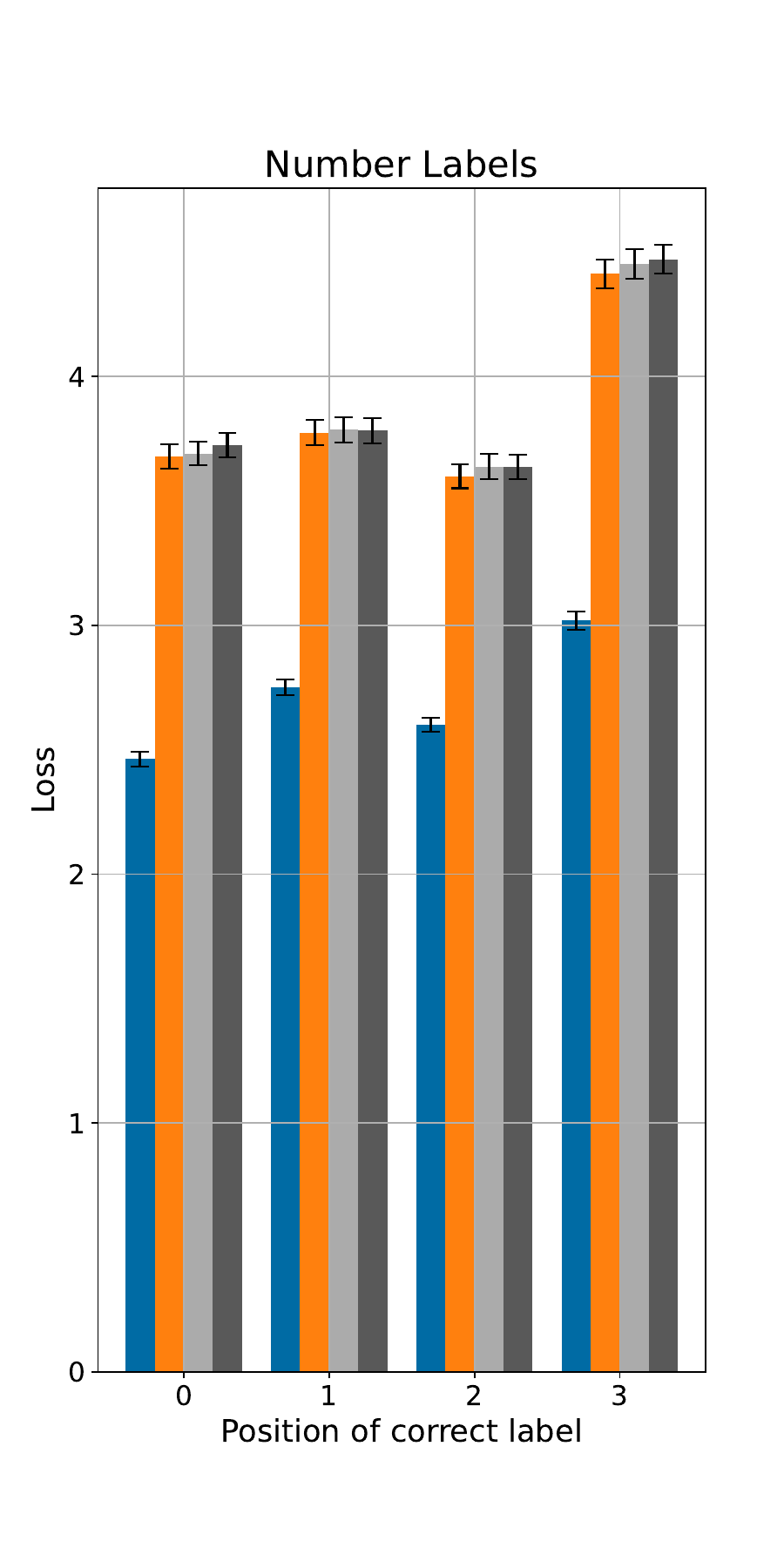}
         \caption{Labels: 1, 2, 3, 4}
         \label{fig:loss_lr_attn_replace_letter_with_numbers}
     \end{subfigure}
     \hfill
     \begin{subfigure}[b]{0.32\textwidth}
         \centering
         \includegraphics[width=\textwidth]{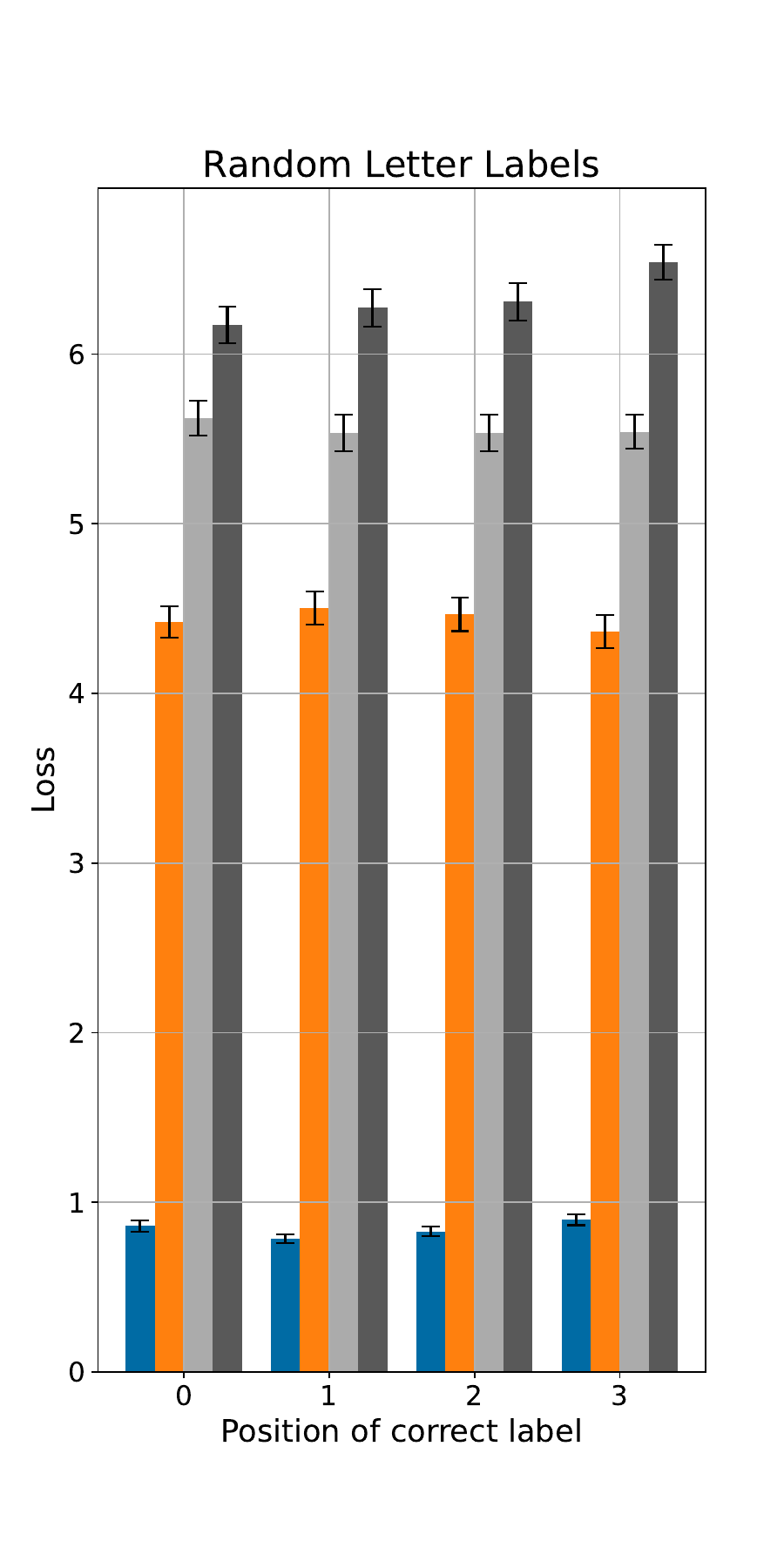}
         \caption{Labels: random, e.g. O, E, B, P}
         \label{fig:loss_lr_attn_randomise_answer_letters}
     \end{subfigure}
    \caption{Loss (negative log probability of correct answer token) when using full rank or low rank attention under various prompt mutations. Note the differing y-axes. As in~\cref{fig:total_effect_all_output_nodes}, `True targets' means running and evaluating the model on $p_{intervention}$, and `Random targets' means running on $p_{intervention}$ but evaluating on $p_{original}$. To judge generalizability we are foremost concerned with comparing the orange and light grey bars. For more results including accuracy see~\cref{fig:mutated_low_rank_loss_full,fig:mutated_low_rank_acc}.}
    \label{fig:mutated_low_rank_loss_maintext}
\end{figure}

We measure the quality of the low-rank approximation in two ways. First, we measure the \emph{direct} effect of the Correct Letter heads at the final token position using the low-rank keys, queries and values. We find in~\cref{fig:low_rank_direct_effects} that there is no substantial difference between the full-rank and low-rank setting. Second, we measure the \emph{total} effect of the low-rank approximation. This must be done with care: because the approximation was constructed based only on the keys at label positions and the queries at the final token position, we wish to leave the heads' operation at other positions undisturbed, since the approximation may not be valid elsewhere. To do this, we replace the keys at the label positions and the query at the final token position in $p_{original}$ with the low-rank keys and queries from corresponding positions in $p_{intervention}$; compute the resulting attention pattern; and patch in the resulting attention only at the final token position.\footnote{We leave values untouched to enable a comparison to results in~\cref{sec:qk_subspace} in which we mutate label tokens -- using low-rank values for mutated prompts results in poor performance, presumably because values appear to encode token identity, such that an approximation based on labels ABCD doesn't work at all for number labels such as 1234.} We compare this with the results when using the \emph{full}-rank keys and queries from $p_{intervention}$ in the same procedure. As shown in~\cref{fig:loss_lr_attn_base}, patching low-rank attention has the same effect as using full-rank attention. Note that we did not patch other nodes besides the Correct Letter heads, so that overall performance is worse than in~\cref{fig:total_effect_all_output_nodes}.

\begin{figure}
    \centering
    \includegraphics{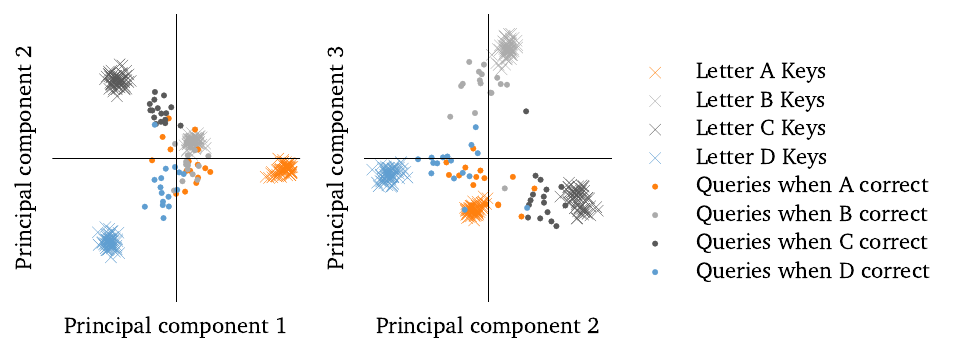}
    \caption{Projection of query and key deltas of \texttt{L40 H62} on the first three singular vectors. See \href{https://sites.google.com/view/does-mechinterp-scale}{\texttt{https://sites.google.com/view/does-mechinterp-scale}} for an interactive 3D plot.}
    \label{fig:3d_proj_qk_delta}
\end{figure}

In~\cref{fig:3d_proj_qk_delta} we show a typical projection of query and key deltas for head \texttt{L40 H62} on the first three singular vectors. We can clearly see that the queries for a given Correct Letter cluster in the same direction as the corresponding keys, and that the key clusters are arranged in a tetrahedron. Furthermore, the queries are much less cleanly separated than the keys. We speculate that the keys are always the same regardless of which answer is correct and the variance in the queries represents the model's uncertainty about which answer is correct. We show the cosine similarity between the mean vectors of the respective query and key clusters in~\cref{fig:cosine_sim_q_and_k_centroids}.

\subsection{QK Subspace Semantics}
\label{sec:qk_subspace}

To understand the semantic meaning of these subspaces, we form a tentative hypothesis and then use mutated prompts to test the hypothesis. We guessed that the subspace might encode ``Nth item in a list'', and therefore used the following prompt mutations:

\begin{itemize}
    \item Replacing ABCD with random capital letters, e.g. OEBP.
    \item Replacing ABCD with random capital letters in alphabetical order, e.g. MNOP.
    \item Replacing ABCD with 1234\footnote{In this case we need to change the prompt structure to use \texttt{X:} as labels, as otherwise \texttt{(X)} would be tokenized as a single token. Furthermore we found that we had to replace \texttt{Answer: (} with \texttt{The correct number is number}, and score against the total logprob on \texttt{ X}, \texttt{ X.}, \texttt{ X:}, \texttt{ X,}, and \texttt{word\_for\_X} (e.g. \texttt{one}), since the model was placing significant probability mass on all of these. }.
    \item Replacing newline separators with periods or semicolons.
    \item Removing the prelude ``A highly knowledgeable and intelligent AI answers multiple-choice questions about \emph{some topic}''.
\end{itemize}

For each of these mutations, we repeat the procedure used to generate~\cref{fig:loss_lr_attn_base}, and examine the change in loss. Using different separators or removing the prelude does not result in a significant difference (see results in the appendix in~\cref{fig:mutated_low_rank_loss_full}). This suggests whatever feature is encoded in the low-rank subspace is not sensitive to the precise formatting of the question. For random capital letters, we do observe a significant difference but still recover one third to half the loss, as depicted in~\cref{fig:loss_lr_attn_randomise_answer_letters}, meaning that part of the subspace does generalize to other letters and part is specific to ABCD. For numbers, we see that Chinchilla is unable to perform the task well (cf.~\cref{fig:loss_lr_attn_replace_letter_with_numbers}), even in the base setting, and that seemingly the Correct Letter heads do not contribute to the performance in this setting\footnote{Investigating that a bit further, we anecdotally found that while the values at the number positions seemed intact, the attention paid to them was near zero, suggesting a failure of the QK circuit, rather than the OV circuit}.

\begin{figure}
    \centering
    \begin{subfigure}[b]{0.38\textwidth}
         \centering
         \includegraphics[width=\textwidth]{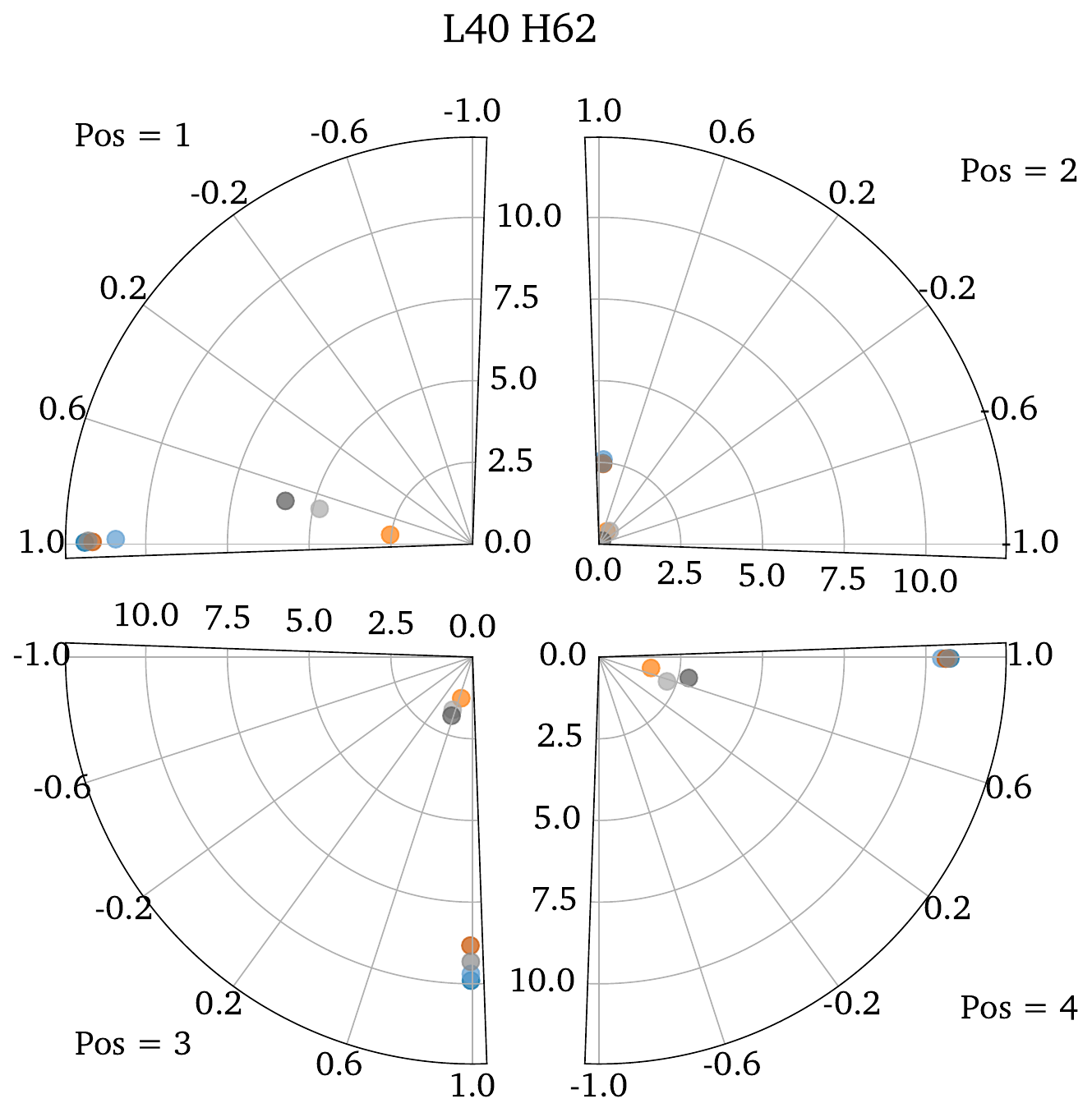}
         \caption{Key deltas $k_\delta$}
         \label{fig:project_k_deltas_on_k_centroids_l40h62}
    \end{subfigure}
    \hfill
    \begin{subfigure}[b]{0.6\textwidth}
         \centering
         \includegraphics[width=\textwidth]{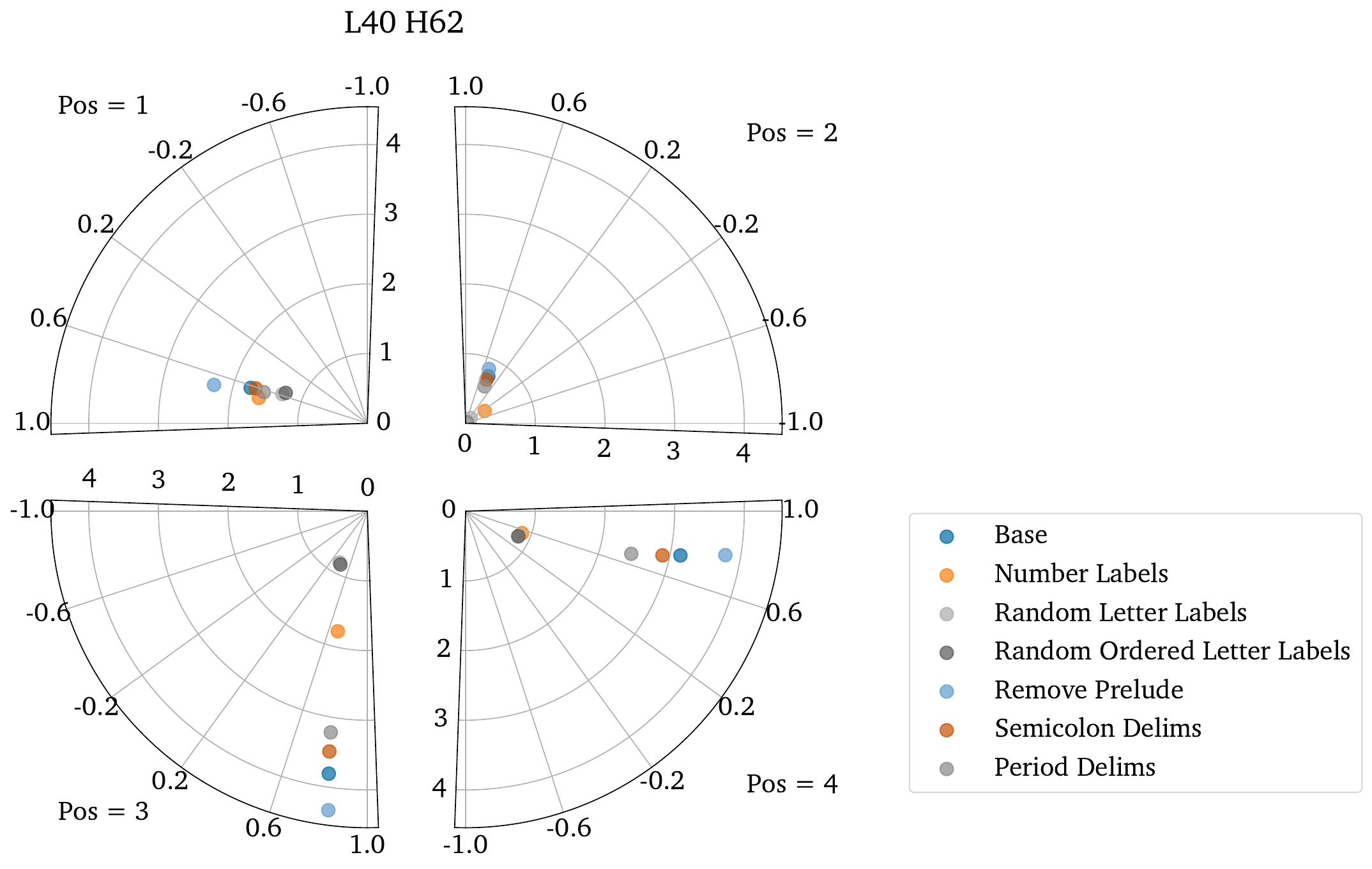}
         \caption{Query deltas $q_\delta$}
         \label{fig:project_q_deltas_on_k_centroids_l40h62}
    \end{subfigure}
    \caption{Cosine similarity and absolute value of the projection of the key and query deltas of head L40 H62 onto the clusters formed by its key deltas in the base case. Cosine similarity is given as angle and projection as radius.}
    \label{fig:project_qk_deltas_on_k_centroids_l40h62}
\end{figure}

In addition to the recovered loss, we can also investigate where the queries and keys from the mutated prompts lie geometrically relative to those from the base case. To do so, we project the $q_\delta$ and $k_\delta$ from different prompt variants onto the key cluster centroids corresponding to the query's or key's label of the base prompt. We chose the key clusters as they seem particularly crisp. Both the magnitude of the projection and the angle between the centroid and the deltas are relevant for the formed attention pattern. We report both quantities for \texttt{L40 H62} in~\cref{fig:project_qk_deltas_on_k_centroids_l40h62}, with results on all Correct Letter heads shown in~\cref{fig:project_k_deltas_on_k_centroids} and~\cref{fig:project_q_deltas_on_k_centroids} in~\cref{sec:more_low_rank_results}. We observe that in virtually all cases, the projections are largest for the base case, for altered separators, and for removed prelude, similar to the total loss graph. Furthermore, the cosine similarity for keys is usually very high, and even for queries is most often above 0.6. From these graphs it seems as if the most dividing factor between the cases with high recovered loss and those with low recovered loss is the magnitude of the projection. In other words, the keys and queries point in roughly the same direction as the original clusters, but are closer to the mean key or query respectively. The fact that the projections are distinct from 0 again suggests that the QK subspace contains some general feature, in addition to more specialized features.

A few possible explanations present themselves to account for the observed differences in losses. Initially one might think that the features stored in the key and query deltas might correspond to a feature like `n-th item in an enumeration'. However, since replacing ABCD with random capital letters works worse when using the low-rank approximation, this suggests that part of it could be related to the specific token identity as well. It is also possible that the model mostly saw enumerations of the form ABCD and much less frequently of the form, say, XPBG, and so the enumeration feature works much better for ABCD than for randomly selected capital letters.

\paragraph{Positional Encoding}
An alternative hypothesis is that part of the encoding is positional. That is, the query encodes ``attend to token at position 110'' and each key encodes ``I am position \texttt{t}''. We can however partially rule out this explanation via~\cref{fig:loss_lr_attn_base}, since the prompts $p_{original}$ and $p_{intervention}$ were independently sampled and so do not have the same length, i.e. the labels are at different token positions in both prompts. Thus, if positional information was crucial, we should see a worse performance for both low rank and full rank. Furhtermore, we would expect to see less tight clustering of keys and queries in~\cref{fig:3d_proj_qk_delta}.

\paragraph{Value Semantics}
So far we have only touched on the semantics of queries and keys. As mentioned above using the low-rank value space for different labels resulted in poor performance. This is to be expected since the Correct Letter heads directly increase the correct token's logit, and the low-rank subspace was constructed on the base case. Thus the low-rank subspace would only coincidentally, if at all, overlap with the subspace used for different letters. It might also be the case that the Correct Letter heads do not have the capability to affect different label tokens, since their expressivity is limited by their rank 128 OV matrices.

\clearpage

\subsection{Correct Letter Head Pseudocode}

Based on the what we've learned about the meaning of the subspaces on which the heads operate, we can now write pseudocode describing the operation of each of these heads. The result is shown in~\cref{fig:pseudocode}.

\begin{figure}[htp]
\begin{center}
\begin{lstlisting}
def head(residual_stream) -> residual:
  item_nums = get_item_nums(residual_stream)  # Keys
  correct_item_num = get_correct_item_num(residual_stream[-1])   # Query
  correct_token_position = argmax(dot(item_nums, correct_item_num))
  token_identities = get_tokens(residual_stream)  # Values
  correct_token = token_identities[correct_token_position]
  return increase_logits_for(correct_token)  # Output weight matrix
\end{lstlisting}
\end{center}
\caption{Pseudocode representation of the Correct Letter heads' operation at the final token position. See text for details.}
\label{fig:pseudocode}
\end{figure}

This pseudocode is essentially a recapitulation of the standard attention mechanism, but with the keys, queries and values given names based on our best attempt at labelling what the features represent: \texttt{item\_nums} for the keys, \texttt{correct\_item\_num} for the query, and \texttt{token\_identities} for the values.

However, these names are only correct to a first approximation. For example, the name \texttt{item\_nums} suggests invariance to whether the items are labelled ABCD, VXYZ, or 1234, but~\cref{fig:mutated_low_rank_loss_maintext,fig:project_qk_deltas_on_k_centroids_l40h62} shows this is not straightforwardly the case. Although the embeddings for, say, the second item label are in a similar direction in feature space regardless of the label, the magnitudes of the embeddings are smaller for random letters, and smaller still for numbers -- suggesting that the representations are only partially invariant to such changes, and more invariant to random letters than to numbers. Furthermore, differences in absolute direction and magnitude are not enough to determine whether the attention, which relies on the relative positions of the embeddings, will also be invariant. We believe these differences are likely to be highly relevant to understanding how the head will behave on off-distribution adversarial inputs such as items labelled 12CD -- but are nuances that are difficult to reflect in code.

\section{Discussion}

\paragraph{Limitations of Causal Interventions and Semantic Analysis}
The main tools for finding the relevant subgraph (`circuit') in this work are analysis of direct and total effect of individual nodes. For semantic analysis we further rely on dimensionality reduction and variation of the inputs. These results come with some caveats. First, as mentioned above, one of the effect of patching a node can be that a downstream node compensates for that change, resulting in a net-zero change. This seems especially concerning in the context of backup behavior~\citep{wang2022interpretability}, where this reaction would be an artifact of the patching process rather than reflective of the computation in the unpatched model. Second, we score results against targets taken from the same prompt from which we inject activation. In other terms, we patch in `clean' activations into a `corrupted' forward pass, i.e. we perform `de-noising'. This means that total effect analyses will find a cross-section of the circuit, i.e. a set of nodes which separate ancestors and descendants of this set, rather than the full set of relevant nodes.
Third, these methods require a base distribution to sample from. Ideally, we would be able to parameterize the input space to the model in terms of features, enabling us to more finely control this base distribution~\citep{chan2022causal}. In absence of that, we can only measure effects of features we can vary. A possible remedy for this would be to add additive noise instead of resampling a node~\citep{meng2022locating} which could however take the model further off-distribution and distort results. Future work could investigate whether these different approaches to identifying causally relevant nodes differ in the circuits they yield. Fourth, it is likely that neurons and attention heads have multiple distinct functions, depending on the context~\citep{elhage2022superposition, gurnee2023finding, jermyn2023attention}. Thus, our results about the meaning of the correct letter heads and the functioning of all nodes we discuss are only applicable to the exact distribution we tested. However, we did observe the same results on a synthetic multiple choice dataset, suggesting some generalization to a more general multiple choice setting is possible. Finally, we focused on the net effect on the correct label relative to other labels. This neglects the part of the circuit which is involved in identifying the subset of possible answer tokens without being concerned about which answer of this set is correct. We did find several such nodes during the exploratory phase of this project. 

\paragraph{Faithfulness of pseudocode} The key difficulty we encountered in trying to write pseudocode descriptions of heads in this work was the trade-off between faithfulness to the original model and having a description that's easy to reason about. Even with a rough hypothesis of what features the head uses, there can be details to those features which is hard to represent in code, such as which subset of possible input mutations a given feature is invariant to. We think the main takeaway from this exercise is that the most suitable description of model components depends on the level of analysis required. Discrete, code-like descriptions of components may still be a reasonable target for coarse analyses of how a circuit as a whole operates, but more detailed analysis likely requires sticking with the unabstracted linear algebra itself.

\paragraph{Open Questions / Future Work}

Our main focus in this work was on the final parts of the circuit which are concerned with the manipulation of the label symbols. While we provide some information on the nodes feeding into the correct letter heads, the rest of the circuit is still undiscovered. Furthermore, it is an open question whether different models will implement the same or similar algorithms. 

As mentioned above, different approaches to perform causal interventions can produce different results and come with different benefits and drawbacks. We welcome more work investigating the effects of different choices in this matter, allowing the interpretability community to agree on a standard set of tools and helping researchers to make more informed decisions.

Manual identification and classification of circuits and their constituting nodes is very labor intensive. As such we believe that future work should place a stronger focus on automating these analyses as much as possible, as long as faithfulness and completeness can be maintained.

There are several interesting node behaviours that we did not investigate as thoroughly as we would have liked. In particular, it would be interesting to investigate whether the Single Letter heads are an instance of attention head superposition~\citep{jermyn2023attention}, and likewise whether the uneven distribution of the direct effects of MLPs are an example of across-layer MLP superposition.

As discussed earlier, teasing out the semantics of features proved quite difficult in this work. While the interpretability community has made significant progress on identifying information flow within circuits, progress on deeply understanding what kind of information is being processed has been comparatively slow. We thus strongly encourage further exploration of this topic in future studies.

Finally, we only examined behavior of the discussed nodes on the narrow distribution of MMLU. We do not provide any evidence about their function on text prediction in general, which could provide insights both on the topic of superposition or conversely whether the nodes implement a more general behavior which naturally supersedes the narrow behavior we sought to explain.

\section{Related work}

\textbf{Understanding circuits.} Broadly, our work continues a recent trend of investigation into the circuits underlying various behaviour. Notable prior works include \citet{meng2022locating}, investigating where in the model key facts are stored; \citet{wang2022interpretability}, analysing the circuit used to identify the grammatical indirect object; \citet{nanda2023progress}, understanding a circuit implementing modular addition; \citet{geva2023dissecting}, tracing out pathways involved in factual recall; and a number of others~\citep{bloom2023decision,hex2023a}.

\label{sec:related_work_circuit_tracing}
\textbf{Identifying relevant circuit nodes.} The most common way of establishing the causal role of any given circuit node is using interventions. Prior work varies in whether to patch in from a corrupted prompt into a clean prompt (`noising') or vice versa (`denoising'), which circuit pathways the patch is allowed to affect, and what metric to use for measuring the effect of the patch. \citet{meng2022locating} uses a denoising approach, starting with a forward pass on noised input embeddings and patching in activations from a normal forward pass. In contrast, \citet{wang2022interpretability} noising, starting with a normal forward pass and patching in activations from a prompt with similar structure but with crucial tokens randomised, and with the intervention only allowed to affect non-attention pathways. \citet{conmy2023towards} also performs noising, examining the effect of knocking out using both zero activations and activations from corrupted prompts, using KL divergence on token predictions to determine which nodes have the least effect. Other techniques include prioritising which nodes to knock out using gradient information~\citep{michel2019sixteen} and learning a mask over circuit nodes using gradient descent~\citep{cao2021low}, 

\textbf{Interpreting intermediate activations.} A key assumption in our analysis is that we can determine any direct contributions each node makes to the logits by unembedding the node's contribution to the residual stream~\citep{nostalgebraist2020interpreting,geva2022transformer,dar2022analyzing,ram2022you}. Recent work~\citep{belrose2023eliciting,din2023jump} suggests there may be complications to this picture, but believing the assumption to still be largely correct, we do not account for these complications in our work.

\textbf{Validating proposed circuits.} The procedure we use to determine what fraction of performance a set of nodes are responsible for is essentially a simplified version of the Causal Scrubbing algorithm proposed in~\citet{chan2022causal}. Other validation techniques are also possible, such as checking whether all possible interventions in a proposed high-level circuit cause changes in the output matching equivalent interventions in the original model~\citep{geiger2021causal,geiger2023causal}. For a comparison of these methods see~\citet{jenner2023comparison}.

\textbf{Understanding circuit nodes.} The main technique used to understand the function of each circuit node in transformer-based language models has been analysis of attention patterns~\citep{elhage2021mathematical,wang2022interpretability}. The most similar method to ours is distributed alignment search~\citep{geiger2023finding,wu2023interpretability}, which directly optimises a direction to correspond to a proposed high-level feature. In contrast, we use SVD to identify a subspace that explains variation in a dataset of activations, and then validate that the resulting subspace matches a proposed high-level feature. This makes it less likely that we find pathological solutions that overfit to the high-level feature, at the cost of making it more likely that we fail to find crucial subspaces that do exist. Other techniques include eigenvalues analysis of the node's weight matrix~\citep{elhage2021mathematical} and search for dataset examples which cause the node to activate strongly~\citep{bills2023language}.

\textbf{Low-rank approximation.} Dimensionality reduction techniques are one of the key building blocks of interpretability research. For example, non-negative matrix factorization was used in~\citet{olah2018the} and~\citet{hilton2020understanding} to identify key directions in activations space. Outside of interpretability, low-rank approximation has also found uses in model compression~\citep{hsu2022language} and resource-efficient fine-tuning~\citep{hu2021lora}.

\section{Conclusion}

In this work we explore the mechanism by which Chinchilla 70B is able to answer multiple choice questions. We establish and categorize a set of attention heads and MLPs which are directly contributing to answering with the correct label. We find that these components form groups which operate seemingly independently from each other, suggesting that the observed general purpose behavior of good performance on a benchmark can be distributed across generalizing and non-generalizing groups of nodes. In particular, we identify `correct letter heads' which attend to the correct label. We are able to compress these heads to a low-rank representation without harming performance on MMLU. Studying the semantics of these low-rank representations, we provide evidence that the attention mechanism of these heads uses both specific (e.g. `token identity') and general (e.g. `position in a list') features of the input. More generally, while we show that existing interpretability techniques yield promising results when applied to very large language models, we also find the results relatively noisy and at times contradictory, highlighting the need for more research into improved tools and methods. Finally, as research of this kind is labour intensive, we are excited about efforts to automate and accelerate future interpretability research.

\subsection*{Acknowledgements}

First, we wish to thank Tom McGrath and Avraham Ruderman for their very valuable input early on in the project.

Second, a huge thank you to Sebastian Borgeaud and Diego de Las Casas for help interfacing with DeepMind's language models, and for being willing to accommodate the modifications needed to enable interpretability work on these models.

Third, thank you to Orlagh Burns for serving as program manager for the team during this project, helping everyone to work together smoothly and generally creating a great working atmosphere.

Finally, thanks to Nicholas Goldowsky-Dill, Stefan Heimersheim, Marius Hobbhahn, Adam Jermyn, Tom McGrath, and Alexandre Variengien for their valuable feedback on an earlier version of this paper.

\subsection*{Author contributions}

Tom Lieberum led the project based on a proposal by Vladimir Mikulik. Tom Lieberum did the majority of the experimental work, with Matthew Rahtz also making contributions to analysis and project management in later stages of the project. For infrastructure, Matthew Rahtz and Vladimir Mikulik designed the first version of the codebase for use with small models, which was then extensively redesigned by J\'{a}nos Kram\'{a}r to be usable for large models, with additional contributions from Tom Lieberum. For the write-up, Tom Lieberum and Matthew Rahtz drafted the paper together, with extensive feedback from Vladimir Mikulik, Rohin Shah and Neel Nanda. Finally, Neel Nanda, Rohin Shah and Geoffrey Irving provided high-level advice and feedback throughout.

\bibliography{main}

\begin{thebibliography}{52}
\providecommand{\natexlab}[1]{#1}
\providecommand{\url}[1]{\texttt{#1}}
\expandafter\ifx\csname urlstyle\endcsname\relax
  \providecommand{\doi}[1]{doi: #1}\else
  \providecommand{\doi}{doi: \begingroup \urlstyle{rm}\Url}\fi

\bibitem[Bai et~al.(2022)Bai, Kadavath, Kundu, Askell, Kernion, Jones, Chen,
  Goldie, Mirhoseini, McKinnon, et~al.]{bai2022constitutional}
Y.~Bai, S.~Kadavath, S.~Kundu, A.~Askell, J.~Kernion, A.~Jones, A.~Chen,
  A.~Goldie, A.~Mirhoseini, C.~McKinnon, et~al.
\newblock Constitutional {AI}: Harmlessness from {AI} feedback.
\newblock \emph{arXiv preprint arXiv:2212.08073}, 2022.

\bibitem[Belrose et~al.(2023)Belrose, Furman, Smith, Halawi, McKinney,
  Ostrovsky, Biderman, and Steinhardt]{belrose2023eliciting}
N.~Belrose, Z.~Furman, L.~Smith, D.~Halawi, L.~McKinney, I.~Ostrovsky,
  S.~Biderman, and J.~Steinhardt.
\newblock Eliciting latent predictions from transformers with the tuned lens.
\newblock \emph{arXiv preprint arXiv:2303.08112}, 2023.

\bibitem[Bills et~al.(2023)Bills, Cammarata, Mossing, Tillman, Gao, Goh,
  Sutskever, Leike, Wu, and Saunders]{bills2023language}
S.~Bills, N.~Cammarata, D.~Mossing, H.~Tillman, L.~Gao, G.~Goh, I.~Sutskever,
  J.~Leike, J.~Wu, and W.~Saunders.
\newblock Language models can explain neurons in language models.
\newblock
  \url{https://openaipublic.blob.core.windows.net/neuron-explainer/paper/index.html},
  2023.

\bibitem[Bloom and Colognese(2023)]{bloom2023decision}
J.~I. Bloom and P.~Colognese.
\newblock Decision transformer interpretability.
\newblock
  \url{https://www.alignmentforum.org/posts/bBuBDJBYHt39Q5zZy/decision-transformer-interpretability},
  2023.

\bibitem[Cammarata et~al.(2020)Cammarata, Carter, Goh, Olah, Petrov, Schubert,
  Voss, Egan, and Lim]{cammarata2020thread}
N.~Cammarata, S.~Carter, G.~Goh, C.~Olah, M.~Petrov, L.~Schubert, C.~Voss,
  B.~Egan, and S.~K. Lim.
\newblock Thread: Circuits.
\newblock \emph{Distill}, 2020.
\newblock \doi{10.23915/distill.00024}.
\newblock https://distill.pub/2020/circuits.

\bibitem[Cao et~al.(2021)Cao, Sanh, and Rush]{cao2021low}
S.~Cao, V.~Sanh, and A.~M. Rush.
\newblock Low-complexity probing via finding subnetworks.
\newblock \emph{arXiv preprint arXiv:2104.03514}, 2021.

\bibitem[Chan et~al.(2022)Chan, Garriga-Alonso, Goldowsky-Dill, Greenblatt,
  Nitishinskaya, Radhakrishnan, Shlegeris, and Thomas]{chan2022causal}
L.~Chan, A.~Garriga-Alonso, N.~Goldowsky-Dill, R.~Greenblatt, J.~Nitishinskaya,
  A.~Radhakrishnan, B.~Shlegeris, and N.~Thomas.
\newblock Causal scrubbing: a method for rigorously testing interpretability
  hypotheses.
\newblock
  \url{https://www.alignmentforum.org/posts/JvZhhzycHu2Yd57RN/causal-scrubbing},
  2022.

\bibitem[Conmy et~al.(2023)Conmy, Mavor-Parker, Lynch, Heimersheim, and
  Garriga-Alonso]{conmy2023towards}
A.~Conmy, A.~N. Mavor-Parker, A.~Lynch, S.~Heimersheim, and A.~Garriga-Alonso.
\newblock Towards automated circuit discovery for mechanistic interpretability.
\newblock \emph{arXiv preprint arXiv:2304.14997}, 2023.

\bibitem[Dai et~al.(2019)Dai, Yang, Yang, Carbonell, Le, and
  Salakhutdinov]{dai2019transformer}
Z.~Dai, Z.~Yang, Y.~Yang, J.~Carbonell, Q.~V. Le, and R.~Salakhutdinov.
\newblock Transformer-{XL}: Attentive language models beyond a fixed-length
  context.
\newblock \emph{arXiv preprint arXiv:1901.02860}, 2019.

\bibitem[Dar et~al.(2022)Dar, Geva, Gupta, and Berant]{dar2022analyzing}
G.~Dar, M.~Geva, A.~Gupta, and J.~Berant.
\newblock Analyzing transformers in embedding space.
\newblock \emph{arXiv preprint arXiv:2209.02535}, 2022.

\bibitem[Din et~al.(2023)Din, Karidi, Choshen, and Geva]{din2023jump}
A.~Y. Din, T.~Karidi, L.~Choshen, and M.~Geva.
\newblock Jump to conclusions: Short-cutting transformers with linear
  transformations.
\newblock \emph{arXiv preprint arXiv:2303.09435}, 2023.

\bibitem[Elhage et~al.(2021)Elhage, Nanda, Olsson, Henighan, Joseph, Mann,
  Askell, Bai, Chen, Conerly, DasSarma, Drain, Ganguli, Hatfield-Dodds,
  Hernandez, Jones, Kernion, Lovitt, Ndousse, Amodei, Brown, Clark, Kaplan,
  McCandlish, and Olah]{elhage2021mathematical}
N.~Elhage, N.~Nanda, C.~Olsson, T.~Henighan, N.~Joseph, B.~Mann, A.~Askell,
  Y.~Bai, A.~Chen, T.~Conerly, N.~DasSarma, D.~Drain, D.~Ganguli,
  Z.~Hatfield-Dodds, D.~Hernandez, A.~Jones, J.~Kernion, L.~Lovitt, K.~Ndousse,
  D.~Amodei, T.~Brown, J.~Clark, J.~Kaplan, S.~McCandlish, and C.~Olah.
\newblock A mathematical framework for transformer circuits.
\newblock \emph{Transformer Circuits Thread}, 2021.
\newblock https://transformer-circuits.pub/2021/framework/index.html.

\bibitem[Elhage et~al.(2022{\natexlab{a}})Elhage, Hume, Olsson, Nanda,
  Henighan, Johnston, ElShowk, Joseph, DasSarma, Mann, Hernandez, Askell,
  Ndousse, Jones, Drain, Chen, Bai, Ganguli, Lovitt, Hatfield-Dodds, Kernion,
  Conerly, Kravec, Fort, Kadavath, Jacobson, Tran-Johnson, Kaplan, Clark,
  Brown, McCandlish, Amodei, and Olah]{elhage2022solu}
N.~Elhage, T.~Hume, C.~Olsson, N.~Nanda, T.~Henighan, S.~Johnston, S.~ElShowk,
  N.~Joseph, N.~DasSarma, B.~Mann, D.~Hernandez, A.~Askell, K.~Ndousse,
  A.~Jones, D.~Drain, A.~Chen, Y.~Bai, D.~Ganguli, L.~Lovitt,
  Z.~Hatfield-Dodds, J.~Kernion, T.~Conerly, S.~Kravec, S.~Fort, S.~Kadavath,
  J.~Jacobson, E.~Tran-Johnson, J.~Kaplan, J.~Clark, T.~Brown, S.~McCandlish,
  D.~Amodei, and C.~Olah.
\newblock Softmax linear units.
\newblock \emph{Transformer Circuits Thread}, 2022{\natexlab{a}}.
\newblock https://transformer-circuits.pub/2022/solu/index.html.

\bibitem[Elhage et~al.(2022{\natexlab{b}})Elhage, Hume, Olsson, Schiefer,
  Henighan, Kravec, Hatfield-Dodds, Lasenby, Drain, Chen, Grosse, McCandlish,
  Kaplan, Amodei, Wattenberg, and Olah]{elhage2022superposition}
N.~Elhage, T.~Hume, C.~Olsson, N.~Schiefer, T.~Henighan, S.~Kravec,
  Z.~Hatfield-Dodds, R.~Lasenby, D.~Drain, C.~Chen, R.~Grosse, S.~McCandlish,
  J.~Kaplan, D.~Amodei, M.~Wattenberg, and C.~Olah.
\newblock Toy models of superposition.
\newblock \emph{Transformer Circuits Thread}, 2022{\natexlab{b}}.
\newblock https://transformer-circuits.pub/2022/toy\_model/index.html.

\bibitem[Geiger et~al.(2021)Geiger, Lu, Icard, and Potts]{geiger2021causal}
A.~Geiger, H.~Lu, T.~Icard, and C.~Potts.
\newblock Causal abstractions of neural networks.
\newblock \emph{Advances in Neural Information Processing Systems},
  34:\penalty0 9574--9586, 2021.

\bibitem[Geiger et~al.(2023{\natexlab{a}})Geiger, Potts, and
  Icard]{geiger2023causal}
A.~Geiger, C.~Potts, and T.~Icard.
\newblock Causal abstraction for faithful model interpretation.
\newblock \emph{arXiv preprint arXiv:2301.04709}, 2023{\natexlab{a}}.

\bibitem[Geiger et~al.(2023{\natexlab{b}})Geiger, Wu, Potts, Icard, and
  Goodman]{geiger2023finding}
A.~Geiger, Z.~Wu, C.~Potts, T.~Icard, and N.~D. Goodman.
\newblock Finding alignments between interpretable causal variables and
  distributed neural representations.
\newblock \emph{arXiv preprint arXiv:2303.02536}, 2023{\natexlab{b}}.

\bibitem[Geva et~al.(2022)Geva, Caciularu, Wang, and
  Goldberg]{geva2022transformer}
M.~Geva, A.~Caciularu, K.~R. Wang, and Y.~Goldberg.
\newblock Transformer feed-forward layers build predictions by promoting
  concepts in the vocabulary space.
\newblock \emph{arXiv preprint arXiv:2203.14680}, 2022.

\bibitem[Geva et~al.(2023)Geva, Bastings, Filippova, and
  Globerson]{geva2023dissecting}
M.~Geva, J.~Bastings, K.~Filippova, and A.~Globerson.
\newblock Dissecting recall of factual associations in auto-regressive language
  models.
\newblock \emph{arXiv preprint arXiv:2304.14767}, 2023.

\bibitem[Glaese et~al.(2022)Glaese, McAleese, Tr{\k{e}}bacz, Aslanides, Firoiu,
  Ewalds, Rauh, Weidinger, Chadwick, Thacker, et~al.]{glaese2022improving}
A.~Glaese, N.~McAleese, M.~Tr{\k{e}}bacz, J.~Aslanides, V.~Firoiu, T.~Ewalds,
  M.~Rauh, L.~Weidinger, M.~Chadwick, P.~Thacker, et~al.
\newblock Improving alignment of dialogue agents via targeted human judgements.
\newblock \emph{arXiv preprint arXiv:2209.14375}, 2022.

\bibitem[Gurnee et~al.(2023)Gurnee, Nanda, Pauly, Harvey, Troitskii, and
  Bertsimas]{gurnee2023finding}
W.~Gurnee, N.~Nanda, M.~Pauly, K.~Harvey, D.~Troitskii, and D.~Bertsimas.
\newblock Finding neurons in a haystack: Case studies with sparse probing.
\newblock \emph{arXiv preprint arXiv:2305.01610}, 2023.

\bibitem[Heimersheim and Janiak(2023)]{hex2023a}
S.~Heimersheim and J.~Janiak.
\newblock A circuit for {Python} docstrings in a 4-layer attention-only
  transformer.
\newblock
  \url{https://www.lesswrong.com/posts/u6KXXmKFbXfWzoAXn/a-circuit-for-python-docstrings-in-a-4-layer-attention-only},
  2023.

\bibitem[Hendrycks et~al.(2020)Hendrycks, Burns, Basart, Zou, Mazeika, Song,
  and Steinhardt]{hendrycks2020measuring}
D.~Hendrycks, C.~Burns, S.~Basart, A.~Zou, M.~Mazeika, D.~Song, and
  J.~Steinhardt.
\newblock Measuring massive multitask language understanding.
\newblock \emph{arXiv preprint arXiv:2009.03300}, 2020.

\bibitem[Hilton et~al.(2020)Hilton, Cammarata, Carter, Goh, and
  Olah]{hilton2020understanding}
J.~Hilton, N.~Cammarata, S.~Carter, G.~Goh, and C.~Olah.
\newblock Understanding {RL} vision.
\newblock \emph{Distill}, 2020.
\newblock \doi{10.23915/distill.00029}.
\newblock https://distill.pub/2020/understanding-rl-vision.

\bibitem[Hoffmann et~al.(2022)Hoffmann, Borgeaud, Mensch, Buchatskaya, Cai,
  Rutherford, Casas, Hendricks, Welbl, Clark, et~al.]{hoffmann2022training}
J.~Hoffmann, S.~Borgeaud, A.~Mensch, E.~Buchatskaya, T.~Cai, E.~Rutherford,
  D.~d.~L. Casas, L.~A. Hendricks, J.~Welbl, A.~Clark, et~al.
\newblock Training compute-optimal large language models.
\newblock \emph{arXiv preprint arXiv:2203.15556}, 2022.

\bibitem[Hsu et~al.(2022)Hsu, Hua, Chang, Lou, Shen, and Jin]{hsu2022language}
Y.-C. Hsu, T.~Hua, S.~Chang, Q.~Lou, Y.~Shen, and H.~Jin.
\newblock Language model compression with weighted low-rank factorization.
\newblock \emph{arXiv preprint arXiv:2207.00112}, 2022.

\bibitem[Hu et~al.(2021)Hu, Shen, Wallis, Allen-Zhu, Li, Wang, Wang, and
  Chen]{hu2021lora}
E.~J. Hu, Y.~Shen, P.~Wallis, Z.~Allen-Zhu, Y.~Li, S.~Wang, L.~Wang, and
  W.~Chen.
\newblock {LoRa}: Low-rank adaptation of large language models.
\newblock \emph{arXiv preprint arXiv:2106.09685}, 2021.

\bibitem[Hubinger et~al.(2019)Hubinger, van Merwijk, Mikulik, Skalse, and
  Garrabrant]{hubinger2019risks}
E.~Hubinger, C.~van Merwijk, V.~Mikulik, J.~Skalse, and S.~Garrabrant.
\newblock Risks from learned optimization in advanced machine learning systems.
\newblock \emph{arXiv preprint arXiv:1906.01820}, 2019.

\bibitem[Irving et~al.(2018)Irving, Christiano, and Amodei]{irving2018ai}
G.~Irving, P.~Christiano, and D.~Amodei.
\newblock {AI} safety via debate.
\newblock \emph{arXiv preprint arXiv:1805.00899}, 2018.

\bibitem[Jenner et~al.(2023)Jenner, Garriga-Alonso, and
  Zverev]{jenner2023comparison}
E.~Jenner, A.~Garriga-Alonso, and E.~Zverev.
\newblock A comparison of causal scrubbing, causal abstractions, and related
  methods.
\newblock
  \url{https://www.alignmentforum.org/posts/uLMWMeBG3ruoBRhMW/a-comparison},
  2023.

\bibitem[Jermyn et~al.(2023)Jermyn, Olah, and Henighan]{jermyn2023attention}
A.~Jermyn, C.~Olah, and T.~Henighan.
\newblock Circuits updates — {M}ay 2023: Attention head superposition.
\newblock \emph{Transformer Circuits Thread}, 2023.
\newblock https://transformer-circuits.pub/2023/may-update/index.html.

\bibitem[Kenton et~al.(2021)Kenton, Everitt, Weidinger, Gabriel, Mikulik, and
  Irving]{kenton2021alignment}
Z.~Kenton, T.~Everitt, L.~Weidinger, I.~Gabriel, V.~Mikulik, and G.~Irving.
\newblock Alignment of language agents.
\newblock \emph{arXiv preprint arXiv:2103.14659}, 2021.

\bibitem[Lightman et~al.(2023)Lightman, Kosaraju, Burda, Edwards, Baker, Lee,
  Leike, Schulman, Sutskever, and Cobbe]{lightman2023lets}
H.~Lightman, V.~Kosaraju, Y.~Burda, H.~Edwards, B.~Baker, T.~Lee, J.~Leike,
  J.~Schulman, I.~Sutskever, and K.~Cobbe.
\newblock Let's verify step by step.
\newblock \emph{arXiv preprint arXiv:2305.20050}, 2023.

\bibitem[Meng et~al.(2022)Meng, Bau, Andonian, and Belinkov]{meng2022locating}
K.~Meng, D.~Bau, A.~Andonian, and Y.~Belinkov.
\newblock Locating and editing factual associations in {GPT}.
\newblock \emph{Advances in Neural Information Processing Systems},
  35:\penalty0 17359--17372, 2022.

\bibitem[Michel et~al.(2019)Michel, Levy, and Neubig]{michel2019sixteen}
P.~Michel, O.~Levy, and G.~Neubig.
\newblock Are sixteen heads really better than one?
\newblock \emph{Advances in neural information processing systems}, 32, 2019.

\bibitem[Nanda et~al.(2023)Nanda, Chan, Liberum, Smith, and
  Steinhardt]{nanda2023progress}
N.~Nanda, L.~Chan, T.~Liberum, J.~Smith, and J.~Steinhardt.
\newblock Progress measures for grokking via mechanistic interpretability.
\newblock \emph{arXiv preprint arXiv:2301.05217}, 2023.

\bibitem[nostalgebraist(2020)]{nostalgebraist2020interpreting}
nostalgebraist.
\newblock interpreting {GPT}: the logit lens.
\newblock
  \url{https://www.lesswrong.com/posts/AcKRB8wDpdaN6v6ru/interpreting-gpt-the-logit-lens},
  2020.

\bibitem[Olah et~al.(2017)Olah, Mordvintsev, and Schubert]{olah2017feature}
C.~Olah, A.~Mordvintsev, and L.~Schubert.
\newblock Feature visualization.
\newblock \emph{Distill}, 2017.
\newblock \doi{10.23915/distill.00007}.
\newblock https://distill.pub/2017/feature-visualization.

\bibitem[Olah et~al.(2018)Olah, Satyanarayan, Johnson, Carter, Schubert, Ye,
  and Mordvintsev]{olah2018the}
C.~Olah, A.~Satyanarayan, I.~Johnson, S.~Carter, L.~Schubert, K.~Ye, and
  A.~Mordvintsev.
\newblock The building blocks of interpretability.
\newblock \emph{Distill}, 2018.
\newblock \doi{10.23915/distill.00010}.
\newblock https://distill.pub/2018/building-blocks.

\bibitem[Olsson et~al.(2022)Olsson, Elhage, Nanda, Joseph, DasSarma, Henighan,
  Mann, Askell, Bai, Chen, Conerly, Drain, Ganguli, Hatfield-Dodds, Hernandez,
  Johnston, Jones, Kernion, Lovitt, Ndousse, Amodei, Brown, Clark, Kaplan,
  McCandlish, and Olah]{olsson2022context}
C.~Olsson, N.~Elhage, N.~Nanda, N.~Joseph, N.~DasSarma, T.~Henighan, B.~Mann,
  A.~Askell, Y.~Bai, A.~Chen, T.~Conerly, D.~Drain, D.~Ganguli,
  Z.~Hatfield-Dodds, D.~Hernandez, S.~Johnston, A.~Jones, J.~Kernion,
  L.~Lovitt, K.~Ndousse, D.~Amodei, T.~Brown, J.~Clark, J.~Kaplan,
  S.~McCandlish, and C.~Olah.
\newblock In-context learning and induction heads.
\newblock \emph{Transformer Circuits Thread}, 2022.
\newblock
  https://transformer-circuits.pub/2022/in-context-learning-and-induction-heads/index.html.

\bibitem[{OpenAI}(2023)]{openai2023gpt4}
{OpenAI}.
\newblock {GPT}-4 technical report.
\newblock \emph{arXiv preprint arXiv:2303.08774}, 2023.

\bibitem[Ouyang et~al.(2022)Ouyang, Wu, Jiang, Almeida, Wainwright, Mishkin,
  Zhang, Agarwal, Slama, Ray, et~al.]{ouyang2022training}
L.~Ouyang, J.~Wu, X.~Jiang, D.~Almeida, C.~Wainwright, P.~Mishkin, C.~Zhang,
  S.~Agarwal, K.~Slama, A.~Ray, et~al.
\newblock Training language models to follow instructions with human feedback.
\newblock \emph{Advances in Neural Information Processing Systems},
  35:\penalty0 27730--27744, 2022.

\bibitem[Pearl(1995)]{pearl1995causal}
J.~Pearl.
\newblock Causal diagrams for empirical research.
\newblock \emph{Biometrika}, 82\penalty0 (4):\penalty0 669--688, 1995.
\newblock ISSN 00063444.
\newblock URL \url{http://www.jstor.org/stable/2337329}.

\bibitem[Pearl(2012)]{pearl2012calculus}
J.~Pearl.
\newblock The do-calculus revisited.
\newblock \emph{arXiv preprint arXiv:1210.4852}, 2012.

\bibitem[Perez et~al.(2022)Perez, Huang, Song, Cai, Ring, Aslanides, Glaese,
  McAleese, and Irving]{perez2022red}
E.~Perez, S.~Huang, F.~Song, T.~Cai, R.~Ring, J.~Aslanides, A.~Glaese,
  N.~McAleese, and G.~Irving.
\newblock Red teaming language models with language models.
\newblock \emph{arXiv preprint arXiv:2202.03286}, 2022.

\bibitem[Ram et~al.(2022)Ram, Bezalel, Zicher, Belinkov, Berant, and
  Globerson]{ram2022you}
O.~Ram, L.~Bezalel, A.~Zicher, Y.~Belinkov, J.~Berant, and A.~Globerson.
\newblock What are you token about? dense retrieval as distributions over the
  vocabulary.
\newblock \emph{arXiv preprint arXiv:2212.10380}, 2022.

\bibitem[Saunders et~al.(2022)Saunders, Yeh, Wu, Bills, Ouyang, Ward, and
  Leike]{saunders2022self}
W.~Saunders, C.~Yeh, J.~Wu, S.~Bills, L.~Ouyang, J.~Ward, and J.~Leike.
\newblock Self-critiquing models for assisting human evaluators.
\newblock \emph{arXiv preprint arXiv:2206.05802}, 2022.

\bibitem[Shevlane et~al.(2023)Shevlane, Farquhar, Garfinkel, Phuong,
  Whittlestone, Leung, Kokotajlo, Marchal, Anderljung, Kolt,
  et~al.]{shevlane2023model}
T.~Shevlane, S.~Farquhar, B.~Garfinkel, M.~Phuong, J.~Whittlestone, J.~Leung,
  D.~Kokotajlo, N.~Marchal, M.~Anderljung, N.~Kolt, et~al.
\newblock Model evaluation for extreme risks.
\newblock \emph{arXiv preprint arXiv:2305.15324}, 2023.

\bibitem[Uesato et~al.(2022)Uesato, Kushman, Kumar, Song, Siegel, Wang,
  Creswell, Irving, and Higgins]{uesato2022solving}
J.~Uesato, N.~Kushman, R.~Kumar, F.~Song, N.~Siegel, L.~Wang, A.~Creswell,
  G.~Irving, and I.~Higgins.
\newblock Solving math word problems with process-and outcome-based feedback.
\newblock \emph{arXiv preprint arXiv:2211.14275}, 2022.

\bibitem[Wang et~al.(2022)Wang, Variengien, Conmy, Shlegeris, and
  Steinhardt]{wang2022interpretability}
K.~Wang, A.~Variengien, A.~Conmy, B.~Shlegeris, and J.~Steinhardt.
\newblock Interpretability in the wild: a circuit for indirect object
  identification in {GPT}-2 small.
\newblock \emph{arXiv preprint arXiv:2211.00593}, 2022.

\bibitem[Wu et~al.(2023)Wu, Geiger, Potts, and Goodman]{wu2023interpretability}
Z.~Wu, A.~Geiger, C.~Potts, and N.~D. Goodman.
\newblock Interpretability at scale: Identifying causal mechanisms in alpaca.
\newblock \emph{arXiv preprint arXiv:2305.08809}, 2023.

\bibitem[Ziegler et~al.(2019)Ziegler, Stiennon, Wu, Brown, Radford, Amodei,
  Christiano, and Irving]{ziegler2019fine}
D.~M. Ziegler, N.~Stiennon, J.~Wu, T.~B. Brown, A.~Radford, D.~Amodei,
  P.~Christiano, and G.~Irving.
\newblock Fine-tuning language models from human preferences.
\newblock \emph{arXiv preprint arXiv:1909.08593}, 2019.

\end{thebibliography}

\appendix

\newpage
\section{Results on synthetic multiple choice}
\label{sec:syntactic_abcd_results}
\begin{figure}[t]
    \centering
     \begin{subfigure}[b]{0.495\textwidth}
         \centering
         \includegraphics[width=\textwidth]{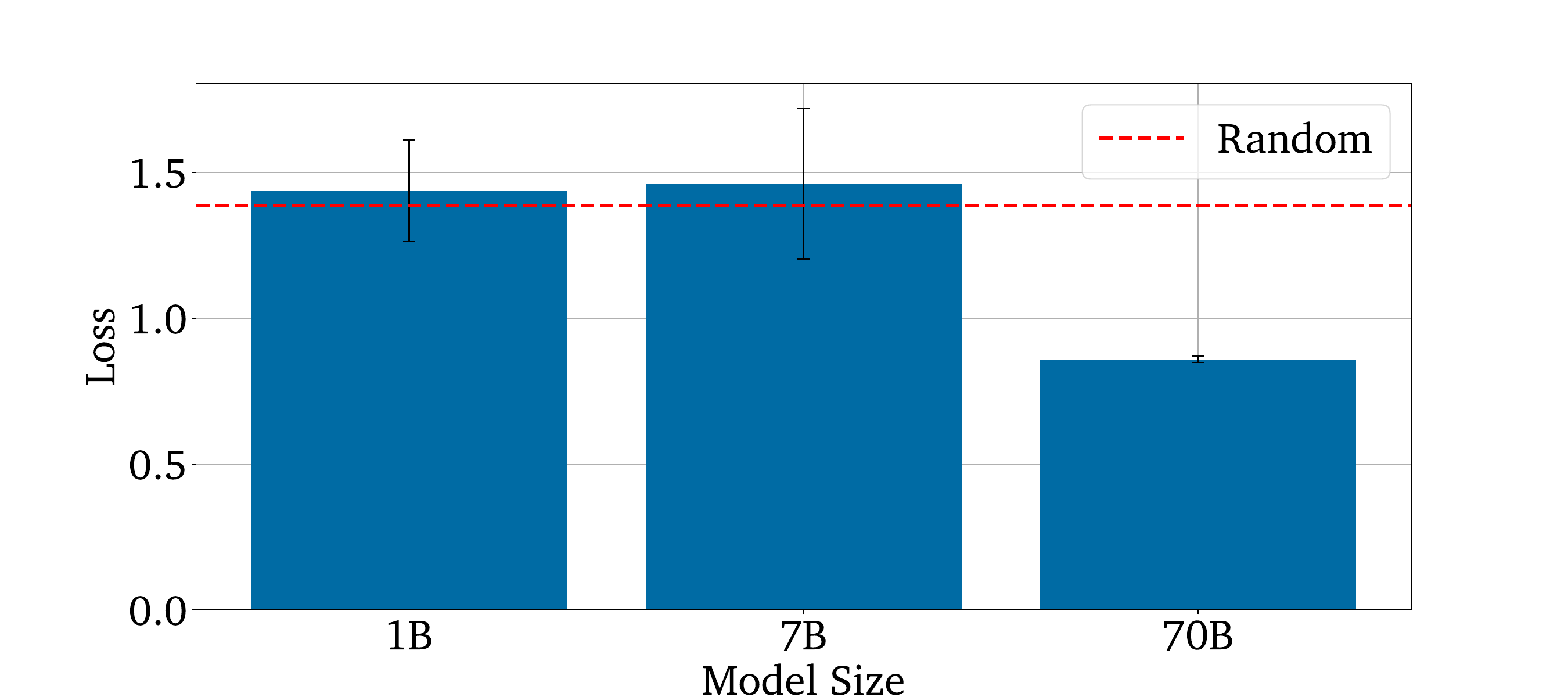}
         \caption{Loss}
     \end{subfigure}
     \hfill
     \begin{subfigure}[b]{0.495\textwidth}
         \centering
         \includegraphics[width=\textwidth]{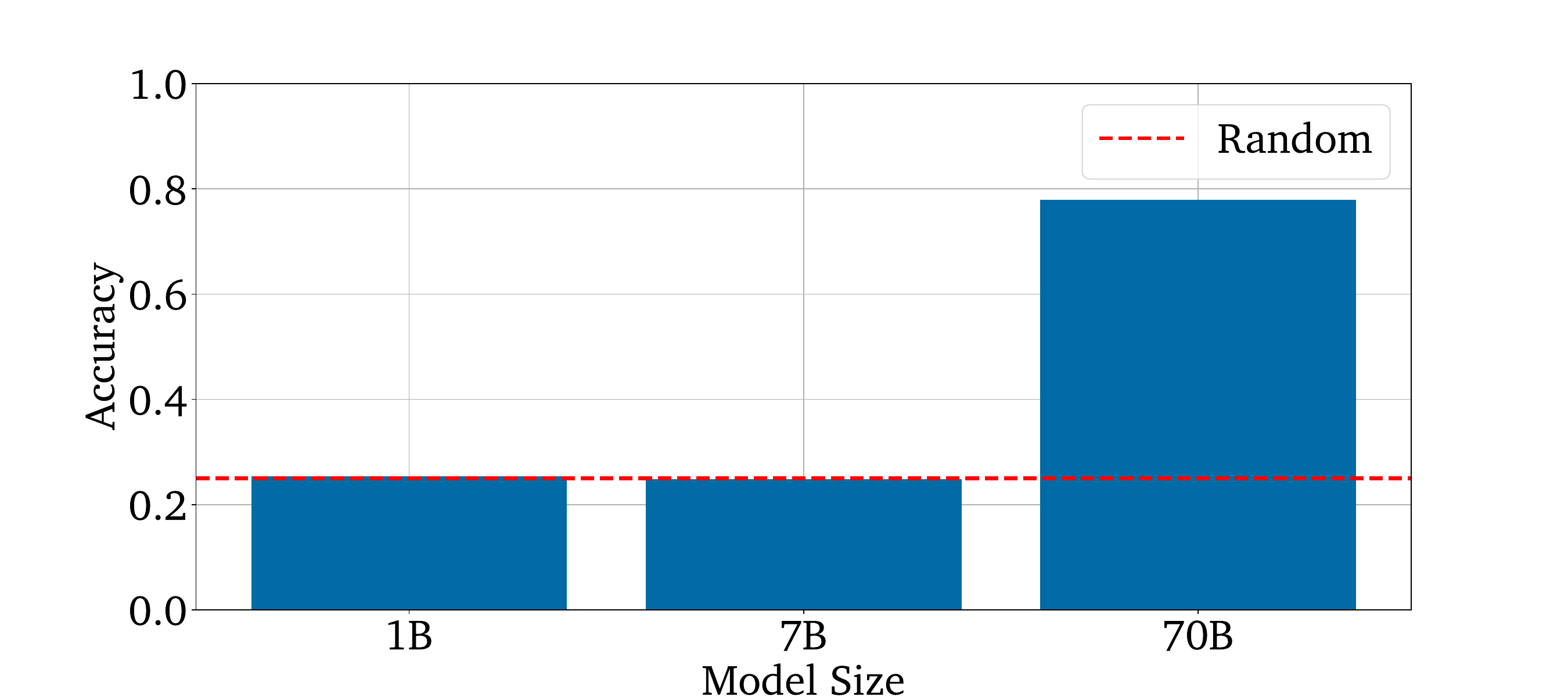}
         \caption{Accuracy}
     \end{subfigure}
    \caption{Performance comparison of model sizes on the synthetic multiple choice dataset.}
    \label{fig:size_perf_synthetic}
\end{figure}

In order to isolate the ability of different model sizes to pick between different answers, we create a synthetic multiple choice dataset which does not rely on factual knowledge. These questions are simply asking which of the options is equal to a specific token. An example question is shown in~\cref{fig:example_synthetic_mcq}. We report loss and accuracy in~\cref{fig:size_perf_synthetic}. Only Chinchilla 70B is able to perform well on this task. Note that the loss (the average negative log probability of the correct answer letter) for smaller models is equal to the entropy of a uniform distribution over a set of four members, meaning that the smaller models are able to discern that they should answer with A, B, C, or D, but are unable to identify which label is the correct one.

Although not shown in this work, when analyzing the circuit of Chinchilla 70B on the synthetic dataset, we largely found the same responsible nodes as on MMLU, with the same set of correct letter, heads, amplification heads, etc.

\begin{figure}[t]
    \centering
    \includegraphics[width=\textwidth]{figures/synth_example.pdf}
    \caption{Example of a synthetic multiple choice question. Boxes denote token boundaries.}
    \label{fig:example_synthetic_mcq}
\end{figure}

\clearpage

\newpage
\section{Preliminary analysis on the rest of the circuit}
\label{sec:rest_of_the_circuit}

\subsection{Content Gatherers}

\begin{figure}[t]
    \centering
     \begin{subfigure}[b]{0.4\textwidth}
         \centering
         \includegraphics[width=\textwidth]{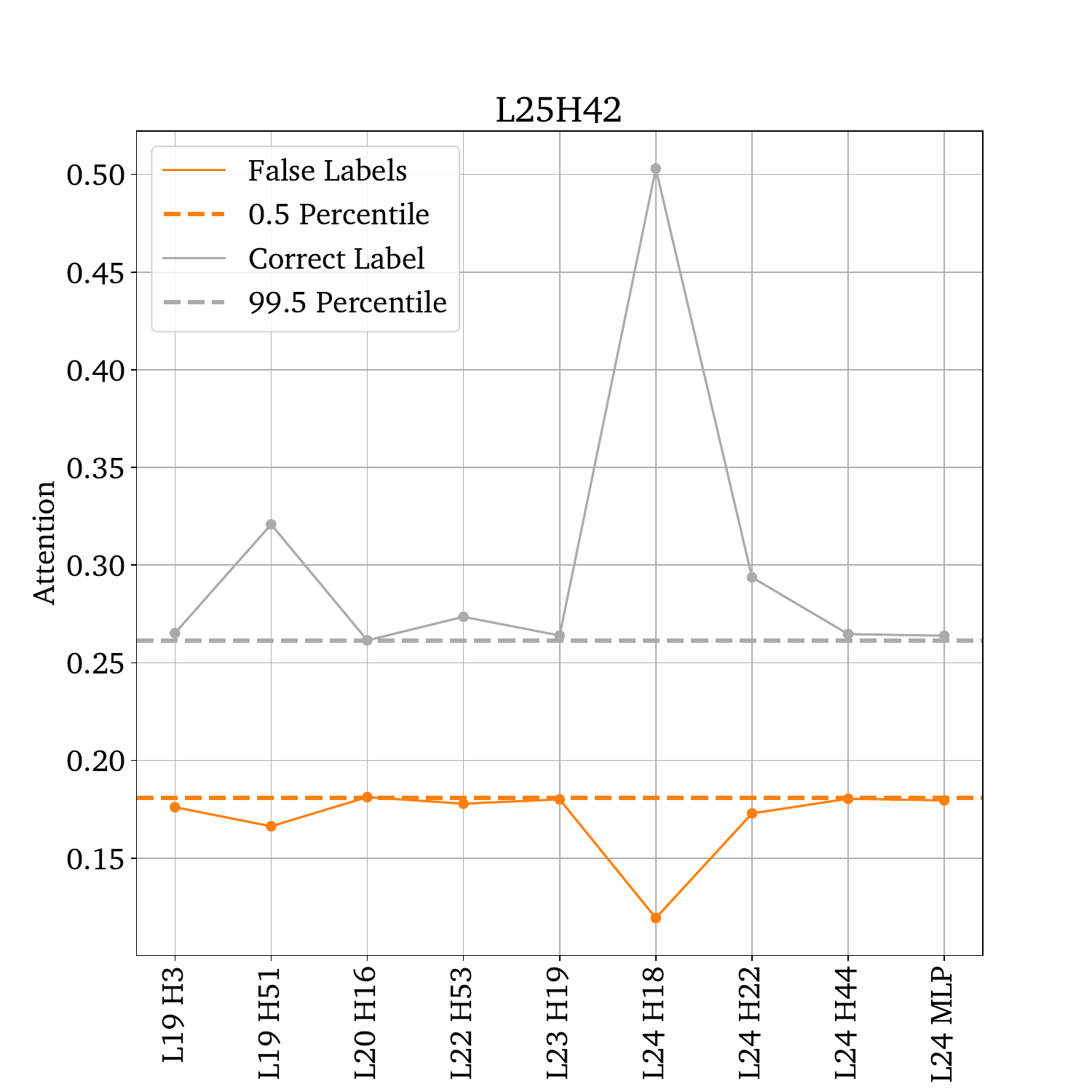}
         \caption{}
     \end{subfigure}
     \hfill
    \begin{subfigure}[b]{0.4\textwidth}
         \centering
         \includegraphics[width=\textwidth]{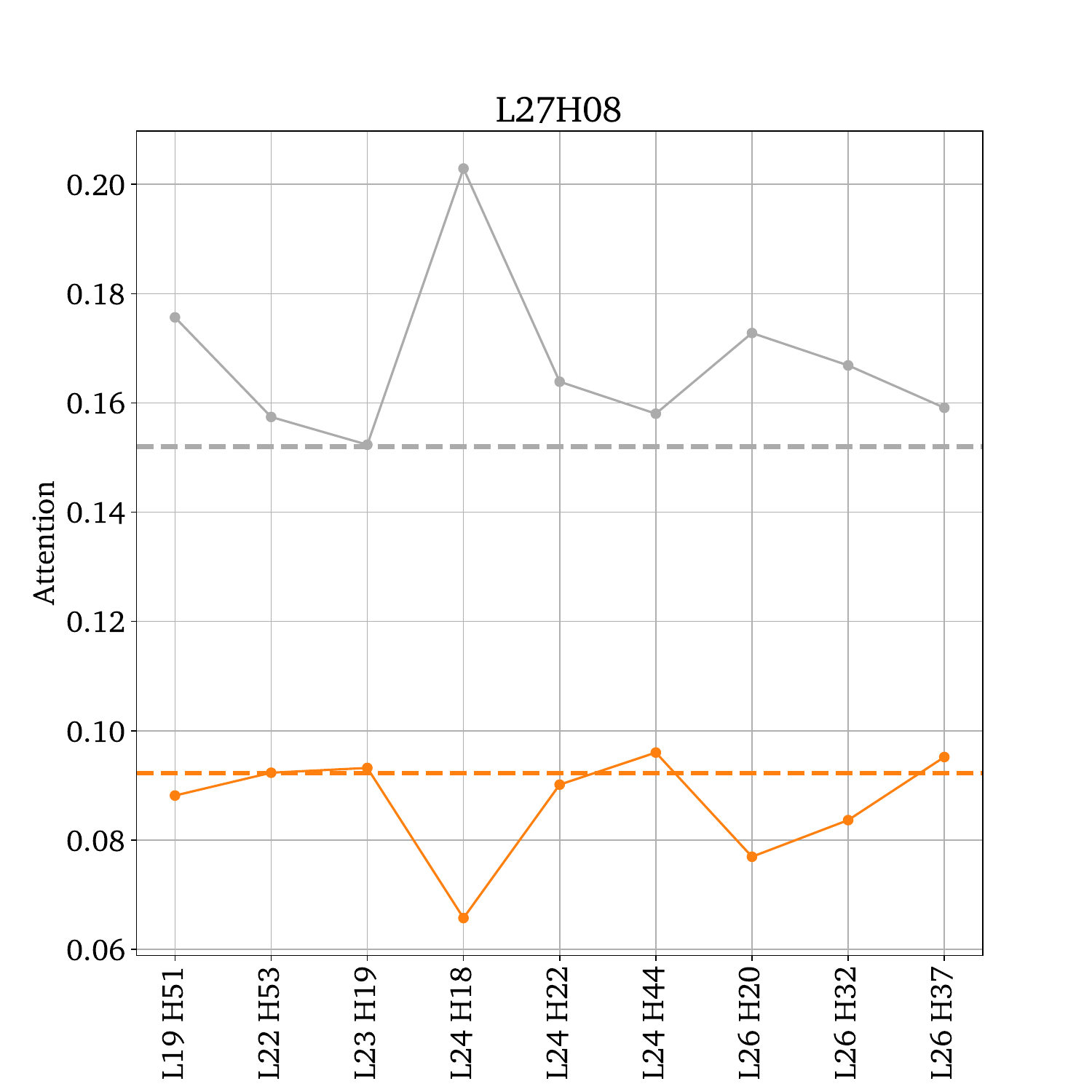}
         \caption{}
     \end{subfigure}
     \hfill
     \begin{subfigure}[b]{0.4\textwidth}
         \centering
         \includegraphics[width=\textwidth]{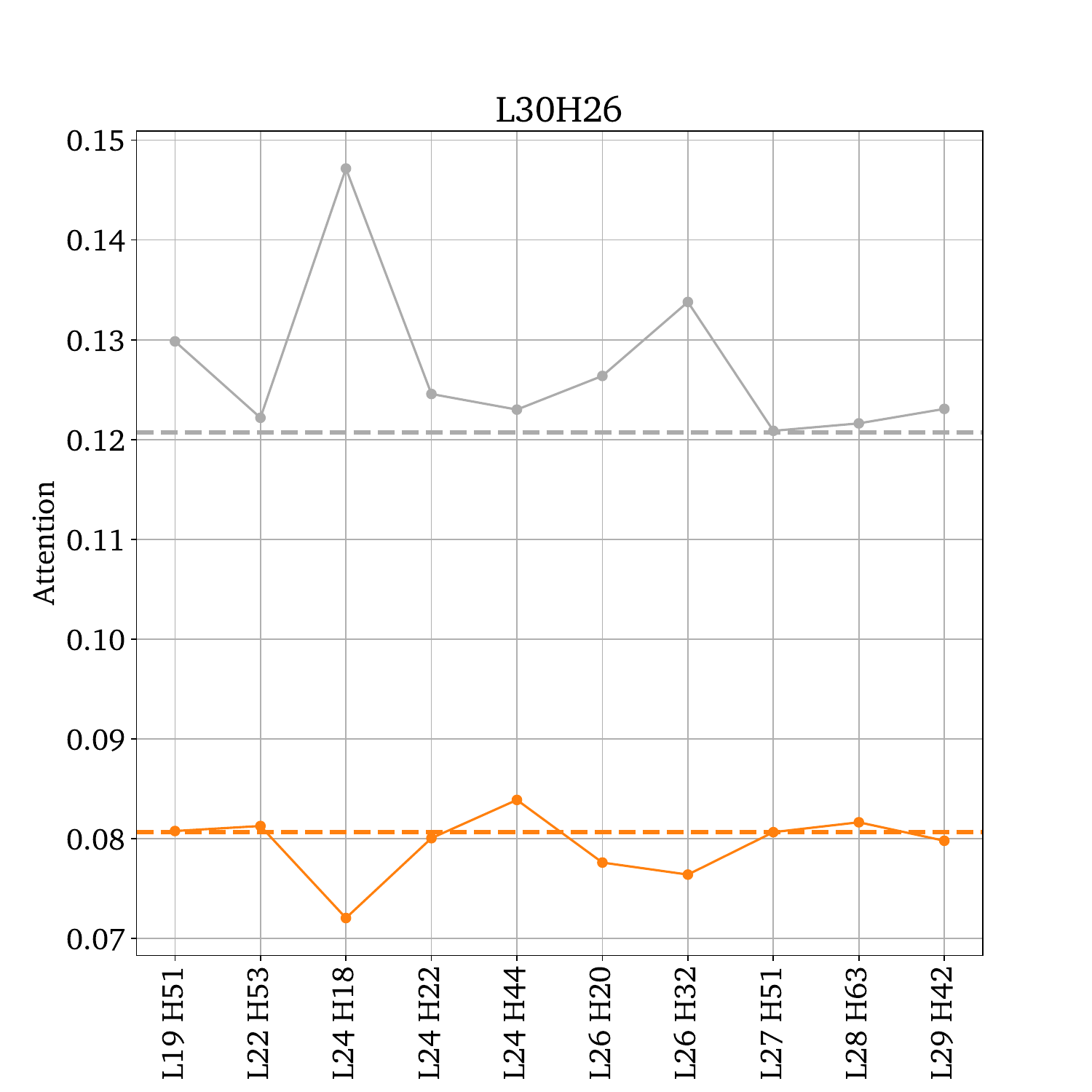}
         \caption{}
     \end{subfigure}
     \hfill
     \begin{subfigure}[b]{0.4\textwidth}
         \centering
         \includegraphics[width=\textwidth]{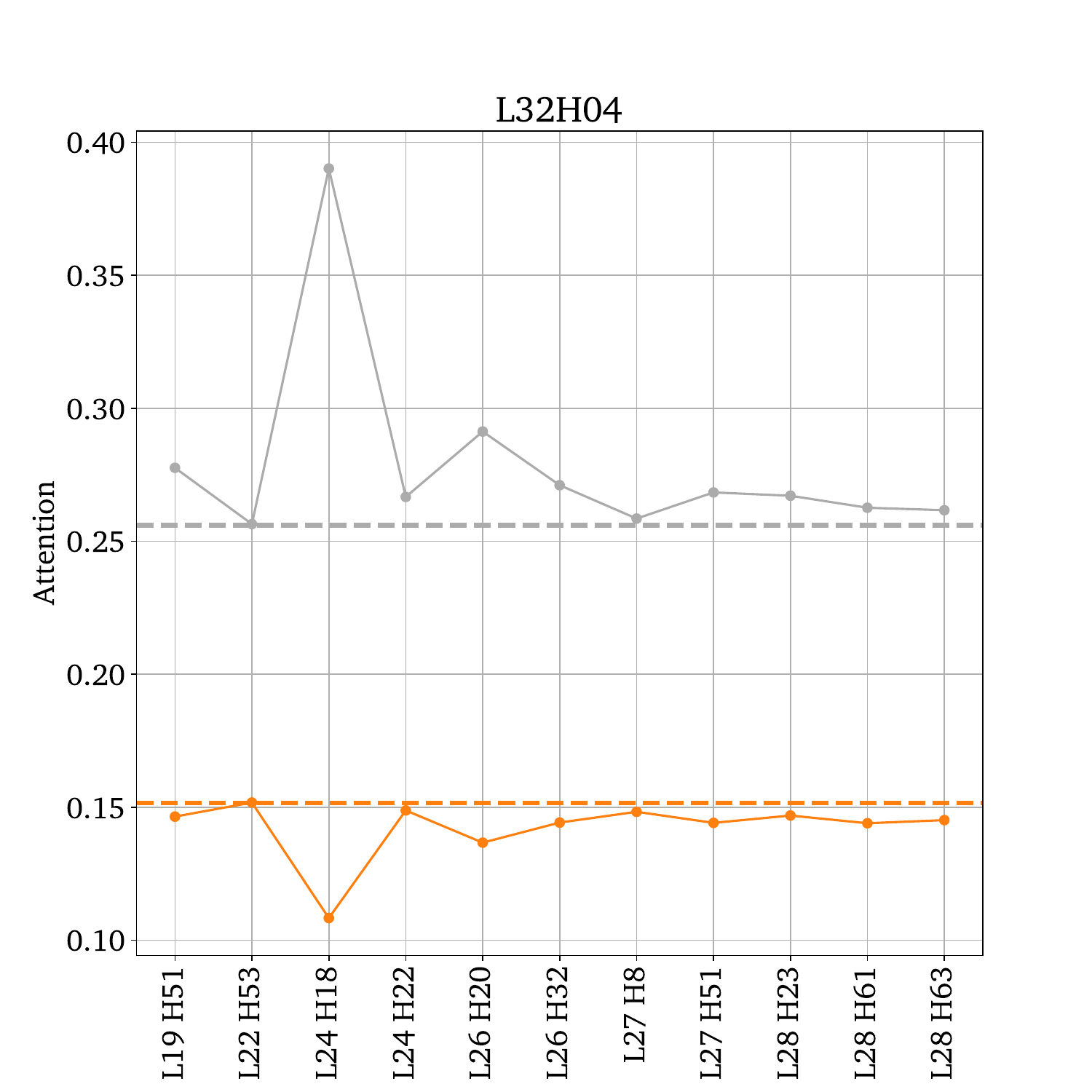}
         \caption{}
     \end{subfigure}
     \hfill
     \begin{subfigure}[b]{0.4\textwidth}
         \centering
         \includegraphics[width=\textwidth]{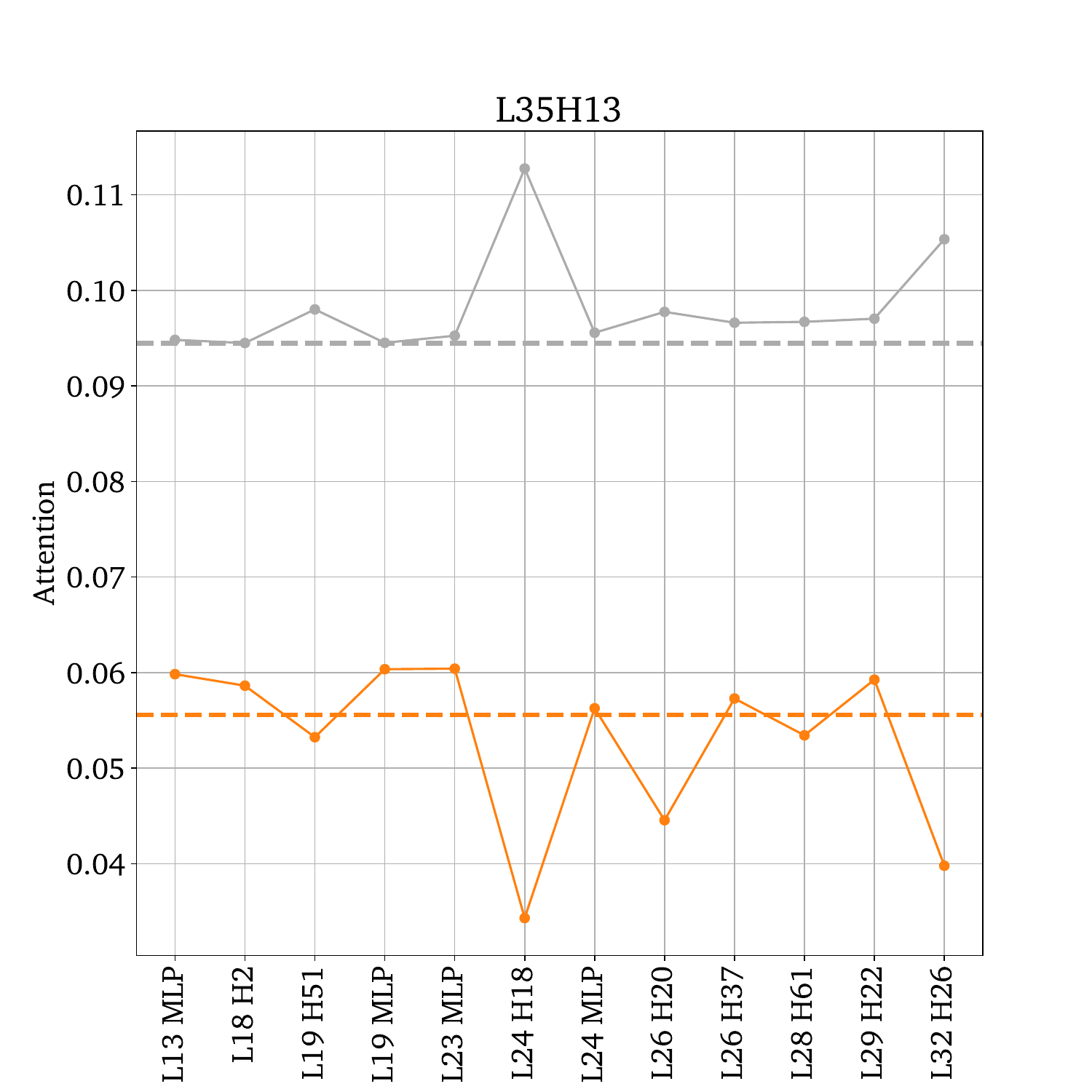}
         \caption{}
     \end{subfigure}
     \hfill
     \begin{subfigure}[b]{0.4\textwidth}
         \centering
         \includegraphics[width=\textwidth]{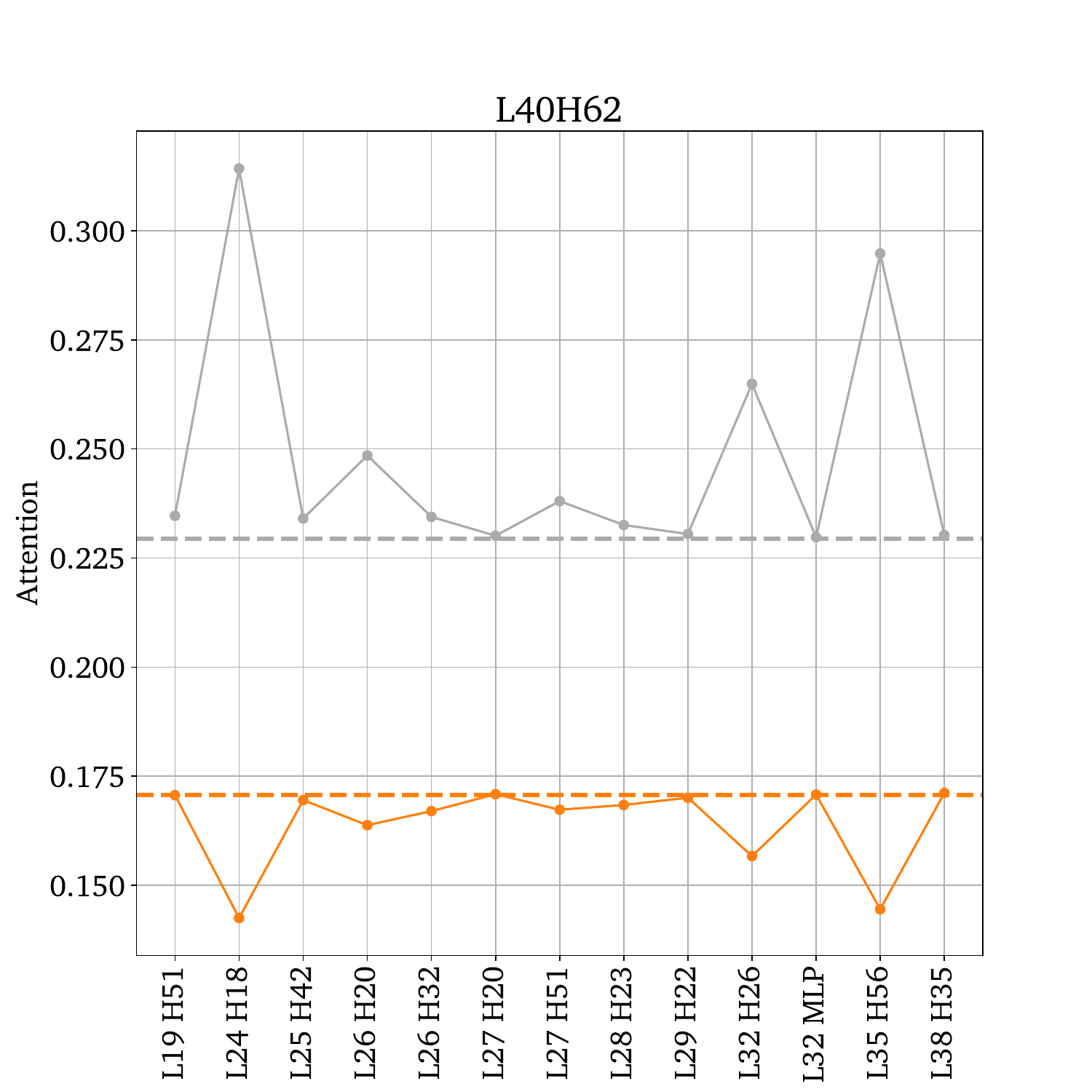}
         \caption{}
     \end{subfigure}
     \hfill
    \caption{Attention from final token to correct or false labels when path patching the outputs of various nodes at the final token.}
    \label{fig:attn_delta_cg}
\end{figure}

In~\cref{sec:understanding_nodes} we identify the correct letter heads as the most interesting heads. Here, we identify the inputs to these heads. Recall that correct letter heads overwhelmingly attend from the final token to the correct letter. First, we identify the nodes which affect this attention via the query, by path patching the edge between each valid node's output at the final token and the attention input at the final token for each correct letter head~\citep{wang2022interpretability}. We patch in from prompt $p_{intervention}$ into prompt $p_{original}$ and record the attention on the correct and incorrect letters according to $p_{intervention}$. In~\cref{fig:attn_delta_cg} we report the 0.5th percentile average attention on the false labels and 99.5th percentile on the correct labels. We show the effects of nodes which cross the 99.5th percentile in terms of increasing attention probability on the correct label.

We want to highlight that the most notable and consistent effect is coming from patching the output of \texttt{L24 H18}, which we identify as having very large total but small direct effect on the loss in~\cref{fig:total_and_direct_effect}. This provides evidence that the mechanism through which it achieves this is by influencing the attention of correct letter heads. 

\begin{figure}[t]
    \centering
    \includegraphics[width=0.7\textwidth]{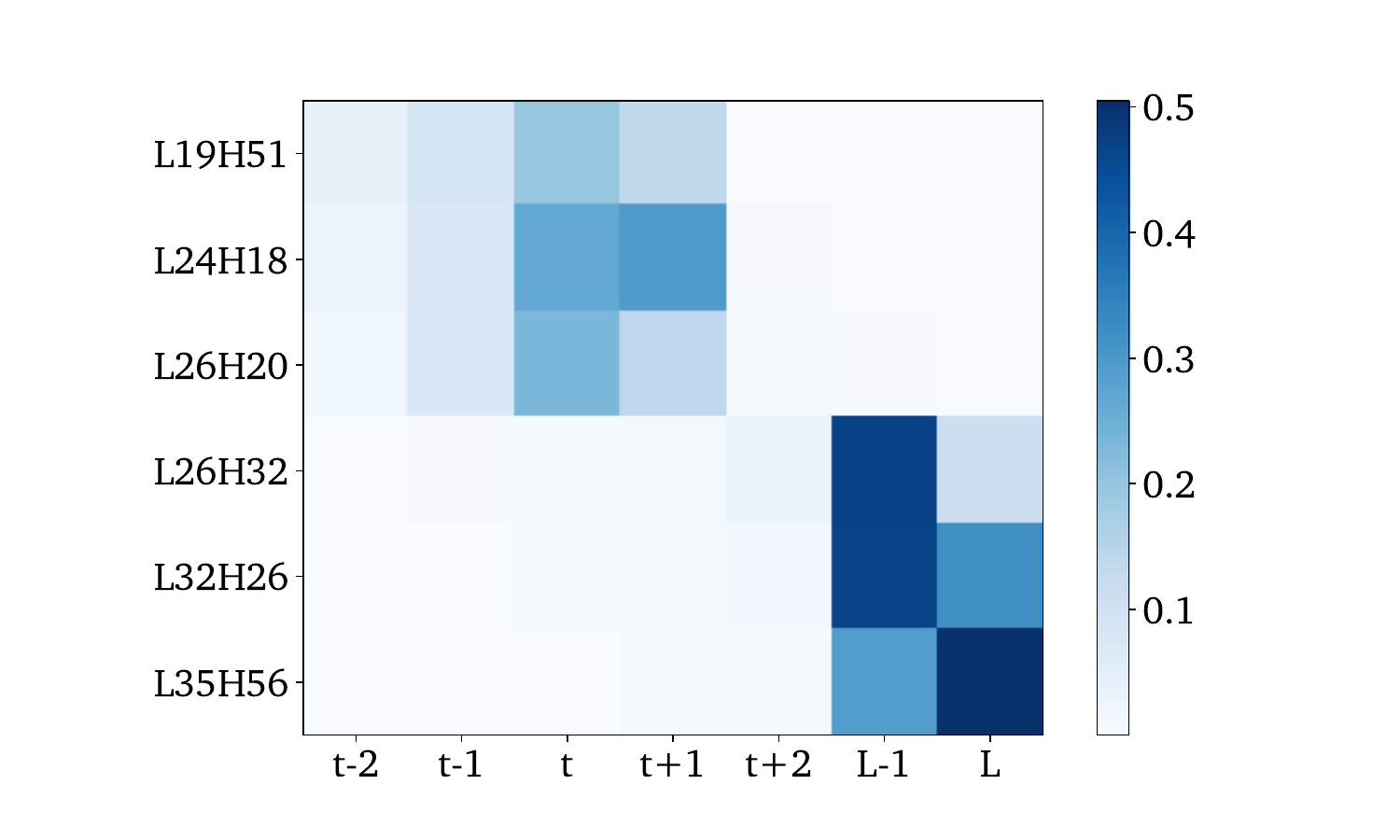}
    \caption{Average attention patterns of content gatherer heads. \texttt{t} denotes the final content token (before the \texttt{\textbackslash n} of the next answer) of the correct answer. \texttt{L} denotes the final token position.}
    \label{fig:content_gather_attention}
\end{figure}

The set of heads with the strongest effects are  \texttt{L19 H51}, \texttt{L24 H18}, \texttt{L26 H20}, \texttt{L26 H32}, \texttt{L32 H26}, and \texttt{L35 H56}. We show a slice of their average attention pattern in~\cref{fig:content_gather_attention}.
We find that the first three heads attend to the final tokens of the content of the correct answer. Sometimes they split their attention between final content tokens of different answers, suggesting uncertainty in the model about which answer is correct. The latter three heads attend mostly to the final two tokens, resembling the so-called amplification heads we identify in~\cref{sec:understanding_nodes}. Speculatively, the latter three heads are cleaning up or amplifying a signal written to the final positions by the former three heads. Based on the attention patterns of the first three heads we refer to this class of heads as content gatherers.

\clearpage

\subsection{Symbol Binding MLPs}

We have now established why the correct letter heads attend to the correct label. Now we turn to the question of how they know to upvote the corresponding label if they attend to it. We find that this is mediated via a diffuse set of MLPs which is inconsistent across correct letter heads.

\begin{figure}[t]
    \centering
     \begin{subfigure}[b]{0.495\textwidth}
         \centering
         \includegraphics[width=\textwidth]{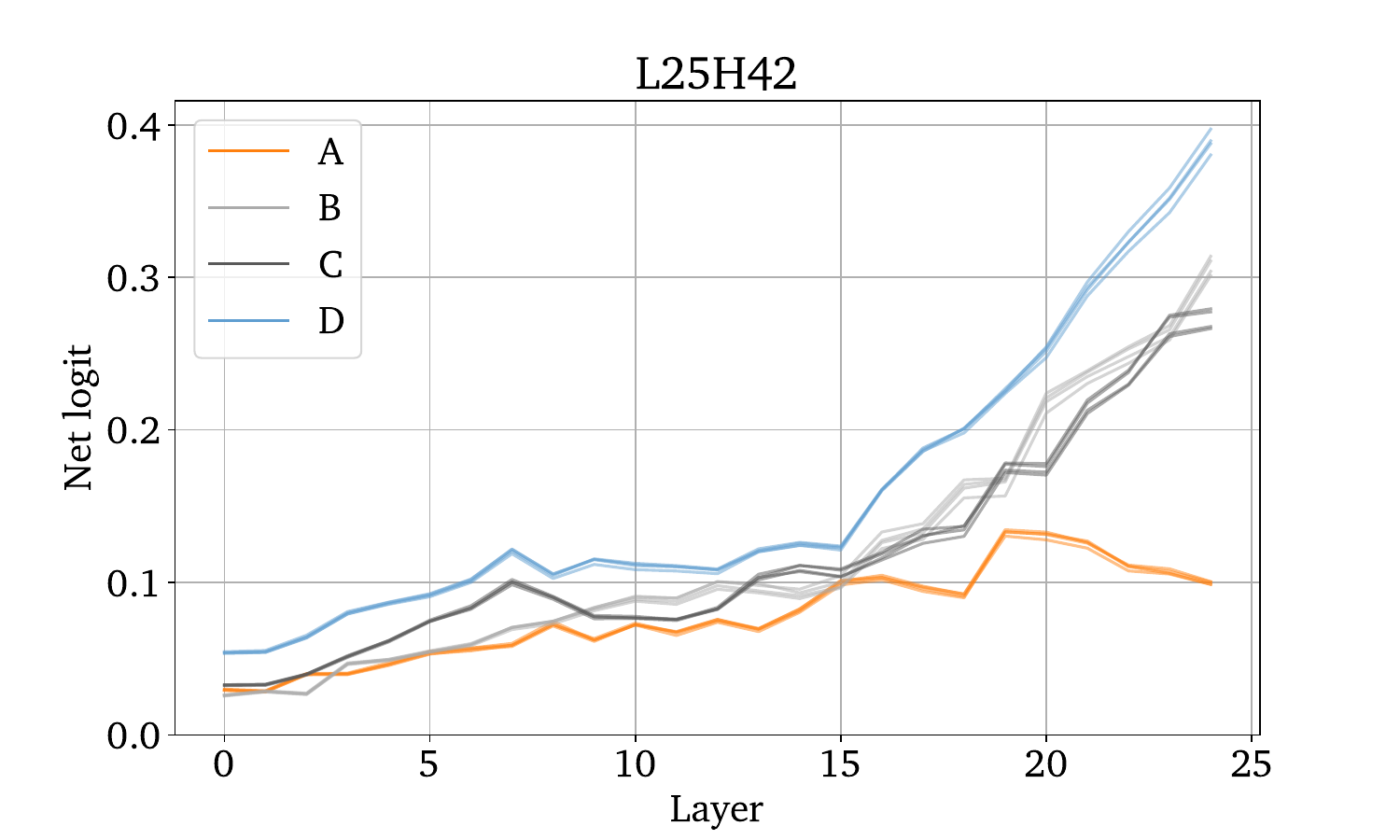}
         \caption{}
     \end{subfigure}
     \hfill
     \begin{subfigure}[b]{0.495\textwidth}
         \centering
         \includegraphics[width=\textwidth]{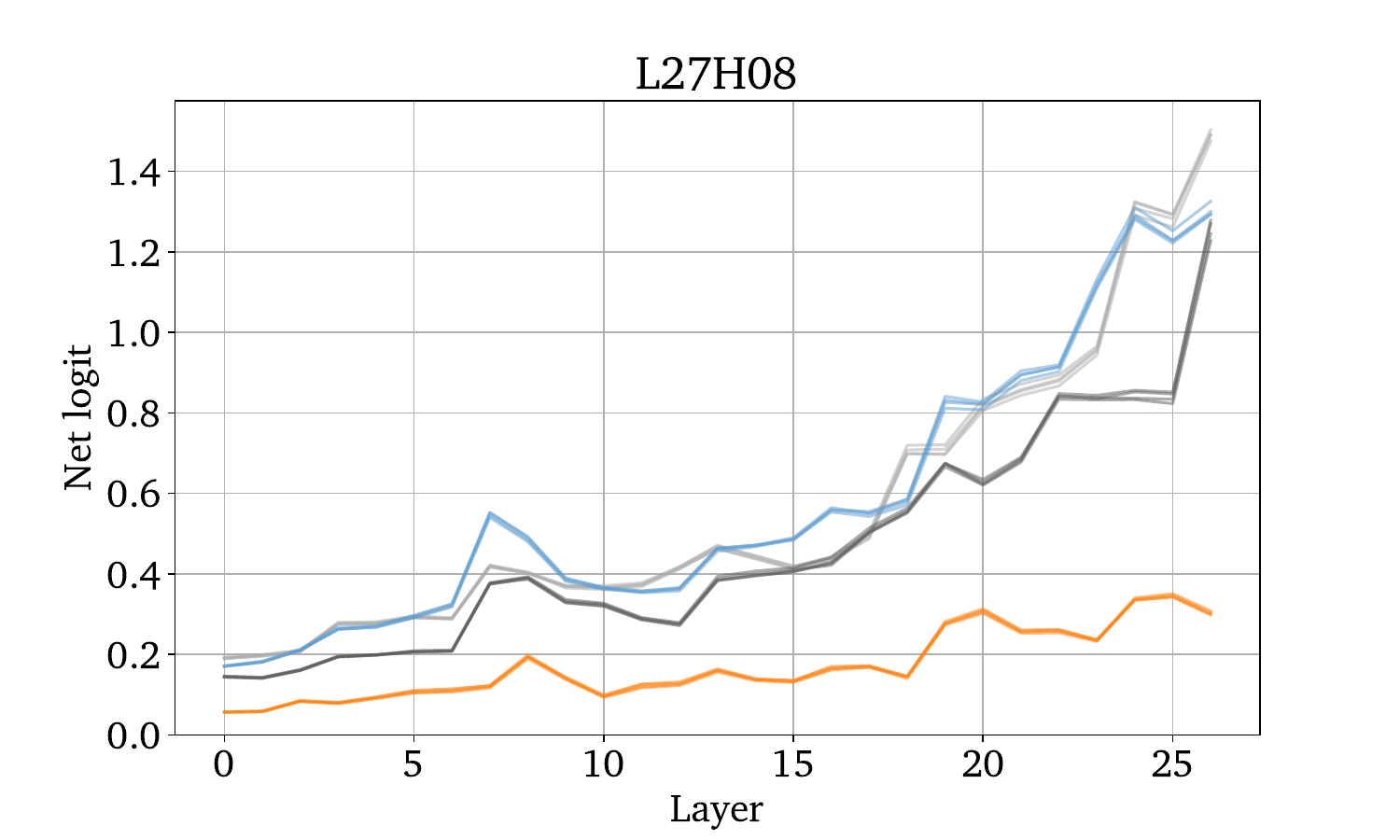}
         \caption{}
     \end{subfigure}
     \hfill
     \begin{subfigure}[b]{0.495\textwidth}
         \centering
         \includegraphics[width=\textwidth]{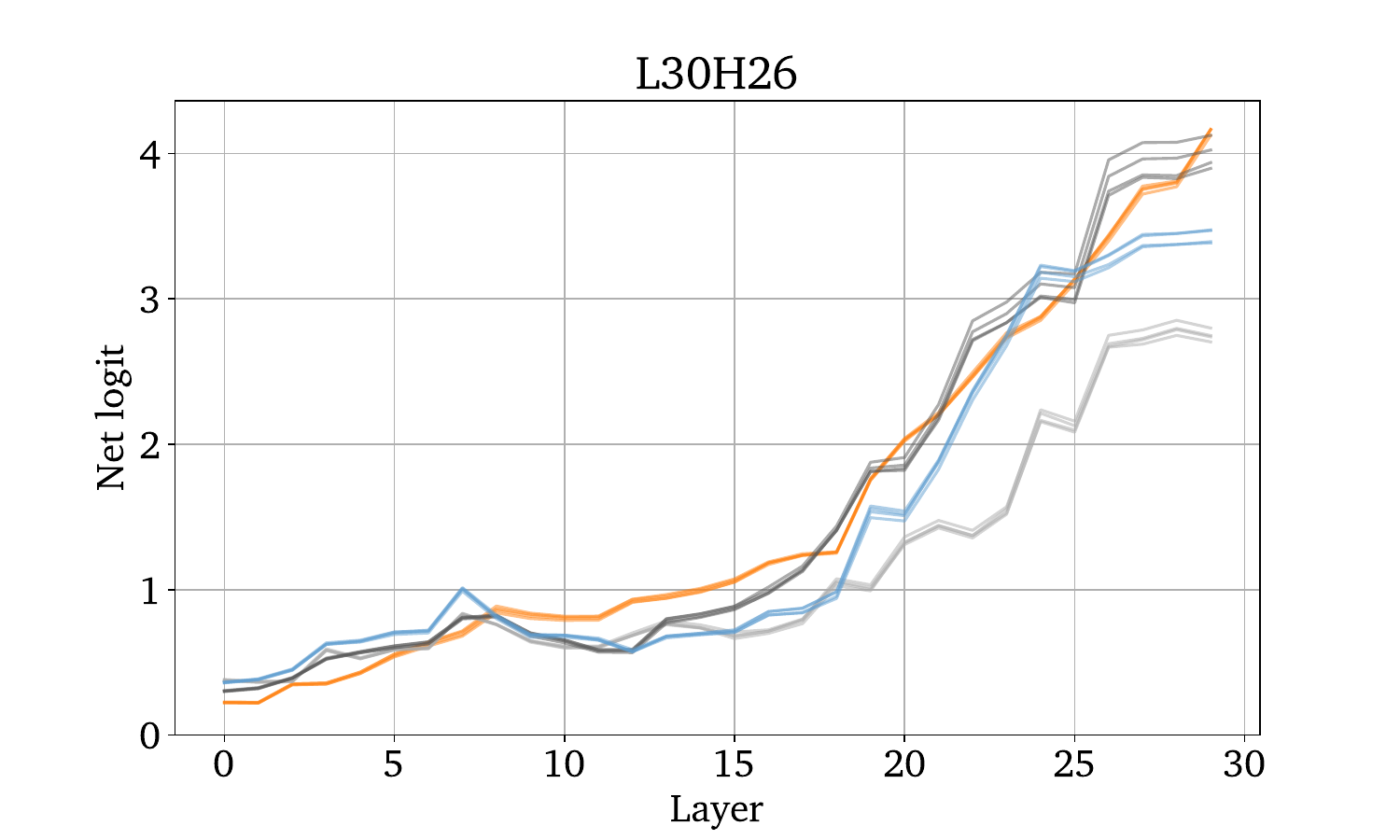}
         \caption{}
     \end{subfigure}
     \hfill
     \begin{subfigure}[b]{0.495\textwidth}
         \centering
         \includegraphics[width=\textwidth]{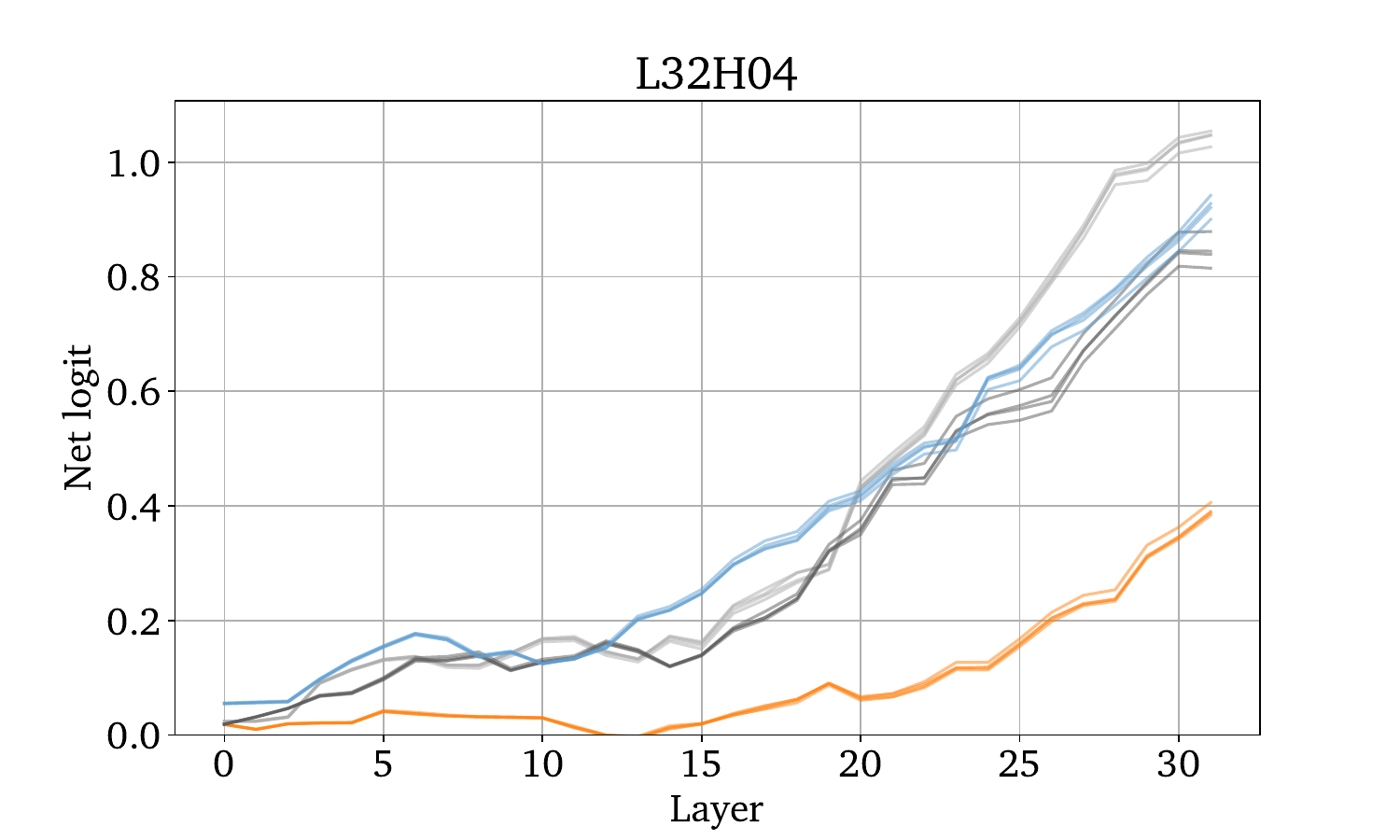}
         \caption{}
     \end{subfigure}
     \hfill
     \begin{subfigure}[b]{0.495\textwidth}
         \centering
         \includegraphics[width=\textwidth]{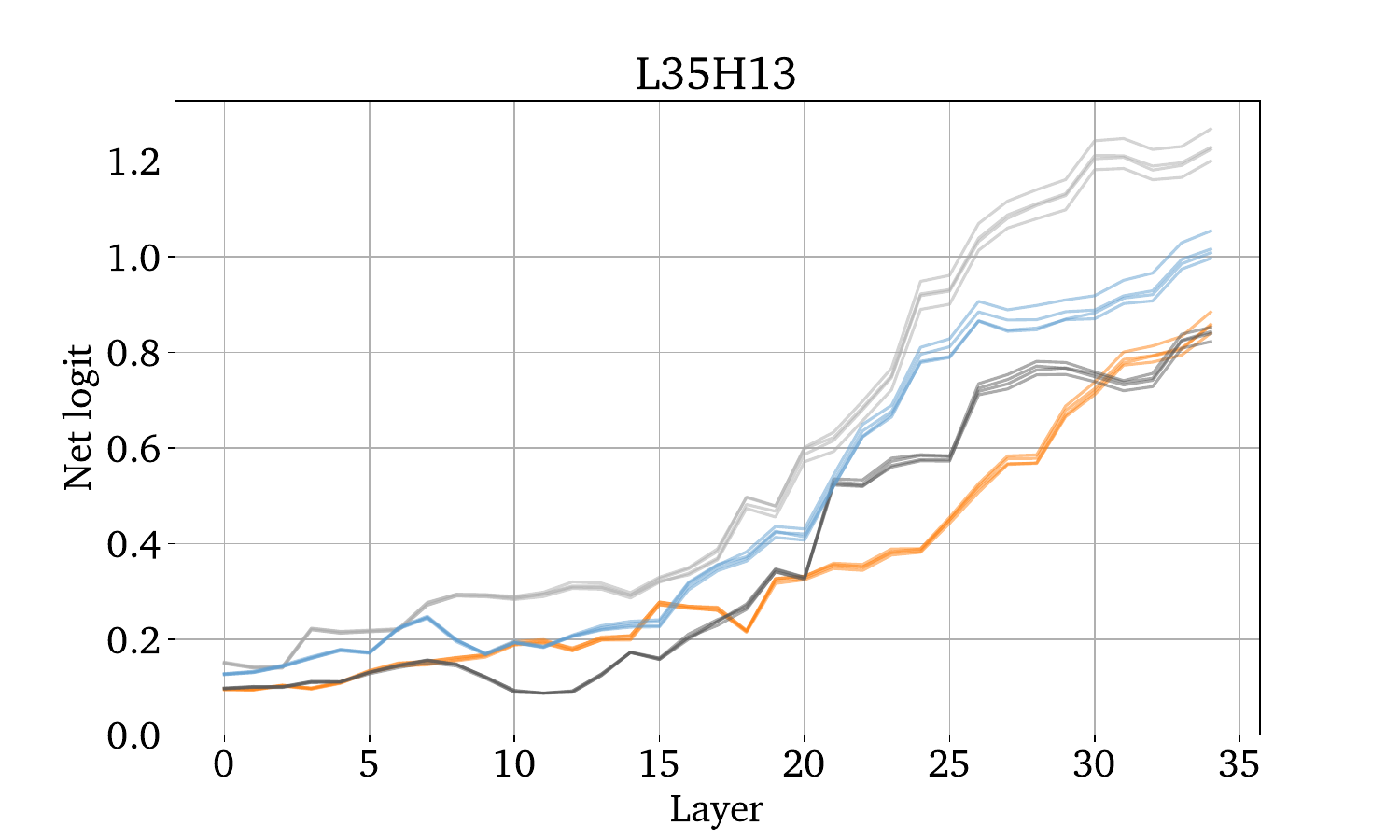}
         \caption{}
     \end{subfigure}
     \hfill
     \begin{subfigure}[b]{0.495\textwidth}
         \centering
         \includegraphics[width=\textwidth]{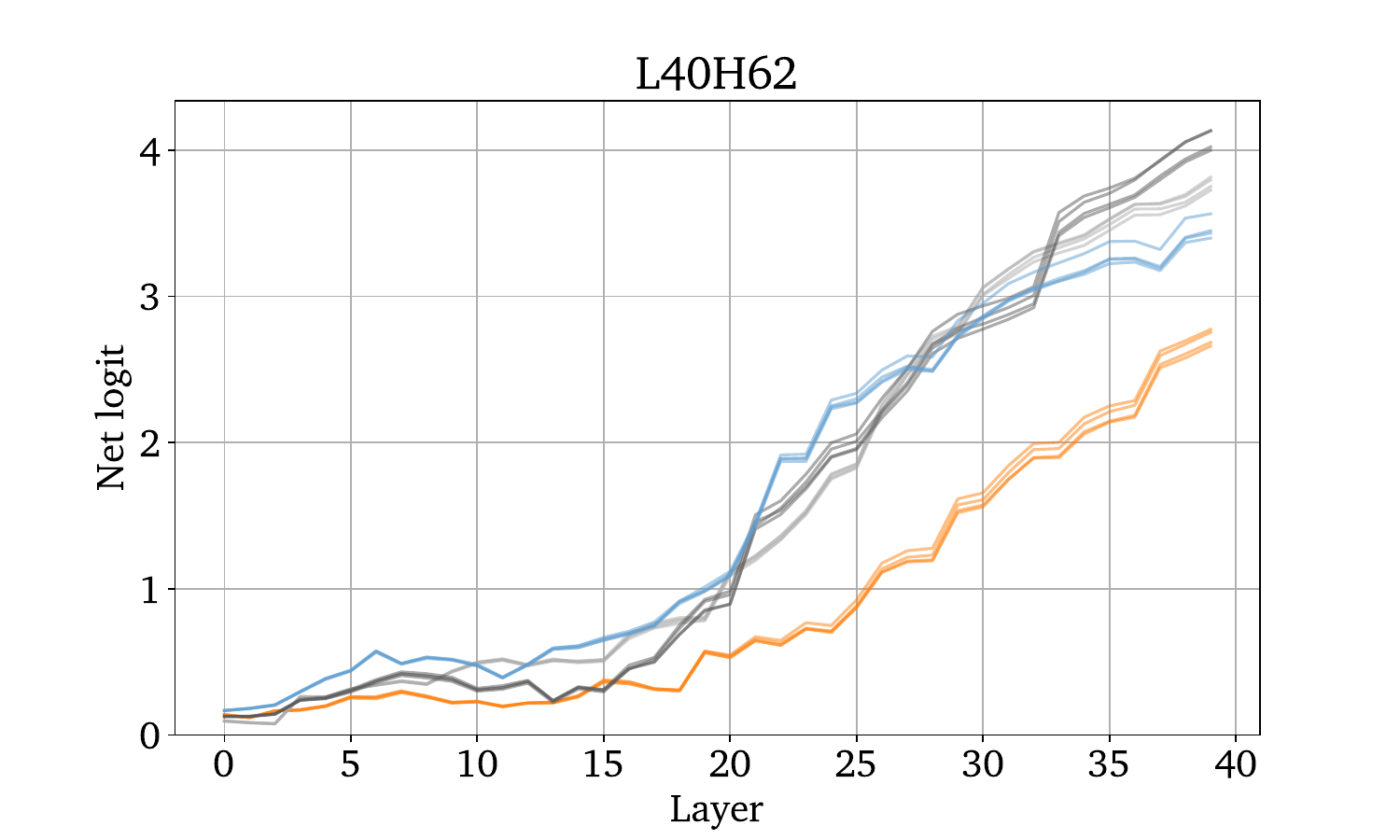}
         \caption{}
     \end{subfigure}
    \caption{Direct effect of correct letter heads when attending to the label positions. Different colors denote different token positions. Different lines of the same color denote different settings, i.e. different correct labels, showing that there is little difference in behavior regardless of the correct label.}
    \label{fig:unembed_residual}
\end{figure}

We define an unembedding function by composing the correct letter head OV matrix with the unembedding matrix and the final RMS norm $\alpha_x$ for a given prompt $x$. This allows us to measure the net effect on the logit of the correct label, relative to the other labels, that fully attending to a position would have. That is, positing a one-hot attention to one of the label positions, what's the direct effect of the correct letter head? This is shown in~\cref{fig:unembed_residual}, where we apply this unembedding function at various network depths, showing that the strength of the `token identity` feature, for lack of a better word, is growing steadily across the depth of the network.

\begin{figure}[t]
    \centering
     \begin{subfigure}[b]{0.495\textwidth}
         \centering
         \includegraphics[width=\textwidth]{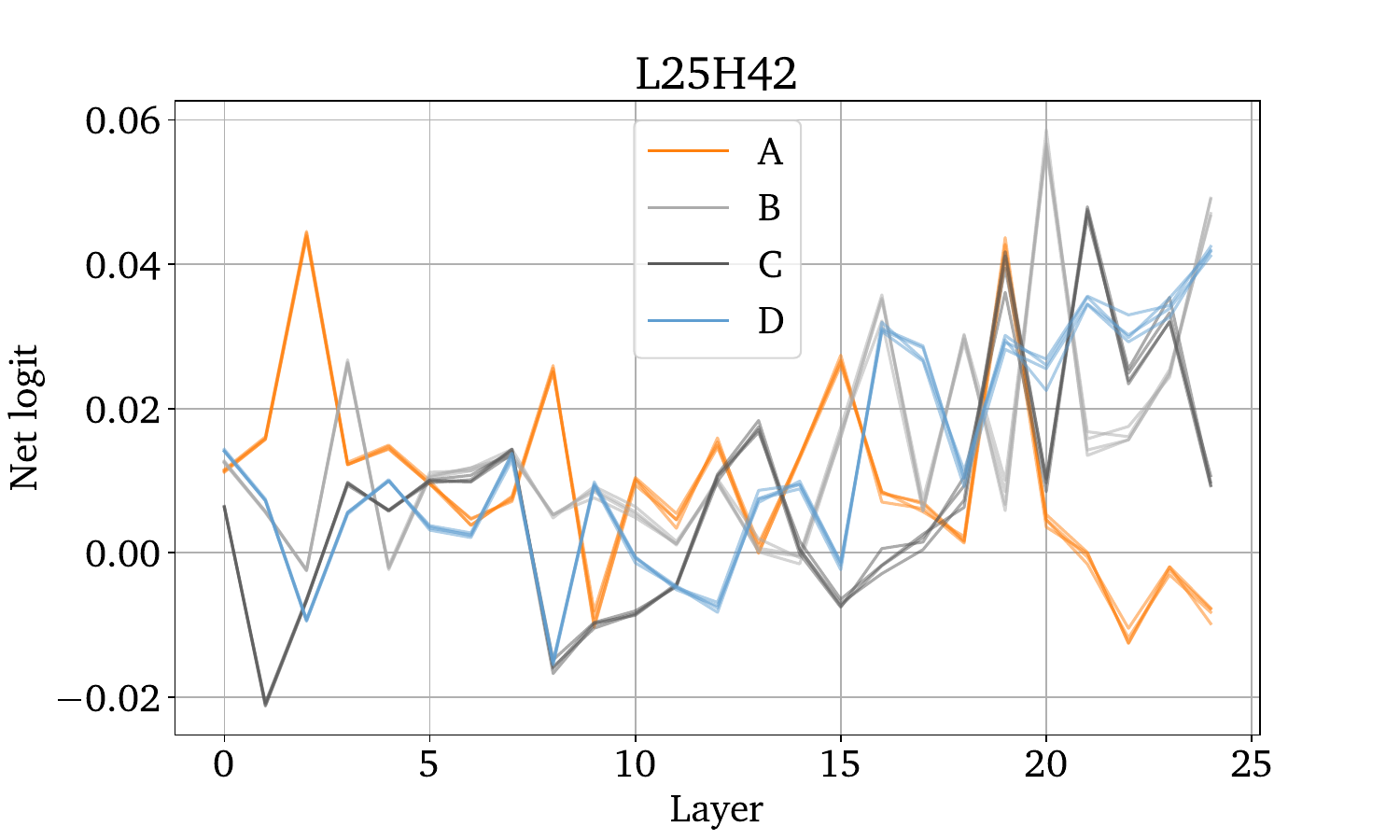}
         \caption{}
     \end{subfigure}
     \hfill
     \begin{subfigure}[b]{0.495\textwidth}
         \centering
         \includegraphics[width=\textwidth]{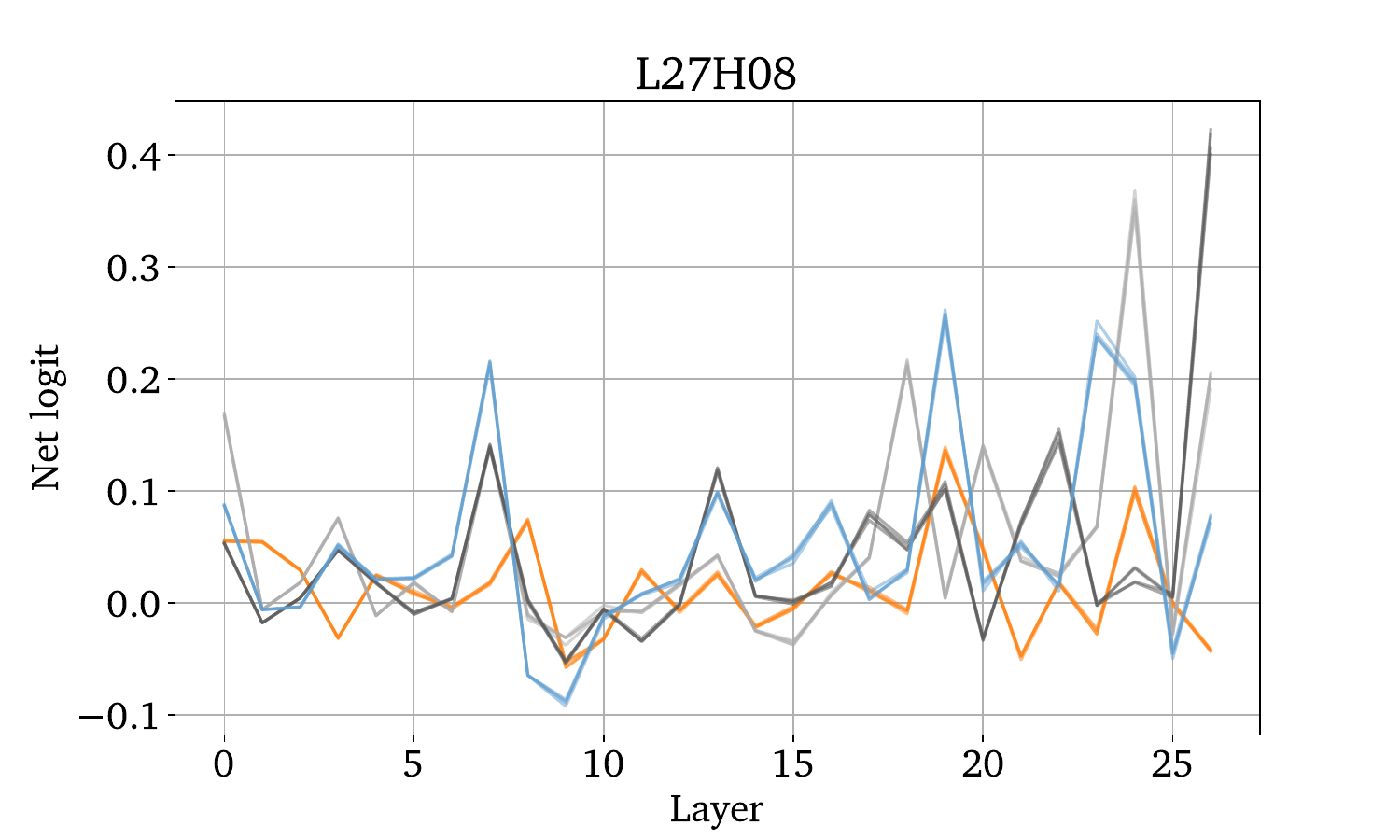}
         \caption{}
     \end{subfigure}
     \hfill
     \begin{subfigure}[b]{0.495\textwidth}
         \centering
         \includegraphics[width=\textwidth]{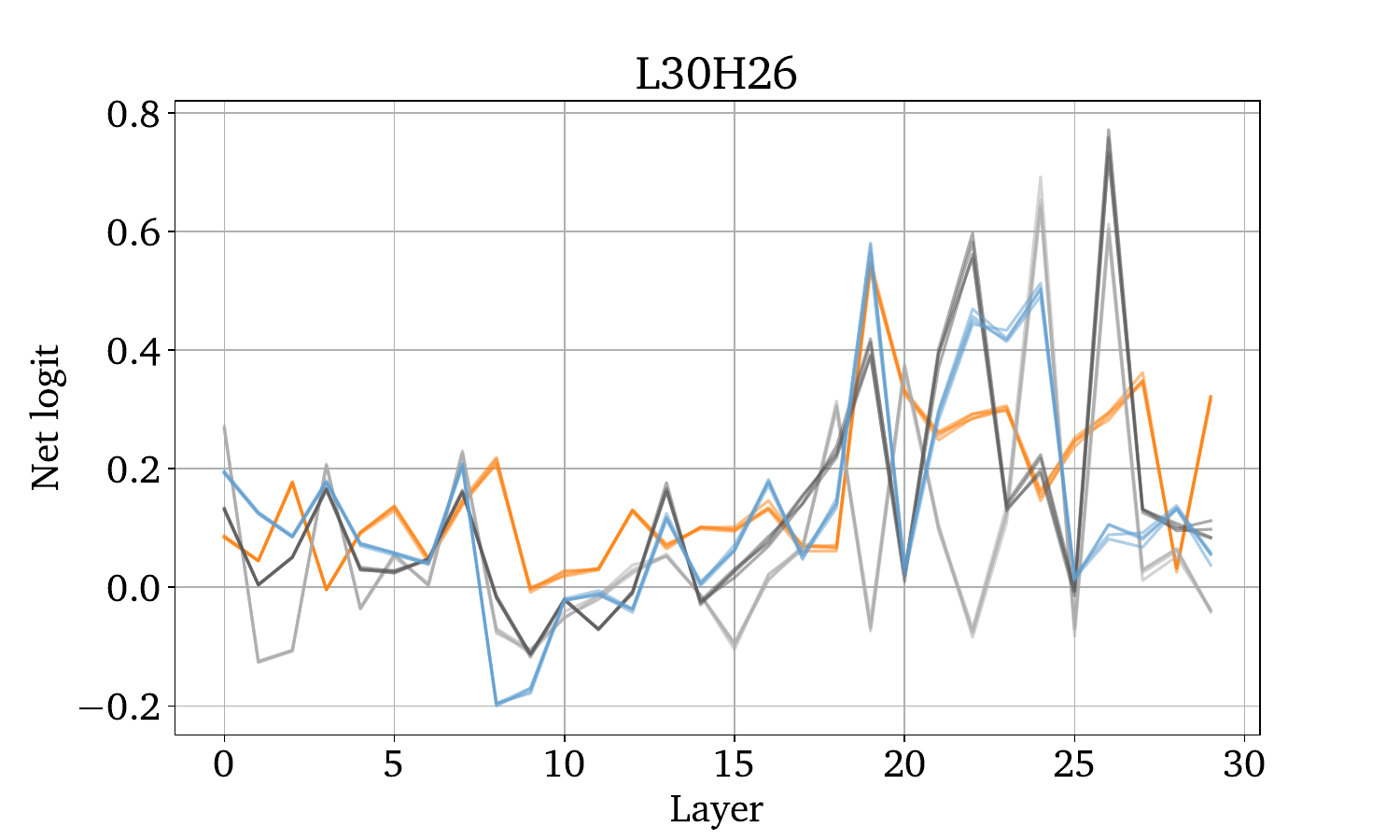}
         \caption{}
     \end{subfigure}
     \hfill
     \begin{subfigure}[b]{0.495\textwidth}
         \centering
         \includegraphics[width=\textwidth]{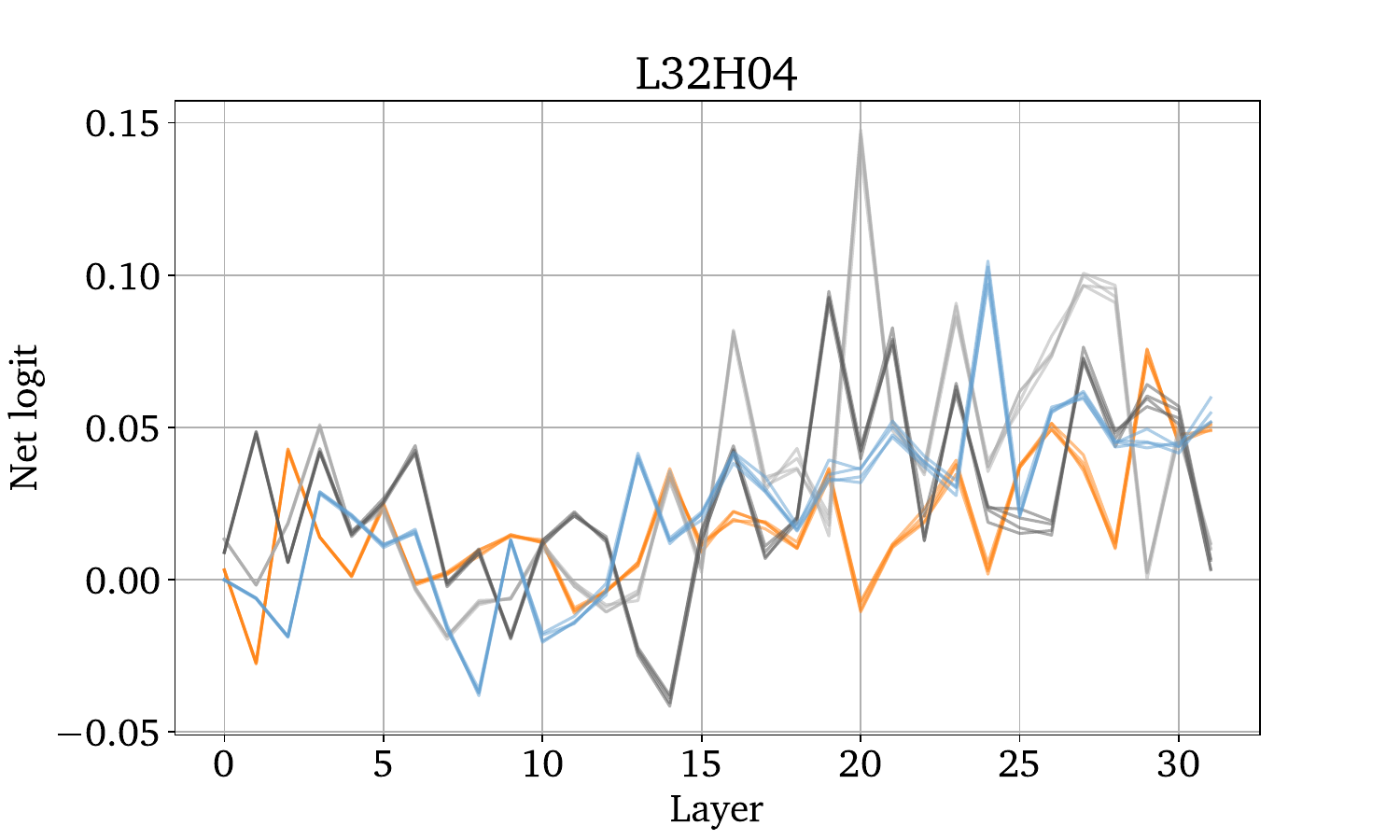}
         \caption{}
     \end{subfigure}
     \hfill
     \begin{subfigure}[b]{0.495\textwidth}
         \centering
         \includegraphics[width=\textwidth]{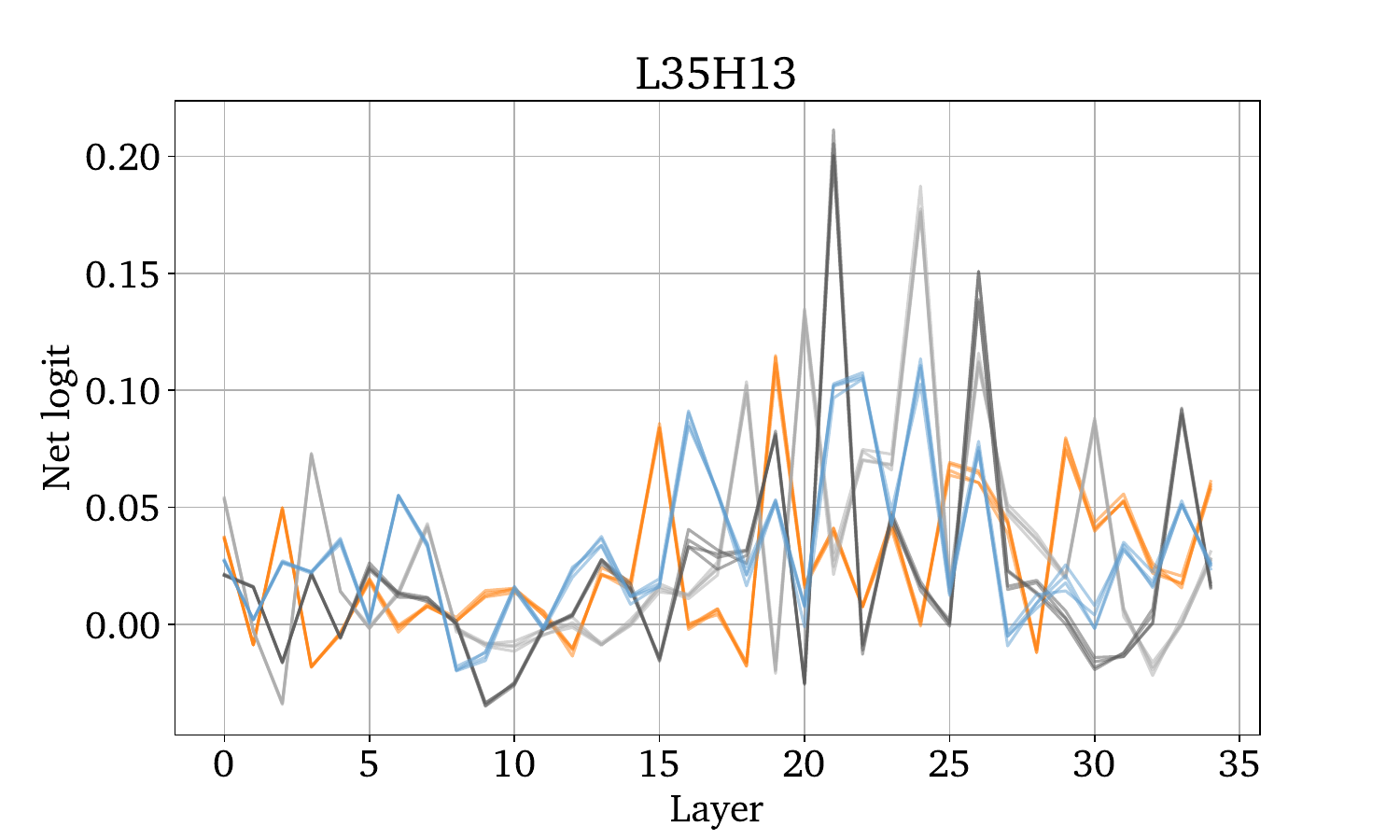}
         \caption{}
     \end{subfigure}
     \hfill
     \begin{subfigure}[b]{0.495\textwidth}
         \centering
         \includegraphics[width=\textwidth]{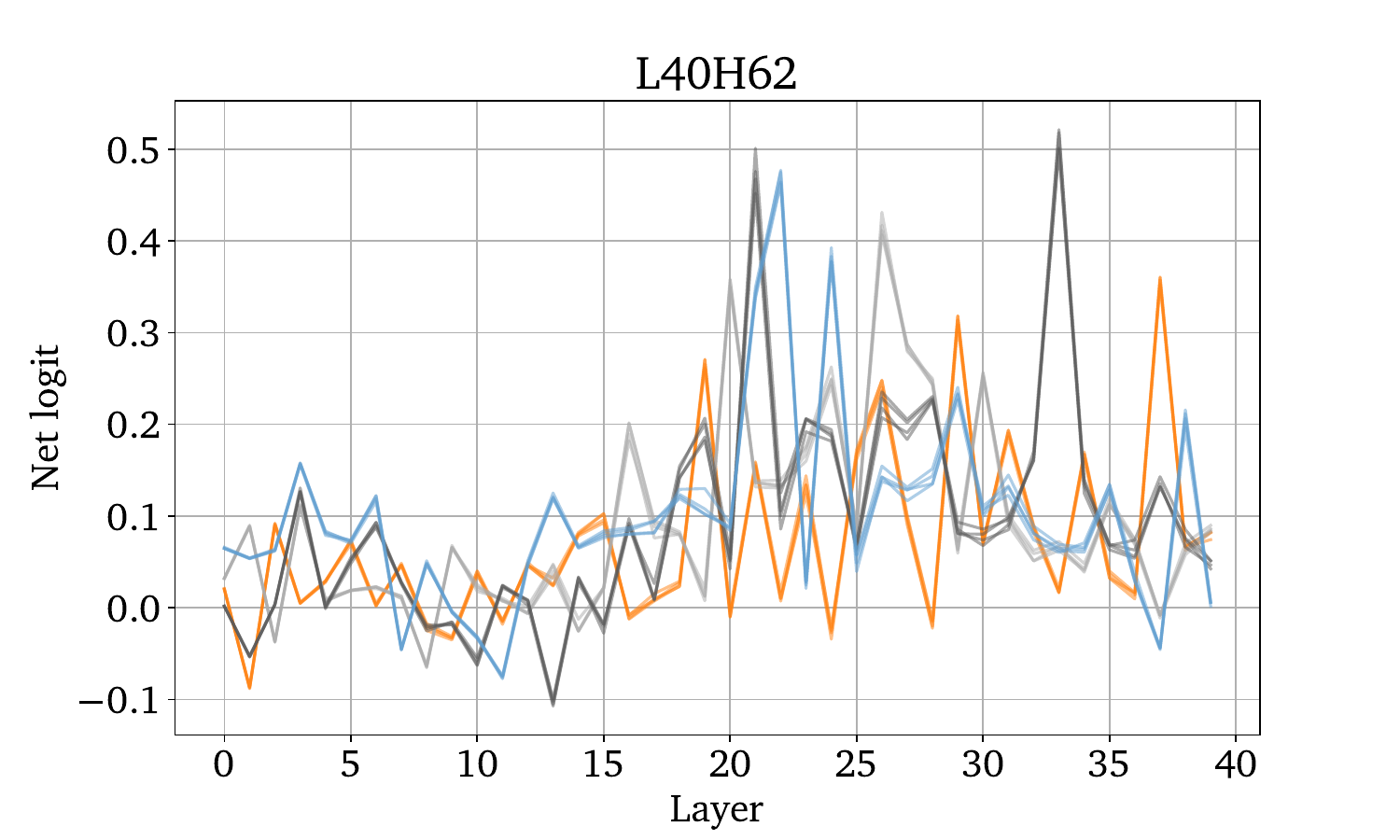}
         \caption{}
     \end{subfigure}
    \caption{Direct effect of MLP's outputs at the label positions \emph{mediated} via the correct letter heads. Different colors denote different token positions. Different lines of the same color denote different settings, i.e. different correct labels, showing that there is little difference in behavior regardless of the correct label.}
    \label{fig:unembed_mlp}
\end{figure}

\begin{figure}[t]
    \centering
     \begin{subfigure}[b]{0.495\textwidth}
         \centering
         \includegraphics[width=\textwidth]{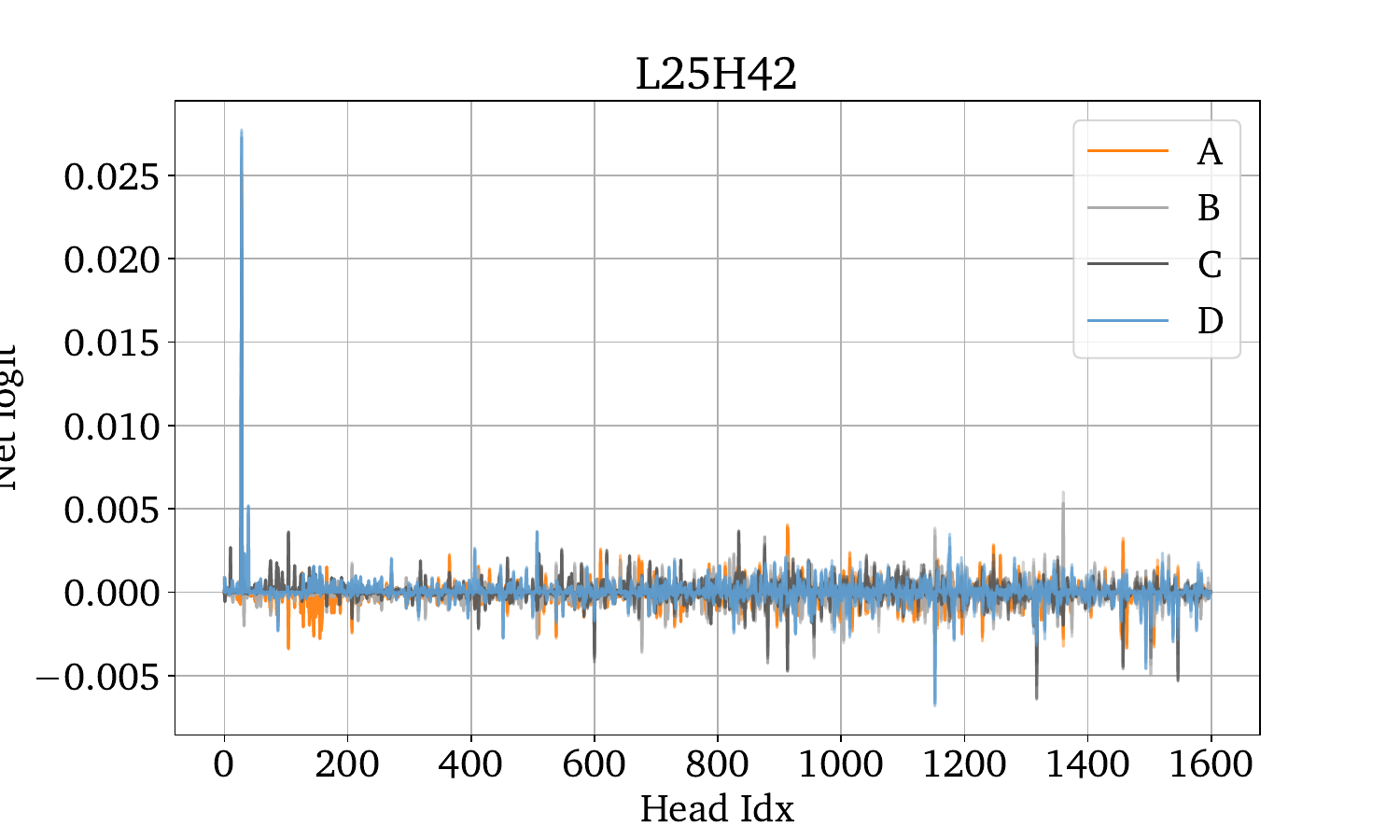}
         \caption{}
     \end{subfigure}
     \hfill
     \begin{subfigure}[b]{0.495\textwidth}
         \centering
         \includegraphics[width=\textwidth]{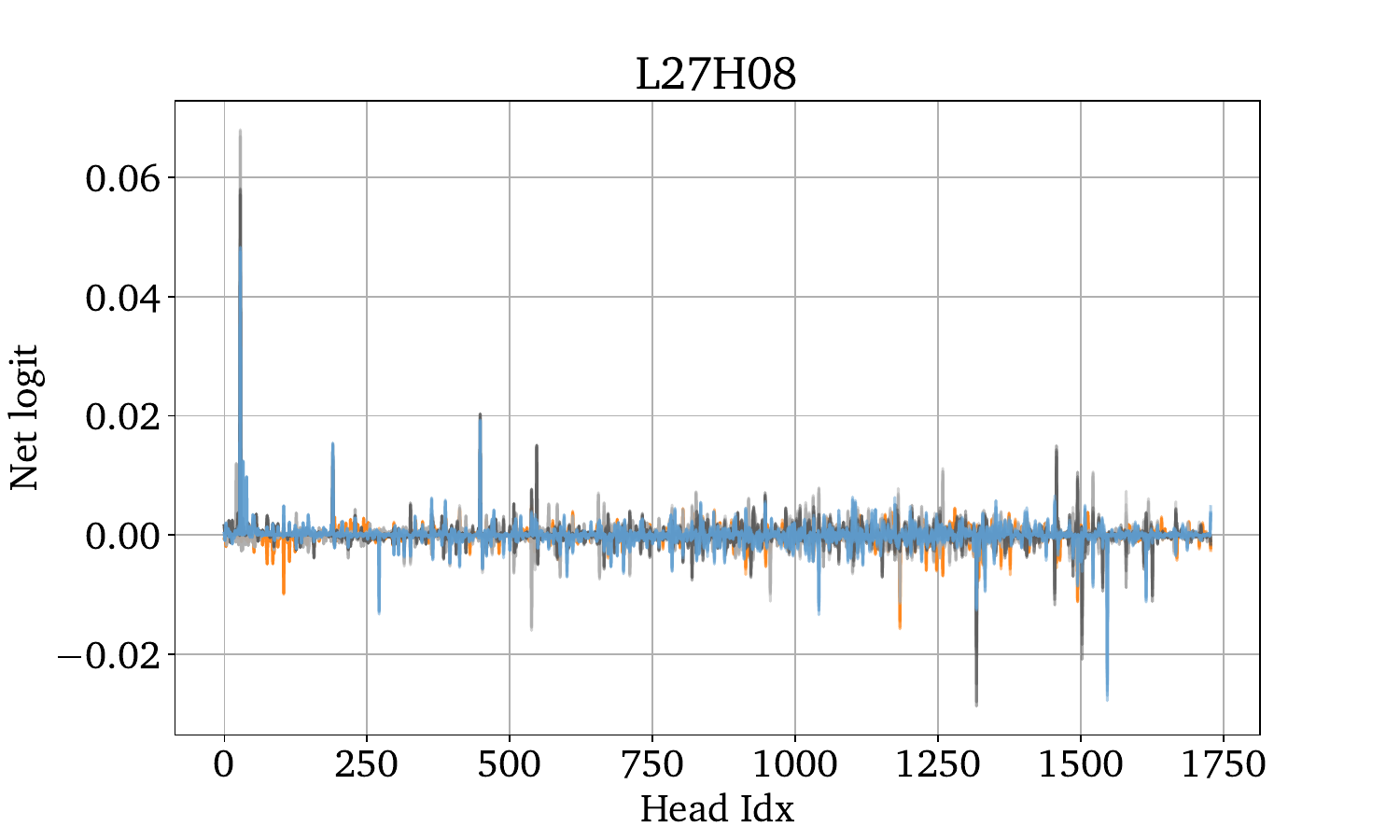}
         \caption{}
     \end{subfigure}
     \hfill
     \begin{subfigure}[b]{0.495\textwidth}
         \centering
         \includegraphics[width=\textwidth]{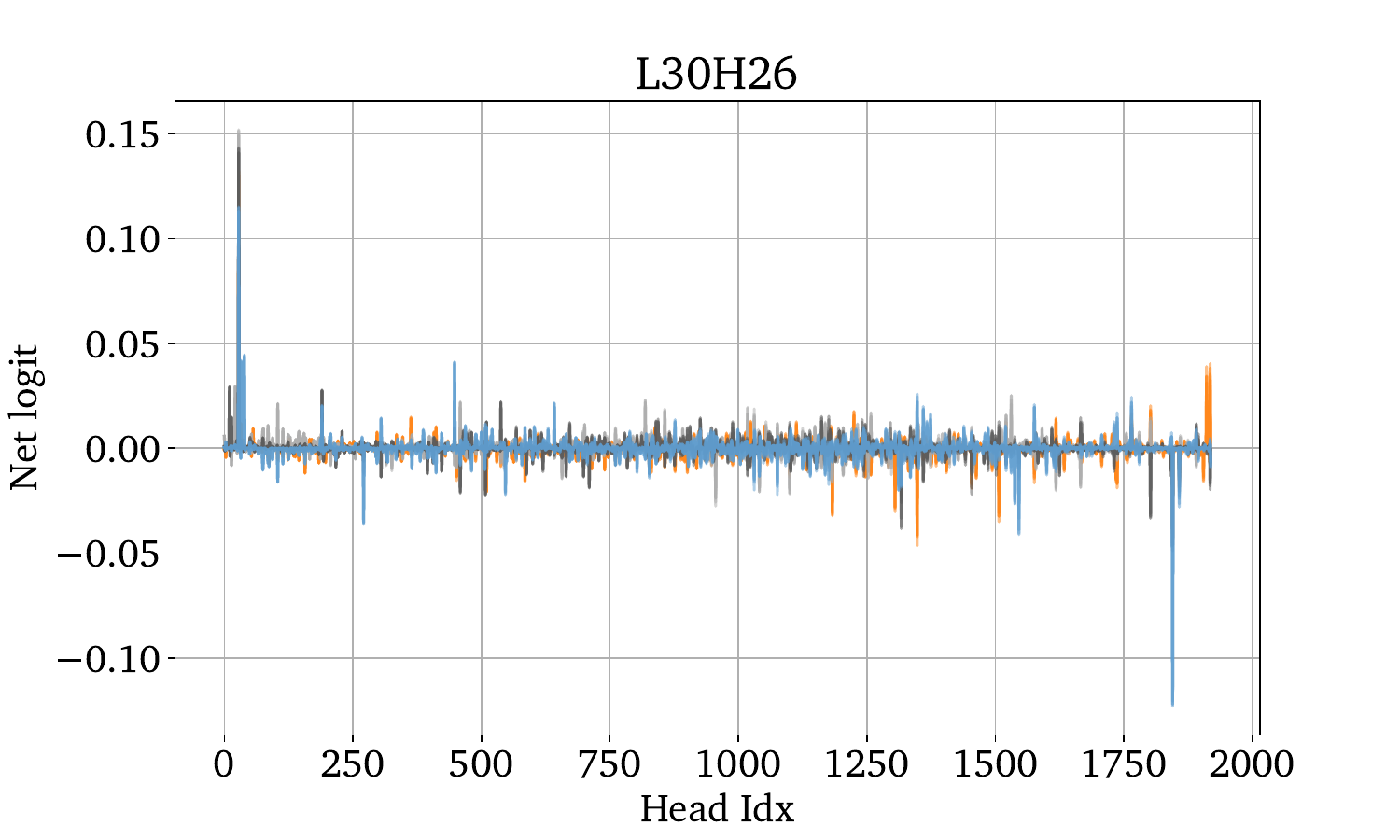}
         \caption{}
     \end{subfigure}
     \hfill
     \begin{subfigure}[b]{0.495\textwidth}
         \centering
         \includegraphics[width=\textwidth]{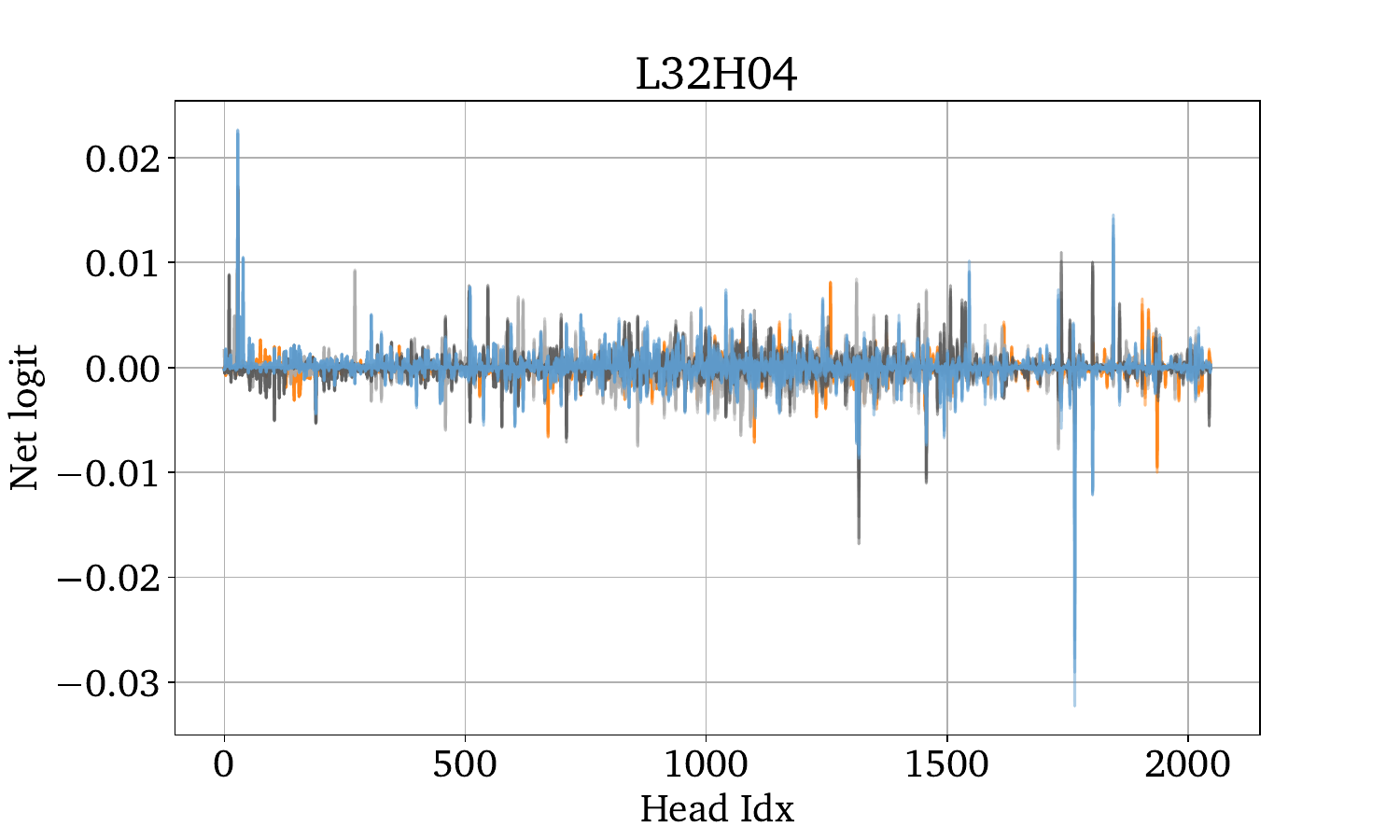}
         \caption{}
     \end{subfigure}
     \hfill
     \begin{subfigure}[b]{0.495\textwidth}
         \centering
         \includegraphics[width=\textwidth]{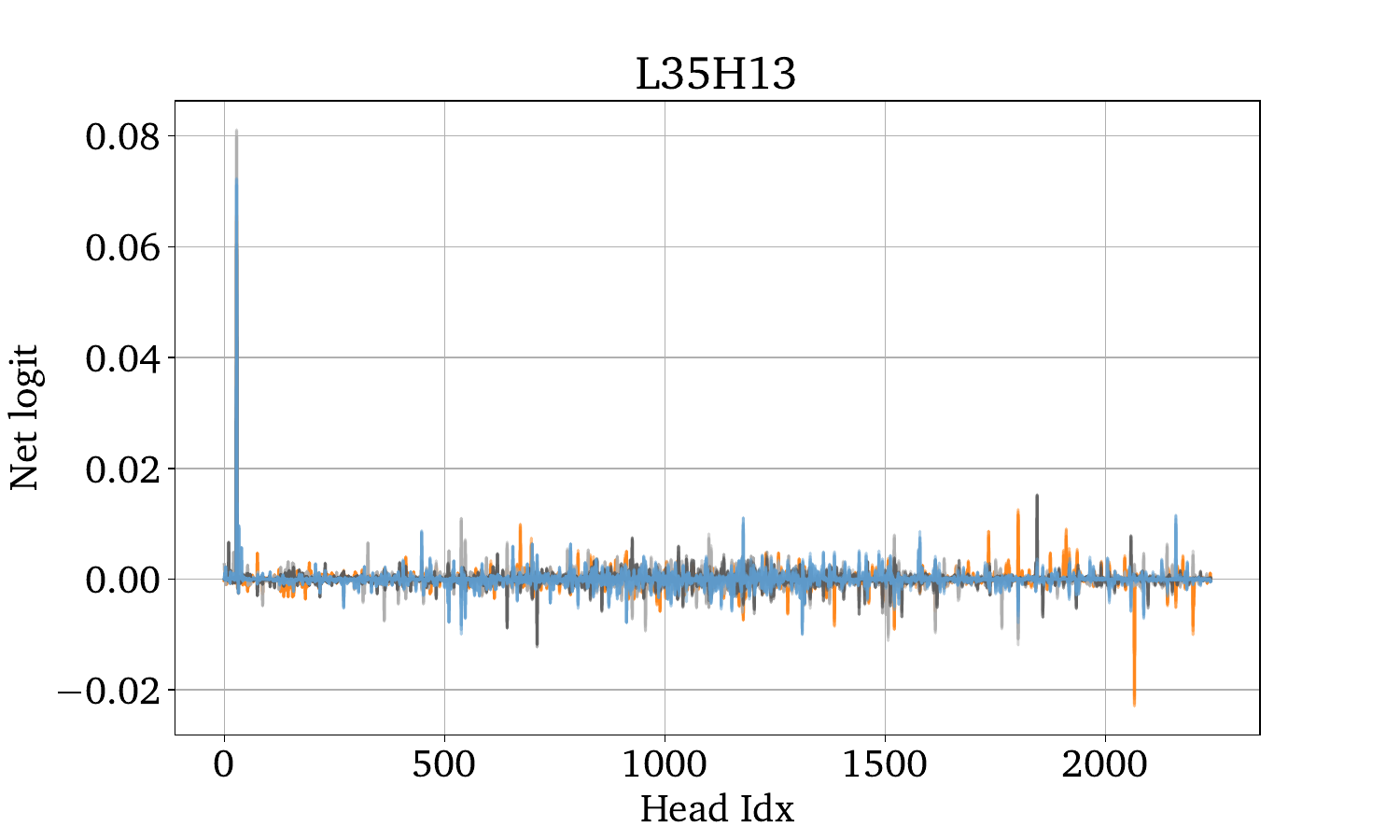}
         \caption{}
     \end{subfigure}
     \hfill
     \begin{subfigure}[b]{0.495\textwidth}
         \centering
         \includegraphics[width=\textwidth]{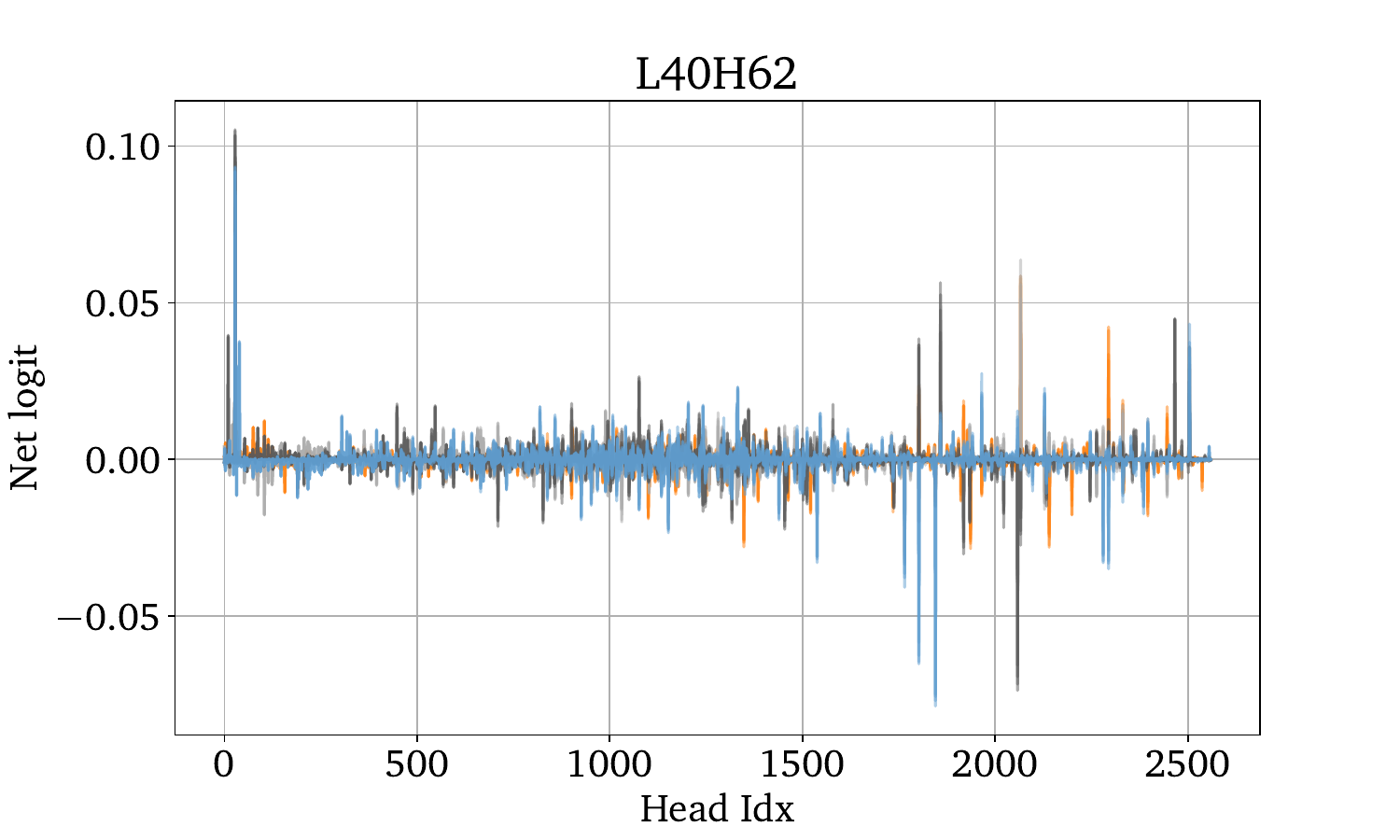}
         \caption{}
     \end{subfigure}
    \caption{Direct effect of attention heads' outputs at the label positions \emph{mediated} via the correct letter heads. Different colors denote different token positions. Different lines of the same color denote different settings, i.e. different correct labels.}
    \label{fig:unembed_head}
\end{figure}

We furthermore apply the composed unembedding function to the outputs of individual attention MLPs and heads at the label positions, displayed in~\cref{fig:unembed_head,fig:unembed_mlp} respectively. Comparing the scales of these contributions we observe that a) only MLPs seem to matter significantly (mostly after layer 15), with the exception of perhaps the very first attention layer, b) the behavior of the MLPs is indifferent to the correct label, and c) contributions vary between correct letter heads, meaning that different correct letter heads do not fully share the subspace from which they read the information about the token identity.

\subsection{Open Questions}

The above discoveries naturally lead to more questions about the circuit, which we did not investigate. For example, how is the content aggregated into the final content token? How is the query for the content gatherers formed?  How do the symbol binding MLPs `know' that they should be reinforcing the token identity? Are the keys of the correct letter heads at the label positions formed via the same process and the same set of MLPs as the values, or does it work via a different mechanism? Do the single letter and uniform heads use the same process to form their values at the label position?

One hypothesis is that the query of the content gatherer is a compressed representation of the original question, whereas each answer content is independently aggregated, such that the content gatherers need only match the question to the content resembling it, but at this point this is merely speculation.

\clearpage

\newpage
\section{Nuances in identifying output nodes}
\label{sec:nuances_identifying_output}

In \cref{sec:finding_nodes}, we describe our methodology for finding `output' nodes -- those that contribute directly to the output logits -- based on unembedding the contributions to the residual stream from each node. However, there is an important caveat to this method, which we detail in this appendix.

Specifically, identifying output nodes based on only the \emph{direct} effect of each node fails to account for the fact that later nodes might cancel out the effects of prior nodes. Indeed, we already see some evidence for this in~\cref{sec:finding_nodes}. Identifying output nodes based only on high direct effect therefore has a chance of producing false-positives: there may some some nodes that have high direct effect but which are prevented from actually contributing to the logits due to interference from later nodes.

To guard against this problem, one should therefore apply additional filtering to the tentative list of output nodes identified based on direct effect. In particular, one should filter the list based on nodes which also have high \emph{total} effect. 

\begin{figure}[t]
    \centering
    \includegraphics[width=\textwidth]{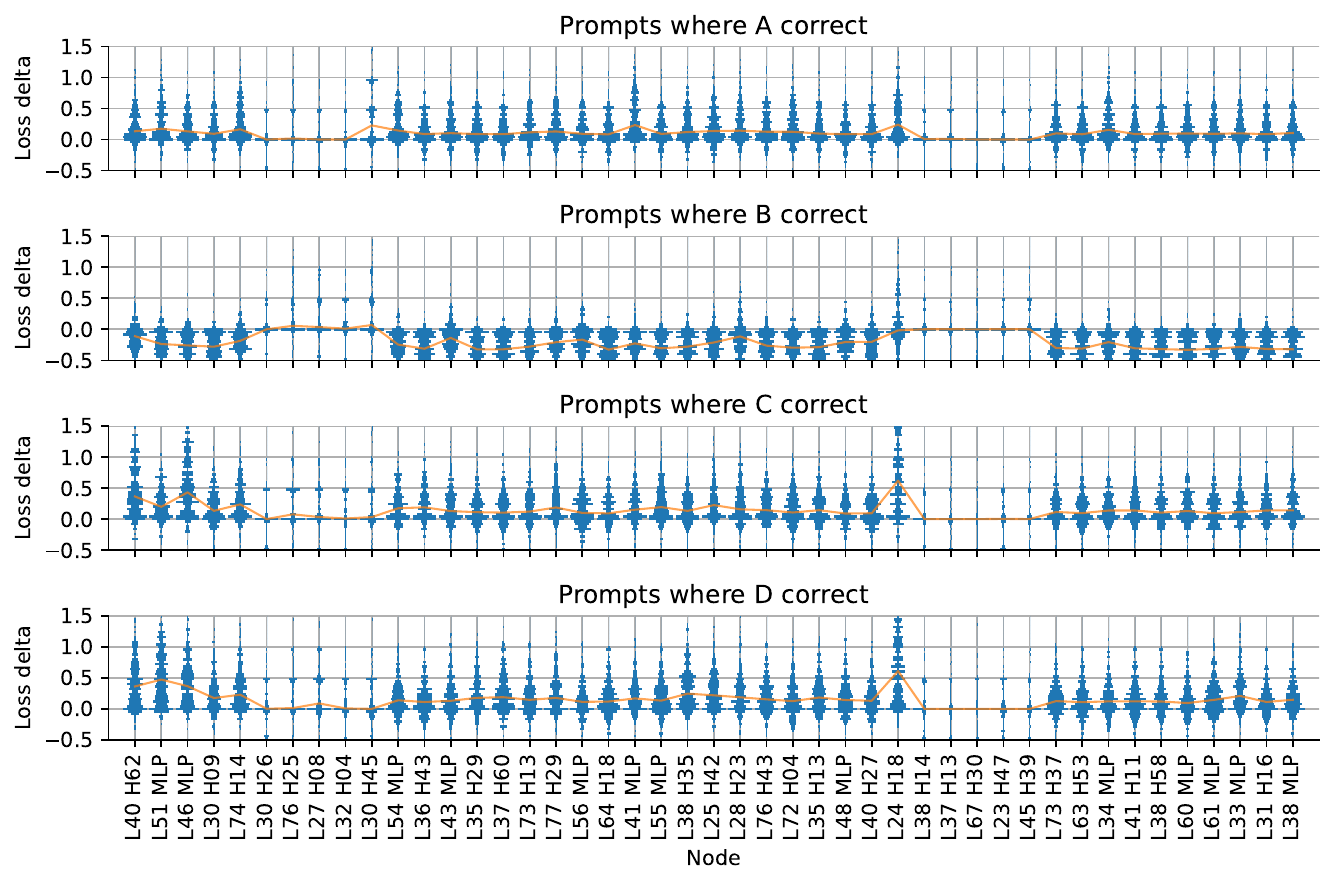}
    \caption{Violin plot of total effects across 1024 prompts of the 45 nodes with highest direct effect identified in \cref{sec:finding_nodes}. Orange line indicates median total effect for each node. Note that each violin is normalised separately, and in contrast to standard violin plots in which density is smoothed, for precision, here we visualise the distribution over 50 discrete bins between -0.5 and 1.5. Key features of interest include a) the high degree of variability in total effect for each node, and b) the fact that almost all nodes show a negative total effect for prompts where B is correct. See text for more details.}
    \label{fig:total_effects_violin}
\end{figure}

However, when we evaluated the total effects of the 45 nodes we identified as having high direct effect, we encountered a number of surprises, shown in~\cref{fig:total_effects_violin}.

\begin{itemize}
    \item \textbf{High variance}. The spread over total effects we see for each node is large enough to complicate the process of ruling out nodes based on low total effect. For example, even nodes like L30 H26 which have a median total effect of zero still have a significant total effect in \emph{some} prompts.
    \item \textbf{Strange behaviour on prompts where B is correct}. Almost all nodes have a negative total effect on these prompts. We're not sure why this is.
\end{itemize}

Due to these complications, we decided against further filtering based on total effects.

\clearpage

\newpage
\section{Net Direct Effect By Letter}
\label{sec:net_direct_effect_facetted}

\begin{figure}[ht]
    \includegraphics[width=\textwidth]{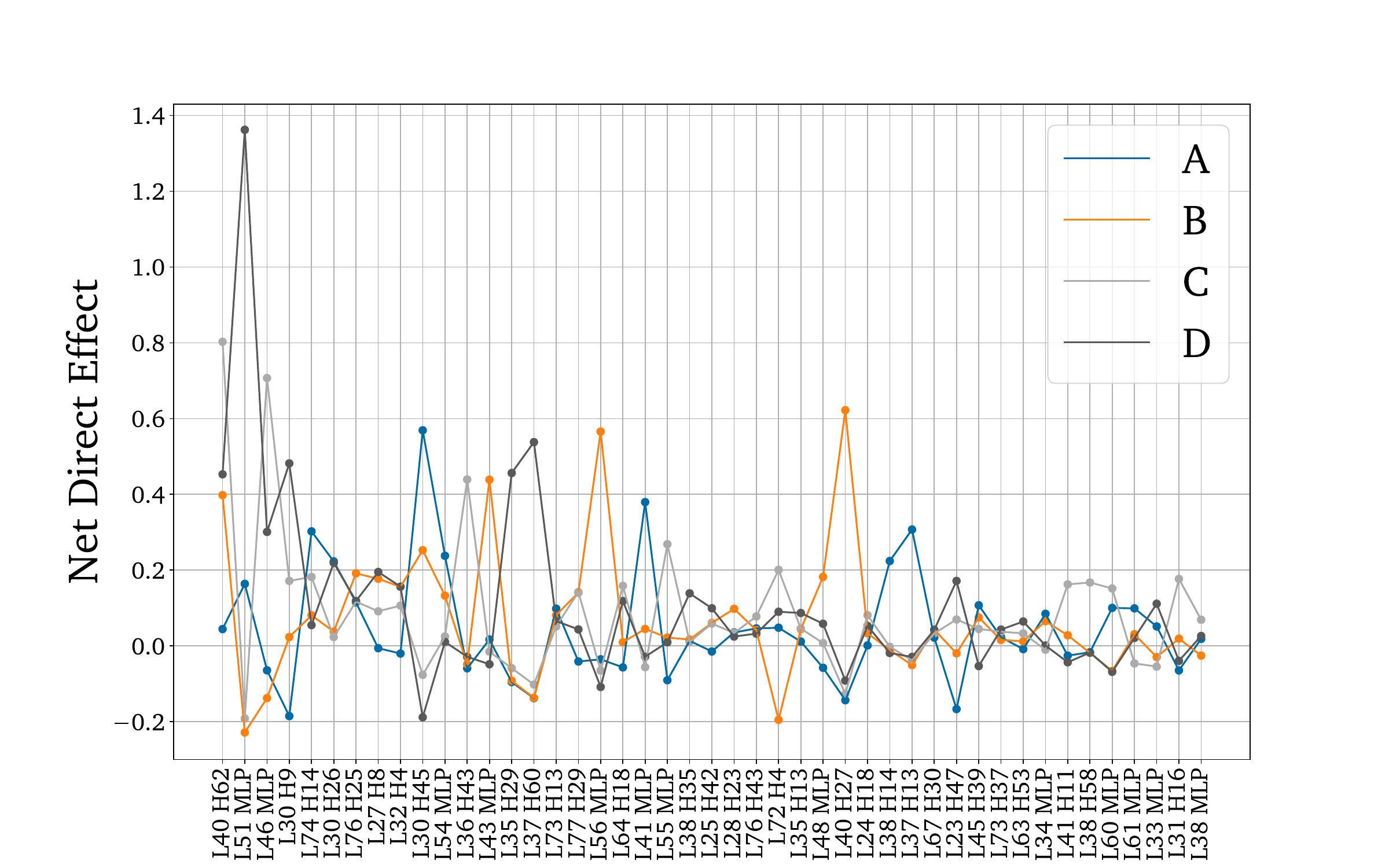}
    \caption{}
    \label{fig:net_direct_effect_all_letters}
\end{figure}

\begin{figure}[ht]
    \includegraphics[width=\textwidth]{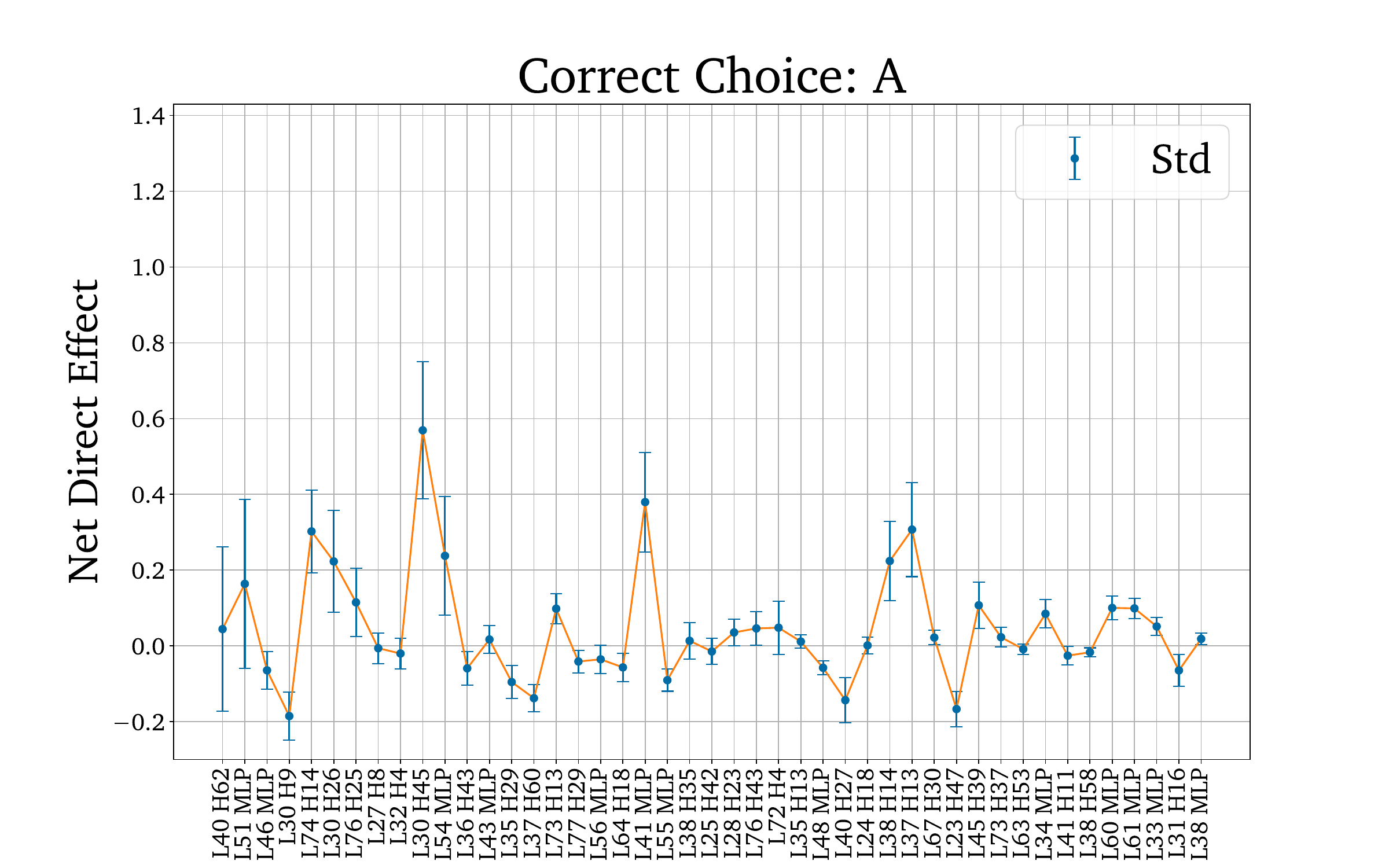}
    \caption{}
    \label{fig:net_direct_effect_A}
\end{figure}

\begin{figure}[ht]
    \includegraphics[width=\textwidth]{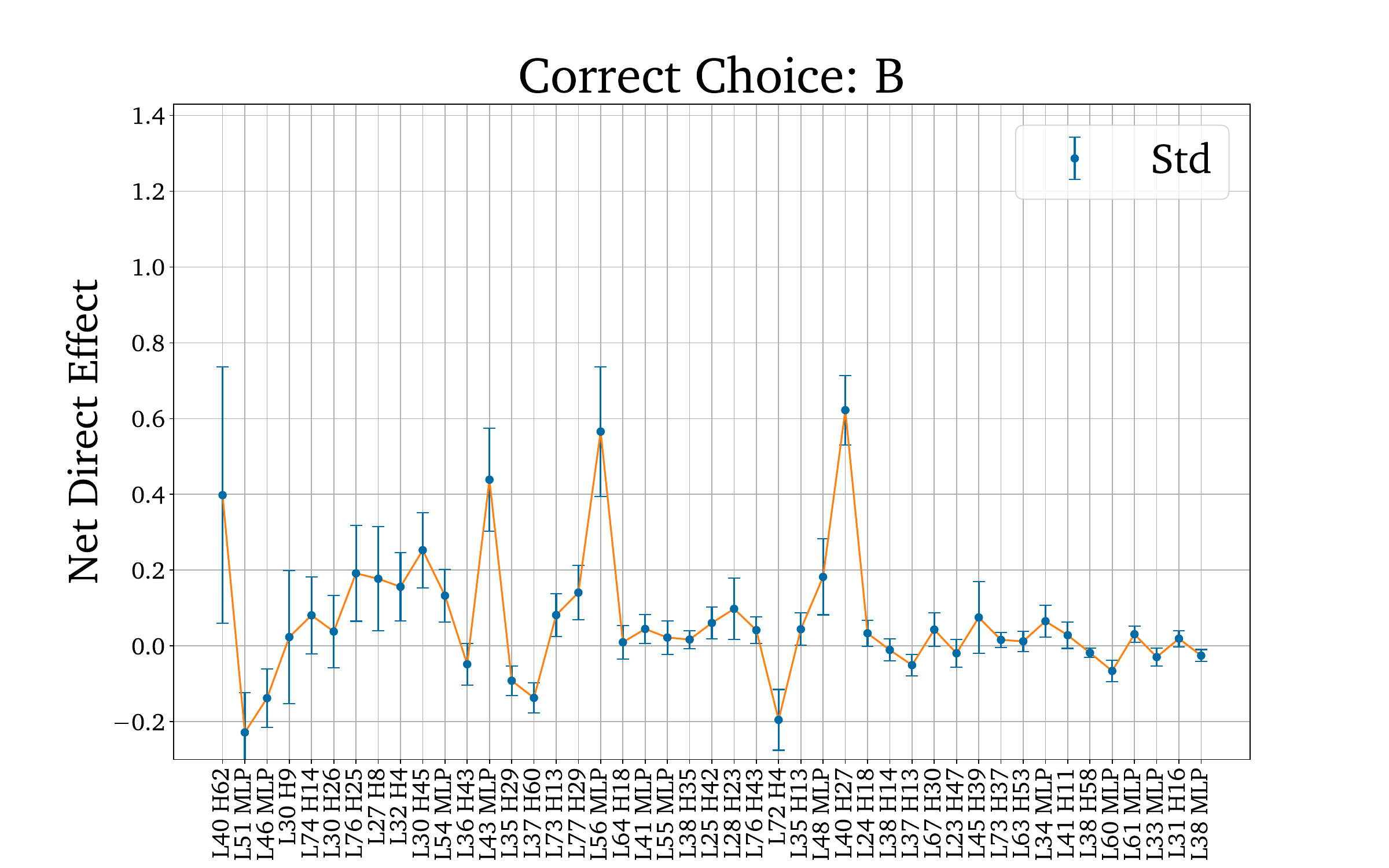}
    \caption{}
    \label{fig:net_direct_effect_B}
\end{figure}

\begin{figure}[ht]
    \includegraphics[width=\textwidth]{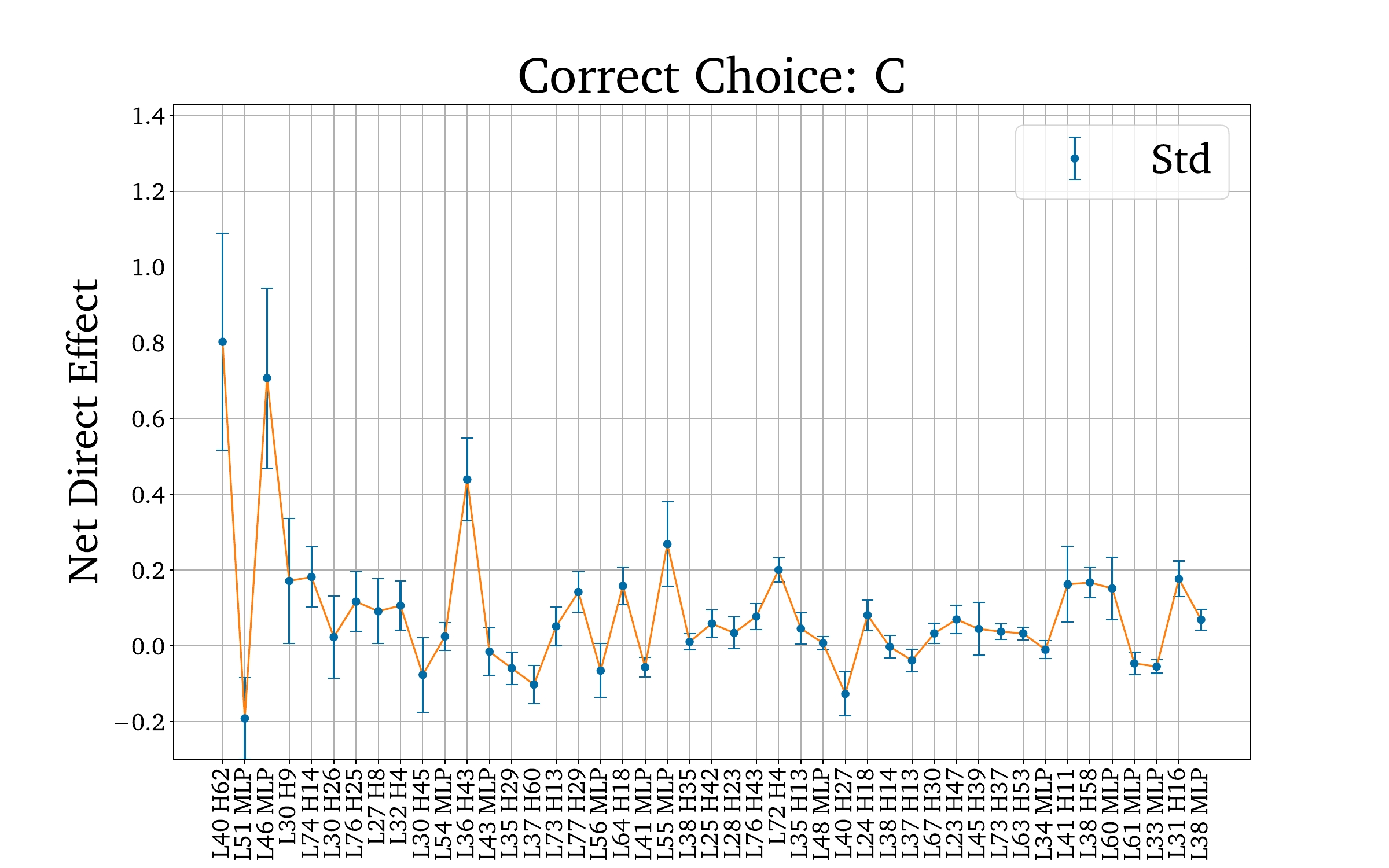}
    \caption{}
    \label{fig:net_direct_effect_C}
\end{figure}

\begin{figure}[ht]
    \includegraphics[width=\textwidth]{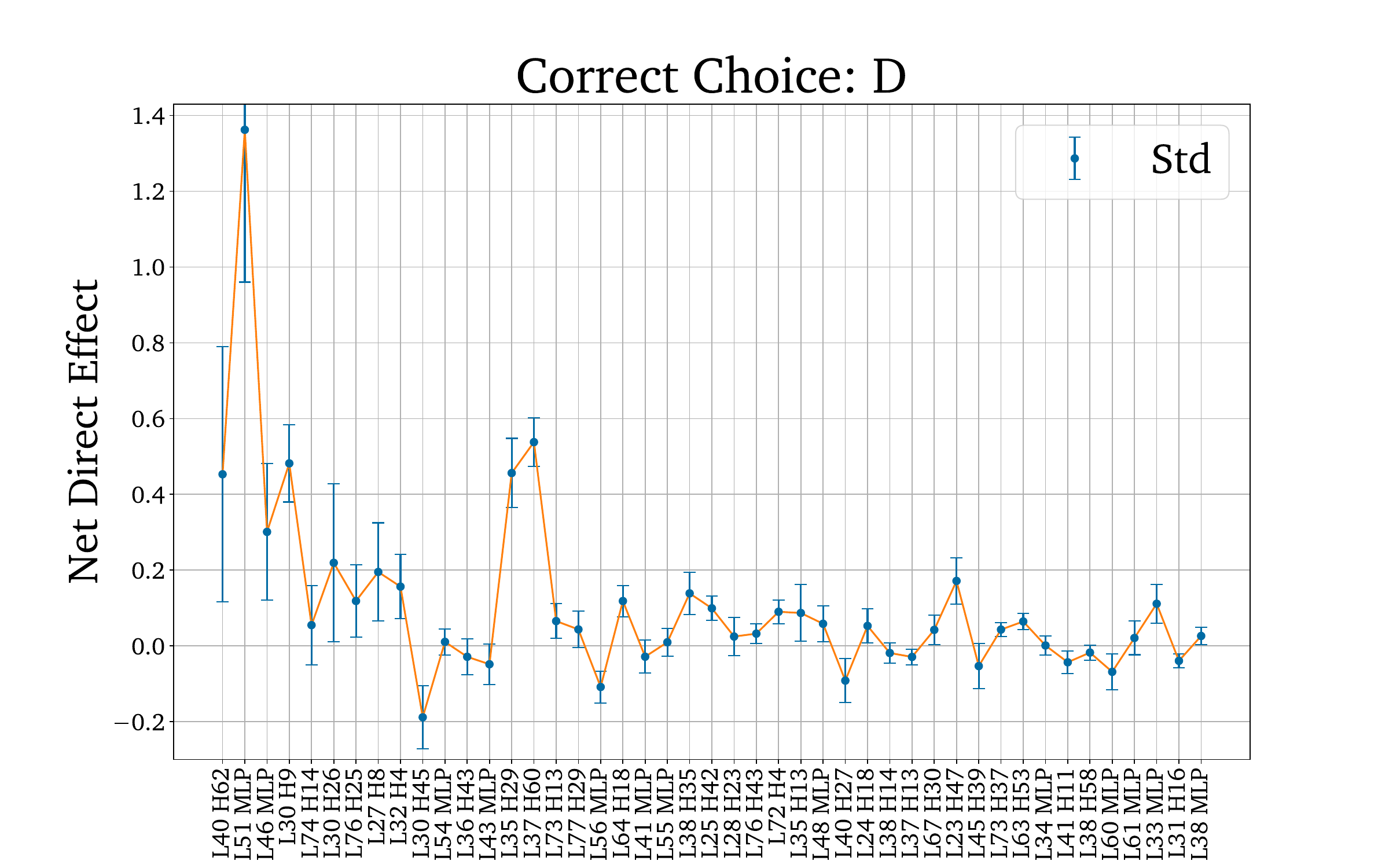}
    \caption{}
    \label{fig:net_direct_effect_D}
\end{figure}

In~\cref{fig:net_direct_effect}, we report the aggregate net direct effect of each node. We did find however large variance between settings (correct letters). We report the results facetted by setting in~\cref{fig:net_direct_effect_all_letters,fig:net_direct_effect_A,fig:net_direct_effect_B,fig:net_direct_effect_C,fig:net_direct_effect_D}.

\clearpage

\newpage
\section{Classification of Output Heads}
\label{sec:classification_output_heads}

Here we show the average attention pattern of the heads with high direct effect. To identify whether attending to a position actually confers information, we report the product of the attention probability and the L2 norm of the value vector at that position. We report the value-weighted attention on the prelude, the label tokens and the final tokens. For the remaining positions we report the maximum in the column "OTHER" in each plot. 

Note that the only two heads which have a meaningful contribution from other token positions are \texttt{L24 H18} and \texttt{L28 H23}, which we identified as a content gatherer in~\cref{sec:rest_of_the_circuit}.

\begin{figure}
    \centering
     \begin{subfigure}[b]{0.49\textwidth}
         \centering
         \includegraphics[width=\textwidth]{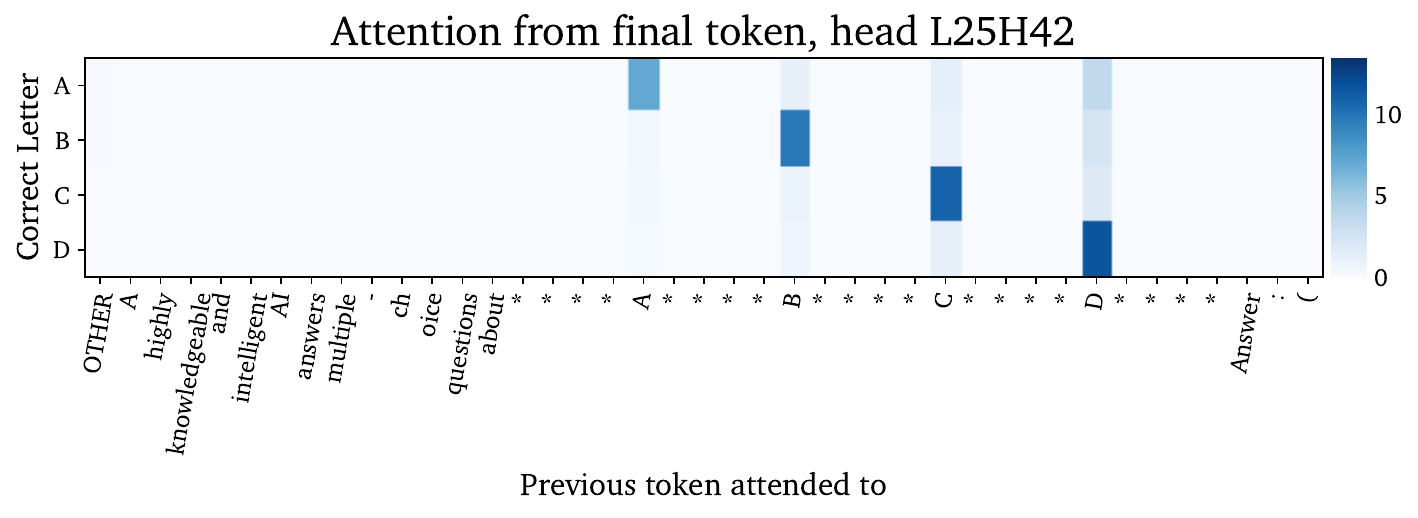}
         \caption{}
     \end{subfigure}
     \hfill
     \begin{subfigure}[b]{0.49\textwidth}
         \centering
         \includegraphics[width=\textwidth]{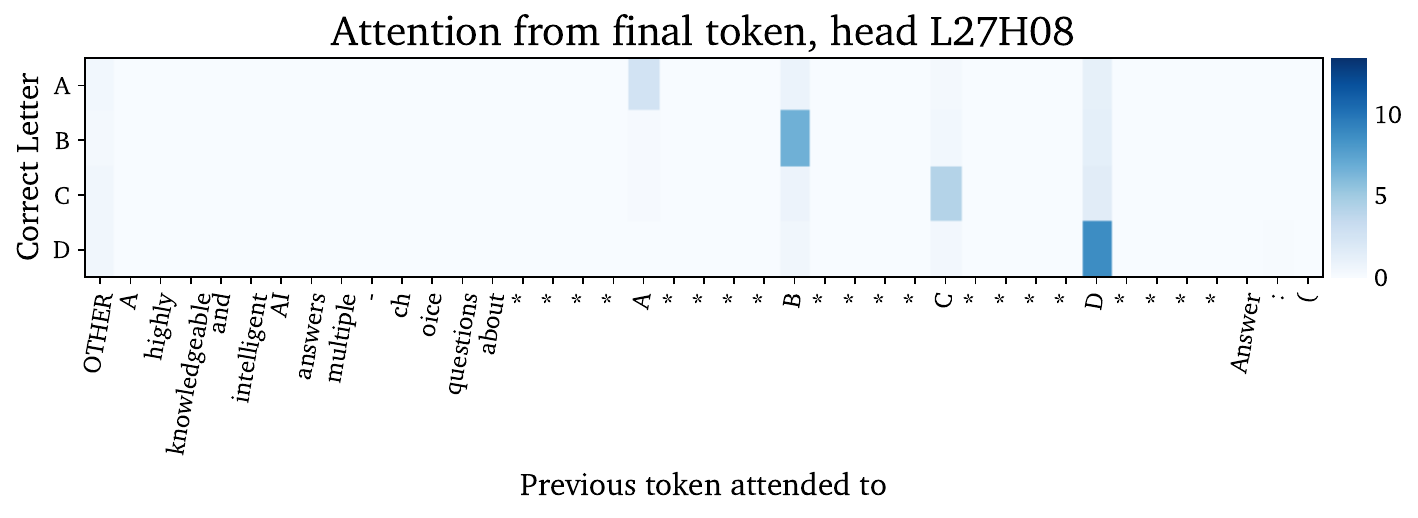}
         \caption{}
     \end{subfigure}
     \hfill
     \begin{subfigure}[b]{0.49\textwidth}
         \centering
         \includegraphics[width=\textwidth]{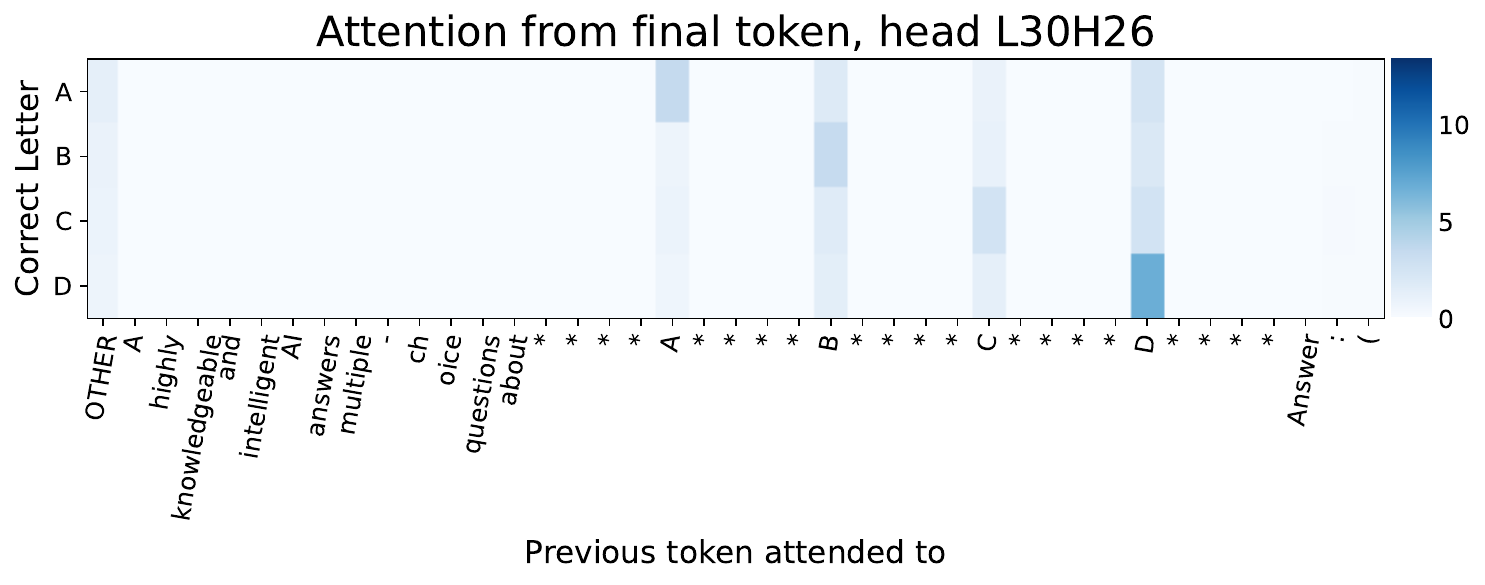}
         \caption{}
     \end{subfigure}
     \hfill
     \begin{subfigure}[b]{0.49\textwidth}
         \centering
         \includegraphics[width=\textwidth]{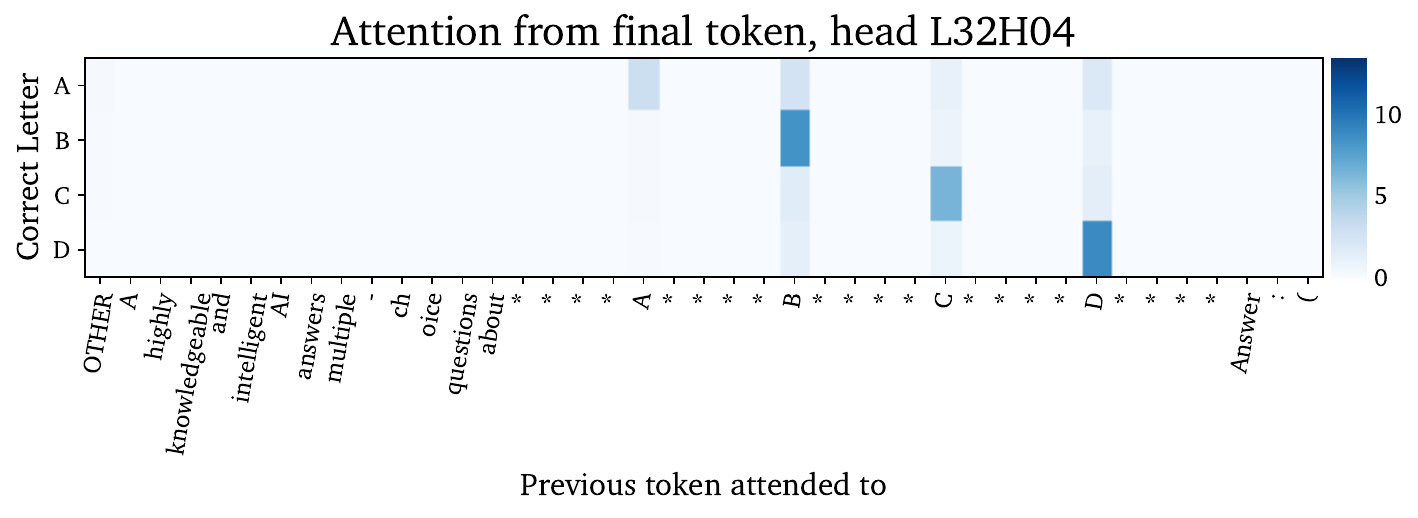}
         \caption{}
     \end{subfigure}
     \hfill
     \begin{subfigure}[b]{0.49\textwidth}
         \centering
         \includegraphics[width=\textwidth]{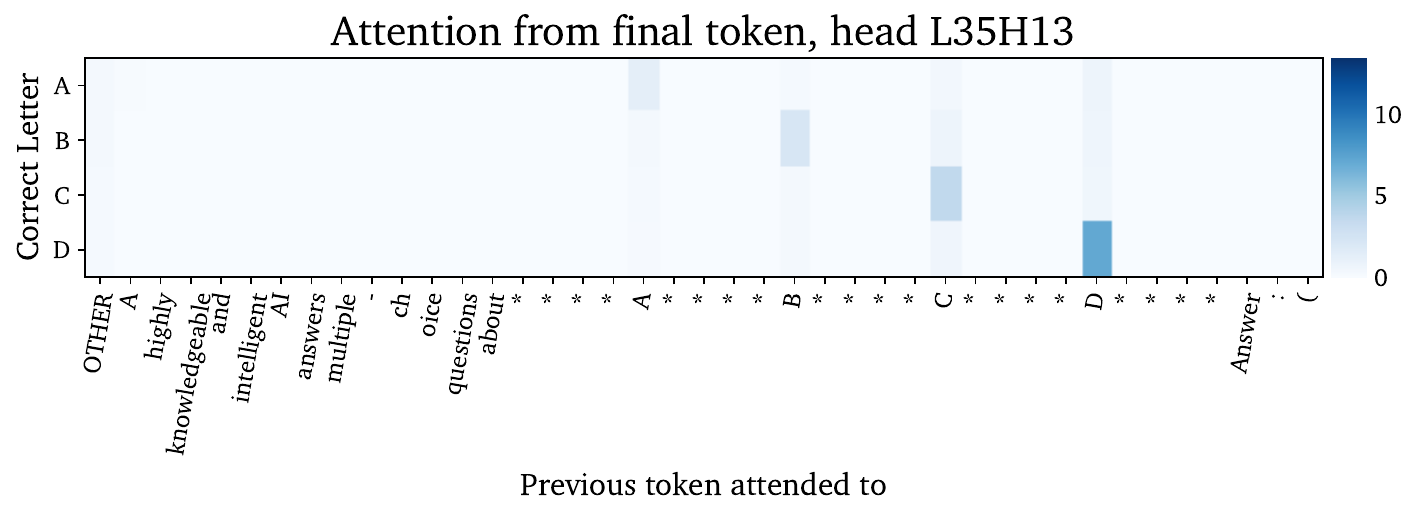}
         \caption{}
     \end{subfigure}
     \hfill
     \begin{subfigure}[b]{0.49\textwidth}
         \centering
         \includegraphics[width=\textwidth]{figures/output_heads_attention/correct_letter/attention_L40_H62.pdf}
         \caption{}
     \end{subfigure}
    \caption{Correct Letter heads}
    \label{fig:correct_letter_heads_val_attn}
\end{figure}

\begin{figure}
    \centering
     \begin{subfigure}[b]{0.49\textwidth}
         \centering
         \includegraphics[width=\textwidth]{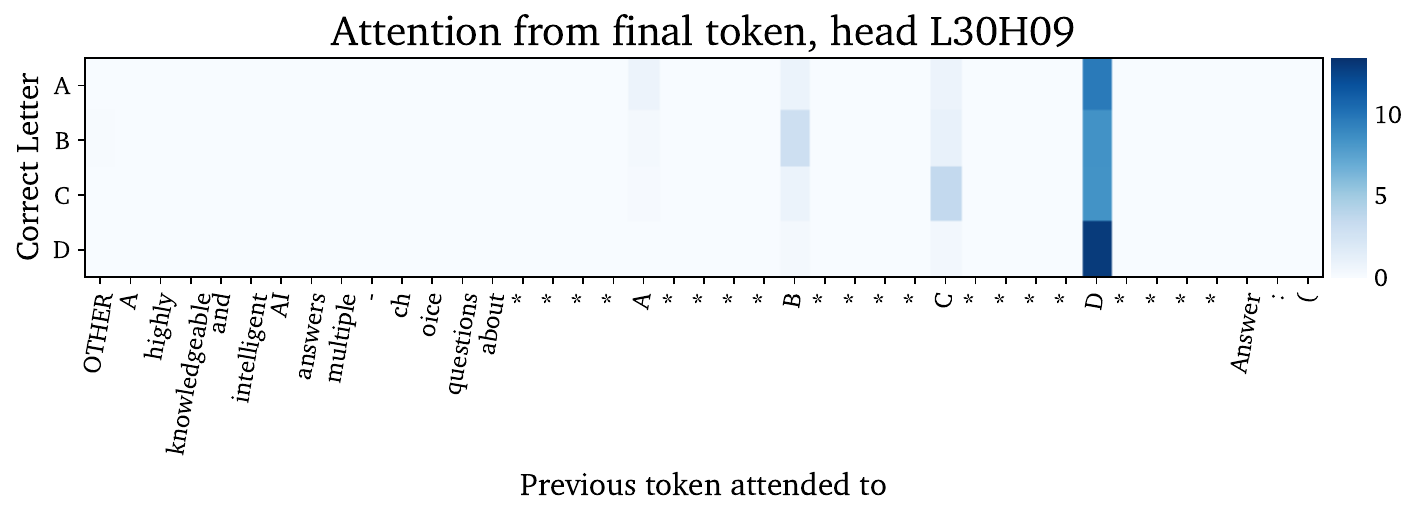}
         \caption{}
     \end{subfigure}
     \hfill
     \begin{subfigure}[b]{0.49\textwidth}
         \centering
         \includegraphics[width=\textwidth]{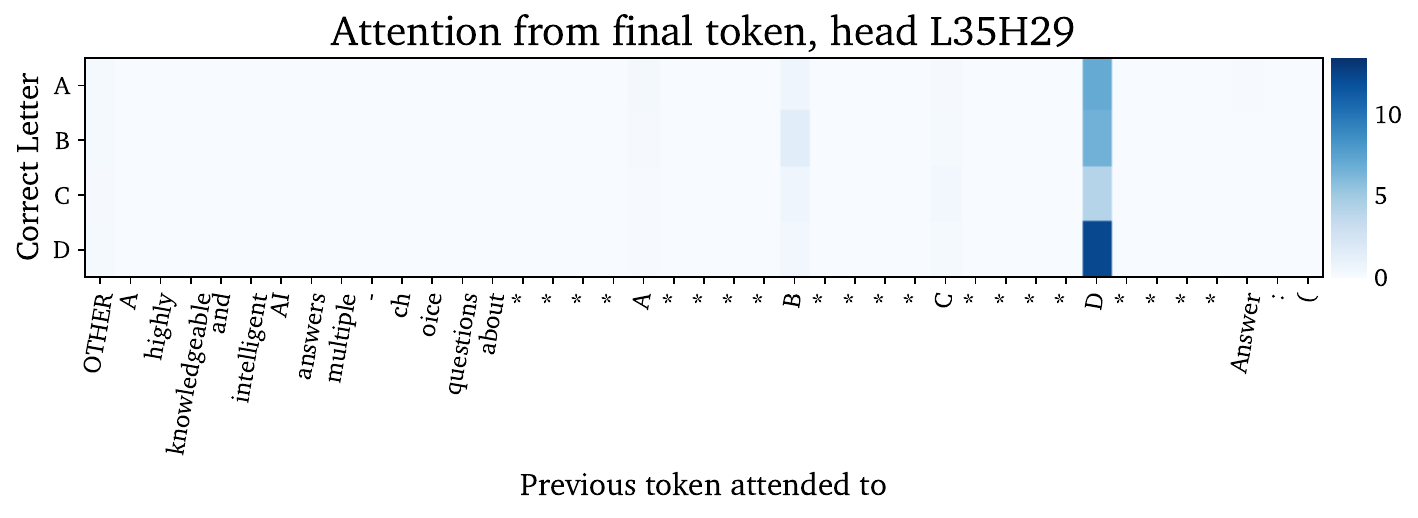}
         \caption{}
     \end{subfigure}
     \hfill
     \begin{subfigure}[b]{0.49\textwidth}
         \centering
         \includegraphics[width=\textwidth]{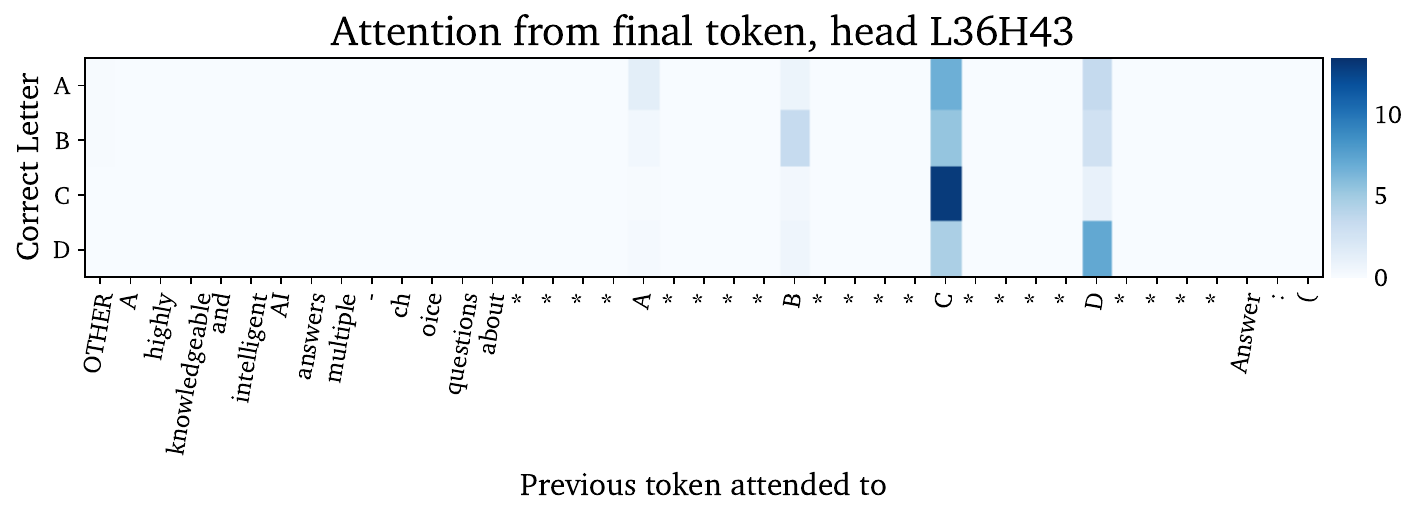}
         \caption{}
     \end{subfigure}
     \hfill
     \begin{subfigure}[b]{0.49\textwidth}
         \centering
         \includegraphics[width=\textwidth]{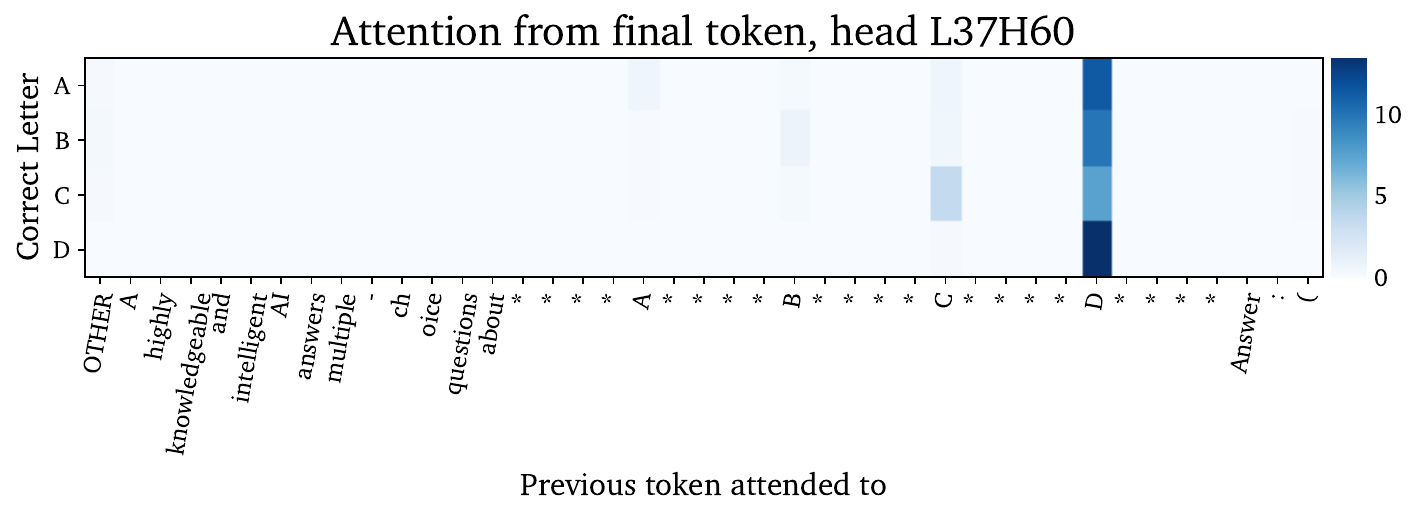}
         \caption{}
     \end{subfigure}
     \hfill
     \begin{subfigure}[b]{0.49\textwidth}
         \centering
         \includegraphics[width=\textwidth]{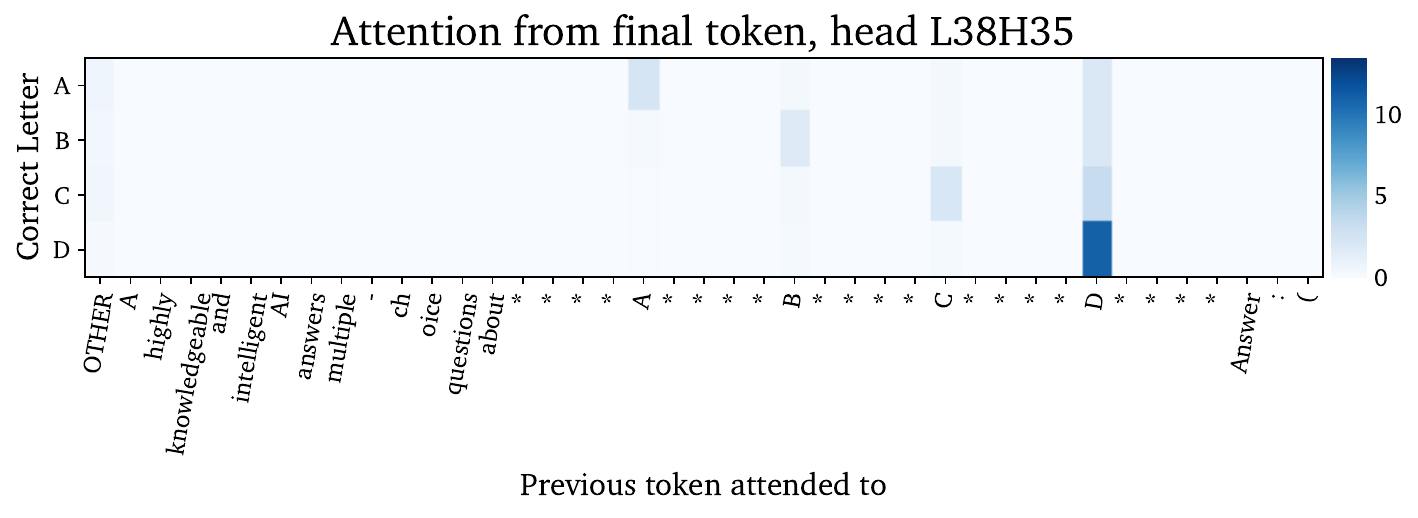}
         \caption{}
     \end{subfigure}
     \hfill
     \begin{subfigure}[b]{0.49\textwidth}
         \centering
         \includegraphics[width=\textwidth]{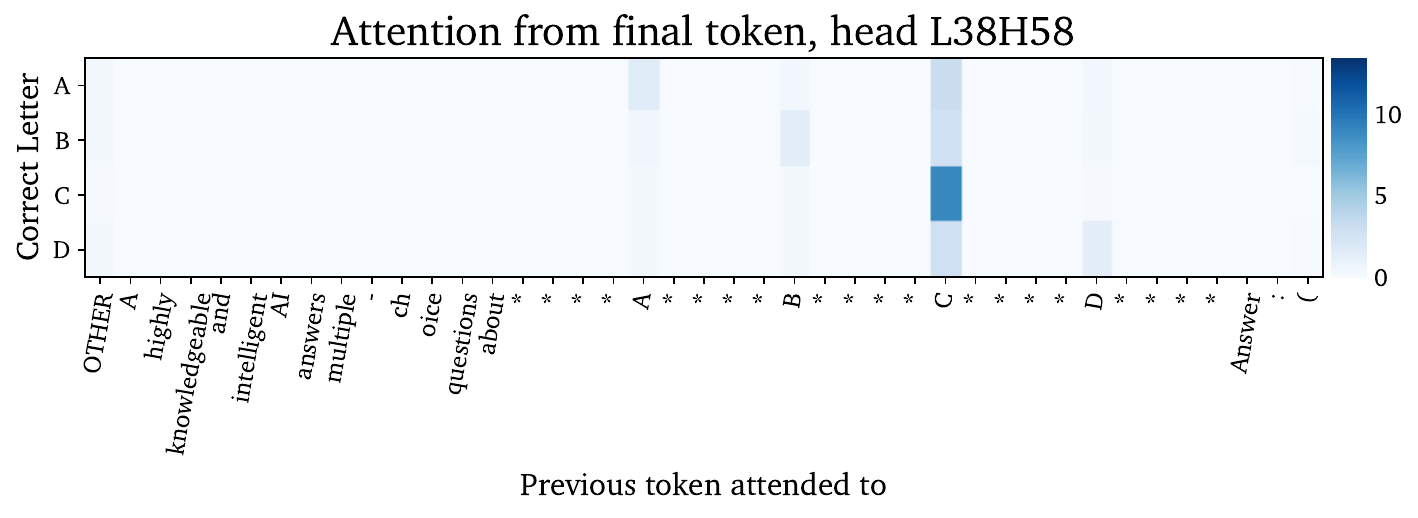}
         \caption{}
     \end{subfigure}
     \hfill
     \begin{subfigure}[b]{0.49\textwidth}
         \centering
         \includegraphics[width=\textwidth]{figures/output_heads_attention/single_letter/attention_L40_H27.pdf}
         \caption{}
     \end{subfigure}
     \hfill
     \begin{subfigure}[b]{0.49\textwidth}
         \centering
         \includegraphics[width=\textwidth]{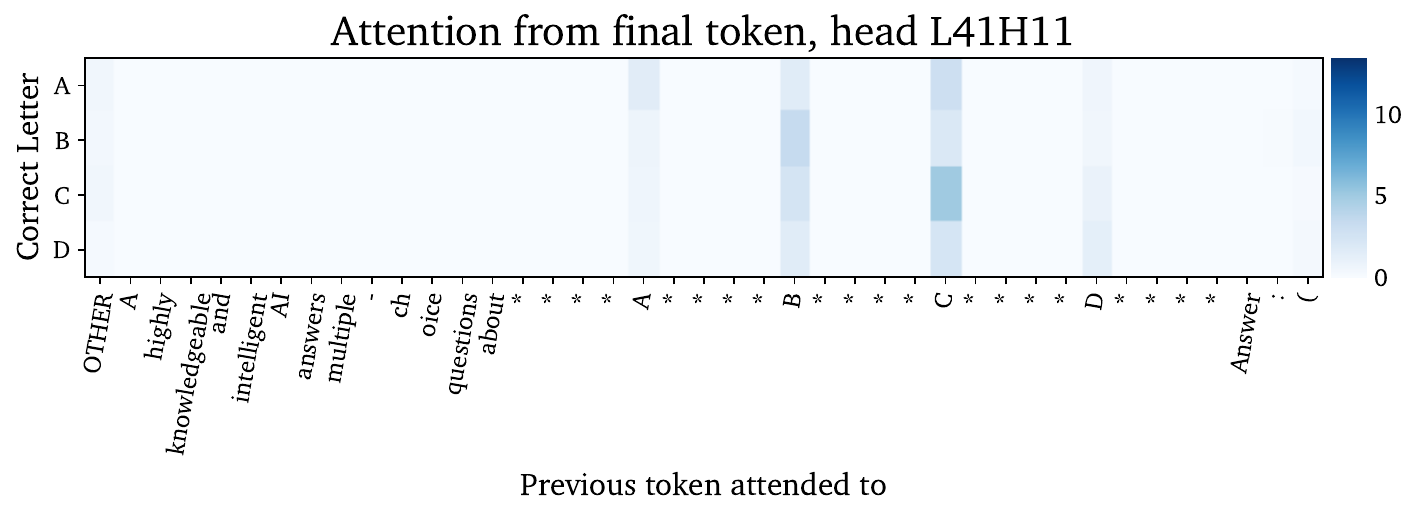}
         \caption{}
     \end{subfigure}
     \hfill
     \begin{subfigure}[b]{0.49\textwidth}
         \centering
         \includegraphics[width=\textwidth]{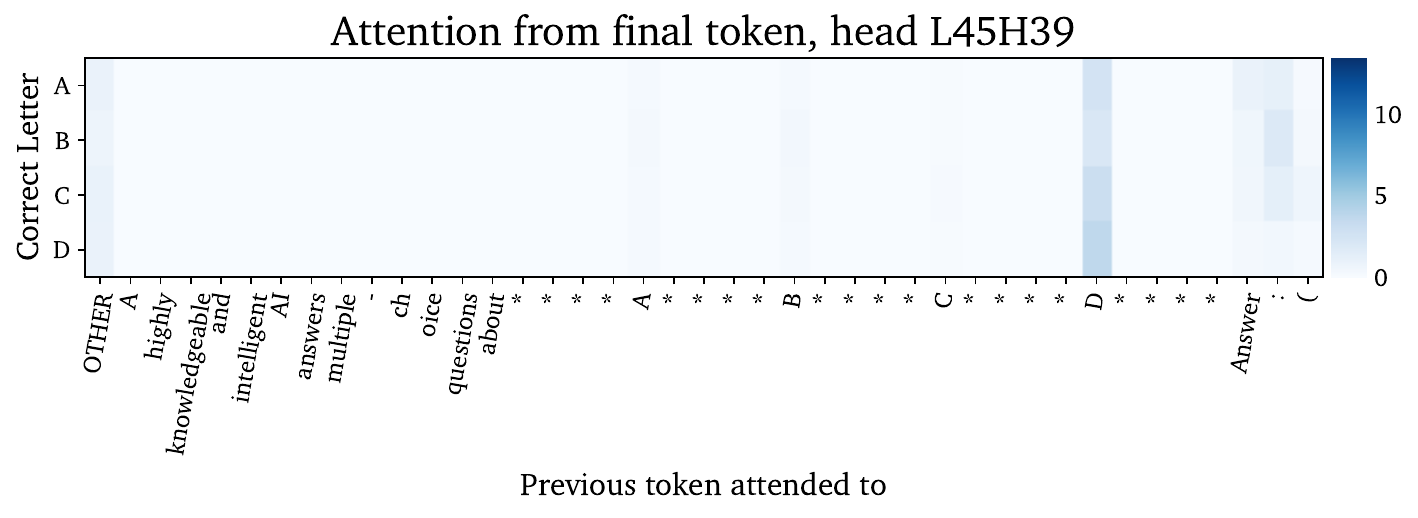}
         \caption{}
     \end{subfigure}
    \caption{Single Letter heads}
    \label{fig:single_letter_heads_val_attn}
\end{figure}

\begin{figure}
    \centering
     \begin{subfigure}[b]{0.49\textwidth}
         \centering
         \includegraphics[width=\textwidth]{figures/output_heads_attention/constant/attention_L23_H47.pdf}
         \caption{}
     \end{subfigure}
     \hfill
     \begin{subfigure}[b]{0.49\textwidth}
         \centering
         \includegraphics[width=\textwidth]{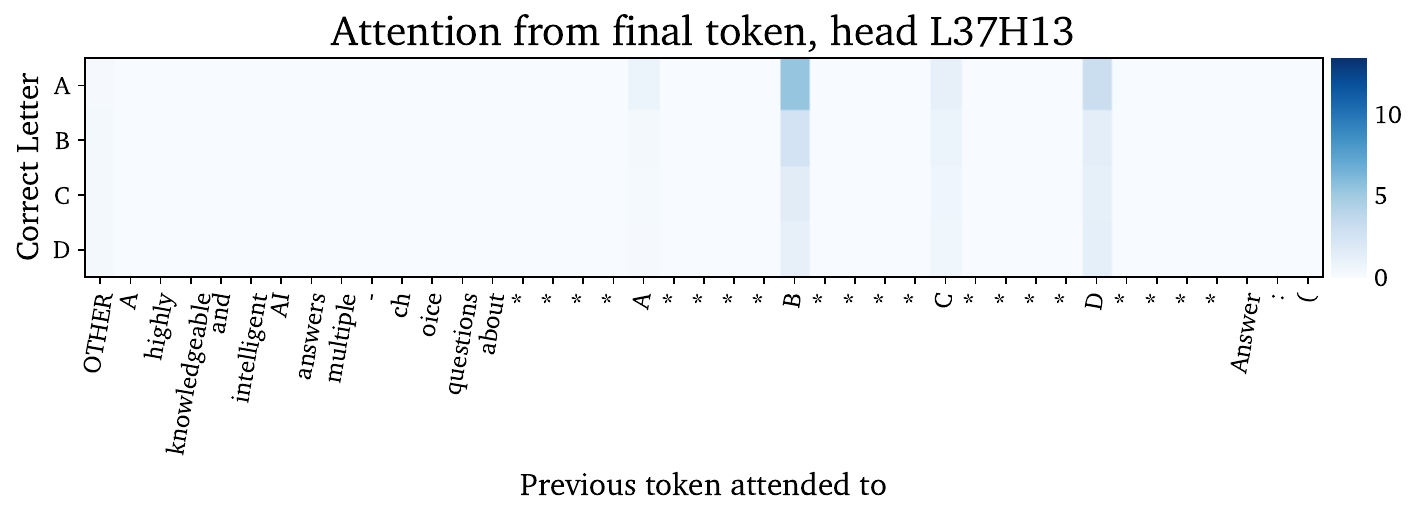}
         \caption{}
     \end{subfigure}
     \hfill
     \begin{subfigure}[b]{0.49\textwidth}
         \centering
         \includegraphics[width=\textwidth]{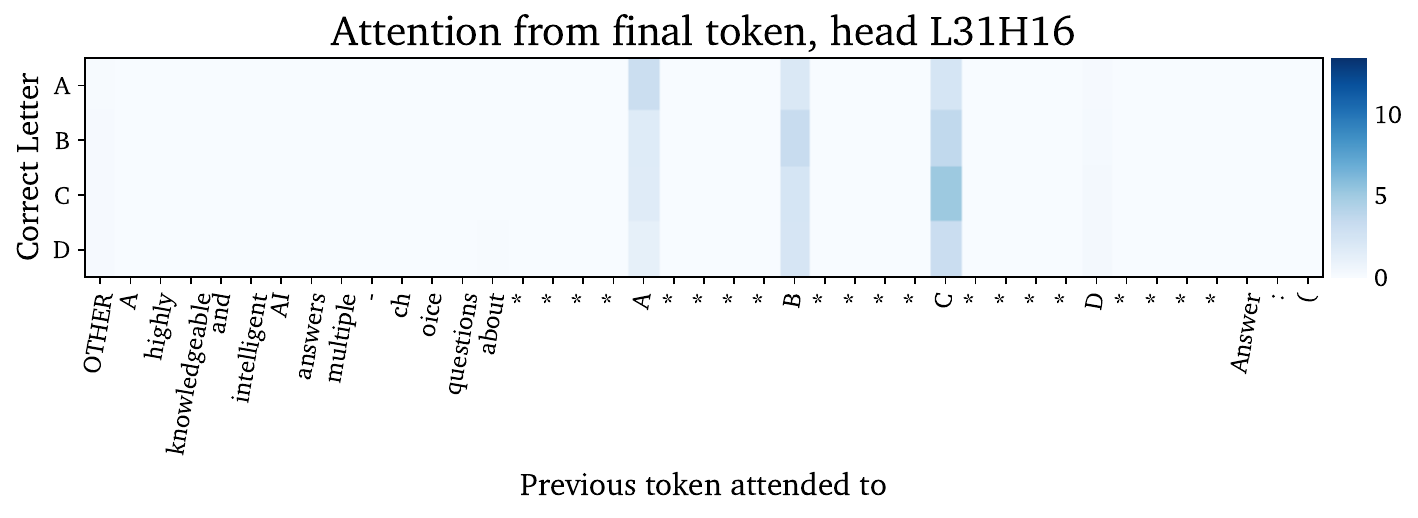}
         \caption{}
     \end{subfigure}
    \caption{Uniform heads}
    \label{fig:constant_heads_val_attn}
\end{figure}

\begin{figure}
    \centering
     \begin{subfigure}[b]{0.49\textwidth}
         \centering
         \includegraphics[width=\textwidth]{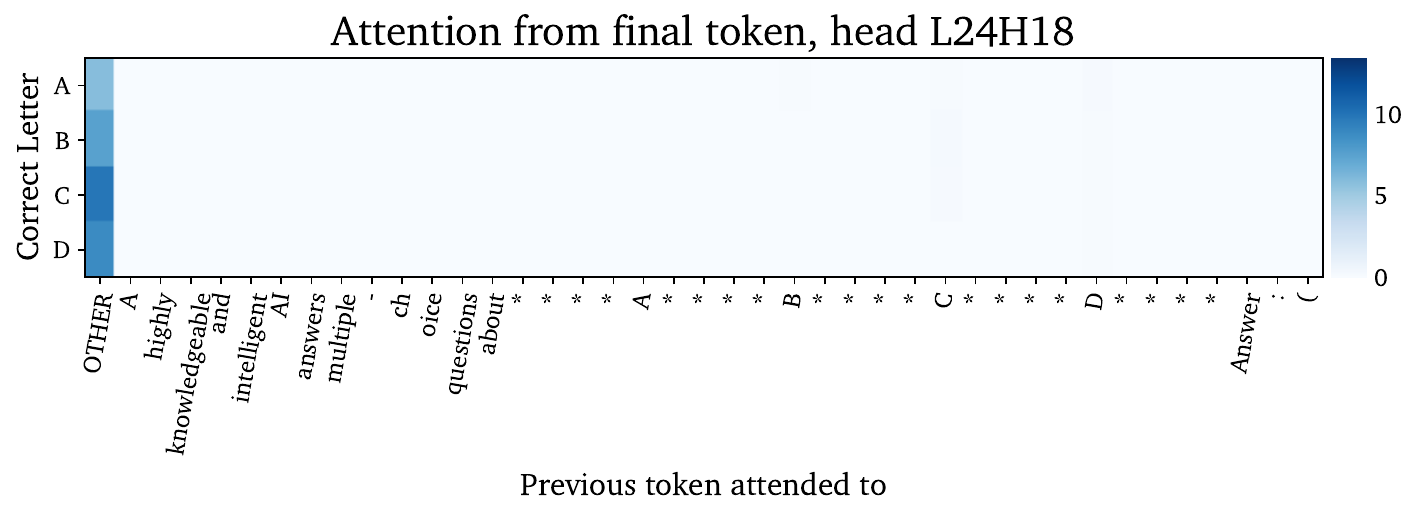}
         \caption{}
     \end{subfigure}
     \hfill
     \begin{subfigure}[b]{0.49\textwidth}
         \centering
         \includegraphics[width=\textwidth]{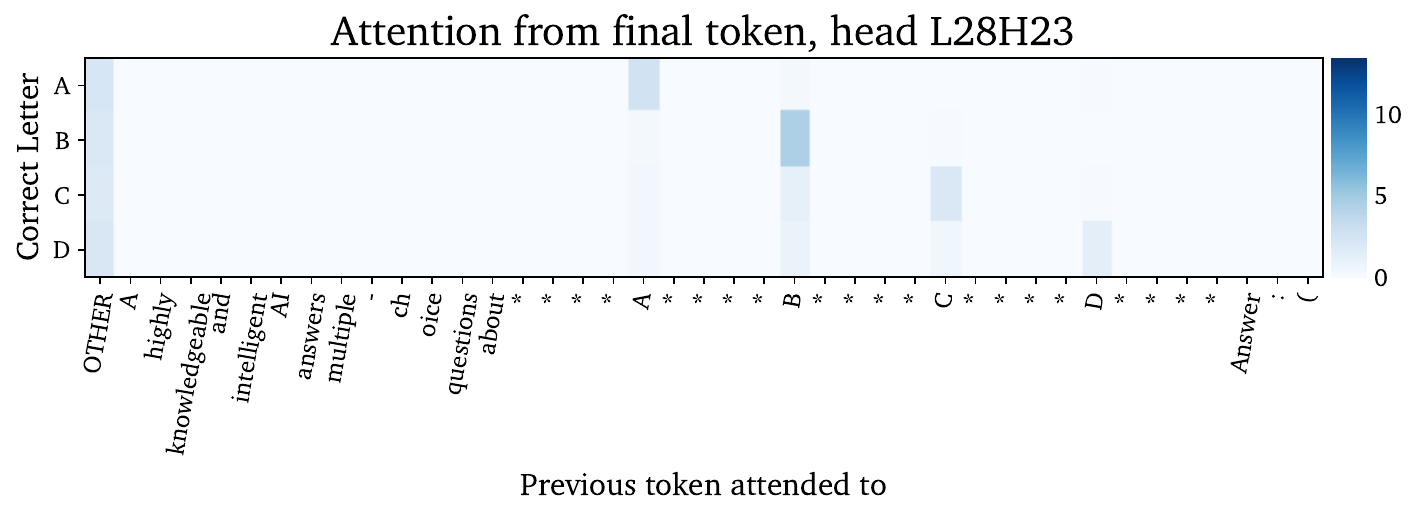}
         \caption{}
     \end{subfigure}
     \hfill
     \begin{subfigure}[b]{0.49\textwidth}
         \centering
         \includegraphics[width=\textwidth]{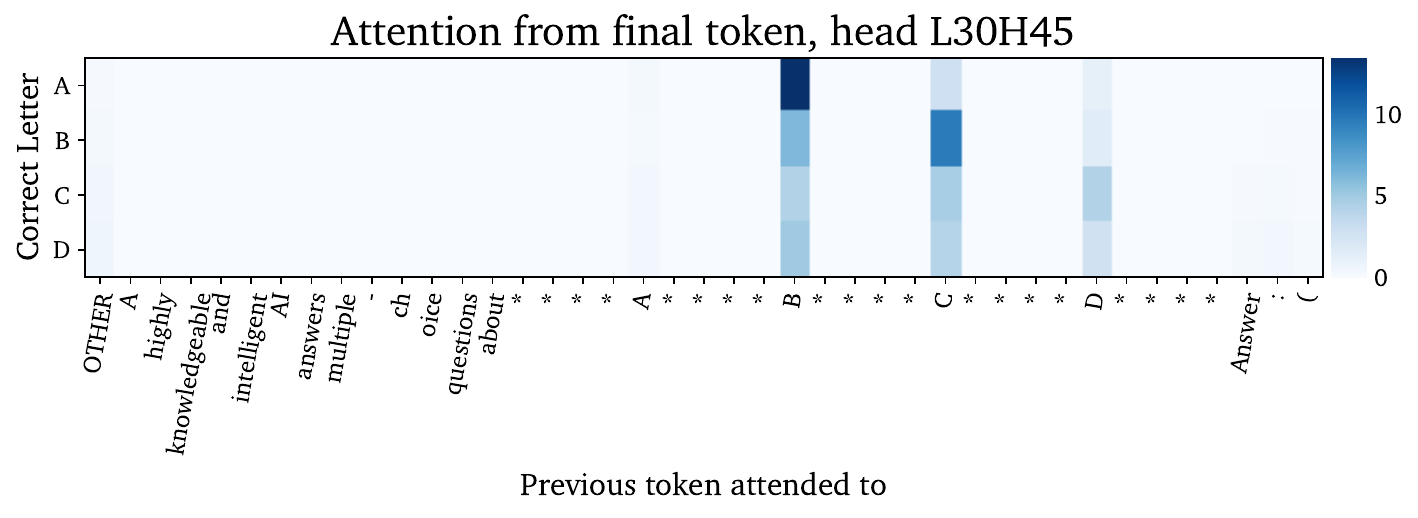}
         \caption{Note that this head attends to the label \emph{after} the correct label}
     \end{subfigure}
    \caption{Misc heads}
    \label{fig:misc_heads_val_attn}
\end{figure}

\begin{figure}
    \centering
     \begin{subfigure}[b]{0.49\textwidth}
         \centering
         \includegraphics[width=\textwidth]{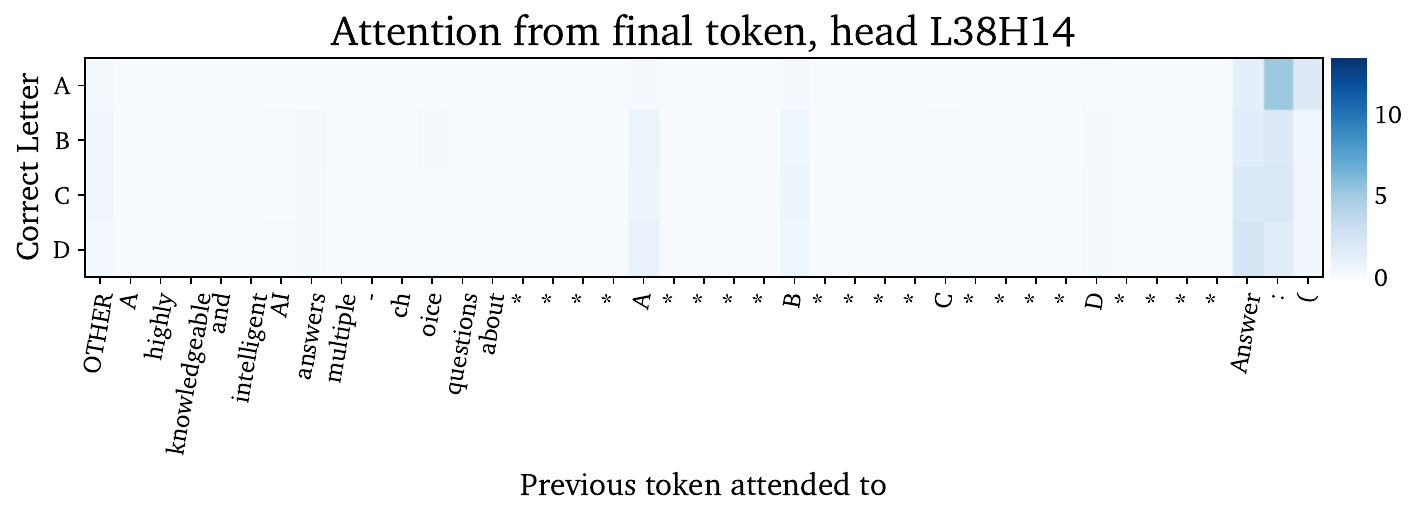}
         \caption{}
     \end{subfigure}
     \hfill
     \begin{subfigure}[b]{0.49\textwidth}
         \centering
         \includegraphics[width=\textwidth]{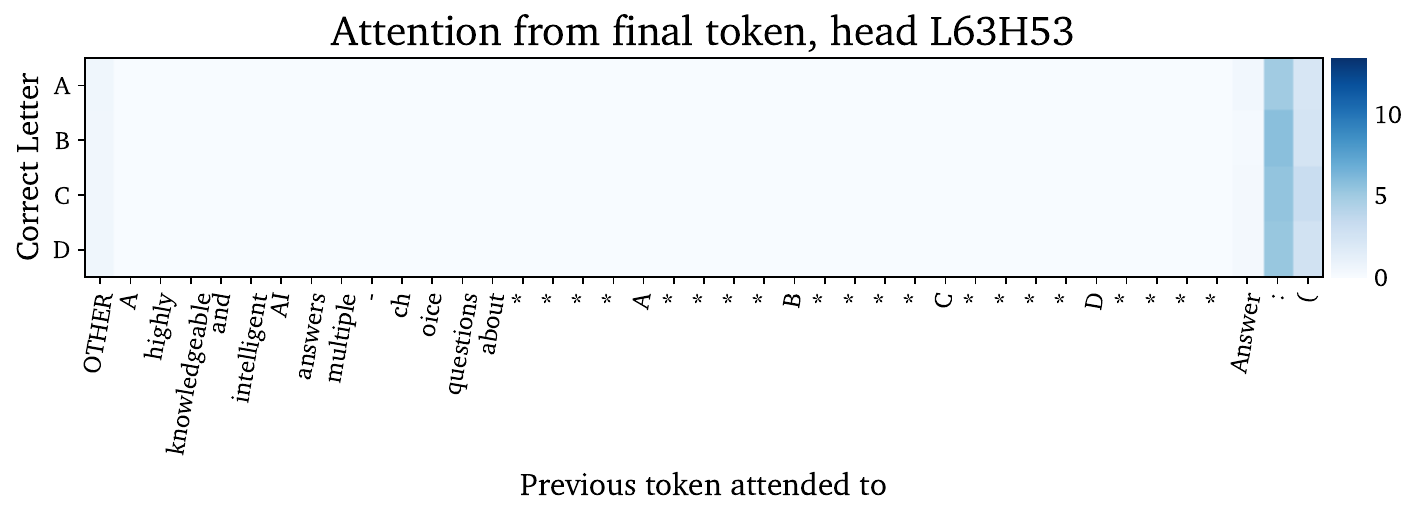}
         \caption{}
     \end{subfigure}
     \hfill
     \begin{subfigure}[b]{0.49\textwidth}
         \centering
         \includegraphics[width=\textwidth]{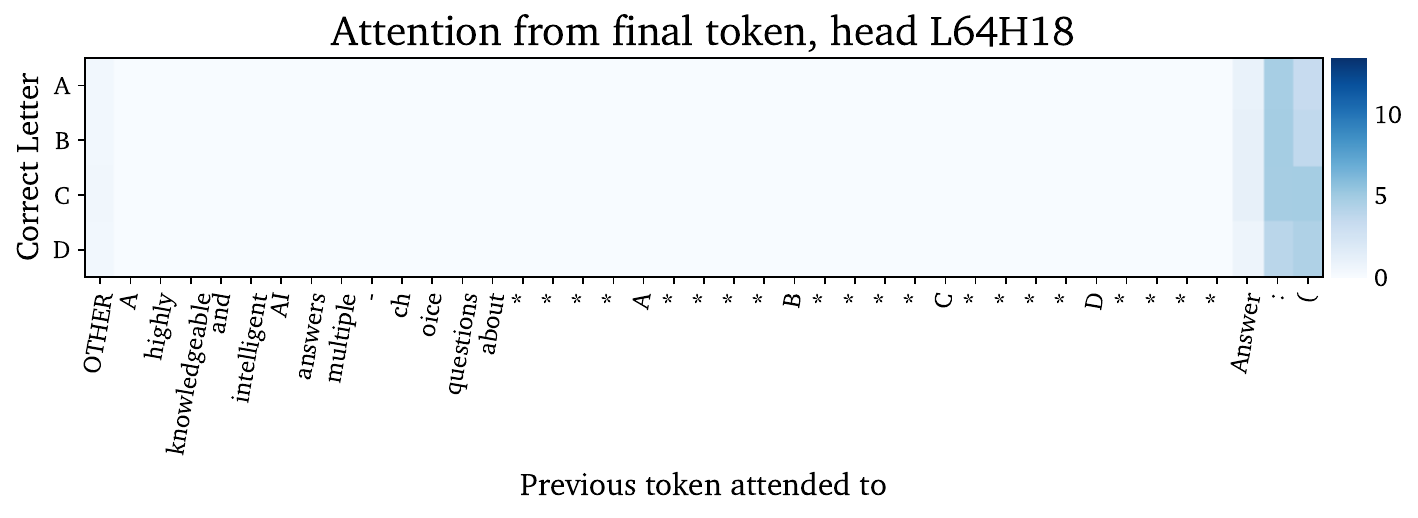}
         \caption{}
     \end{subfigure}
     \hfill
     \begin{subfigure}[b]{0.49\textwidth}
         \centering
         \includegraphics[width=\textwidth]{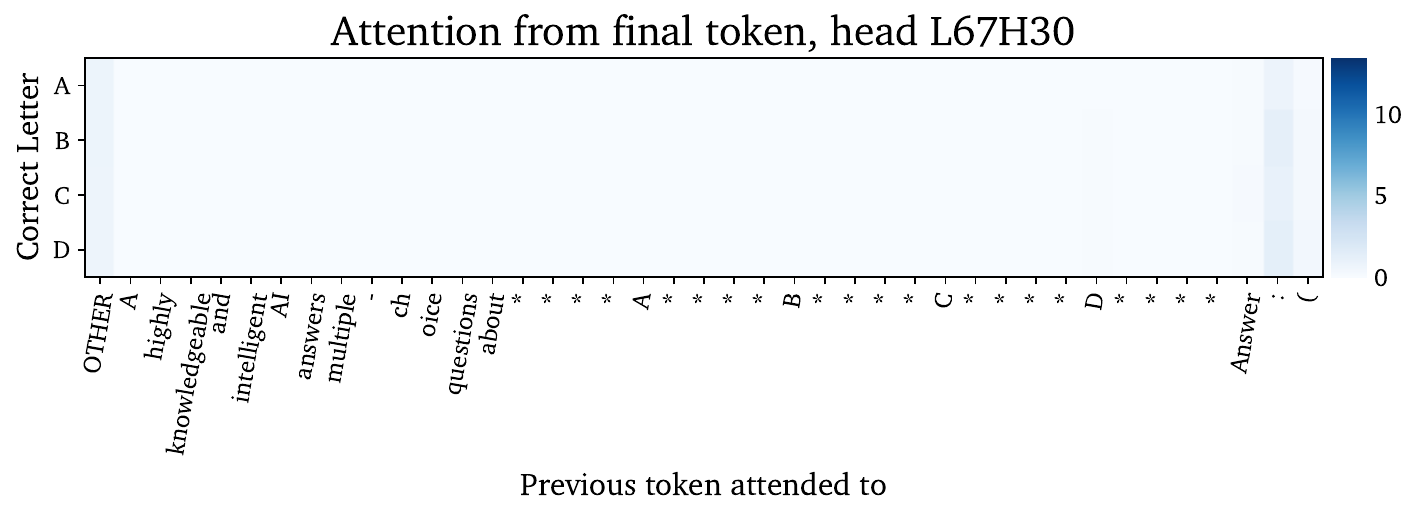}
         \caption{}
     \end{subfigure}
     \hfill
     \begin{subfigure}[b]{0.49\textwidth}
         \centering
         \includegraphics[width=\textwidth]{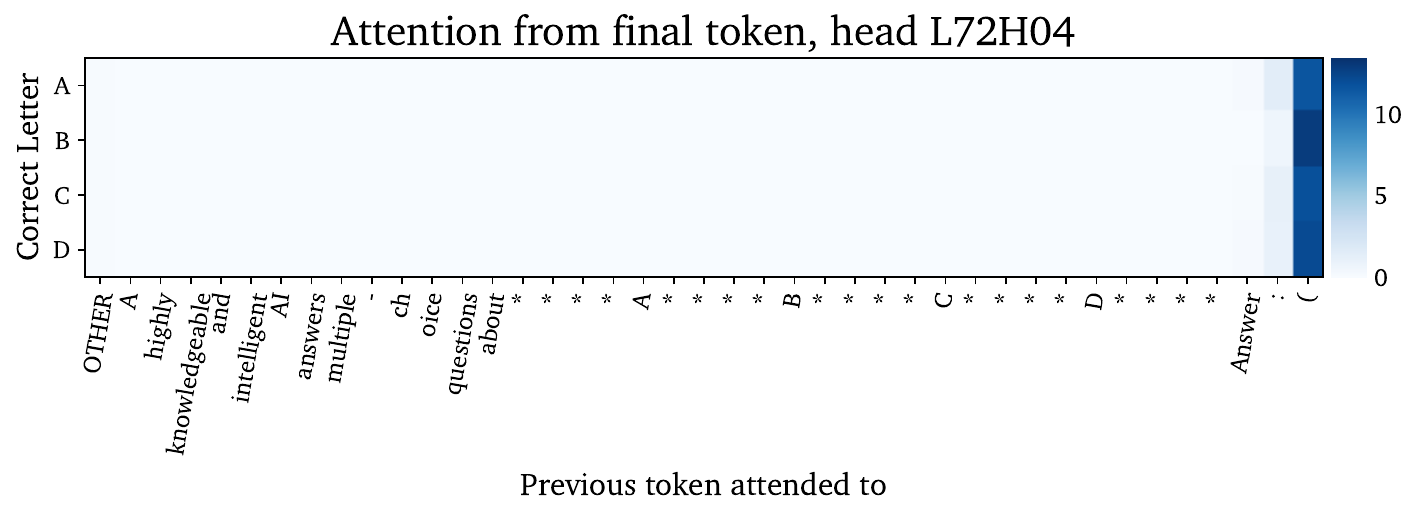}
         \caption{}
     \end{subfigure}
     \hfill
     \begin{subfigure}[b]{0.49\textwidth}
         \centering
         \includegraphics[width=\textwidth]{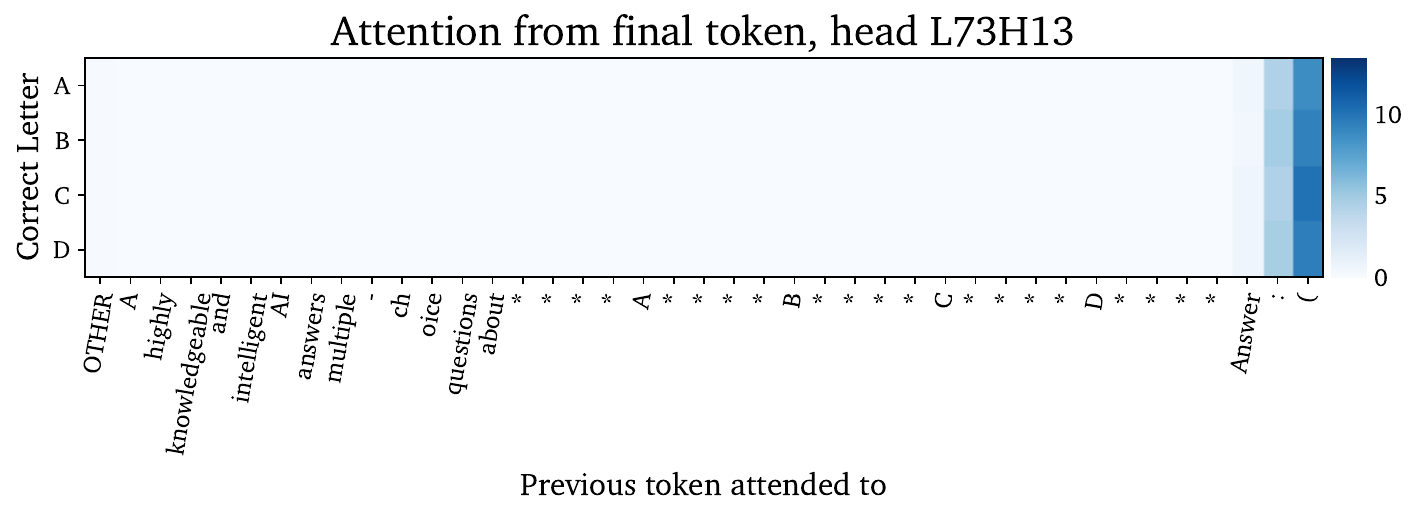}
         \caption{}
     \end{subfigure}
     \hfill
     \begin{subfigure}[b]{0.49\textwidth}
         \centering
         \includegraphics[width=\textwidth]{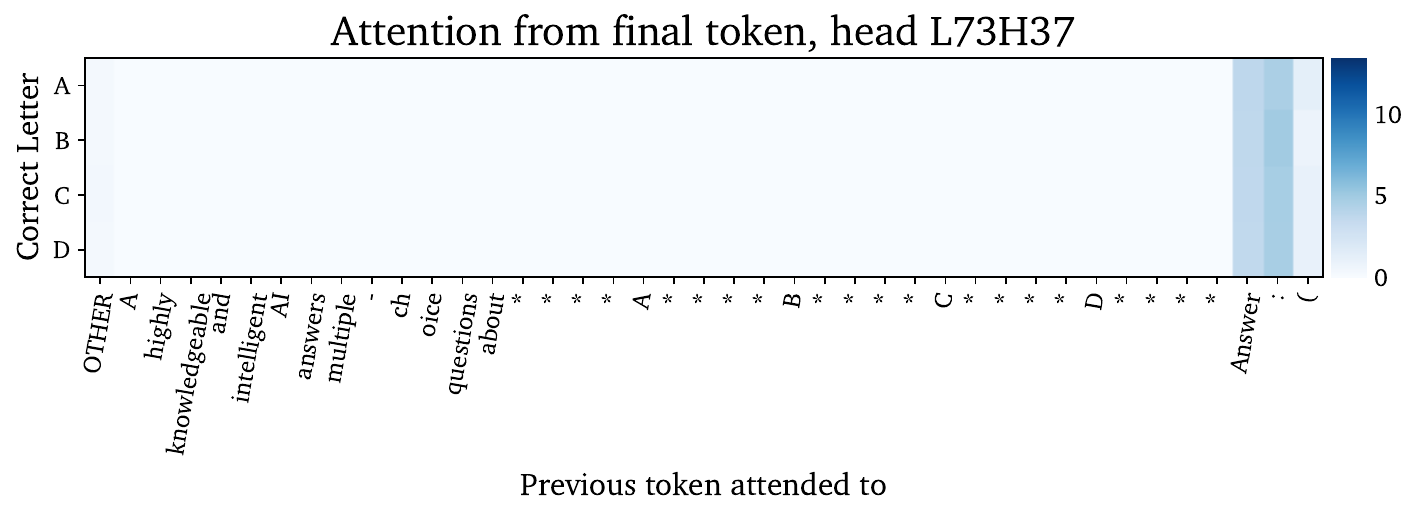}
         \caption{}
     \end{subfigure}
     \hfill
     \begin{subfigure}[b]{0.49\textwidth}
         \centering
         \includegraphics[width=\textwidth]{figures/output_heads_attention/amplification/attention_L74_H14.pdf}
         \caption{}
     \end{subfigure}
     \hfill
     \begin{subfigure}[b]{0.49\textwidth}
         \centering
         \includegraphics[width=\textwidth]{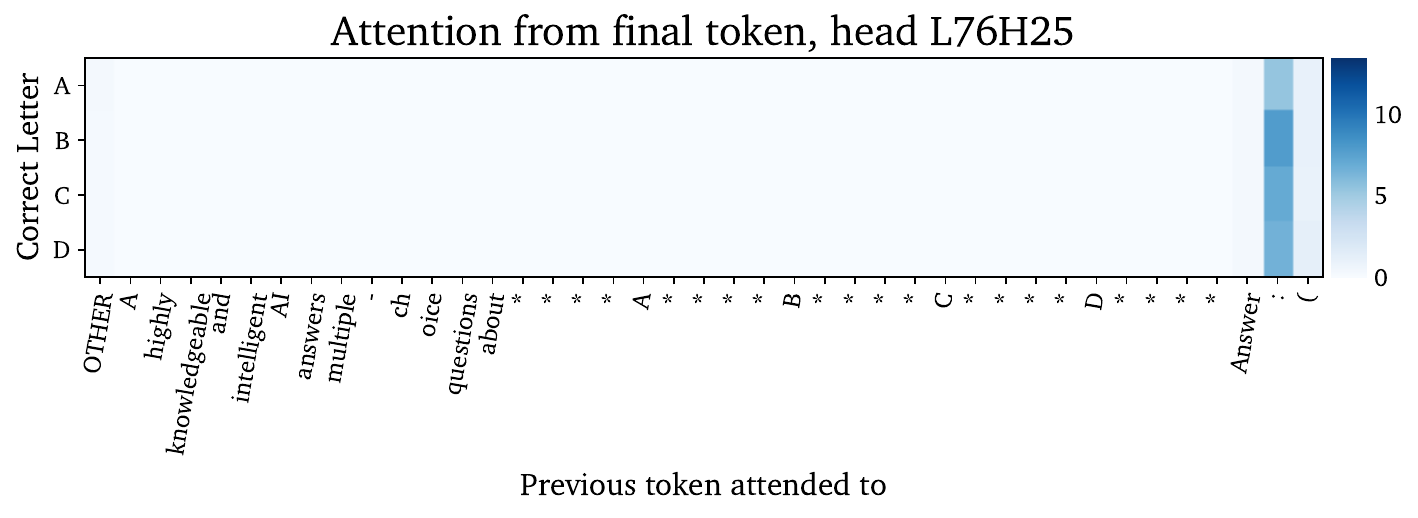}
         \caption{}
     \end{subfigure}
     \hfill
     \begin{subfigure}[b]{0.49\textwidth}
         \centering
         \includegraphics[width=\textwidth]{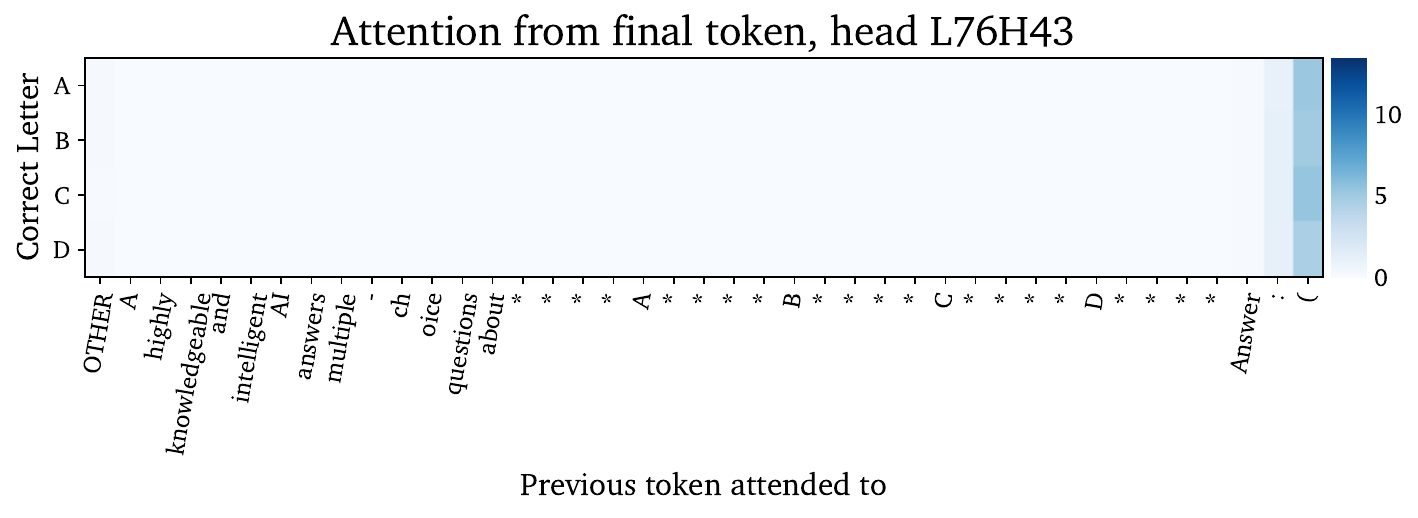}
         \caption{}
     \end{subfigure}
     \hfill
     \begin{subfigure}[b]{0.49\textwidth}
         \centering
         \includegraphics[width=\textwidth]{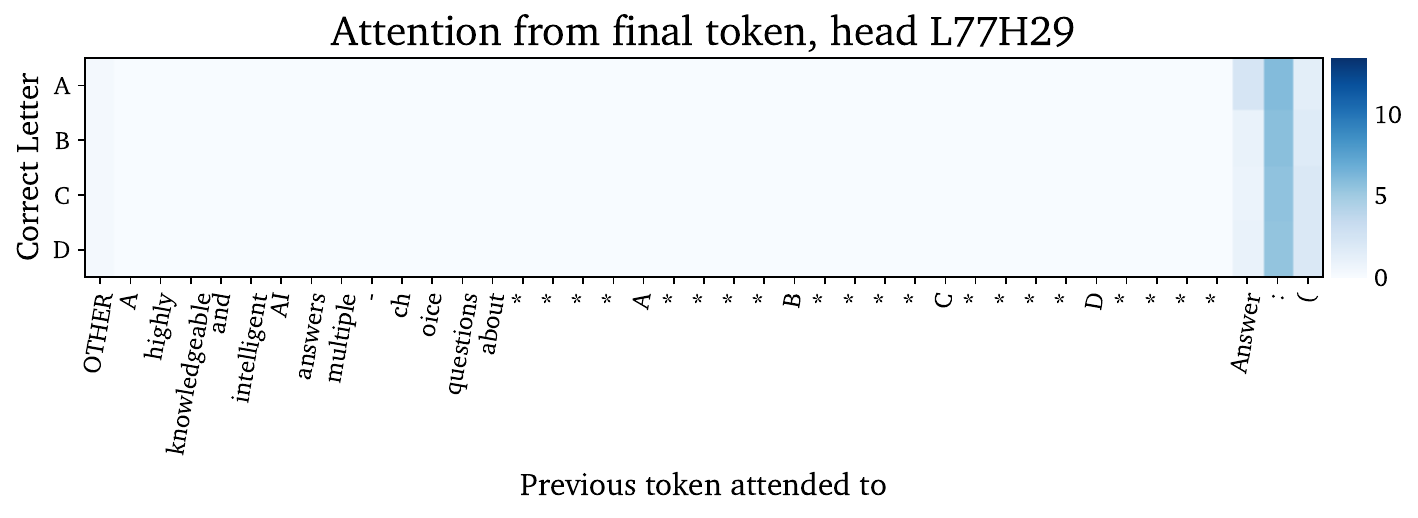}
         \caption{}
     \end{subfigure}
    \caption{Amplification heads}
    \label{fig:amplification_heads_val_attn}
\end{figure}
\clearpage

\newpage
\section{Do the keys alone encode the correct answer?}
\label{sec:key_deltas}

In~\cref{sec:understanding_nodes} we focus on the \emph{centred} keys and queries of the Correct Letter heads. However, the dot product between the centred keys and queries is only one of four components of the full dot product -- the dot product between the \emph{uncentred} keys and queries. In this section, we illustrate that the remaining three terms really are uninformative as to which answer is correct.

In each of the following plots, we show all four components of the dot product separately for each head, averaged over 1024 prompts -- 256 prompts where A is correct, 256 prompts where B is correct, etc. As in~\cref{sec:understanding_nodes}, we denote the key mean $k_\mu$, query mean $q_\mu$, centred keys $k_\delta$, and centred queries $q_\delta$.

Note that, for all heads, only the dot products between the centred queries and centred keys seem to identify which letter is correct.

\includegraphics[width=\textwidth]{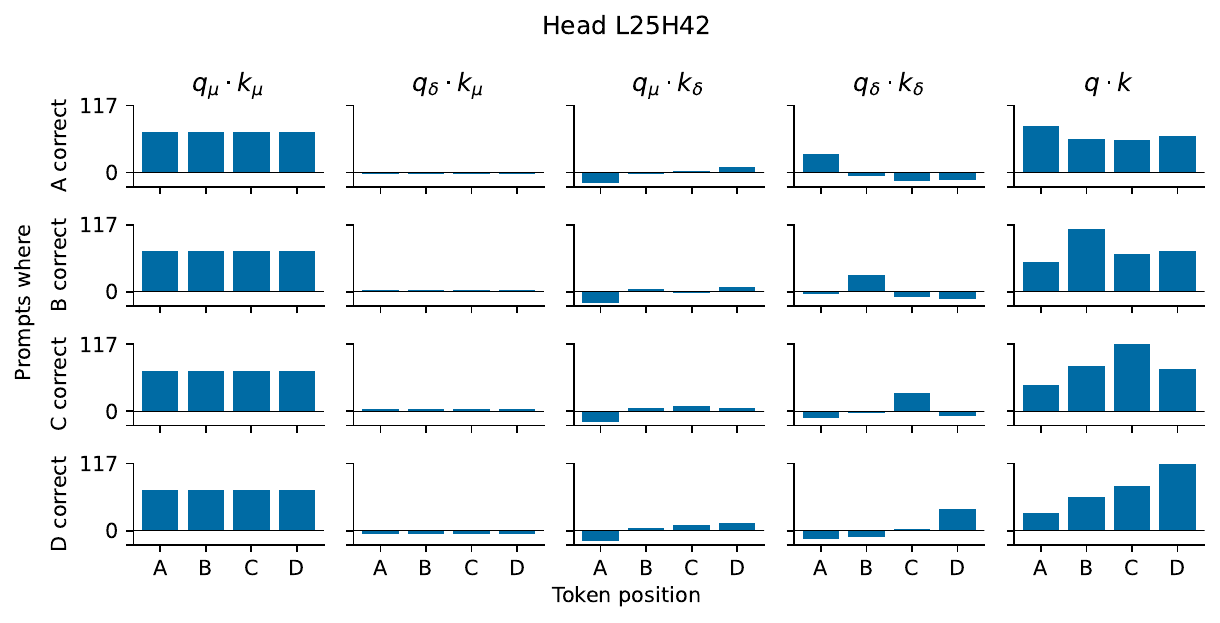}

\includegraphics[width=\textwidth]{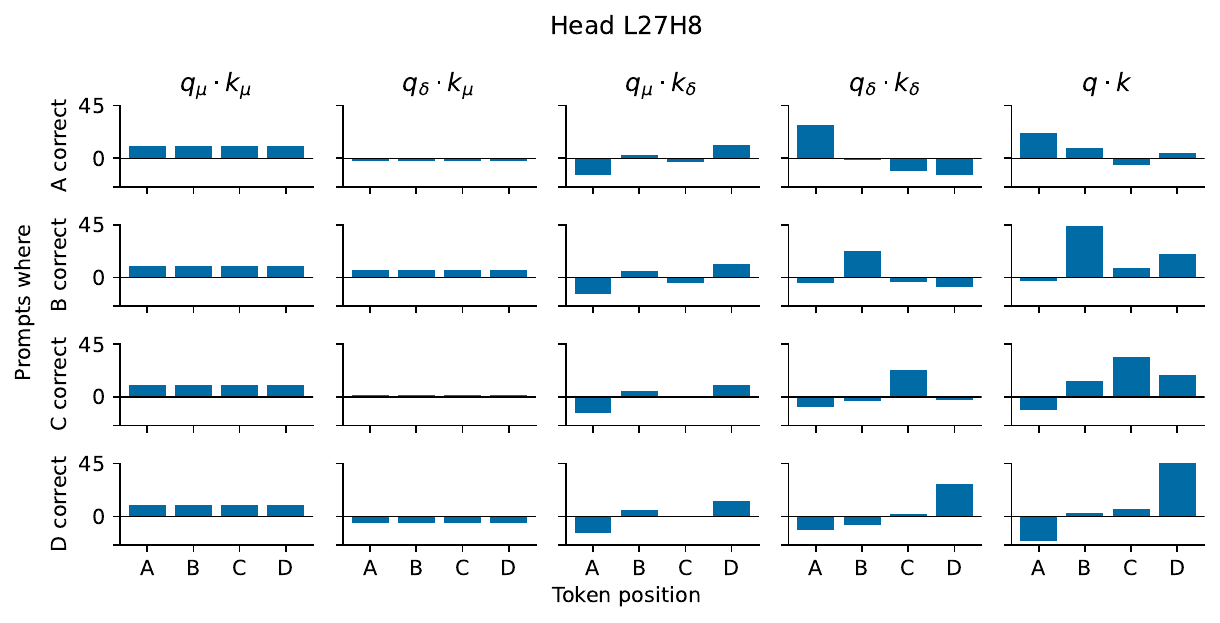}

\includegraphics[width=\textwidth]{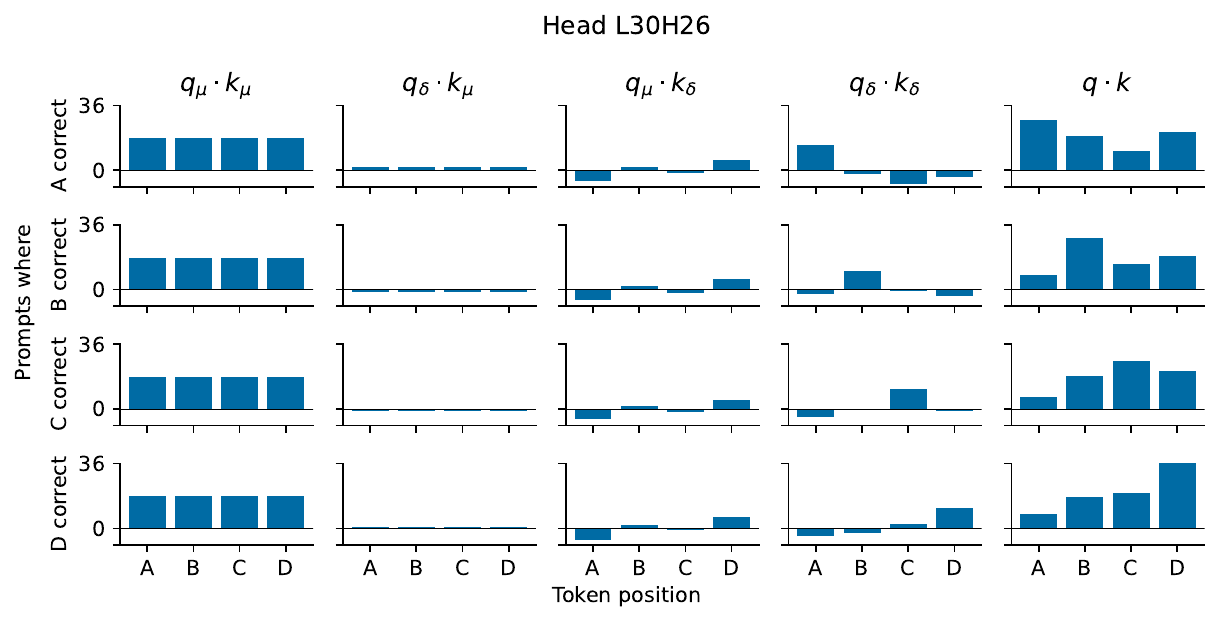}

\includegraphics[width=\textwidth]{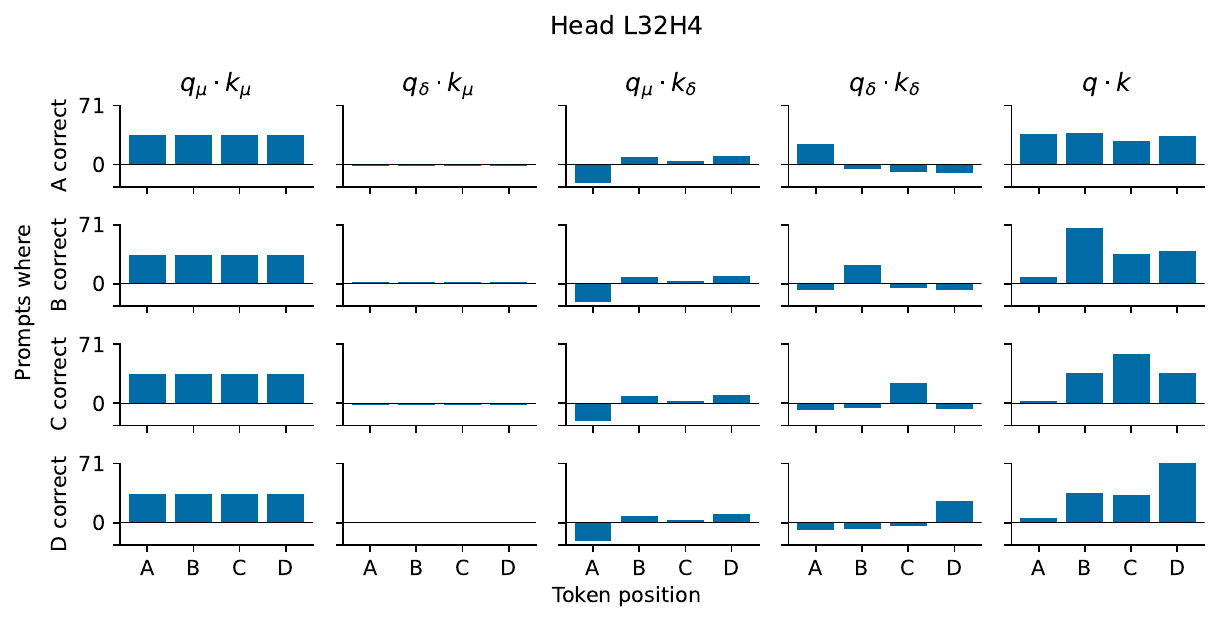}

\includegraphics[width=\textwidth]{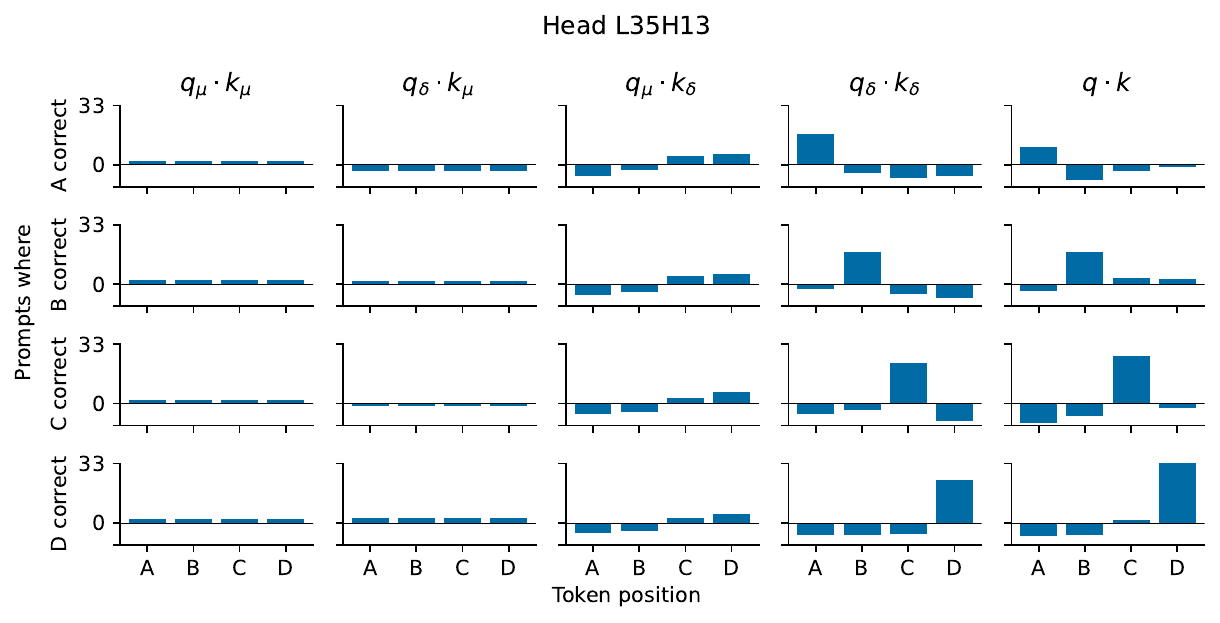}

\includegraphics[width=\textwidth]{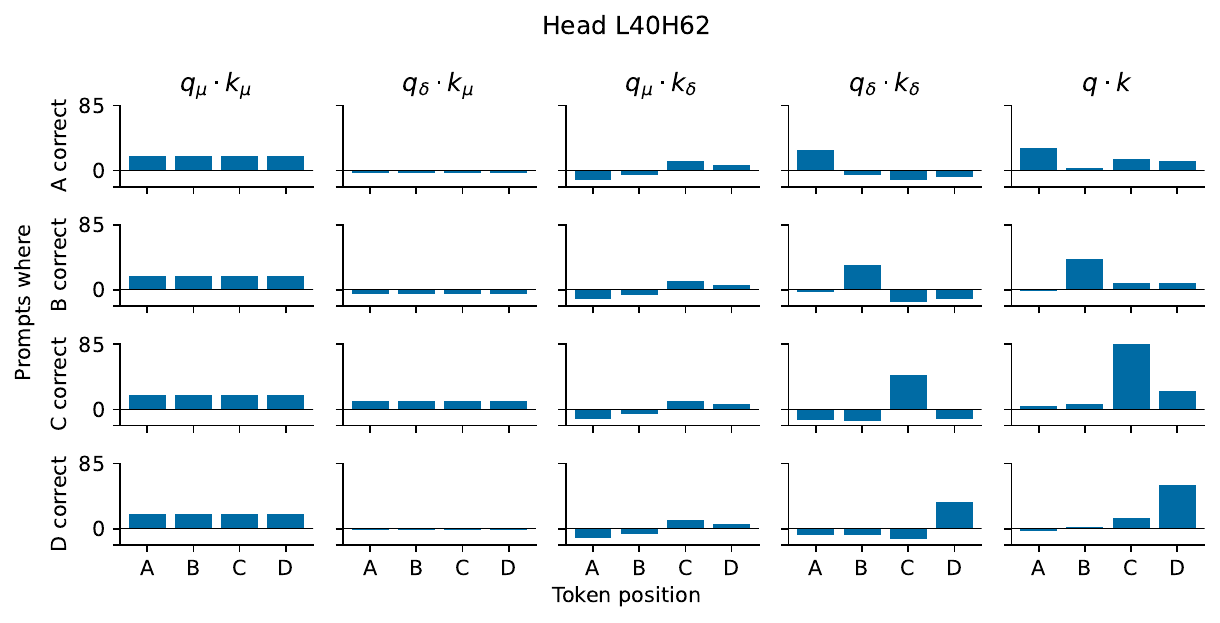}

\clearpage

\newpage
\section{More Low Rank Results}
\label{sec:more_low_rank_results}

\begin{figure}[t]
    \centering
    \includegraphics{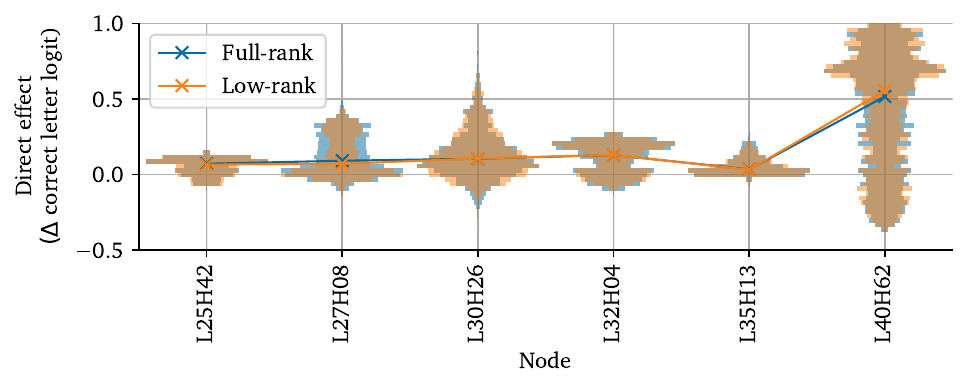}
    \caption{Direct effect of correct letter heads when replacing their attention and values with the corresponding low-rank versions.}
    \label{fig:low_rank_direct_effects}
\end{figure}

\begin{figure}[t]
    \centering
     \begin{subfigure}[b]{0.245\textwidth}
         \centering
         \includegraphics[width=\textwidth]{figures/loss_LR_attn/loss_lr_attn_base.pdf}
         \caption{}
     \end{subfigure}
     \hfill
     \begin{subfigure}[b]{0.245\textwidth}
         \centering
         \includegraphics[width=\textwidth]{figures/loss_LR_attn/loss_lr_attn_replace_letter_with_number.pdf}
         \caption{}
     \end{subfigure}
     \hfill
     \begin{subfigure}[b]{0.245\textwidth}
         \centering
         \includegraphics[width=\textwidth]{figures/loss_LR_attn/loss_lr_attn_randomise_answer_letters.pdf}
         \caption{}
     \end{subfigure}
     \hfill
     \begin{subfigure}[b]{0.245\textwidth}
         \centering
         \includegraphics[width=\textwidth]{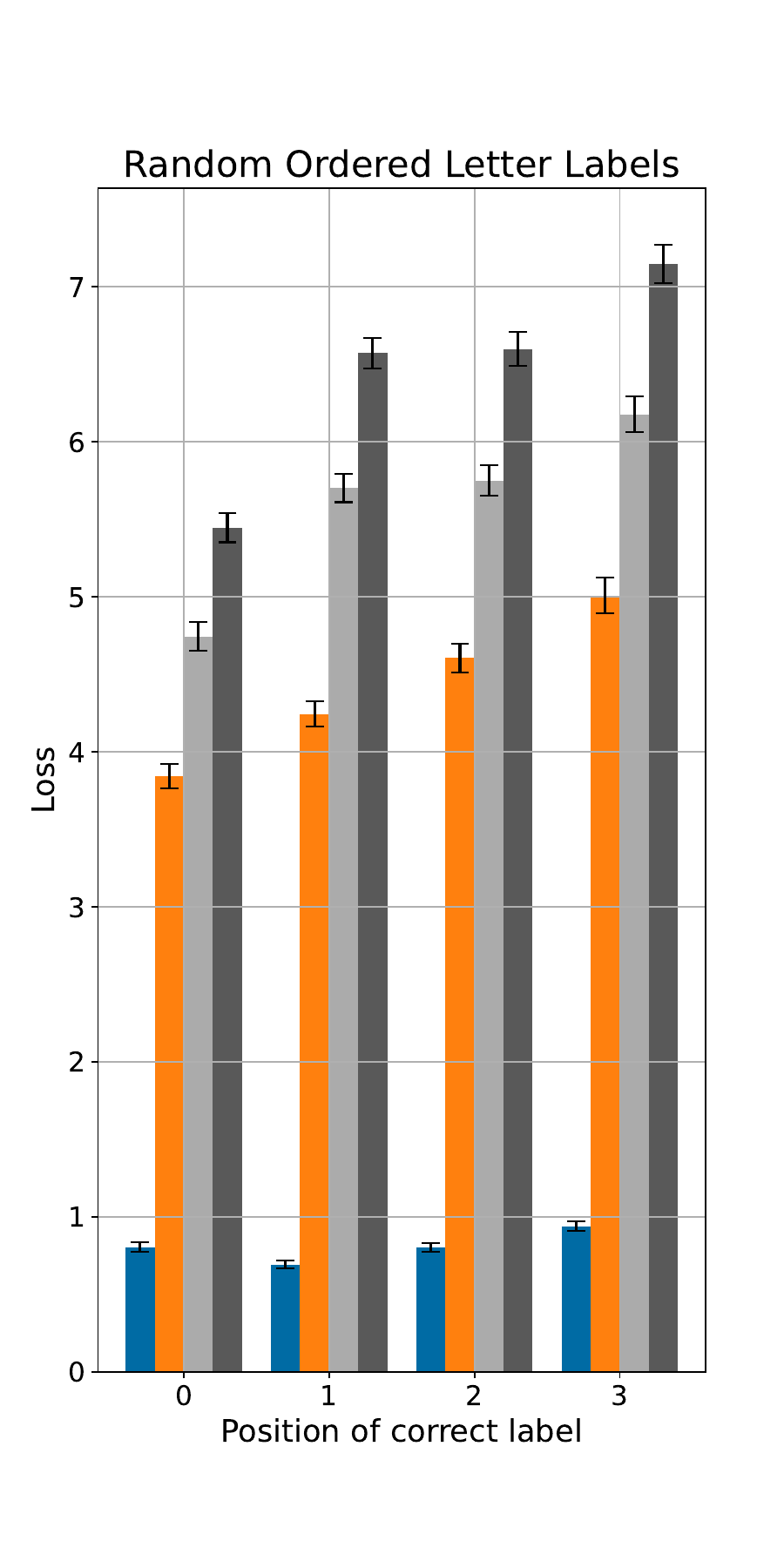}
         \caption{}
         \label{fig:loss_lr_attn_randomise_answer_letters_in_order}
     \end{subfigure}
     \hfill
     \begin{subfigure}[b]{0.245\textwidth}
         \centering
         \includegraphics[width=\textwidth]{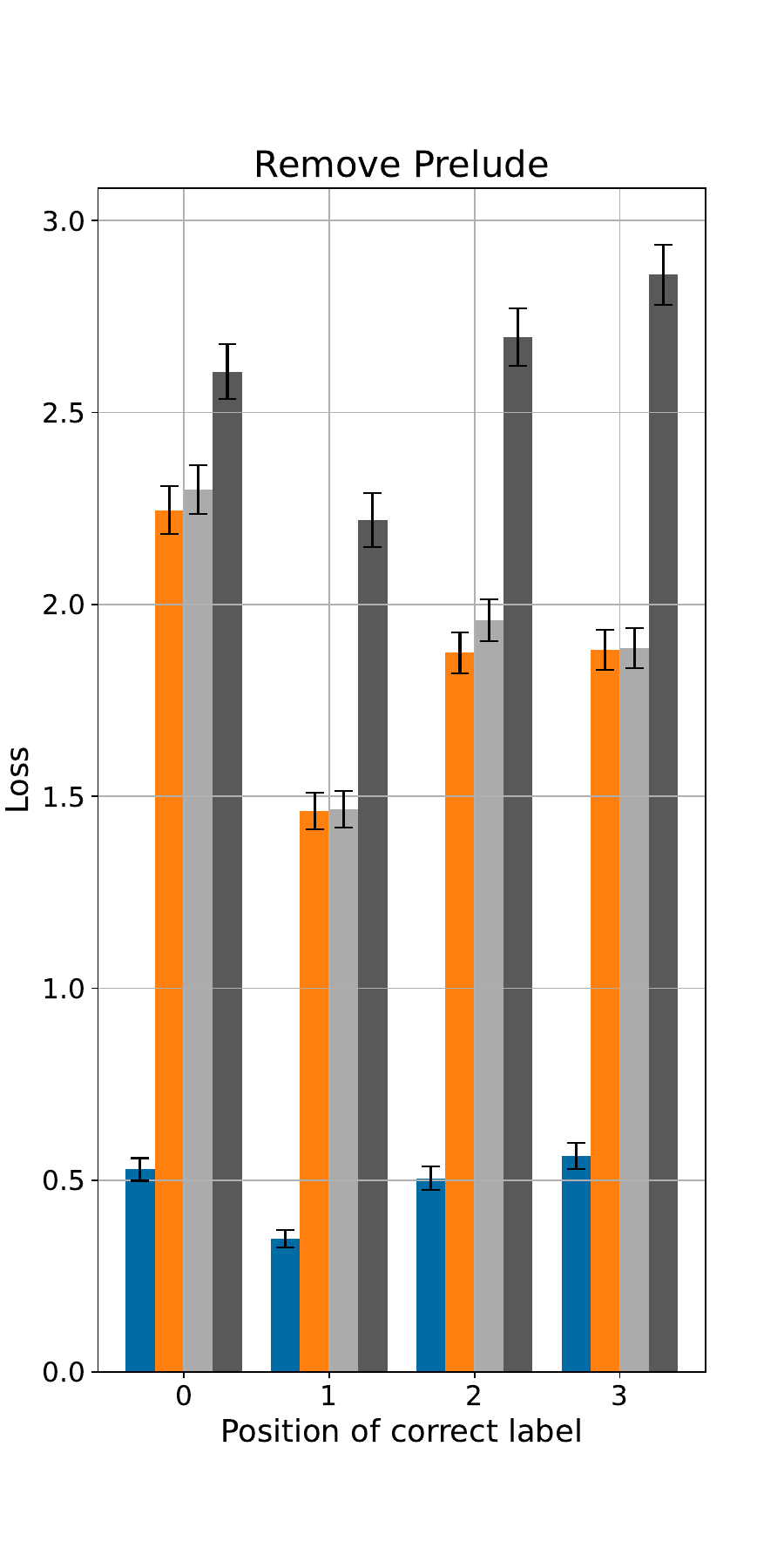}
         \caption{}
         \label{fig:loss_lr_attn_remove_prelude}
     \end{subfigure}
     \hfill
     \begin{subfigure}[b]{0.245\textwidth}
         \centering
         \includegraphics[width=\textwidth]{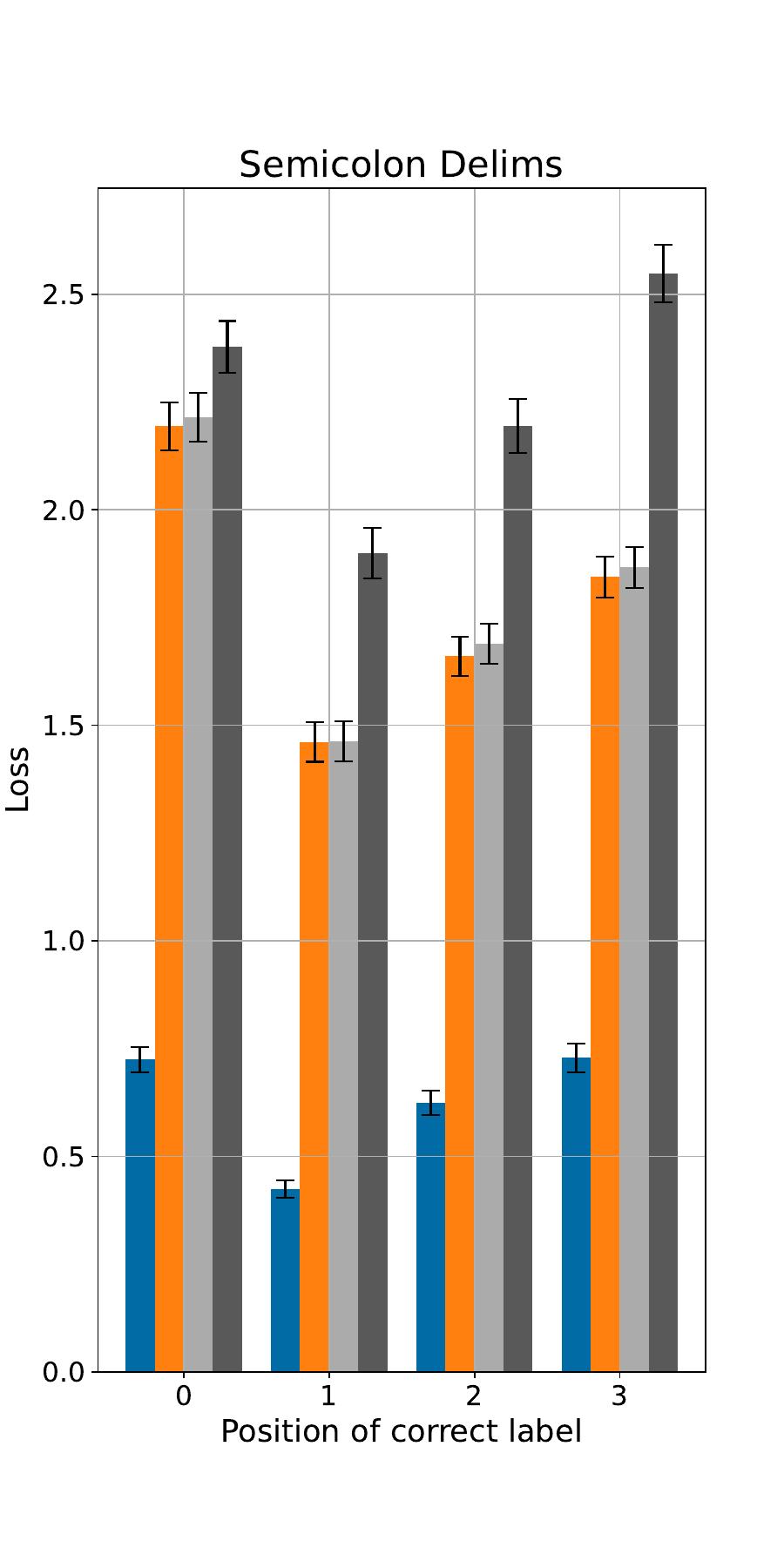}
         \caption{}
         \label{fig:loss_lr_attn_semicolons}
     \end{subfigure}
     \hfill
     \begin{subfigure}[b]{0.245\textwidth}
         \centering
         \includegraphics[width=\textwidth]{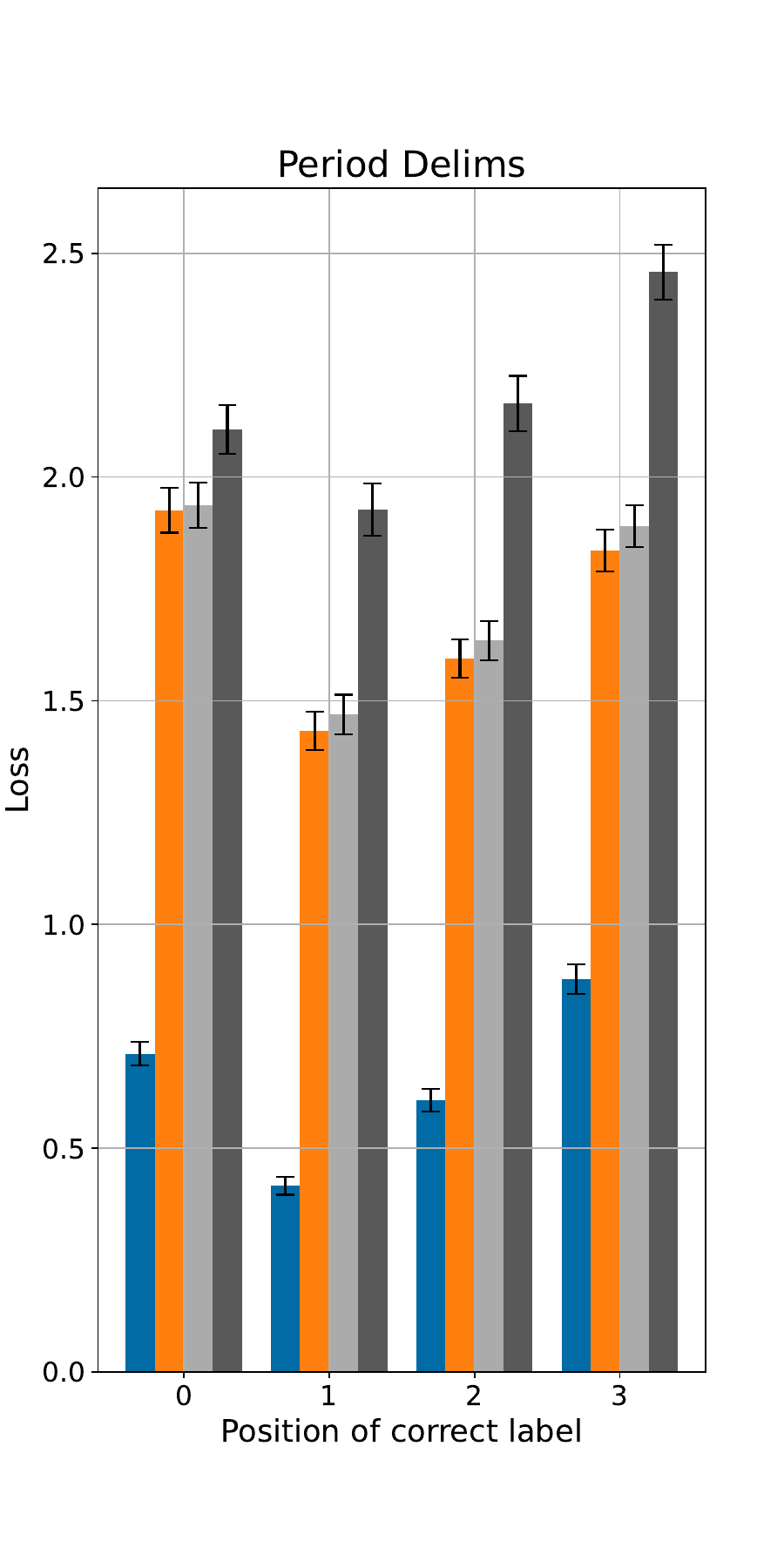}
         \caption{}
         \label{fig:loss_lr_attn_periods}
     \end{subfigure}
    \caption{Loss when using full rank or low rank attention under various prompt mutations. Note the differing y-axes. `True targets' means running and evaluating the model on $p_{intervention}$, and `Random targets' means running on $p_{intervention}$ but evaluating on $p_{original}$.}
    \label{fig:mutated_low_rank_loss_full}
\end{figure}

\begin{figure}[t]
    \centering
     \begin{subfigure}[b]{0.5\textwidth}
         \centering
         \includegraphics[width=\textwidth]{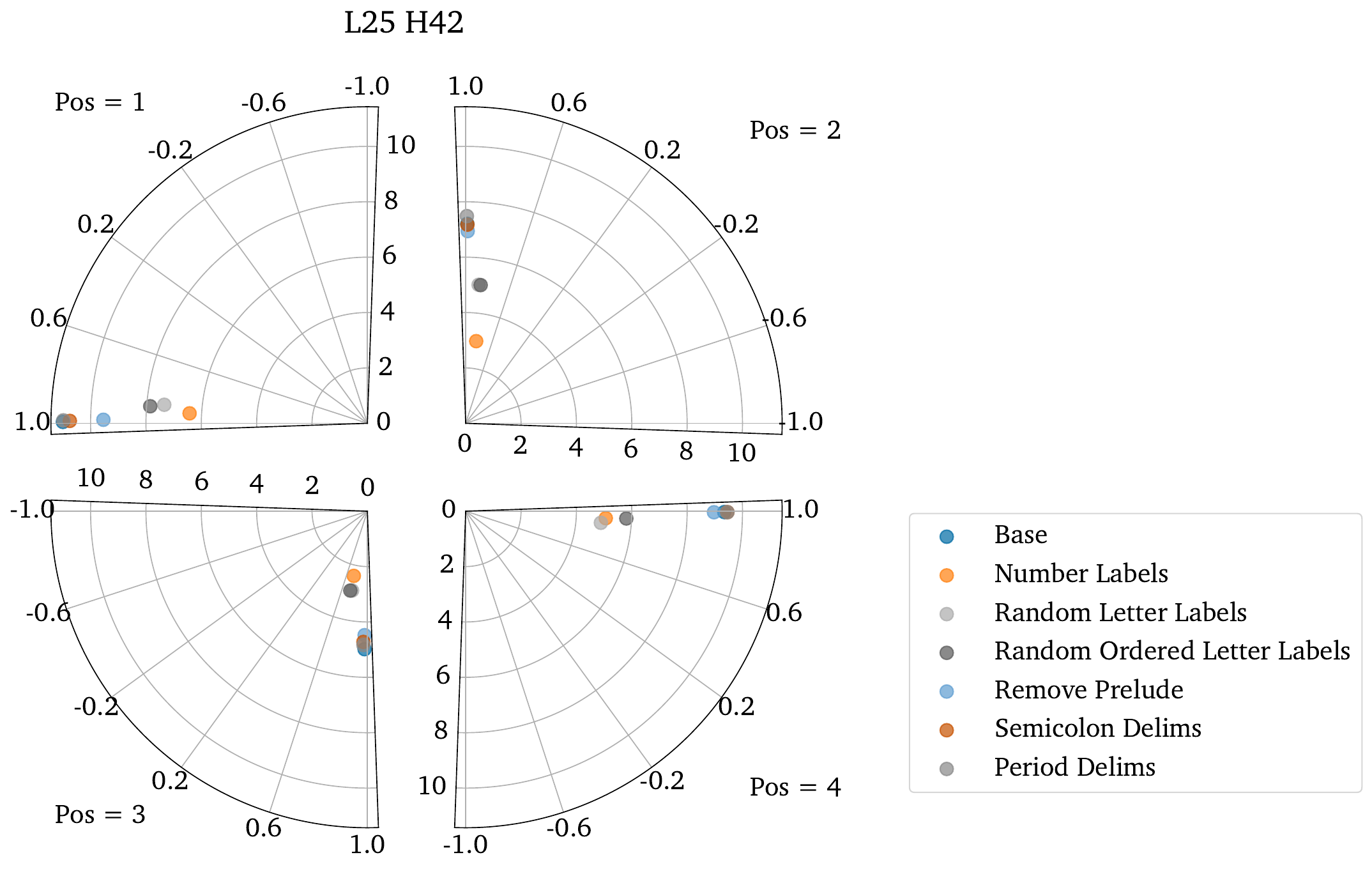}
         \caption{}
         \label{fig:project_k_deltas_on_k_centroids_l25h42}
     \end{subfigure}
     \hfill
     \begin{subfigure}[b]{0.3\textwidth}
         \centering
         \includegraphics[width=\textwidth]{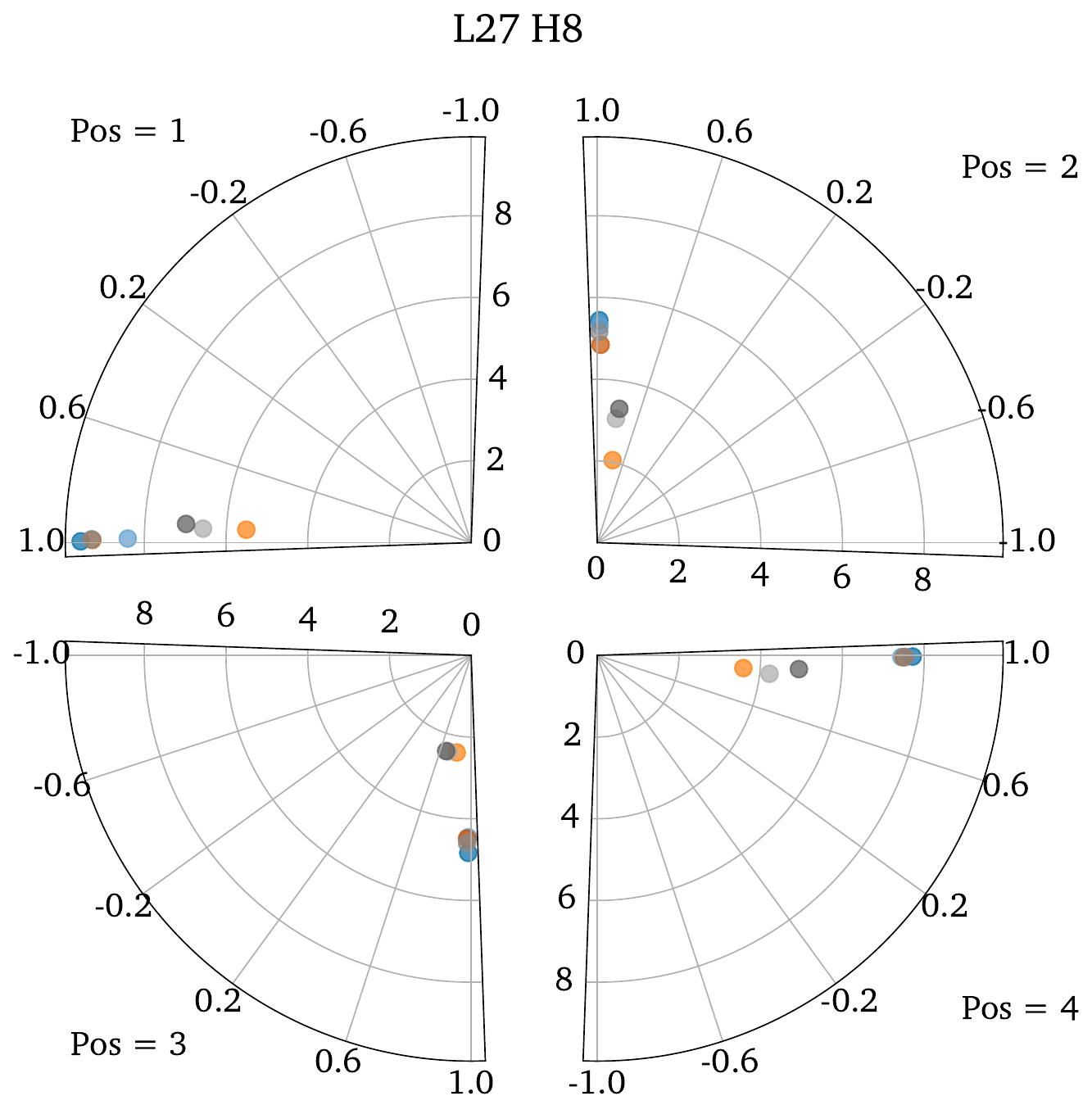}
         \caption{}
         \label{fig:project_k_deltas_on_k_centroids_l27h08}
     \end{subfigure}
     \hfill
     \begin{subfigure}[b]{0.3\textwidth}
         \centering
         \includegraphics[width=\textwidth]{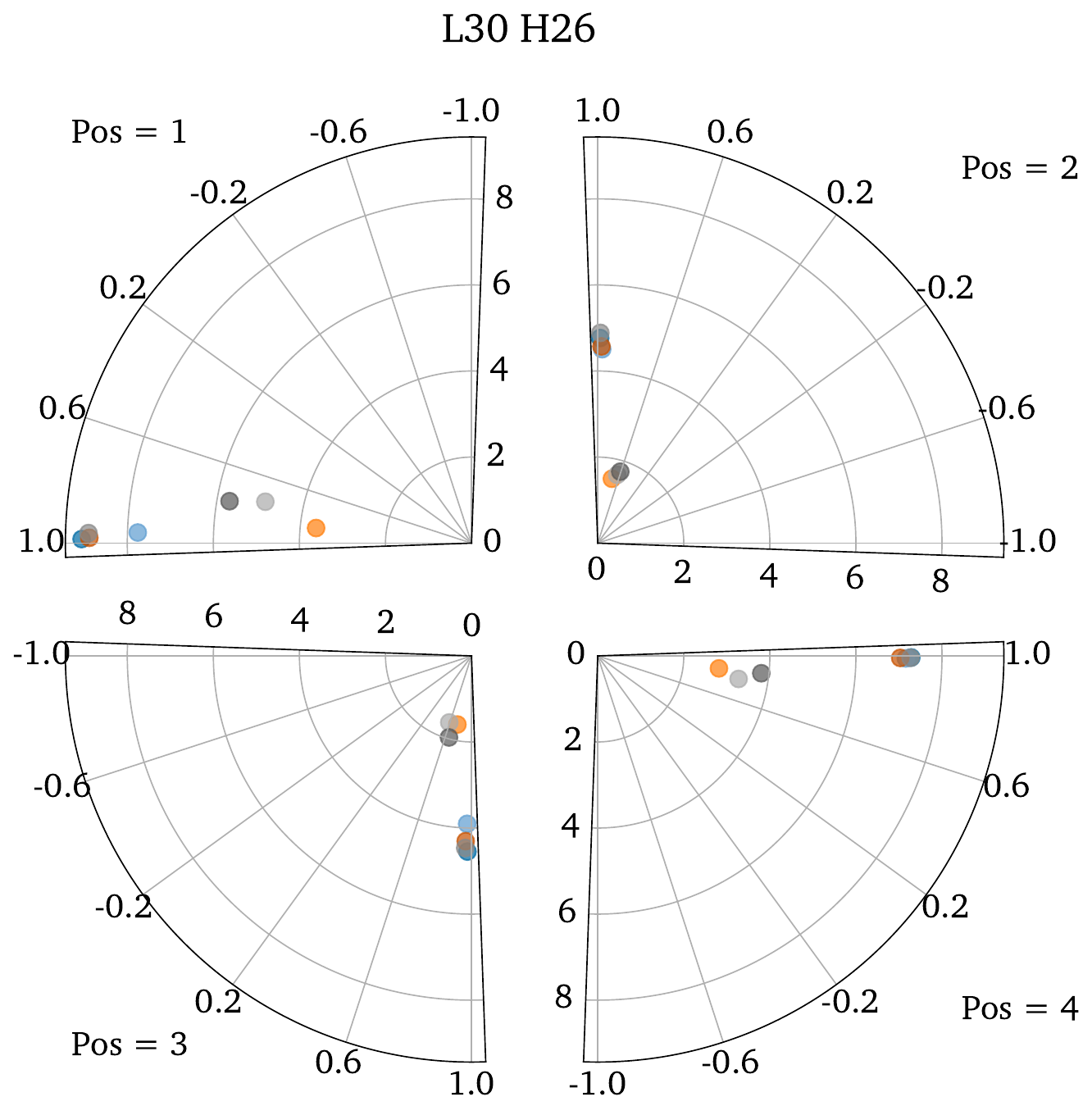}
         \caption{}
         \label{fig:project_k_deltas_on_k_centroids_l30h26}
     \end{subfigure}
     \hfill
     \begin{subfigure}[b]{0.3\textwidth}
         \centering
         \includegraphics[width=\textwidth]{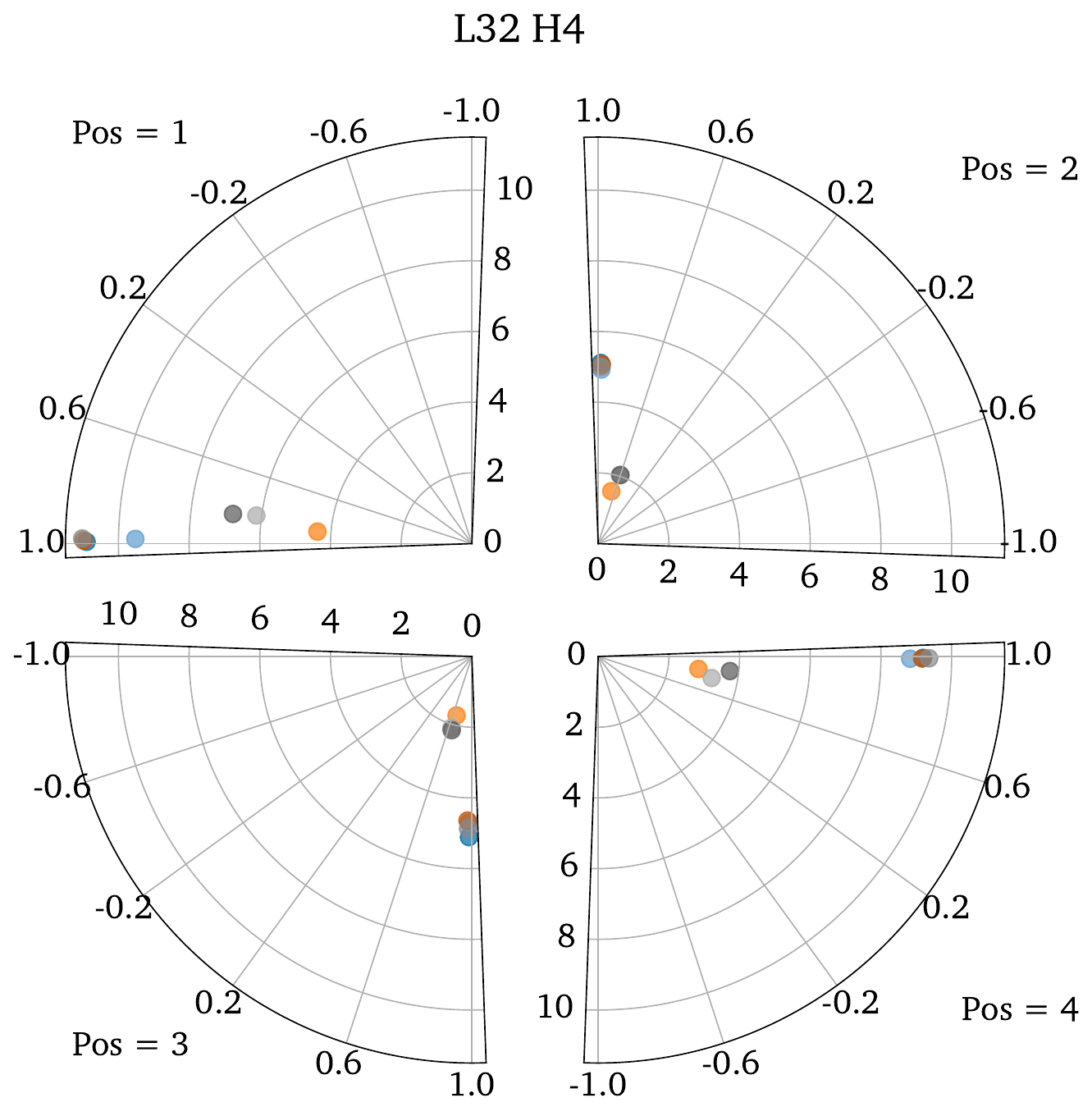}
         \caption{}
         \label{fig:project_k_deltas_on_k_centroids_l32h04}
     \end{subfigure}
     \hfill
     \begin{subfigure}[b]{0.3\textwidth}
         \centering
         \includegraphics[width=\textwidth]{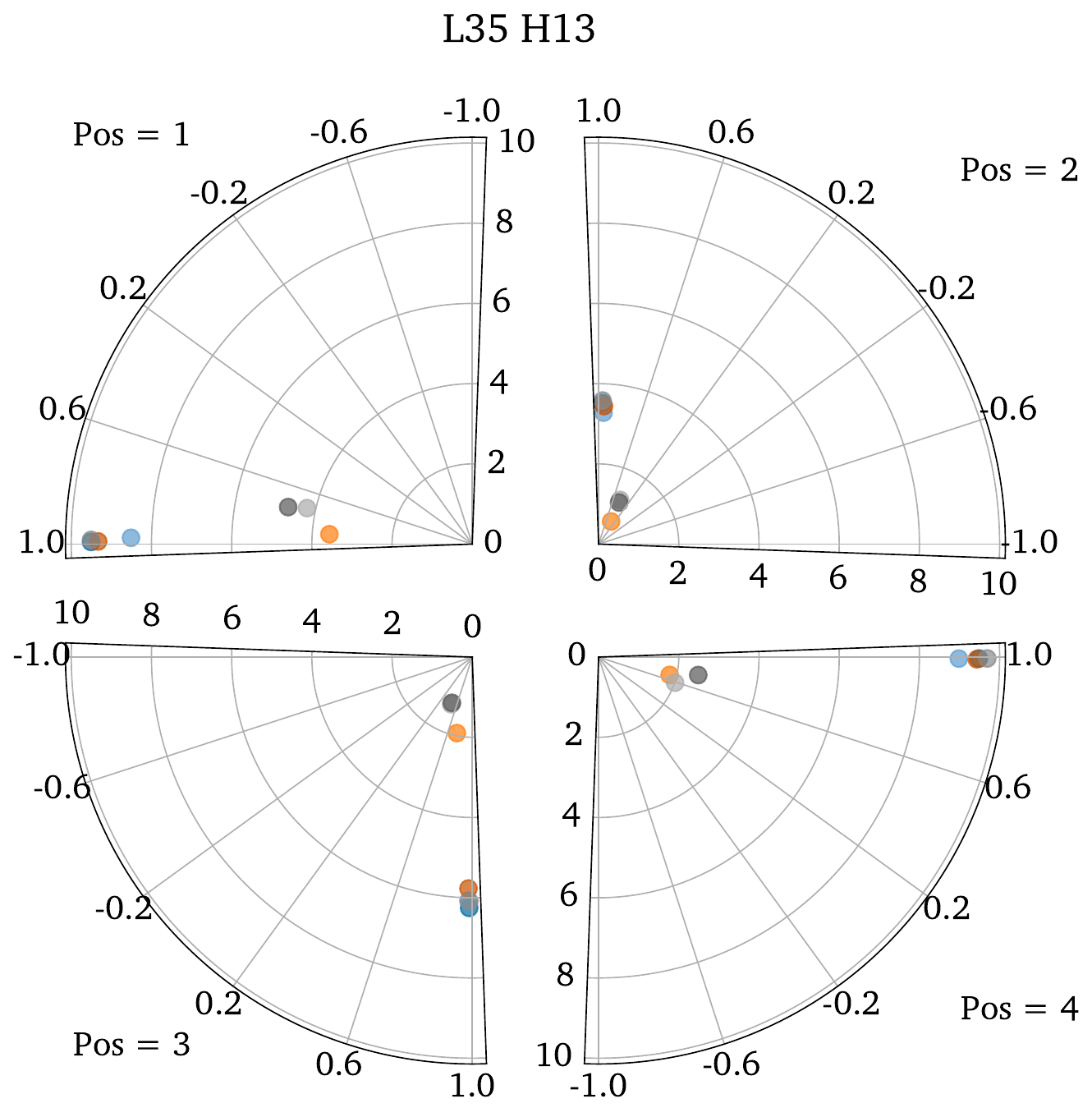}
         \caption{}
         \label{fig:project_k_deltas_on_k_centroids_l35h13}
     \end{subfigure}
     \hfill
     \begin{subfigure}[b]{0.3\textwidth}
         \centering
         \includegraphics[width=\textwidth]{figures/polarplots/kdeltas/polarplot_kdelta_kcentroid_L40H62.pdf}
         \caption{}
     \end{subfigure}
    \caption{Cosine similarity and absolute value of the projection of $k_\delta$ onto the centroids of the base $k_\delta$.}
    \label{fig:project_k_deltas_on_k_centroids}
\end{figure}

\begin{figure}[t]
    \centering
     \begin{subfigure}[b]{0.5\textwidth}
         \centering
         \includegraphics[width=\textwidth]{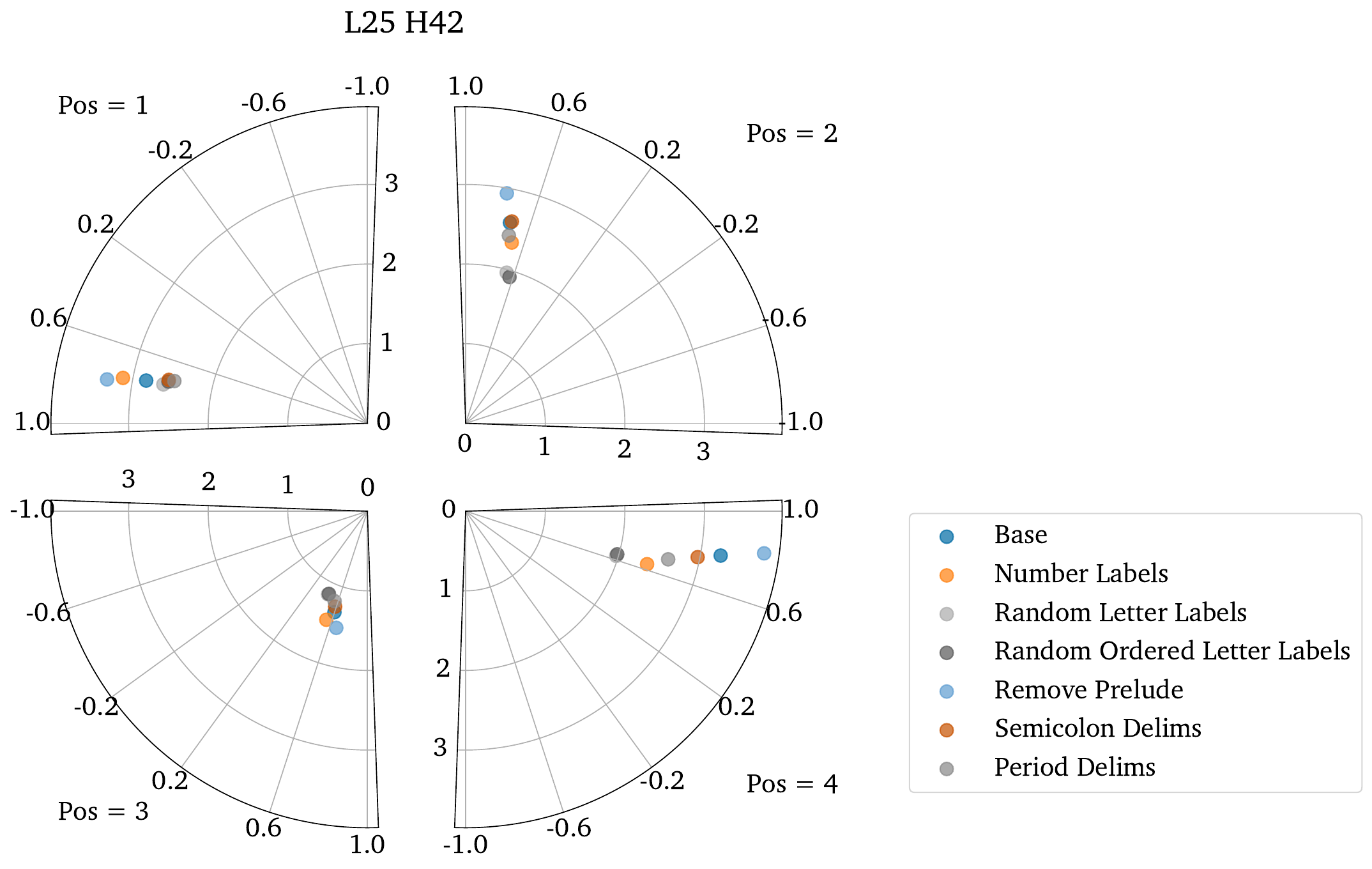}
         \caption{}
         \label{fig:project_q_deltas_on_k_centroids_l25h42}
     \end{subfigure}
     \hfill
     \begin{subfigure}[b]{0.3\textwidth}
         \centering
         \includegraphics[width=\textwidth]{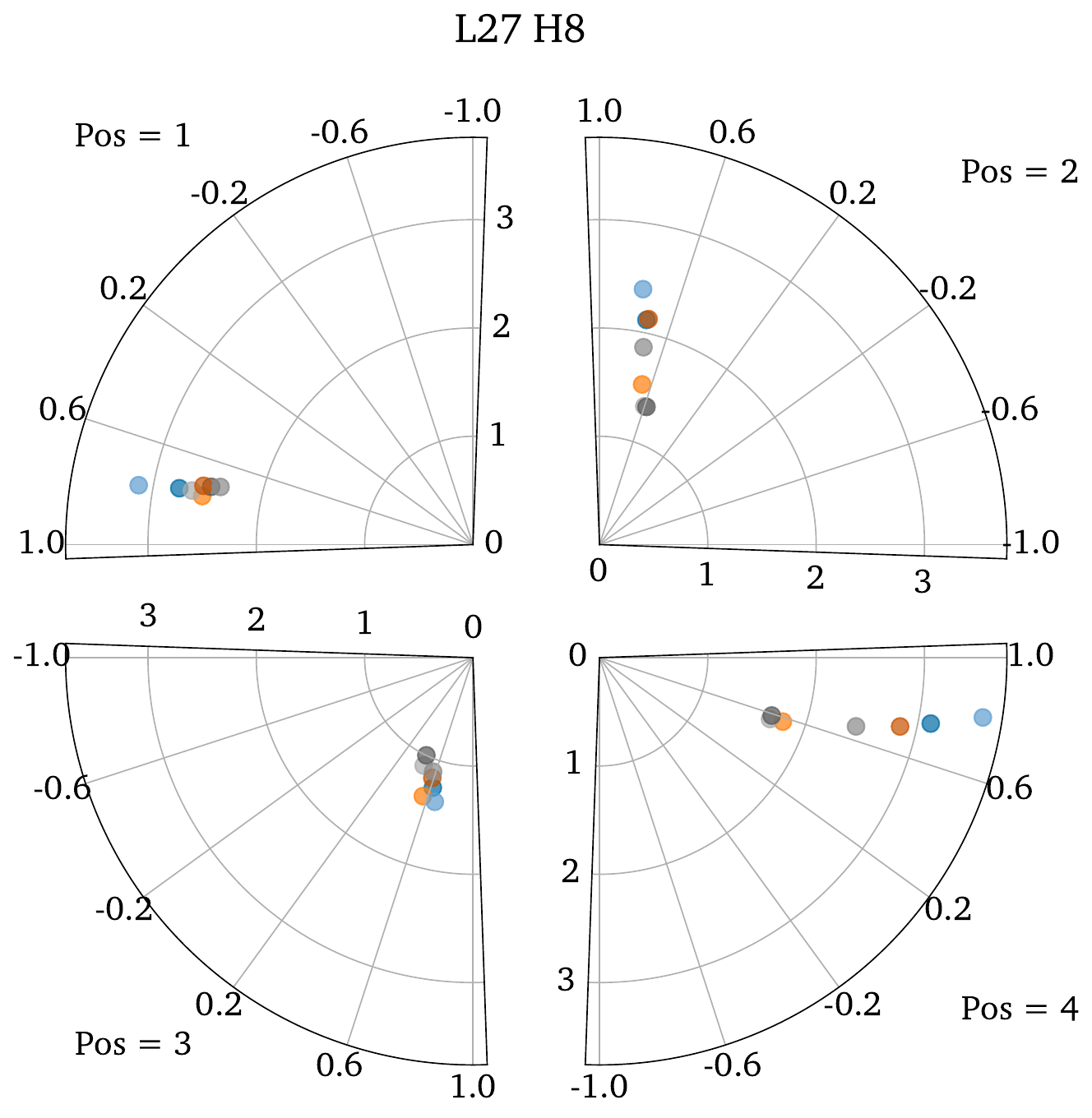}
         \caption{}
         \label{fig:project_q_deltas_on_k_centroids_l27h08}
     \end{subfigure}
     \hfill
     \begin{subfigure}[b]{0.3\textwidth}
         \centering
         \includegraphics[width=\textwidth]{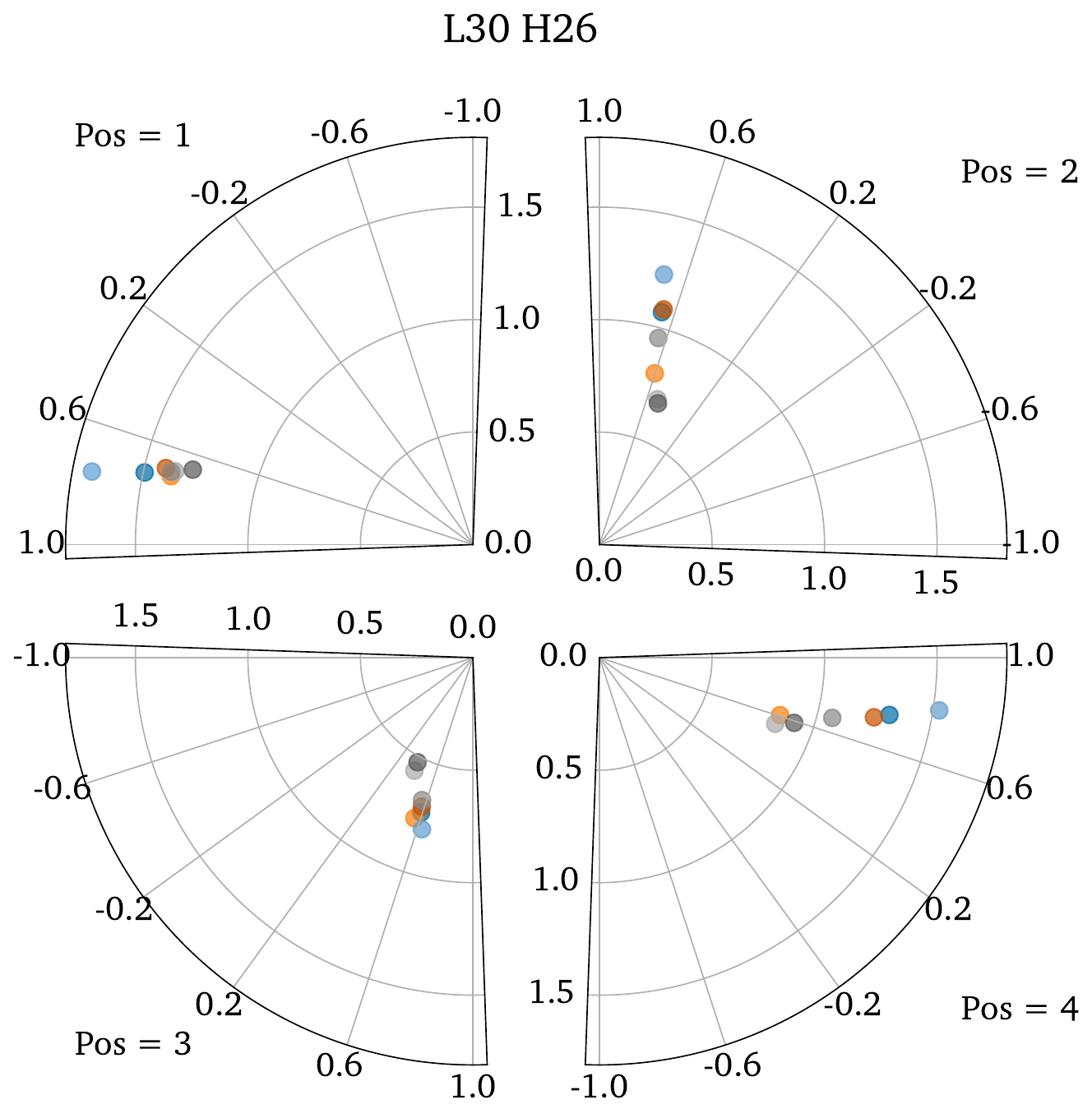}
         \caption{}
         \label{fig:project_q_deltas_on_k_centroids_l30h26}
     \end{subfigure}
     \hfill
     \begin{subfigure}[b]{0.3\textwidth}
         \centering
         \includegraphics[width=\textwidth]{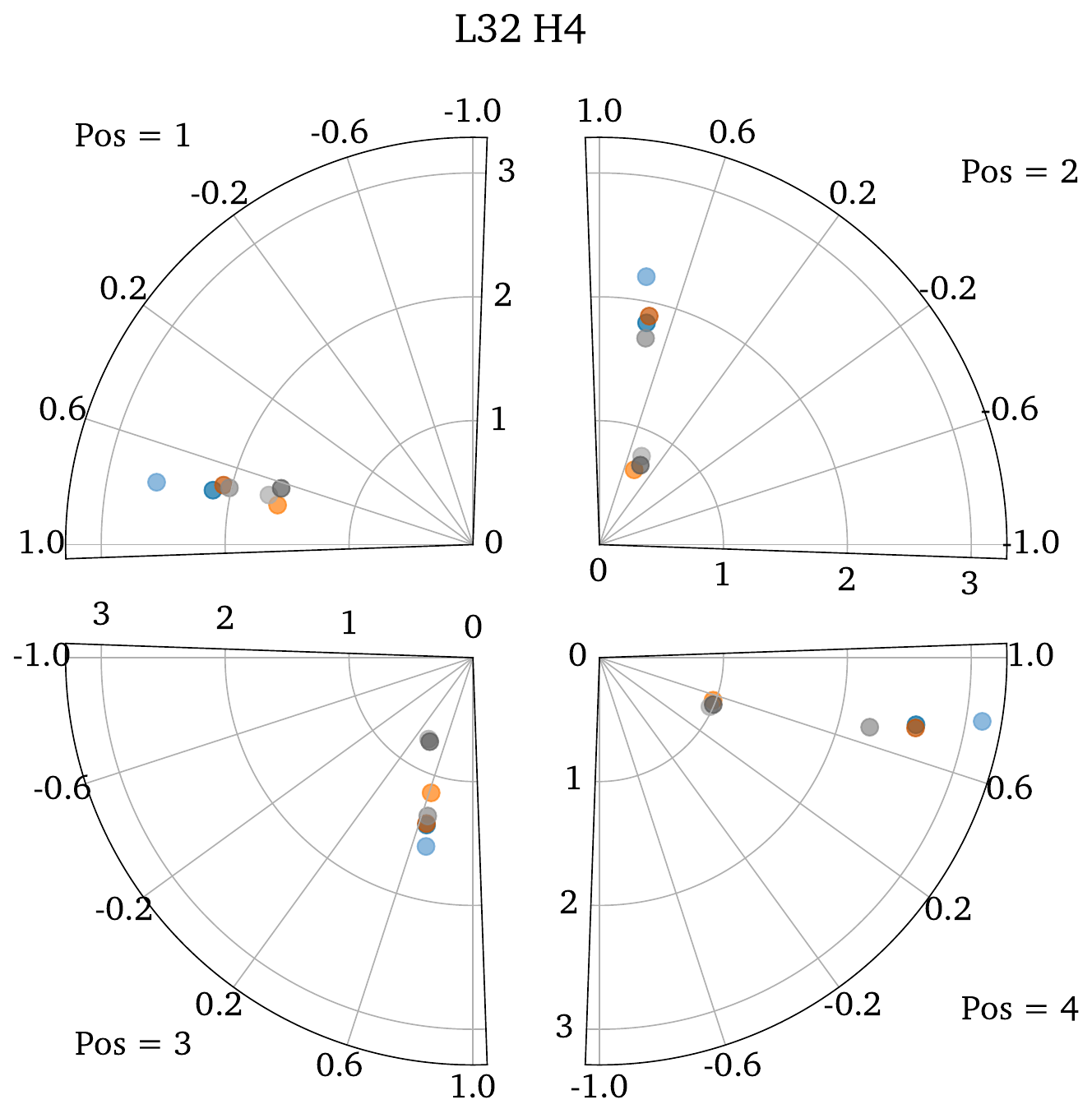}
         \caption{}
         \label{fig:project_q_deltas_on_k_centroids_l32h04}
     \end{subfigure}
     \hfill
     \begin{subfigure}[b]{0.3\textwidth}
         \centering
         \includegraphics[width=\textwidth]{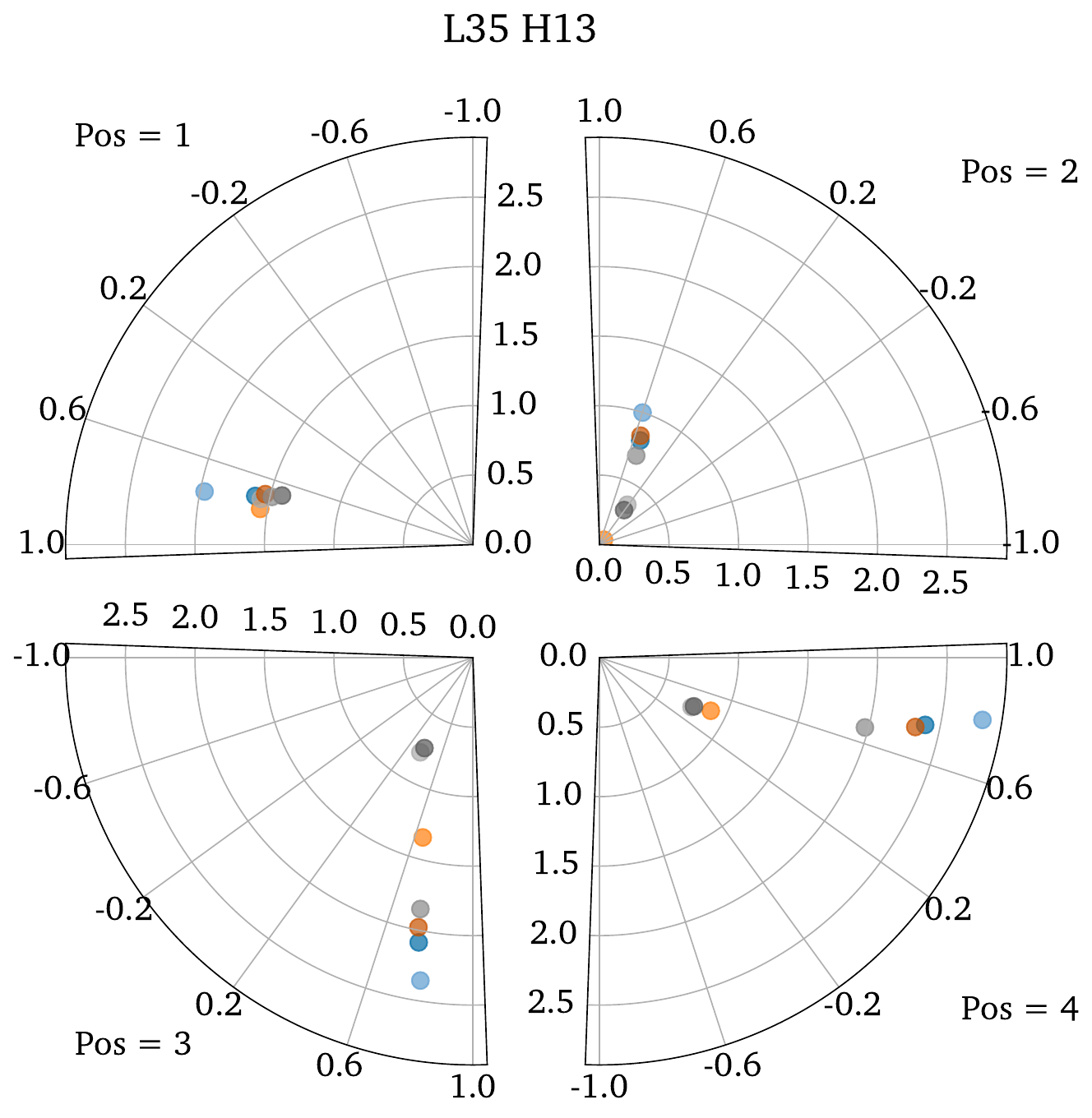}
         \caption{}
         \label{fig:project_q_deltas_on_k_centroids_l35h13}
     \end{subfigure}
     \hfill
     \begin{subfigure}[b]{0.3\textwidth}
         \centering
         \includegraphics[width=\textwidth]{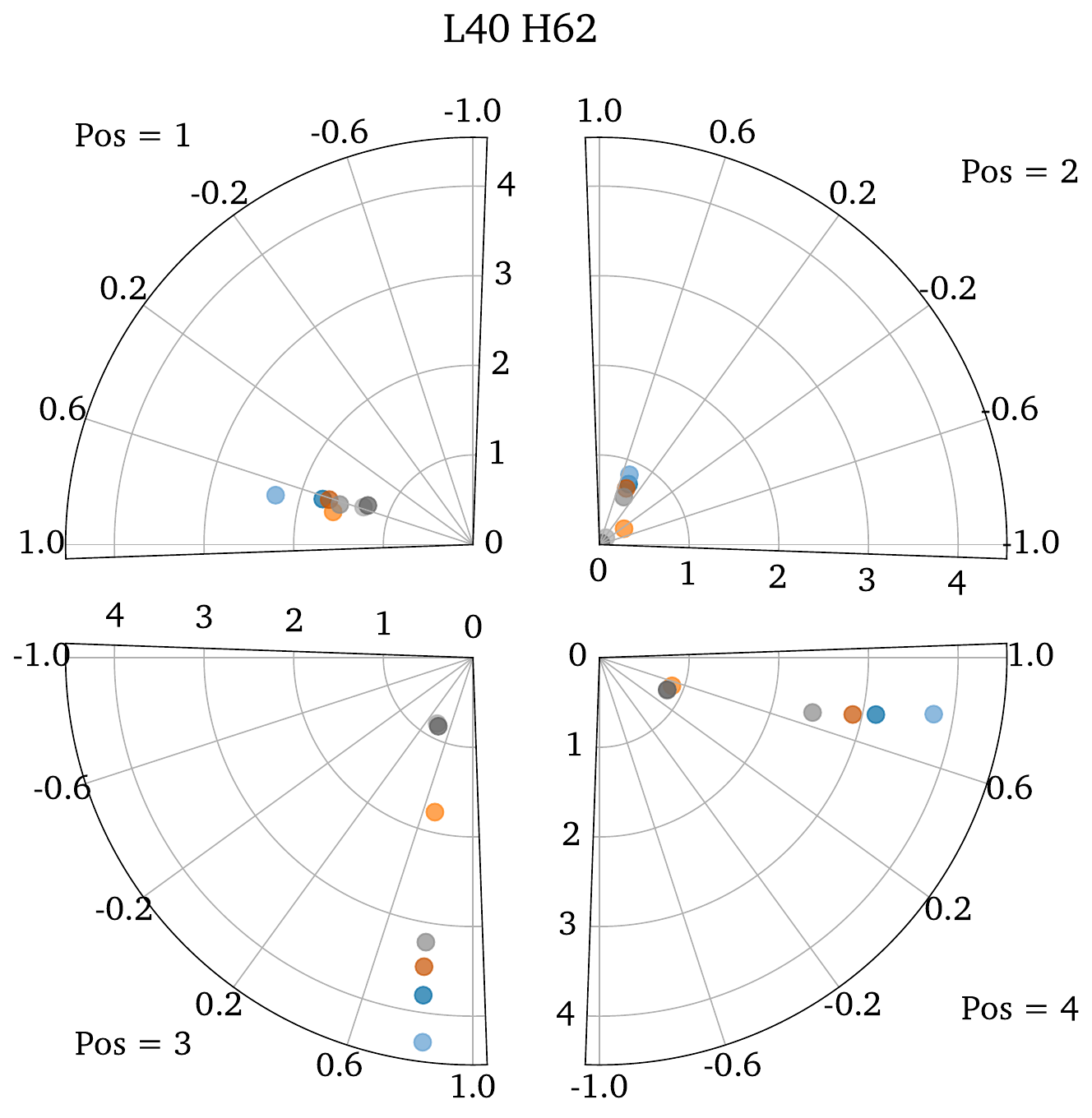}
         \caption{}
     \end{subfigure}
    \caption{Cosine similarity and absolute value of the projection of $q_\delta$ onto the centroids of the base $k_\delta$.}
    \label{fig:project_q_deltas_on_k_centroids}
\end{figure}

\begin{figure}[t]
    \centering
    \includegraphics[width=0.7\textwidth]{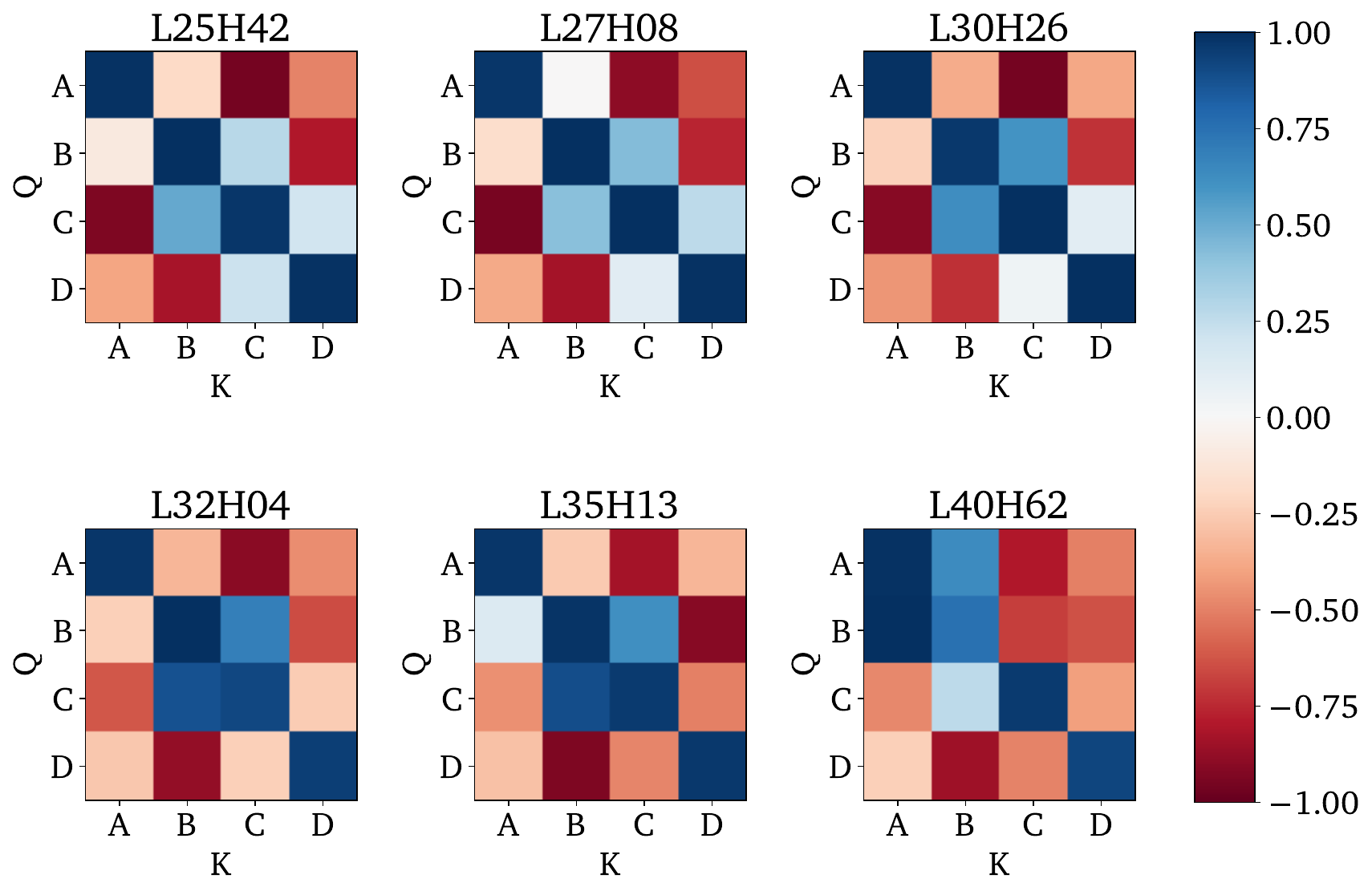}
    \caption{Cosine similarity between the centroids of the $k_\delta$ and $q_\delta$ on the base prompt settings. All heads exhibit a significant overlap of centroids between at least two letters.}
    \label{fig:cosine_sim_q_and_k_centroids}
\end{figure}

In this section we provide more detailed results on analyzing the low-rank approximation of the correct letter heads. \Cref{fig:low_rank_direct_effects} shows the change in direct effect when using low-rank queries, keys and values. In~\cref{fig:mutated_low_rank_loss_full} we report the change in loss when patching low-rank or full-rank attention under various prompt mutations. \Cref{fig:project_k_deltas_on_k_centroids,fig:project_q_deltas_on_k_centroids} display cosine similarity between the query and key deltas and key centroids and the projection of the deltas onto the centroids for all heads. Finally, we report the cosine similarity between the query and key centroids in~\cref{fig:cosine_sim_q_and_k_centroids}.

\begin{figure}[t]
    \centering
     \begin{subfigure}[b]{0.245\textwidth}
         \centering
         \includegraphics[width=\textwidth]{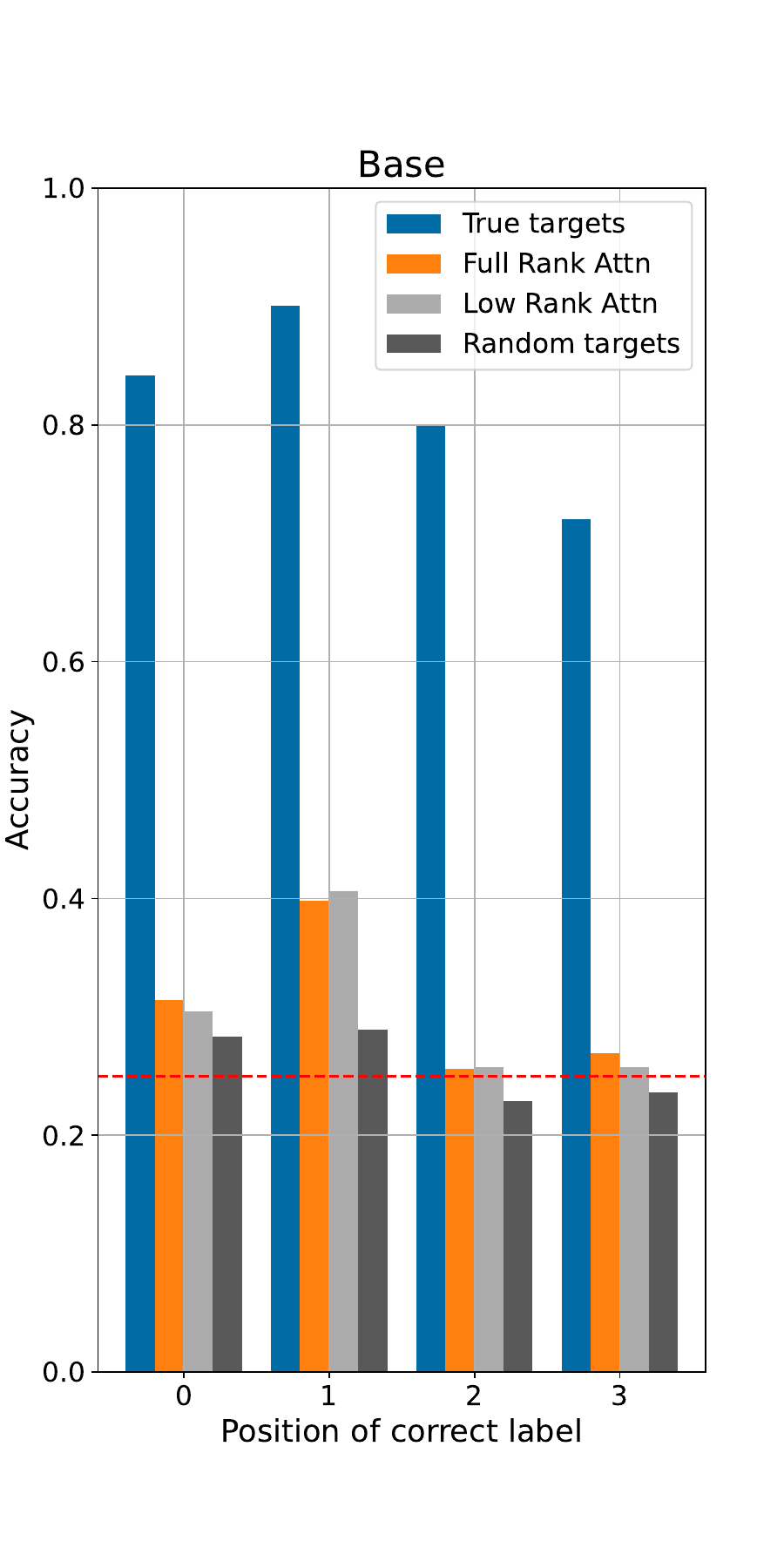}
         \caption{}
     \end{subfigure}
     \hfill
     \begin{subfigure}[b]{0.245\textwidth}
         \centering
         \includegraphics[width=\textwidth]{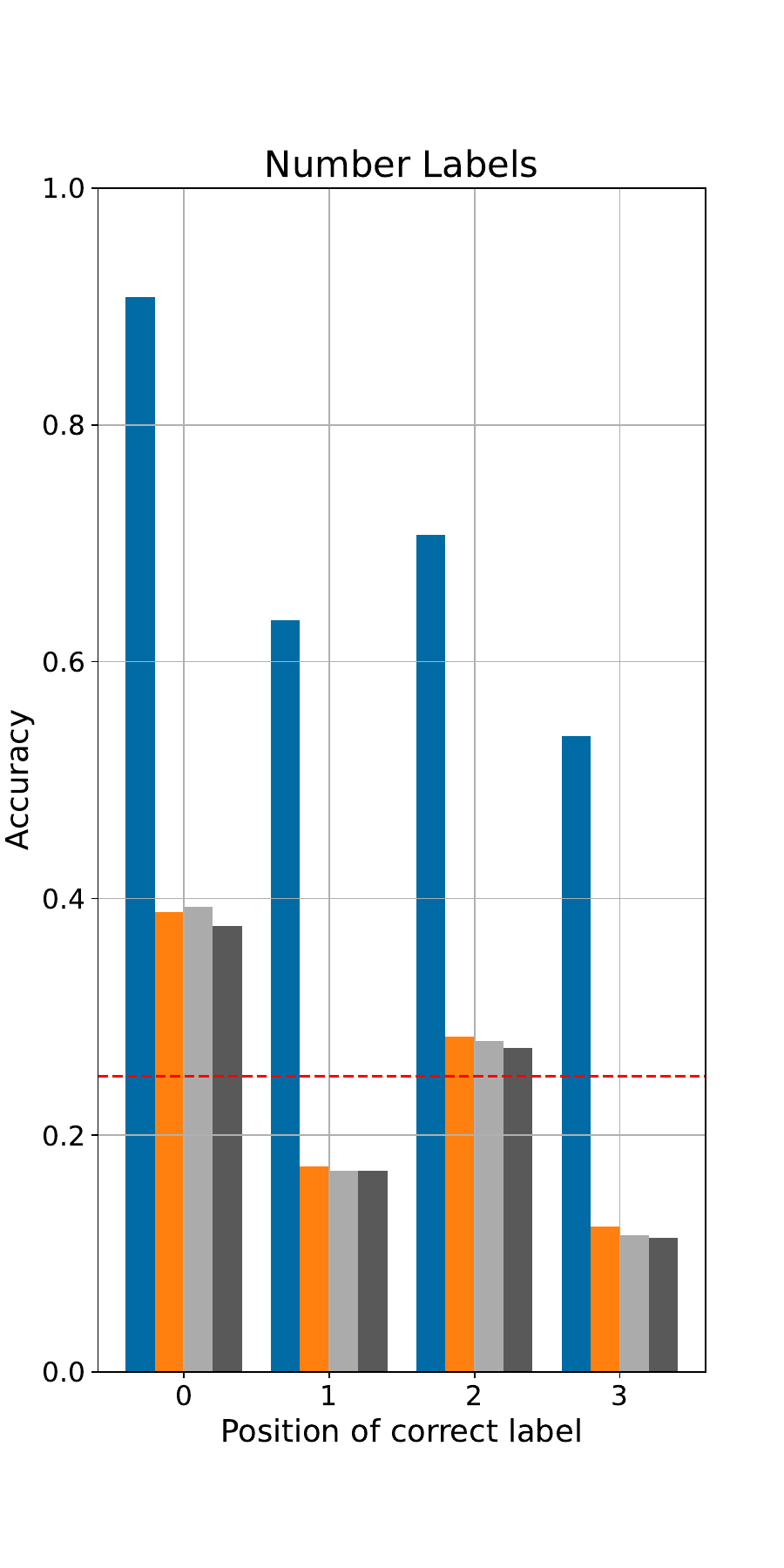}
         \caption{}
     \end{subfigure}
     \hfill
     \begin{subfigure}[b]{0.245\textwidth}
         \centering
         \includegraphics[width=\textwidth]{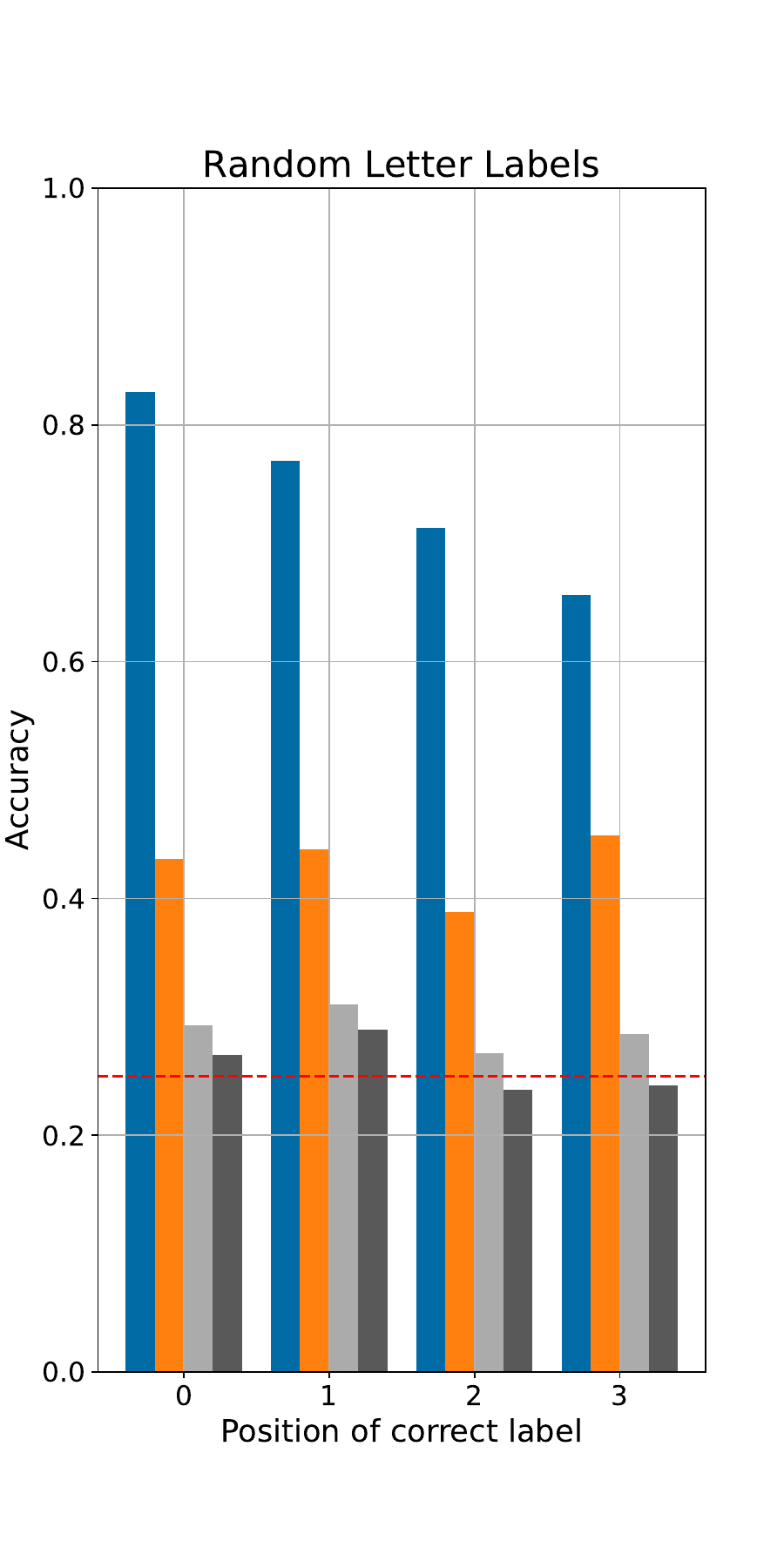}
         \caption{}
     \end{subfigure}
     \hfill
     \begin{subfigure}[b]{0.245\textwidth}
         \centering
         \includegraphics[width=\textwidth]{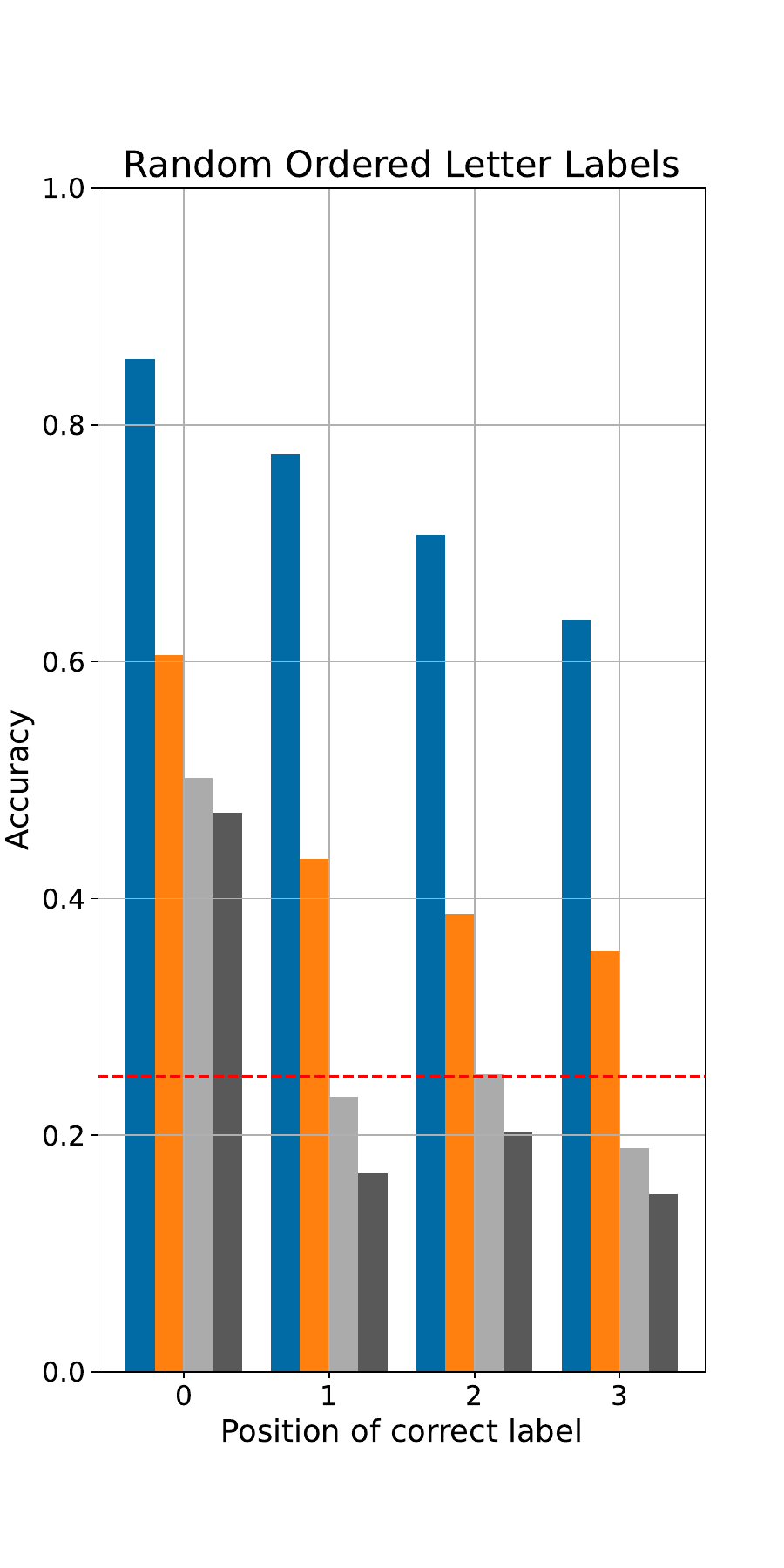}
         \caption{}
     \end{subfigure}
     \hfill
     \begin{subfigure}[b]{0.245\textwidth}
         \centering
         \includegraphics[width=\textwidth]{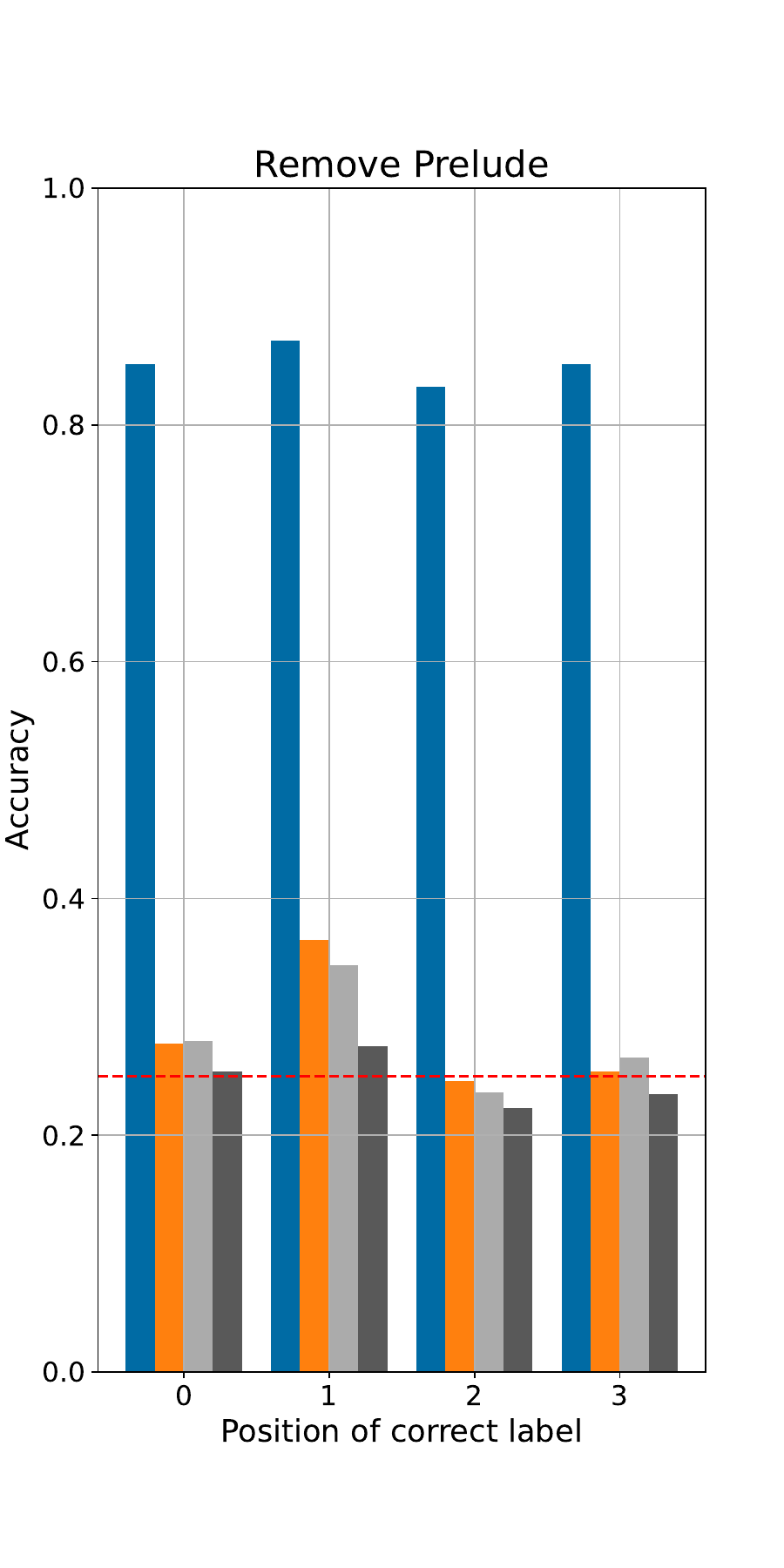}
         \caption{}
     \end{subfigure}
     \hfill
     \begin{subfigure}[b]{0.245\textwidth}
         \centering
         \includegraphics[width=\textwidth]{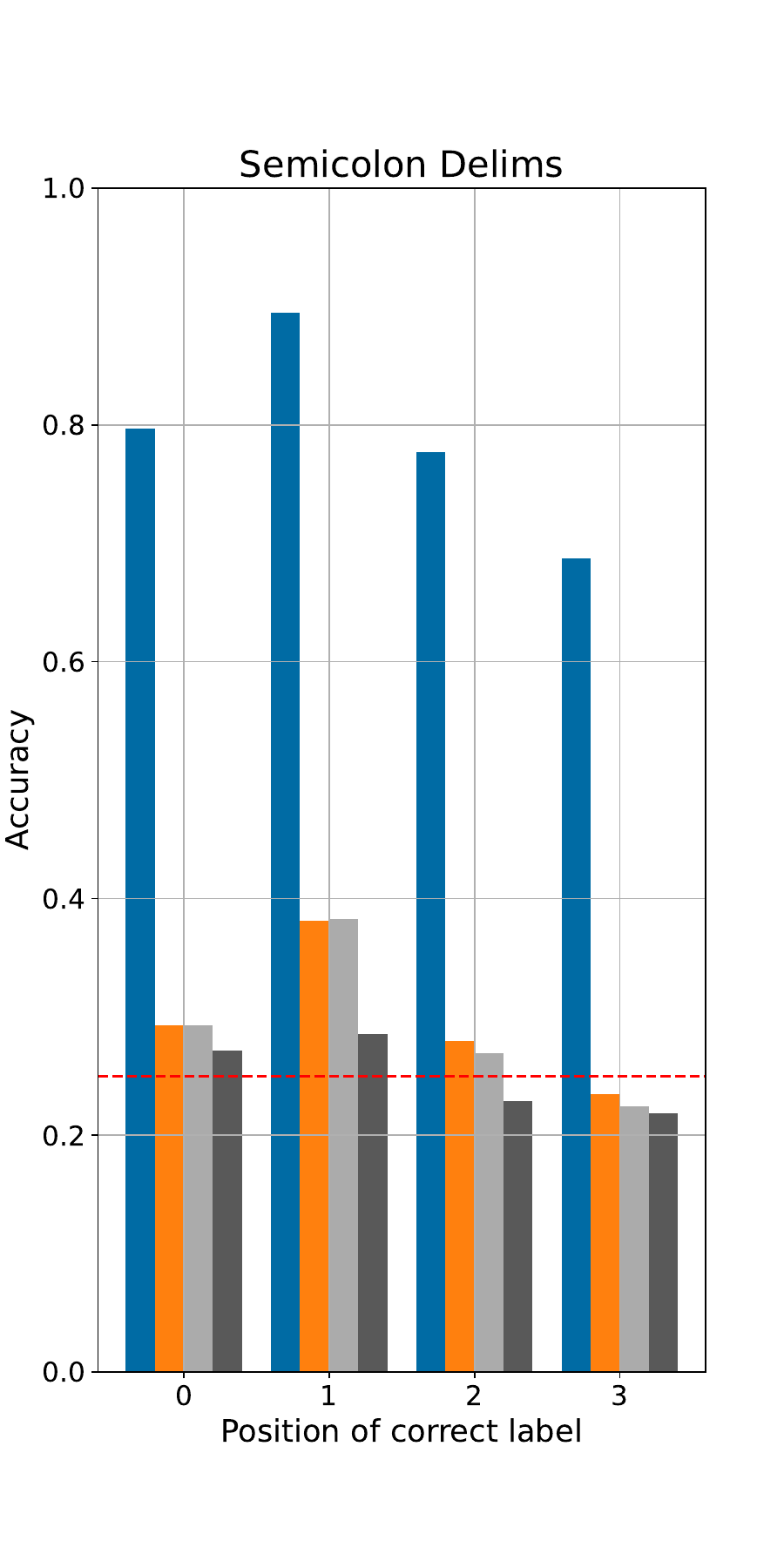}
         \caption{}
     \end{subfigure}
     \hfill
     \begin{subfigure}[b]{0.245\textwidth}
         \centering
         \includegraphics[width=\textwidth]{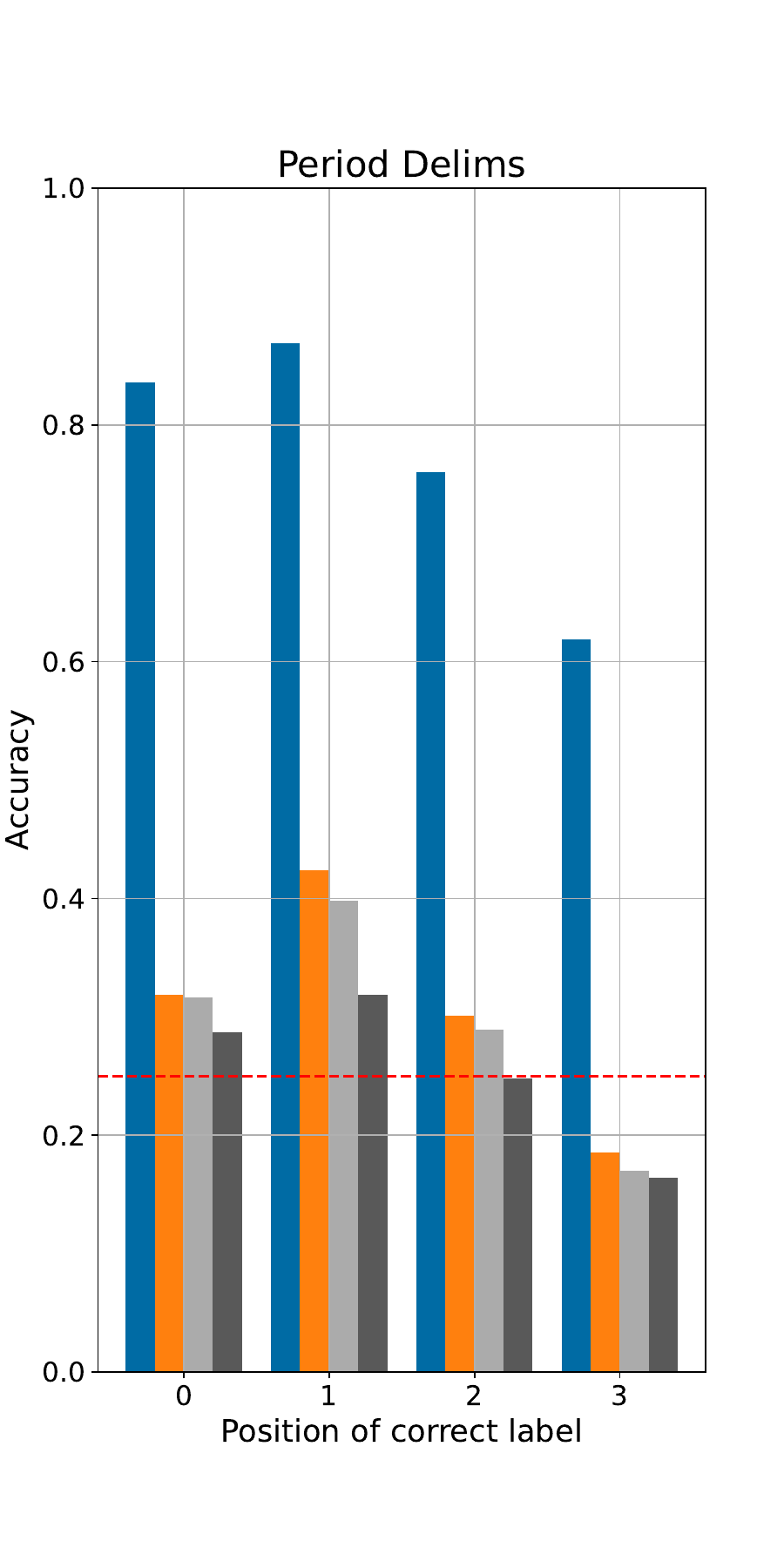}
         \caption{}
     \end{subfigure}
    \caption{Accuracy over the answer set when using full rank or low rank attention under various prompt mutations. `True targets' means running and evaluating the model on $p_{intervention}$, and `Random targets' means running on $p_{intervention}$ but evaluating on $p_{original}$.}
    \label{fig:mutated_low_rank_acc}
\end{figure}

\clearpage

\end{document}